\documentclass[]{final_report}

\usepackage[hidelinks]{hyperref}

\usepackage{algpseudocode}
\usepackage{algorithm}

\usepackage[nocompress]{cite}
\usepackage{amsfonts}
\usepackage{amsmath}
\usepackage{tocloft}

\usepackage{subcaption}

\usepackage{afterpage}
\usepackage{multicol}
\usepackage{footmisc}
\usepackage{units}
\usepackage{multirow}
\usepackage{booktabs}

\usepackage{soul}
\usepackage{color}
	\definecolor{celadon}{rgb}{0.67, 0.88, 0.69}
    \definecolor{flamingopink}{rgb}{0.99, 0.56, 0.67}

\usepackage{newfloat}

\usepackage{makecell}

\usepackage{enumitem}

\usepackage[table,xcdraw]{xcolor}
\usepackage{listings}
\usepackage{caption}

\definecolor{LightGray}{HTML}{444444}

\usepackage{graphicx}
\usepackage{hyperref}
\usepackage{lipsum}
\usepackage{refcount}

\def\studentname{Congcong Wang}

\def\projecttitle{Coping with low data availability for social media crisis message categorisation}
\def\supervisorname{Dr. David Lillis}
\def\moderatorname{Assoc. Prof. Neil Hurley}
\def\dspmembers{\\ Assoc. Prof. John Dunnion \\ Assoc. Prof. Mark Scanlon}
\begin{document}

\maketitle
\def\baselinestretch{1.5}\selectfont


\begingroup
  \hypersetup{hidelinks}
  \tableofcontents
\endgroup

\chapter*{Statement of Original Authorship}

``I hereby certify that the submitted work is my own work, was completed while registered as a candidate for the degree stated on the Title Page, and I have not obtained a degree elsewhere on the basis of the research presented in this submitted work.''

\chapter*{Dedication}
\vspace*{\fill}
\begin{center}
\begin{minipage}{.7\linewidth}

To my grandmother for her love.

\end{minipage}
\end{center}
\vfill 

\chapter*{Acknowledgements}

The pursuit of a Ph.D. degree can be a lengthy and isolating journey, one that would not have been possible without the help and support of numerous individuals. Firstly, I would like to extend my sincere gratitude to my supervisor, David Lillis. His unwavering support, patience, and encouragement have been invaluable. As a mentor, he provides me with constructive feedback on my research and as a friend, he inspires me when I experience feelings of self-doubt. Without his assistance, it would have been difficult to imagine how I could have made it this far on this journey.

I am grateful for the guidance and supervision provided by Paul Nulty in some of my research projects. His deep understanding of the domain and innovative ideas have been invaluable to my research work. Without his dedicated efforts, I would have faced significant challenges in completing some of my research work.

I am also grateful to my research committee members, John Dunnion and Mark Scanlon, for their guidance and support throughout my research. Their constructive feedback and suggestions have been invaluable in moving my work forward. Additionally, I would like to express my appreciation to the professors and staff in the school who have provided timely assistance and support whenever I have reached out to them.

Furthermore, I extend my gratitude to Tri Kurniawan Wijaya, Gonzalo Fiz Pontiveros, Steven Derby, and Puchao Zhang from Huawei IRC. Their valuable insights and suggestions have contributed to the improvement of my research work.


Lastly, I am deeply appreciative of my friends and family, whose unwavering support has been instrumental in my journey. I would like to extend my heartfelt gratitude to my colleague and dear friend Chidubem Iddianozie, who has been like a brother to me and has provided invaluable assistance not only in my research but also in my personal life. I am also grateful for the presence of my girlfriend Xiaoyu Du in my life since the beginning of this journey. It has been a wonderful experience to have her by my side, offering kindness, support, and inspiration during the tough times. I wish to express my special thanks to my parents for their unrelenting efforts in providing me with a good education since childhood. Their financial and emotional support have made a significant difference in shaping the person I am today. Lastly, I am indebted to my grandmother for her upbringing and nurturing during my formative years. I realise that words alone cannot fully convey my gratitude for her kindness and care.



\newlist{abbrv}{itemize}{1}
\setlist[abbrv,1]{label=,labelwidth=1in,align=parleft,itemsep=0.1\baselineskip,leftmargin=!}
 
\chapter*{\normalfont \textbf{List of Abbreviations}}

\chaptermark{List of Notations}
 
 
 \begin{abbrv}
 \item [$s$] An input sequence or instance
\item [$t$] Token or word
\item [$\textbf{x}$] Vector representing input features or representations of a sequence or sample
\item [$x_i$] The i-th feature of a sequence or sample
\item [$\textbf{X}$] Matrix representing input features or representations of a dataset
\item [$Y$] The class or label space
\item [$y$] A class or label
\item [$n$] Number of sequences or samples in a dataset
\item [$m$] Number of classes
\item [$k$] Feature or embedding size
\item [$T$] Number of tokens of a sequence
\item [$\theta$] Model parameters
\item [$p$] Probability density function
\item [$\textbf{W}$] Weight matrix
\item [$\textbf{b}$] Bias
\item [$\alpha$] Learning rate
\item [$J$] Loss or objective function
\item [$\mathcal{T}$] Task
\item [$D$] Domain

 \end{abbrv}

\chapter*{\normalfont \textbf{List of Publications}}

The thesis mainly discusses the following published articles:

\begin{itemize}
\item \textbf{Congcong Wang}, and David Lillis. ISA: Iterative Self-controlled Augmentation for Few Shot Text
Classification. \textit{Experiments finished and to be submitted} 2023.

\item \textbf{Congcong Wang}, Gonzalo Fiz Pontiveros, Steven Derby and Tri Kurniawan Wijaya. STA: Self-controlled Text Augmentation for Improving Text Classification. \textit{Submitted and under review,} 2023.

\item \textbf{Congcong Wang}, Paul Nulty, and David Lillis. Using Pseudo-Labelled Data for Zero-Shot Text Classification. In Proceedings of \textit{the 27th International Conference on Natural Language \& Information Systems (NLDB 2022)}, Valencia, Spain, June 2022.

\item \textbf{Congcong Wang} and David Lillis. UCD-CS at TREC 2021 Incident Streams Track. In \textit{Proceedings of the Thirty Text REtreival Conference (TREC 2021),} Gaithersburg, MD, USA, 2022.
\item \textbf{Congcong Wang}, Paul Nulty, and David Lillis. Crisis Domain Adaptation Using Sequence-to-Sequence Transformers. In \textit{Proceedings of 18th International Conference on Information Systems for Crisis Response and Management, pages 655--666 (ISCRAM 2021)}, Blacksburg, VA (USA), Virginia Tech, 2021.
\item \textbf{Congcong Wang}, Paul Nulty, and David Lillis. Transformer-based Multi-task Learning for Disaster Tweet Categorisation. In \textit{Proceedings of 18th International Conference on Information Systems for Crisis Response and Management, pages 705--718 (ISCRAM 2021)}, Blacksburg, VA (USA), Virginia Tech, 2021. 
\item \textbf{Congcong Wang} and David Lillis. Multi-task transfer learning for finding actionable information from crisis-related messages on social media. In \textit{Proceedings of the Twenty-Ninth Text REtreival Conference (TREC 2020),} Gaithersburg, MD, USA, 2021.
\item \textbf{Congcong Wang} and David Lillis. Classification for Crisis-Related Tweets Leveraging Word Embeddings and Data Augmentation. In \textit{Proceedings of the Twenty-Eighth Text REtreival Conference (TREC 2019),} Gaithersburg, MD, USA, 2020.
\end{itemize}

The following articles are related, but will not be thoroughly discussed in this
thesis:
\begin{itemize}
\item \textbf{Congcong Wang} and David Lillis. A Comparative Study on Word Embeddings in Deep Learning for Text Classification. In \textit{Proceedings of the 4th International Conference on Natural Language Processing and Information Retrieval (NLPIR 2020)}, Seoul, South Korea, Dec. 2020.
\item \textbf{Congcong Wang} and David Lillis. UCD-CS at W-NUT 2020 Shared Task-3: A Text to Text Approach for COVID-19 Event Extraction on Social Media. In \textit{Proceedings of the Sixth Workshop on Noisy User-generated Text (W-NUT 2020), pages 514--521}, Online, Nov. 2020. Association for Computational Linguistics.


\end{itemize}


\begin{abstract}

During crisis situations, social media allows people to quickly share information, including messages requesting help. This can be valuable to emergency responders, who need to categorise and prioritise these messages based on the type of assistance being requested. However, the high volume of messages makes it difficult to filter and prioritise them without the use of computational techniques. Fully supervised filtering techniques for crisis message categorisation typically require a large amount of annotated training data, but this can be difficult to obtain during an ongoing crisis and is expensive in terms of time and labour to create.

This thesis focuses on addressing the challenge of low data availability when categorising crisis messages for emergency response. It first presents domain adaptation as a solution for this problem, which involves learning a categorisation model from annotated data from past crisis events (source domain) and adapting it to categorise messages from an ongoing crisis event (target domain). In many-to-many adaptation, where the model is trained on multiple past events and adapted to multiple ongoing events, a multi-task learning approach is proposed using pre-trained language models. This approach outperforms baselines and an ensemble approach further improves performance. In one-to-one or many-to-one adaptation, this research studies which combination of past events to include in the model to achieve the best adaptation performance for a particular target event. An approach using sequence-to-sequence pre-trained language models is proposed that incorporates event information for crisis message categorisation, and it is found to outperform existing state-of-the-art methods. The study also finds that using past events that are more similar to the target event tends to lead to better adaptation performance, while using dissimilar events does not improve performance.

 However, crisis domain adaptation is only effective when the categorisation task is the same for both the source and target event and there is sufficient annotated data available from the source event. To address the situation where there is very limited labelled data is available relating to the target event, the research presents a self-controlled augmentation approach and an optimised iterative self-controlled augmentation approach to generate additional crisis data for model training. These approaches are able to generate high quality crisis data, leading to better classification performance compared to other methods in the few-shot learning scenario. Additionally, the research presents a method for training a categorisation model in a zero-shot setting, where there is no time to annotate any data for the new event. This involves matching label names with the unlabelled data of the target event and creating a pseudo-labelled dataset with high confidence for model training. The results show that this approach is able to effectively pseudo-label the unlabelled data, resulting in better performance compared to other zero-shot methods. The proposed few-shot and zero-shot approaches are also tested in other domains such as emotion and topic classification, and demonstrate superior generalisation performance compared to baselines in these domains.

This thesis contributes to the crisis informatics research by coping with low annotated data availability of emerging events for crisis message categorisation on social media. The approaches presented in the thesis are developed in close association with real-world situations and show top performance in experiments. The approaches have the potential to be used in practice for timely and effective humanitarian aid response.

\end{abstract}

\newpage

\pagenumbering{arabic}

\chapter{Introduction}
\label{ch:introduction}
\section{Research context}
\label{sec:intro-res-context}

The widespread use of social media and the abundance of user-generated content in recent years has made it a valuable tool for connecting people. In the context of emergency situations, social media has gained attention as a means of communication and coordination for emergency response agencies and humanitarian organisations, which aim to provide timely and efficient responses to time-sensitive events such as natural disasters or human-induced hazards~\cite{fraustino2012social}.


According to~\cite{guha2011annual}, natural disasters result in an average of 50,000 deaths worldwide each year. Delays in obtaining critical information during an emergency can not only cause additional property damage but also put lives at risk. Therefore, the timely delivery of emergency response is essential in minimising the impact of such events. Social media's ability to provide real-time communication between those affected by an incident and emergency aid centers can help improve situational awareness~\cite{vieweg2010microblogging,vieweg2012situational,endsley2017toward}, which refers to the effective and accurate understanding of how an incident is unfolding, enabling response services to take timely preventive measures to address the situation. To be specific, emergency services can use social media in several ways to improve situational awareness, for example by monitoring social media platforms for messages and posts about the crisis to gain real-time information about the situation on the ground, or by analysing the sentiment and emotional content of social media posts to understand the overall impact of the crisis on the community, or by capturing location information associated with social media posts to identify and respond to specific areas in need of assistance.



According to a study, 69\% of people believe that emergency response operators should monitor their social media accounts and respond promptly during a crisis~\cite{jolie2021}. Another study found that about 10\% of emergency-related posts on Twitter (called ``tweets'') are important and about 1\% are critical~\cite{McCreadie2019}, highlighting the potential of social media for emergency response. Traditionally, tracking emergencies on social media has been done manually, with human workers filtering and classifying messages as critical or not~\cite{lambert2005internet,palen2007citizen}. While this approach can achieve high precision, it is very labour-intensive and costly in terms of time, especially during crises when the number of relevant messages often increases rapidly. This has motivated the development of computational techniques for emergency tracking and response.~\cite{McCreadie2019,mccreadie2020trec,imran2013extracting,olteanu2014crisislex,olteanu2015expect,zahra2020automatic}.

\begin{figure}[!h]
    \centering
    \includegraphics[width=\linewidth]{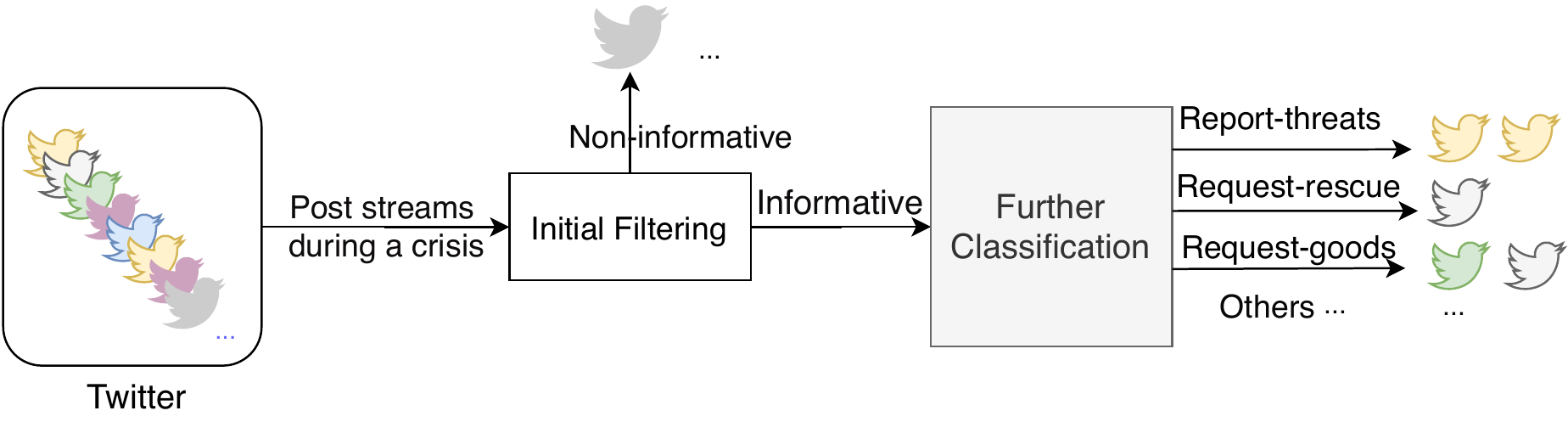}
    \caption{The workflow of automatic crisis message categorisation for emergency response on social media.}
    \label{fig:sys-workflow}
\end{figure}

Taking Twitter as an example\footnote{Throughout this thesis, unless stated otherwise, Twitter is the default social media platform being discussed due to its high popularity and timeliness.}, this research introduces the workflow of automatic crisis message categorisation for emergency response with two phases, namely initial filtering and further classification, as presented in Figure~\ref{fig:sys-workflow}. It begins with a ``post stream'' of messages describing an ongoing crisis. These tend to be noisy and numerous, and are initially filtered by monitoring hashtags or keywords related to the crisis. This initial filtering aims to find those that are informative to users aid needs (i.e. what assistance the users are trying to seek from emergency responders). 
However, only finding the informative messages is insufficient for efficient emergency response because the categories of aid needs have different priorities (e.g. search and rescue is more critical than weather and location). This motivates the next phase: further classification is needed to identify these categories. The categories refer to information types~\cite{mccreadie2019trec} that represent different user needs during a crisis. For example, a message could be requesting rescue (for oneself or others) or asking for donations of goods or money. Based on the definition, each informative message after the initial filtering is further classified into one or more of the information types and finally pushed to corresponding emergency operators in accordance with the requirements of their roles.

The problem of social media crisis message categorisation falls into the area of text classification, which is a form of natural language processing (NLP). Text classification has been widely studied from traditional machine learning approaches~\cite{sebastiani2002machine} such as Na\"ive Bayes to more recent transfer learning methods using pre-trained language models such as BERT and T5~\cite{vaswani2017attention,devlin2018bert,ruder2019neural,raffel2020exploring}. However, most of these methods are based on fully-supervised learning, which requires a large amount of annotated data for model building. For example, to create a model that can classify the topic of a new piece of news, a dataset of news articles that have been labelled with their topics are needed to train the model. In the domain of crisis response, it is often difficult to use fully-supervised learning techniques because annotated data of emerging events is not readily available. This is because crises happen suddenly and it can be costly and time-consuming to manually label a dataset of crisis-related social media messages. As a result, there is a need for methods that can categorise crisis messages effectively even when there is limited labelled data available. This research aims to address this problem by developing methods for categorising social media crisis messages.


\section{Research approaches}
\label{sec:research-approches}

As mentioned above, the difficult and time-consuming nature of annotating data from ongoing crisis events makes it challenging to create categorisation models for emergency response using fully-supervised approaches. This research addresses this challenge by proposing three scenarios that explore different sources of data that is feasible to use for model development.

The first scenario, called ``crisis domain adaptation'', involves using annotated data from past crisis events (source events) to create a categorisation model that is capable of operating when no annotated data from ongoing events (target events) is available. This is feasible because although it still requires time and human labour, annotating data for past crises is easier and less pressing than annotating data for ongoing events and in most cases, new events have strong similarities to some past event or events.

The second scenario, called ``crisis few-shot learning'', involves developing the model using a small amount of labelled data from a target event. This is feasible because it requires little effort to annotate a small amount of data for ongoing crises.

\begin{figure}[!ht]
    \centering
\includegraphics[width=0.8\linewidth]{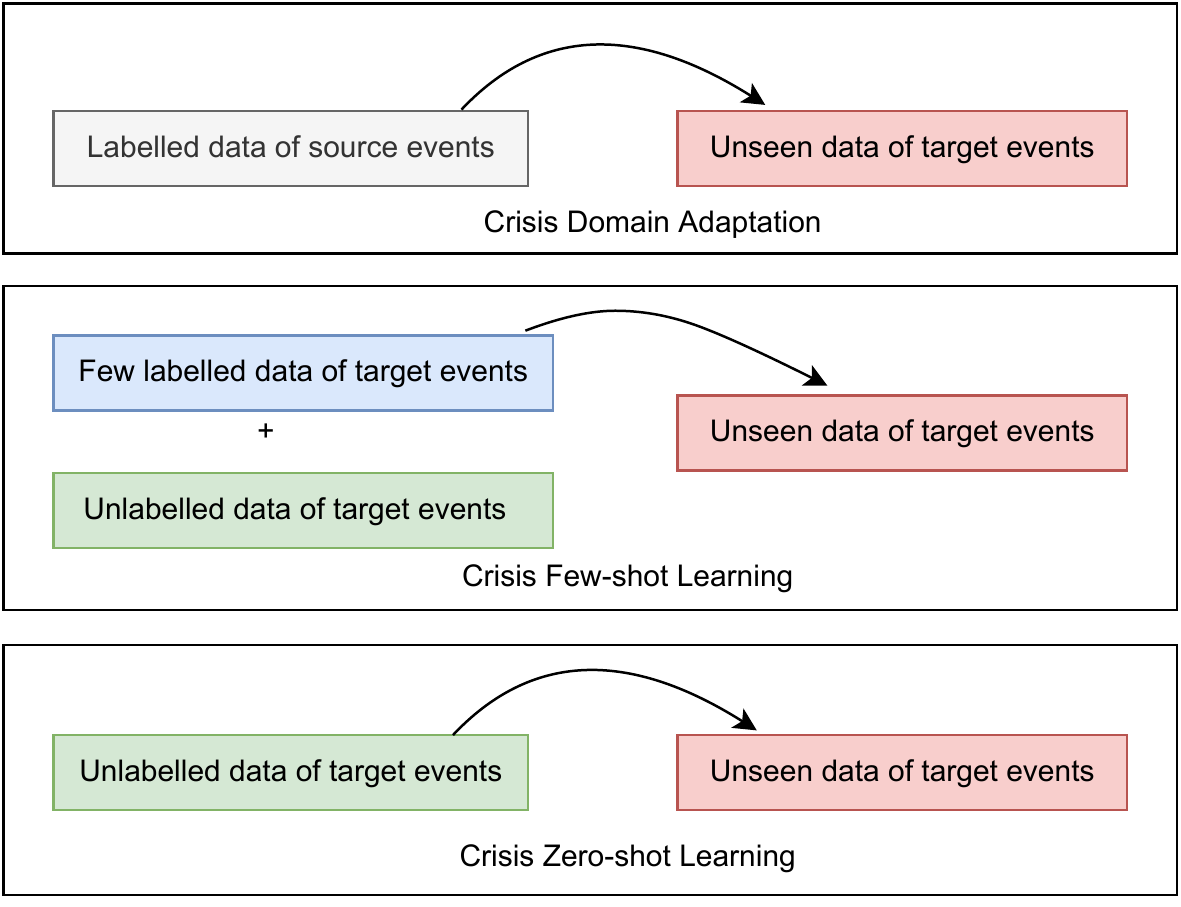}
    \caption{Three research scenarios of this research: crisis domain adaptation, crisis few-shot learning and crisis zero-shot learning.}
    \label{fig:research-problems}
\end{figure}

The third scenario, called ``crisis zero-shot learning'', involves creating the model using only unlabelled data from a target event, which is easy to obtain as the event unfolds. Figure~\ref{fig:research-problems} presents the three research scenarios that are discussed in the following text.


Crisis domain adaptation involves adapting a model trained on labelled data from source events (source domain) to classify messages from target events (target domain) that were not seen during training. This raises two research challenges. The first, referred to as many-to-many adaptation, arises when responders want to deploy a categorisation model to target events without knowing the details of the crisis events. It involves training the model on the source domain of multiple events and adapting it to the target domain of any event. Additionally, it is also of interest to determine the best way to select suitable past events to be the source domain and adapt the model to a target event when there is basic knowledge about the events, such as the type (flood or fire) and location. This is known as many/one-to-one adaptation, and is a more targeted adaptation to a specific target event than many-to-many adaptation. In this scenario, the goal of the research is to create models that are based on pre-trained language models in order to improve performance in both types of adaptation.


In emergency response, responders take appropriate action based on the categorisation of crisis messages into specific types of aid needs (information types). In crisis domain adaptation, messages from ongoing events are categorised into information types by a model that is trained on annotated data from past events with the same set of information types. However, in reality, responders may need to look for new information types for emerging events. For example, in a fire event, responders may ask for information about ``firefighting'', but this type of information may not have been included in the annotations for past events such as terrorist attacks and floods. This makes it difficult to adapt the model trained on past events, which do not include the new information types, to a target event with the new information types. This motivates the research to go beyond crisis domain adaptation.



The crisis few-shot learning scenario poses the question of how to build a categorisation model that can operate with only a limited amount of labelled target data, but that may have access to larger quantities of unlabelled data relating to the target event. Since it is not practical to annotate a large volume of target data for a new event, a small amount of annotated training data can be easily collected at the start of the new event. The crisis few-shot learning approach uses this small labelled dataset to build the model, making it a more feasible solution with regard to annotation efforts. As the crisis event progresses, a growing quantity of unlabelled data from the event becomes available and can be used to further refine the model. In this scenario, the goal of the research is to improve the performance of a categorisation model by using pre-trained language models to augment the small labelled dataset, then fine-tuning the model with unlabelled target data.


The crisis zero-shot learning scenario poses the question of how to classify unseen crisis messages using only unlabelled target data, which is inexpensive to obtain. This is challenging because the model must operate without any reliance on supervision resources in the domain to map an information type to an unseen crisis message. Despite the difficulty, this approach has the advantage of not requiring any labelled data to build the categorisation model, making it the most efficient solution in terms of annotation efforts. The goal of this research in this scenario is to use information types and a corpus of unlabelled target data to improve the performance of crisis message categorisation.

\section{Research questions}

The following are the primary research questions that this thesis aims to address based on the presented research scenarios:

\begin{itemize}
    \item Regarding crisis domain adaptation, how can pre-trained language models be utilised to enhance the performance of crisis messages categorisation? For many-to-many adaptation, what are the machine learning models and pre-trained language models that can improve the categorisation performance? For many/one-to-one adaptation, what factors should be considered when selecting past events to serve as the source domain and adapting the model to a target event?
    \item In crisis few-shot learning, what techniques can be used to identify new information types that may be required for crisis messages categorisation in emerging events? How can unlabelled target data be used to improve the performance of categorisation in crisis few-shot learning settings?
    \item In crisis zero-shot learning settings, how can information types and a corpus of unlabelled target data be used to improve the performance of crisis message categorisation?
    \item Last, what are the limitations of using each of the proposed scenarios for building categorisation models in crisis response, and how can these limitations be addressed?
\end{itemize}

\section{Research contributions}

Based on the aforementioned research scenarios and questions, the contributions of this thesis are summarised as follows.

\begin{itemize}
    \item In practice, it is important to address the challenge of low data availability for categorising social media crisis messages during emerging events. To study this issue systematically, in this thesis, three research scenarios are proposed based on the level of data availability: crisis domain adaptation, crisis few-shot learning, and crisis zero-shot learning. Each of these scenarios plays a crucial role in effective and timely emergency response.
    \item In the crisis domain adaptation scenario, a multi-task learning (MTL) approach using pre-trained language models like BERT is proposed for many-to-many crisis domain adaptation. The results indicate that the MTL method performs better than state of the art baseline methods, and an ensemble approach based on the MTL approach further improves performance.
    \item For many/one-to-one adaptation, a method called CAST is proposed that uses sequence-to-sequence pre-trained language models like T5 incorporating event information for crisis message categorisation. Comparison with existing state-of-the-art crisis domain adaptation approaches shows that CAST, which has the least dependence on target data, also outperforms these methods in terms of effectiveness. In addition, the combination of similar events is found to be more effective for adaptation compared to the addition of dissimilar events to the training set.
    \item For crisis few-shot learning, a self-controlled augmentation (STA) approach is proposed to generate training crisis messages using a few labelled seed messages for crisis messages categorisation. When tested in low data regimes, STA outperforms state-of-the-art augmentation approaches. The generated messages produced by STA demonstrate better quality compared to state-of-the-art approaches. To further improve STA, an iterative mechanism and a de-duplication mechanism (referred to as ISA) are added to the STA pipeline. ISA is very effective in improving the categorisation performance when tested in few-shot settings and its performance can be further improved by leveraging a target unlabelled to refine the categorisation model. STA and ISA not only show state-of-the-art performance in the crisis domain, but also exhibit strong performance on other text classification domains such as emotion or news topic classification.
    \item For crisis zero-shot learning, an approach called P-ZSC is proposed that pseudo-labels an unlabelled corpus for model training. P-ZSC not only performs better than other crisis-related approaches, but also has better generalisation to non-crisis-related text classification tasks. When the pseudo-labelled data produced by P-ZSC is examined, it is found to be as effective as a supervised approach using hundreds of manually annotated samples in both crisis categorisation and non-crisis classification tasks, suggesting the potential to save significant time and effort in training a model for these tasks.
    \item While the proposed methods still have some distance to cover before they can match the performance of fully-supervised approaches with abundant training data, these latter approaches serve as an interesting benchmark, representing the upper limit of what can currently be achieved under ideal circumstances. However, the proposed methods do outperform their corresponding baselines, making them a significant improvement over current state-of-the-art adaptation, zero, and few-shot approaches, and bringing the community closer to closing the gap between these approaches and their fully-supervised counterparts.
    
\end{itemize}

    
    



\section{Thesis outline}

The outline of this thesis is summarised as follows.

\begin{itemize}
    \item \textbf{Chapter~\ref{ch:background}: Background.} As crisis message categorisation falls under the umbrella of text classification, this chapter provides a general background review of text classification. It begins by introducing the basics of machine learning and relevant methods, followed by a discussion of deep neural network methods such as word embeddings and pre-trained language models. These methods are essential for building computational models for text classification.
    \item \textbf{Chapter~\ref{ch:literature}: Crisis Message Categorisation in Low-data Settings.} This chapter focuses on a review of the literature on crisis message categorisation in low-data settings. It covers existing methods in the literature for crisis domain adaptation, few-shot learning, and zero-shot learning. While some of these methods were initially developed for general text classification, they can be easily adapted for use in the crisis domain, making them the major works used as baselines in this research. The chapter also reviews relevant datasets and evaluation metrics used in the research.
    \item \textbf{Chapter~\ref{ch:adaptation}: Many-to-many Crisis Domain Adaptation.} This chapter presents various methods for adapting to crisis domain data in a many-to-many setting. It begins by introducing single-task methods and then moves on to multi-task methods that utilise pre-trained language models. The chapter provides detailed explanations of each method and offers insights into how machine learning and deep learning techniques can be applied to achieve many-to-many adaptation.
    \item \textbf{Chapter~\ref{ch:adaptation2}: Many-to-one and One-to-one Crisis Domain Adaptations.}
    This chapter introduces CAST, a method that uses sequence-to-sequence models and incorporates event information for many-to-one and one-to-one domain adaptations. The details of the methods, the experimental setup for testing its effectiveness, and key insights are included in this chapter.
    \item \textbf{Chapter~\ref{ch:sta-isa}: Augmentation for Crisis Few-shot Learning.} This chapter provides details on the self-controlled (STA) and iterative self-controlled (ISA) augmentation techniques for crisis few-shot learning. It compares these techniques to existing baselines, including both augmentation and other few-shot approaches in low data regimes. The chapter also includes ablation studies and an examination of the generated data produced by the methods and baselines.
    \item \textbf{Chapter~\ref{ch:pzsc}: Using Pseudo-labelled Data For Crisis Zero-shot Learning.} This chapter introduces the P-ZSC method for crisis zero-shot learning, which involves the use of pseudo-labelled data. The chapter discusses the process of label expansion, pseudo label assignment and refinement on unlabelled target data. In addition to categorisation results, the chapter includes the results of an ablation study and a study of the quality of the pseudo-labelled data.
    \item \textbf{Chapter~\ref{ch:conclusion}: Conclusion.} This chapter summarises the achievements of this research, outlines the major contributions of the study, and discusses future directions for addressing the challenge of low data availability in the categorisation of social media crisis messages.
\end{itemize}




\chapter{Background}
\label{ch:background}

This chapter presents the background knowledge that the subsequent chapters will build upon. As the core task in this research is to apply machine learning methods for social media crisis messages categorisation, this chapter focuses on the background of machine learning for text classification in the area of natural language processing (NLP). Regarding the content structure, the basic elements of machine learning are reviewed and linear regression (a preliminary form of neural networks) for supervised text classification is introduced as the first part of this chapter (Section~\ref{sec:ml}). Neural Networks are a specific category of machine learning models that have been widely and successfully applied to text classification. They are introduced as the second part of this chapter, including convolutional neural networks, recurrent networks and pre-trained language models (Section~\ref{sec:nns}).

\section{Machine learning}
\label{sec:ml}
This section first reviews key concepts of machine learning at a high level and then introduces a specific example of machine learning approaches---linear regression that can be regarded as a simple neutral network. In the subsection of linear regression, the process of building a logistic regression model for text classification is detailed.

\subsection{Key concepts}

According to the definition by Mitchell (1997)~\cite{mitchell1997machine}, machine learning is viewed as: ``A computer program is said to learn from experience $E$ with respect to some class of tasks $T$ and performance measure $P$, if its performance at tasks in $T$, as measured by $P$, improves with experience $E$.'' From this definition therefore, there are major three concepts regarding machine learning: the task $T$, the performance $P$ and the experience $E$.

\textbf{Text classification as the task}: The tasks of machine learning ($T$) are various. It can be a task of enabling a robot to play a game, a task of finding human faces in a picture or a task of identifying emotion in human speech. For example, in this research, the task is to classify aid types for user-posted messages during crises on social media. In formal terms, the crisis message is described as the \textit{example}, \textit{instance}, \textit{sample} or \textit{input} that is processed by the machine learning system to perform the task. To enable the machine to understand the example, it is represented as a vector $\textbf{x} \in \mathbb{R}^k$ of $k$ features, where each feature $x_i$ refers to the value for a particular attribute of the example. For instance, at its simplest, the features can be occurrence counts of words in the crisis message. This thesis concentrates on crisis message categorisation, which can be considered to be a text classification task. The following provides a high-level description of how a machine learning system is applied for text classification. Basically, in text classification, the machine program is asked to identify which of $m$ categories the example $\textbf{x}$ belongs to. The $m$ categories are known as the \textit{classes} or \textit{labels} comprising the \textit{label space} $Y$ that are defined as the \textit{output} of the machine program. Depending on $Y$, text classification exists in multiple forms. If the number of classes $m$ is $2$, it is called as a \textit{binary classification} task. If $m\textgreater2$, it is defined as a \textit{multi-class classification} task. In cases where multiple classes may be identified for the example, it is referred to as a\textit{ multi-label classification} task. To achieve the classification, the learning algorithm needs to produce a \textit{hypothesis} function that maps the input to the output, i.e., $y=h(\textbf{x})$ where $y\in Y$. One typical type of mapping function is to estimate the probability distribution over the classes given the input i.e., $h(\textbf{x})=p(y|\textbf{x})$. To categorise the example, the classes with high probabilities are usually assigned.  

\textbf{Performance evaluation in text classification}: When the machine learning system is carrying out on a task, it is important to measure how well the system performs so that its performance ($P$) can be compared to another system. To measure its performance, some evaluation metrics are required. In the context of text classification, the most straightforward metric to evaluate the performance is \textit{accuracy}. It is obtained by the number of true predictions divided by the number of all predictions by the system, which is formulated as follows:

\begin{equation}
    \text{Acc} = \frac{1}{n}\sum_{i}^{n} I(y_i=\hat{y_i})
\end{equation}

where $n$ stands for the number of all predictions and $I$ is the indicator function that is $1$ if the actual label $y_i$ is matched with the predicted label $\hat{y_i}$ and $0$ otherwise. In some cases, as an alternative to accuracy, the \textit{error rate} is used to measure the performance, which is calculated as the proportion of incorrect predictions out of all predictions. For multi-class classification, another metric \textit{F1 score} is usually used to measure the performance on a per class basis, which is the harmonic mean of \textit{precision} and \textit{recall}, defined as follows.

\begin{equation}
  \begin{aligned}
    \text{Precision} = \frac{TP}{TP+FP}, \,
    \text{Recall} = \frac{TP}{TP+FN} \\ \\
    \text{F1} = \frac{2\times\text{Precision}\times \text{Recall}}{\text{Precision}+\text{Recall}} = \frac{TP}{TP+\frac{1}{2}(TP+FN)}
    \end{aligned}
\end{equation}

where $TP$, $FP$ and $FN$ stands for the number of true positives, false positives and false negatives respectively. At a per-class level, an example is said to be a true positive when it is classified to the class that it actually belongs to, a false positive when it is classified to the class that it does not belong to and a false negative when it is not classified to the class that it actually belongs to. Hence, it is easy to know from the equation that precision measures the proportion of true positives out of all positive predictions and recall measures the proportion of true positives out of actual positive examples of the class. To consider both precision and recall, the F1 score is used as the harmonic mean of precision and recall for a more balanced summarisation of the system's performance. Since the aforementioned F1 is calculated on a per class basis, some averaging methods are normally used to get a global performance evaluation for all classes. There are three commonly-used averaging methods for this purpose: \textit{micro-F1}, \textit{macro-F1} and \textit{weighted-F1}. The micro-F1 metric computes a global average F1 score by counting the sums of the TPs, FNs, and FPs, formulated as follows:

\begin{equation}
    \text{micro-F1} = \frac{\sum_{i=0}^{m}TP_i}{\sum_{i=0}^{m}TP_i+\frac{1}{2}(\sum_{i=0}^{m}TP_i+\sum_{i=0}^{m}FN_i)}
\end{equation}

Where $m$ is the number of classes. The macro-F1 metric is straightforward, calculated by taking the mean of all the per-class F1 scores.

\begin{equation}
    \text{macro-F1} = \frac{1}{m}\sum_{i=0}^{m} \text{F1}_i
\end{equation}

As seen from the equation, the macro-F1 metric simply averages the per-class F1 scores without taking into account the number of examples of each class (also known as \textit{support}). The weighted-F1 is a variant of the macro-F1, and it is computed by taking the mean of all per-class F1 scores while considering each class’s support.

\begin{equation}
    \text{weighted-F1} = \sum_{i=0}^{m} r_i\text{F1}_i
\end{equation}

where $r_i$ refers to the ratio of examples of the $i$th class out of all examples. Regarding the choice of the averaging methods, it is dependent on the characteristics of the examples that are classified by the system. If the examples are imbalanced across all classes, the micro average is a straightforward choice for reporting the overall performance of the system regardless of the class. If the examples tend to be balanced across all classes and the classes are equally important, the macro average would be a good choice as it treats all classes equally. In cases where the examples are imbalanced and the classes with more examples need to be paid more importance, the weighted averaging is preferred.

As described above, for a machine learning system, it is able to perform a specific task such as assigning classes to textual input examples and its performance on the task can be measured by the evaluation metrics. To enable the system perform the task, it has to learn from the experience $E$.

\textbf{Learning from the experience}: In the learning process of a machine program, the experience normally refers to a dataset consisting of the examples or data points that the program can learn from. For examples, in the crisis messages categorisation task, the experience indicates the dataset containing crisis messages. Depending on whether the data is annotated or not, machine learning can be broadly divided into two categories: \textit{supervised learning} and \textit{unsupervised learning}~\cite{Goodfellow-et-al-2016}. The former says that every example in the dataset is assigned with one or more labels that are used to teach the program what to output given the example. It is termed ``supervised'' since the labels are annotated by humans (i.e., human supervision for teaching the program to learn from the examples). In contrast, in unsupervised learning, the learning algorithm only sees the examples without any labels being designated and the goal is to identify patterns within the dataset containing the data points on its own, such as clustering users into groups of similar interest in a recommendation system. It is termed ``unsupervised'' since there is no external guidance for the program to perform that task. Situated between supervised learning and unsupervised learning, \textit{semi-supervised learning} is also widely studied in the literature of machine learning where the learning algorithm is provided with a combination of labelled data and unlabelled data to perform a task. 

As discussed in Section~\ref{sec:research-approches}, this research proposes three scenarios to solve the low data problem for crisis messages categorisation. Although the focus of this research is on settings where there is a lack of available annotated data, the proposed methods are in essence based on supervised learning. For crisis domain adaptation, the algorithm is learnt on the labelled data of source events and then adapted to a new event. In crisis few-shot learning, the proposed methods apply augmentation techniques to obtain more labelled examples based on a small quantity of originally-provided labelled examples. Hence, the algorithm is learnt on the augmented data that is labelled. Although initially there is no labelled data for crisis zero-shot learning, this research proposes a method that uses label names with the unlabelled data to obtain a pseudo-labelled dataset on which the algorithm is learnt. Overall, the idea is to convert a semi-supervised or unsupervised task into a situation where supervised approaches can be used. As such, the remaining background review will be centered around supervised machine learning algorithms. Machine learning has developed with various algorithms proposed in the literature, from traditional machine learning algorithms such as linear regression or na\"ive bayes to neural networks algorithms such as pre-trained language models. The proposed methods in this research are mainly built upon the neural networks algorithms due to their strong performance in text classification. Hence, the focus of the discussion of neural network models is on their use within text classification. As the cornerstone of neural networks, the introduction to linear regression comes first as follows.






\subsection{Linear regression}

This section presents a linear model in the context of regression tasks, and then demonstrates how a similar approach can be applied to classification tasks. To better describe the process of applying linear regression for text classification, an entire dataset is represented by $\textbf{X} \in \mathbb{R}^{n \times k}$ containing $n$ examples (data points) of $k$ features where $\textbf{x}_i$ stands for the input features of the i-th example in the dataset. In supervised learning, the input $\textbf{x}_i$ is associated with a label $y_i \in Y$. For a machine learning algorithm in text classification, the objective is to build a model that predicts the label $y$ given the input $\textbf{x}$ through an estimation of conditional probability distribution using a set of \textit{parameters} or \textit{weights} $\theta$, denoted as $p(y|\textbf{x};\theta)$. The common way to achieve the estimation is known as \textit{maximum likelihood estimation} (MLE) where the optimal parameters $\hat{\theta}$ are obtained by maximising the likelihood of the data $p(y|\textbf{X};\theta)$

\begin{equation}
    \hat{\theta} = \underset{
    \theta}{\text{arg max }} p(y|\textbf{X};\theta)
\end{equation}

which can be transformed into:

\begin{equation}
    \hat{\theta} = \underset{
    \theta}{\text{arg max }} \sum_{i=1}^n \text{log} p(y|\textbf{x}_i;\theta)
    \label{eq:lr-transformation-function}
\end{equation}

For a linear regression model, the objective is to predict a numeric value $\hat{y} \in \mathbb{R}$ given the input features $\textbf{x} \in \mathbb{R}^k$ using a vector $\theta \in \mathbb{R}^k$ of parameters and a \textit{bias} $b \in \mathbb{R}$, formulated as follows.

\begin{equation}
    \hat{y} = \theta^\top\textbf{x} + b
\end{equation}

where $\hat{y}$ is the predicted value of $y$. To learn the parameters $\theta$ in regression tasks, the mean square error (MSE) function is commonly used to measure the difference between the model's prediction $\hat{y}$ and the true value $y$, formulated as follows~\cite{chai2014root}.

\begin{equation}
    \text{MSE} = \frac{1}{n}\sum_{i=1}^n(\hat{y_i}-y_i)^2
\end{equation}

The \textit{error} function is also known as the \textit{objective} or \textit{cost} function. That says, to find the optimal $\hat{\theta}$, the objective is to minimise the error or cost.

\begin{equation}
      \hat{\theta} = \underset{
    \theta}{\text{arg min }} \frac{1}{n}\sum_{i=1}^n(\theta^\top\textbf{x}_i + b-y_i)^2
    \label{eq:lr-sigmoid}
\end{equation}

\textbf{Logistic regression}: The aforementioned describes linear regression in the regression problem where the output is a numeric value. It can be easily adapted to classification tasks by converting the predicted numeric value to a value between $0$ and $1$ implying the conditional probability of $y$ given $\textbf{x}$. Linear regression becomes \textit{logistic regression} when it is applied for classification in this way. The conversion function is known as the \textit{sigmoid} or \textit{logistic} function denoted as $\sigma$, which is defined as follow.

\begin{equation}
    \sigma(x) = \frac{1}{1+e^{-x}}
    \label{eq:sigmoid-function}
\end{equation}

The sigmoid function takes an input $x$ and converts it into a value between $0$ and $1$. The converted value is close to $1$ when the input goes for positive infinity and $0$ when it approaches negative infinity. In binary classification, there are two classes: class $0$ and class $1$. Logistic regression models the conditional probability $\hat{p}(y|\textbf{x};\theta)$ as follows:

\begin{equation}
\begin{aligned}
     \hat{p}(y=1|\textbf{x};\theta) = \sigma(\theta^\top\textbf{x}) = \frac{1}{1+e^{-\theta^\top\textbf{x}}}  \\
     \hat{p}(y=0|\textbf{x};\theta) = 1-  \hat{p}(y=1|\textbf{x};\theta)
\end{aligned}
\label{eq:prob-binary-lr}
\end{equation}

which can be rewritten as:

\begin{equation}
 \hat{p}(y|\textbf{x};\theta)=(\hat{y})^{y}(1-\hat{y})^{1-y}
\end{equation}

By applying maximum likelihood estimation, the parameters $\theta$ can be estimated as follows:

\begin{equation}
\begin{aligned}
\hat{\theta} &=\underset{\theta}{\operatorname{argmax}} \prod_{i=1}^{n} \hat{p}\left(y_{i} \mid \mathbf{x}_{i} ; \theta\right) \\
&=\underset{\theta}{\operatorname{argmin}} \frac{1}{n} \sum_{i=1}^{n}-y \log (\hat{y})-\left(1-y\right) \log \left(1-\hat{y}_{i}\right)
\end{aligned}
\label{eq:cross-entropy-lr}
\end{equation}

In linear regression, MSE is used as the loss function to find the optimal $\hat{\theta}$. Similarly, the last part of Equation~\ref{eq:cross-entropy-lr} known as the \textit{binary cross entropy} (BCE) loss~\cite{janocha2016loss} is minimised to find the optimal $\hat{\theta}$. Once the optimal $\hat{\theta}$ are obtained, the classifier is finished building and can then be used to predict class $1$ or $0$ for an example via Equation~\ref{eq:prob-binary-lr}. For multi-label multi-class classification involving $m$ classes, it can be done by building $m$ such binary classifiers. Basically, for each classifier, a separate set of parameters $\theta_i$ for the i-th class are learnt through Equation~\ref{eq:prob-binary-lr} first. Then in predictions, the i-th binary classifier with the learnt $\theta_i$ applies Equation~\ref{eq:prob-binary-lr} to decide whether the i-th class is assigned to an example (assigned if being predicted to be class $1$ and not if class $0$). For single-label multi-class classification, instead of the sigmoid function in Equation~\ref{eq:prob-binary-lr}, the \textit{softmax} function is used to obtain a categorical distribution:

\begin{equation}
    \hat{p}\left(y_{i} \mid \mathbf{x} ; \theta\right)=\frac{e^{\theta_{i}^{\top} \mathbf{x}}}{\sum_{j=1}^{m} e^{\theta_{j}^{\top} \mathbf{x}}}
    \label{eq:lr-softmax}
\end{equation}

The softmax function normalises the model output for i-th class to be a value between $0$ and $1$, indicating the conditional probability of the i-th class given the example $\textbf{x}$. Given the example $\textbf{x}$ over all classes, the \textit{categorical cross-entropy} function (CCE)~\cite{janocha2016loss} is used as the loss function to calculate the difference between the empirical conditional probability $p(y|\textbf{x})$ and the model predicted probability $\hat{p}(y|\textbf{x})$, defined as follows:

\begin{equation}
    J(\theta)=-\sum_{i=1}^{m} p\left(y_{i} \mid \mathbf{x}\right) \log \hat{p}\left(y_{i} \mid \mathbf{x} ; \theta\right)
\end{equation}

The optimal $\theta$ can be found by minimising the loss/objective function on all the $n$ examples, formulated as follows:

\begin{equation}
    \hat{\theta}=\underset{\theta}{\operatorname{argmin}}\frac{1}{n} \sum_{i=1}^{n}\sum_{j=1}^{m} -p\left(y_{j} \mid \mathbf{x_i}\right) \log \hat{p}\left(y_{j} \mid \mathbf{x_i} ; \theta\right)
\end{equation}

In describing logistic regression for classification, one question remained unanswered is how to update the parameters $\theta$ so they can be the optimal. Denoting the objective function as $J(\theta)$, one common optimisation procedure is \textit{gradient descent} that updates the parameters $\theta$ in the opposite direction of the gradients $\nabla_{\theta} J(\theta)$ with a \textit{learning rate} $\alpha$ in\textit{ mini-batches}. For details on gradient descent and more optimisation procedures such as Adam~\cite{kingma2014adam} and Adagrad~\cite{duchi2011adaptive}, an overview of these procedures can be found in~\cite{ruder2016overview}. 

\textbf{Training and generalisation}: So far, the process of how to apply logistic regression for classification has been described. It is important to note that the aforementioned entire dataset $\textbf{X}$ refers to the \textit{observed data} (also known as \textit{training data}) for building the classifier. In machine learning, the process of learning the parameters $\theta$ for the classifier using the observed data is called as the \textit{training} stage. When applying the classifier to make predictions for the unobserved test data, it is called as the \textit{testing} or \textit{inference} stage. In practice for building a machine learning model, the dataset $\textbf{X}$ is usually split into two subsets: training set and development/validation set. They are drawn from $\textbf{X}$ and hence they are assumed to have the same distribution. The training set is used as the seen data to learn the model's parameters $\theta$, the development set is treated as the unseen data to develop the model in choosing hyper-parameters such as the learning rate or batch size and reflecting the model's generalisation capability. The model's error on the training set is called \textit{training error} and on the validation set is called \textit{test} or \textit{generalisation error}. For a machine learning model, the objective is to make the training error small and the gap between training and test error small. A high training error indicates that the model is \textit{underfitting} on the data or the model has a high \textit{bias} and a big gap between training and test error implies the model is \textit{overfitting} on the data or the model has a high \textit{variance} (poor generalisation capability). To prevent underfitting, common solutions include increasing the training data or increasing the capacity of the model (its ability to fit on various functions). To prevent overfitting, the solution is to decrease the capacity of the model. One common way is through \textit{regularisation}, such as $l_1$ and $l_2$ norm (check~\cite{Goodfellow-et-al-2016} for details).

\section{Neural networks}
\label{sec:nns}

\begin{figure}[H]
    \centering
    \includegraphics{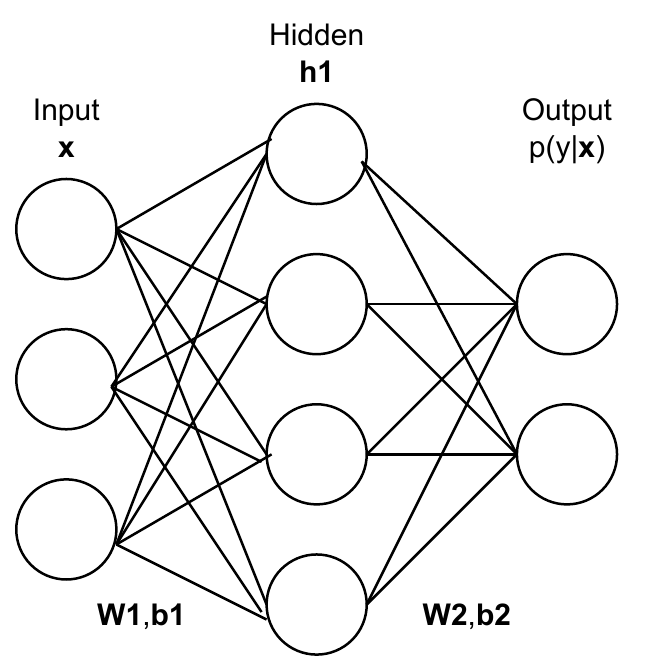}
    \caption{A basic neural network for classification}
    \label{fig:basic-nn}
\end{figure}

To describe neural networks, Figure~\ref{fig:basic-nn} presents an example of a basic neural network for classification. Recalling linear regression, for input $\textbf{x}$, the output $\hat{y}$ is estimated by a linear transformation function using the parameters $\theta$ and bias $b$ (Equation~\ref{eq:lr-transformation-function}). When it is adapted to logistic regression for classification, the conditional probability $p(y|\textbf{x})$ is estimated by applying the sigmoid function or the softmax function on the output of linear regression (Equation~\ref{eq:lr-sigmoid} and~\ref{eq:lr-softmax}). Logistic regression can be viewed as a simple neural network with one \textit{layer} where the sigmoid or softmax function is known as the \textit{activation function}. 

For a neural network, a layer refers to a function that maps an input to an output with learnable parameters $\textbf{W}$~\footnote{The $\textbf{W}$ notation is used here as an alternative for $\theta$ used in linear regression.} and bias $\textbf{b}$. To increase the capacity of a neural network model, there are usually multiple layers stacked together where the output of a previous layer is used as the input of the next layer. To interpret the term ``neural'', each layer consists of many units used to process the information from the previous layer and forward it to the next layer, which acts similarly to a biological neuron. As the number of layers increases to a big number, the network is said to be deep in its architecture and in this case neural networks are deep neural networks.

The number of layers is defined as the \textit{depth} of a neural network. For example, Figure~\ref{fig:basic-nn} presents a neural network of three layers. The first layer refers to the \textit{input layer}, the middle layer is the \textit{hidden layer}, and last layer is the \textit{output layer}. In the literature, a neural network of this form is known as the multilayer perceptron network (MLP). In mathematics, it can be described as follows:

\begin{equation}
\begin{aligned}
\textbf{h}_1 = g_1(\textbf{W}_1\textbf{x}+\textbf{b}_1)\\
p(y|\textbf{x}) = g_2(\textbf{W}_2\textbf{h}_1+\textbf{b}_2)
\end{aligned}
\label{eq:basic-nn}
\end{equation}

In this basic neural network, the input $\textbf{x}$ is fed forward to next layer to obtain a hidden representation $\textbf{h}_1$ that is then fed forward to next layer to obtain the output $p(y|\textbf{x})$, which is known as \textit{forward propagation}. In the equation, $g_1$ and $g_2$ refer to the activation function of the hidden layer and output layer respectively. The activation function is normally a non-linear function to apply a non-linear transformation on the output of that layer. Commonly-used activation functions of the hidden layer include \textit{Rectified Linear Unit} (ReLU) and the \textit{tanh} function, which are defined as follows.

\begin{equation}
\begin{aligned}
\text{ReLU}(x)= \text{max}(0,x)\\
\text{tanh}(x)= \frac{e^x-e^{-x}}{e^x+e^{-x}}
\end{aligned}
\label{eq:mlp}
\end{equation}

Regarding the activation function of the output layer for classification, it is usually a sigmoid function (binary or multi-label) or a softmax function (multi-class). In a neural network, the output of last layer before applying the activation function is known as the \textit{logits} output.

In Equation~\ref{eq:mlp}, $\textbf{W}s$ refer to the learnable parameters or weights and $\textbf{b}s$ are the bias vectors of the neural network model used to estimate the final conditional probability output $p(y|\textbf{x})$. In this example, $\textbf{x}\in \mathbb{R}^n$ where $n=3$, there are four nodes in the middle hidden layer and the output is binary, hence $\textbf{W}_1\in \mathbb{R}^{3\times4}$, $\textbf{b}_1\in \mathbb{R}^4$ and $\textbf{W}_2\in \mathbb{R}^{4\times2}$, $\textbf{b}_2\in \mathbb{R}^2$. Similar to linear regression, to find the optimal weights $\textbf{W}s$ and $\textbf{b}s$, an objective function is first defined and then the weights are updated through gradient descent on the objective function. In regression, MSE is usually used as the objective function. In classification, binary cross entropy is used as the objective function (binary or multi-label) and categorical cross entropy is used as the objective function (multi-class). Like linear regression, the optimisation is done by a process of gradient descent, which takes derivatives of the objective function with respect to the weights for changing the weights. This is usually referred to as \textit{back-propagation} in the literature~\cite{ruder2016overview}. 

\begin{figure}[!h]
    \centering
    \includegraphics[width=\linewidth]{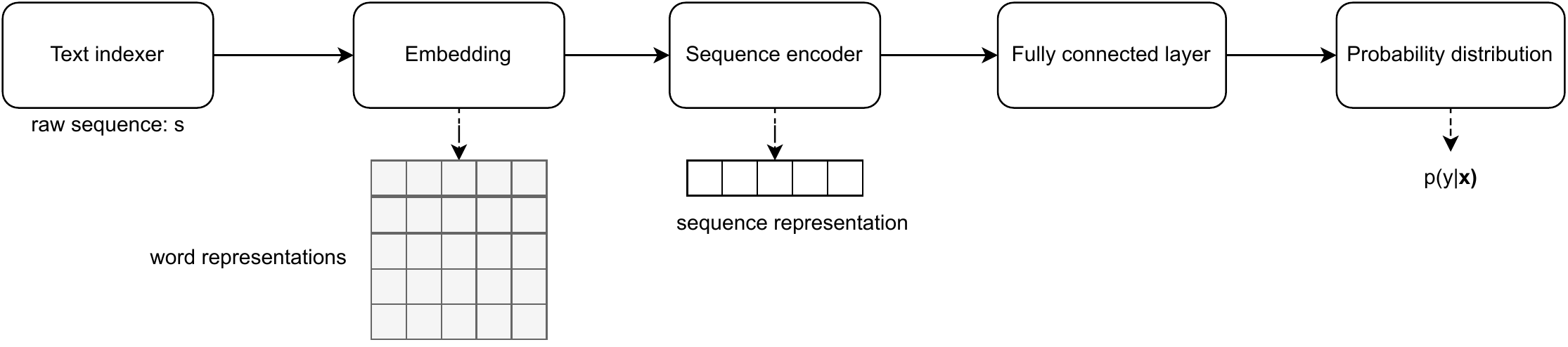}
    \caption{The general architecture of neural network models for text classification}
    \label{fig:arch-deep-for-tc}
\end{figure}

\textbf{Neural networks in text classification}: When applying neural networks for text classification, in order to enable the model to understand texts, a piece of text has to be represented numerically. To achieve this, each word in the input text is represented by a vector. The vector representations for each word are known as \textit{word embeddings}, which are used as the initial features of the input text. Word embeddings can be classified into two categories: sparse and dense. A one-hot embedding, which represents a word in a vocabulary with a vector that has a 1 at the position corresponding to the word and 0s elsewhere, is the most basic example of a sparse embedding. On the other hand, dense word embeddings are typically pre-trained and represent a word with a fixed-length vector of continuous real numbers. Once the individual words of the input text are represented by word embeddings, the representation of the whole text is then learnt based on the initial word features through the neural network models. Figure~\ref{fig:arch-deep-for-tc} demonstrates the general architecture of neural network models for text classification where a textual input example is viewed as a sequence of tokens $s: \{t_1,t_2,...,t_T\}$ (the tokens usually indicate words or word pieces). To begin, a vocabulary is constructed that includes all the tokens in the training examples, where each token in the vocabulary is associated with an id. The  text indexer maps the tokens to their corresponding ids before they are passed to the embedding module. The embedding module then converts the tokens into vectors based on their corresponding ids, denoted by $\textbf{x}\in\mathbb{R}^{T\times k}$ where $k$ implies the embedding dimension. In the case of spare one-hot embedding, $\textbf{x}\in\mathbb{R}^{k}$ where the feature dimension $k$ is the vocabulary size. However, in neural network models, dense pre-trained word embeddings are commonly used, introduced later in section~\ref{subsubsec:pre-trained word-embedding}.

After the tokens are embedded, the word representations are fed to the next module which here is called the sequence encoder. Given the classification objective, the sequence encoder aims to learn a good deep representation for the sequence or loosely speaking a good summary of the individual word representations. After being encoded, the sequence representation is forwarded to a fully connected layer (i.e., a linear transformation layer) to output the logits across all classes. This is then converted by the activation function to a probability distribution to capture the probability of the input $s$ represented by $\textbf{x}$ belonging to class $y$. In a neural network, the layers can be customised and connected in different ways to learn representations of data for a target problem. This brings many variations of deep models for sequence representation learning. The variations primarily include convolutional neural networks (CNN) and long short-term memory based recurrent neural network (LSTM-RNN) and pre-trained language models. The details of them are covered in Section~\ref{subsec:cnn}~\ref{subsec:rnn} and~\ref{subsec:plms}.

\subsection{Pre-trained word embeddings}
\label{subsubsec:pre-trained word-embedding}

\begin{figure}[!h]
        \centering
        \includegraphics[width=0.7\linewidth]{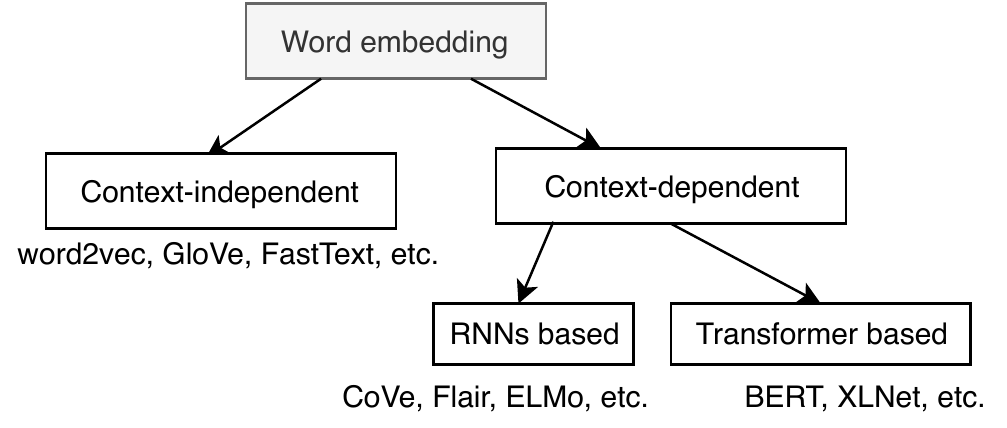}
        \caption{A taxonomy of pre-trained word embeddings}
        \label{fig:tax_of_word_embedding}
    \end{figure}

Pre-trained word embedding is a way to represent a word with fixed-length vectors of continuous real numbers~\cite{mikolov2013efficient,mikolov2013distributed,bojanowski2017enriching,pennington2014glove}. It maps a word in a vocabulary to a latent vector space where words with similar contexts are in proximity. Through word embedding, a word is converted to a vector that summarises both the word's syntactic and semantic information. Word embeddings are used as input feature representations for neural network models in text classification, which corresponds to the embedding module in Figure~\ref{fig:arch-deep-for-tc}. Figure~\ref{fig:tax_of_word_embedding} presents a taxonomy of pre-trained word embeddings. Broadly speaking, there are two main types of word embeddings, namely context-independent and context-dependent embeddings. The typical examples of the former are word2vec~\cite{mikolov2013distributed}, GloVe~\cite{pennington2014glove} and FasText~\cite{bojanowski2017enriching}. These are known as classic word embeddings, which learn representations through language model (LM) based shallow neural networks or co-occurrence matrix factorisation~\cite{zhang2018deep}. The learned representations are characterised by being distinct for each word without considering the word's context. Hence, they are usually pre-trained and stored in the form of downloadable files which can be directly applied to text classification tasks~\footnote{For example, GloVes can be downloaded from: \url{https://nlp.stanford.edu/projects/glove/}}.

In comparison to context-independent word embeddings, context-dependent methods learn different embeddings for the same word with different contextual use. For example, for the homonym ``bank'', its embedding changes depending on whether it is used in a river-related context or finance-related one. This feature has made this type of embeddings to become the mainstream. Examples of context-dependent include ELMo~\cite{peters2018deep},  Flair~\cite{akbik2018contextual},  BERT~\cite{devlin2018bert}, etc. Below gives a short introduction to some specific word embedding approaches.


\textbf{Classic embeddings}: This type of embeddings are pre-trained over very large corpora and shown to capture latent syntactic and semantic features.
\begin{itemize}
    \item \textbf{word2vec} \cite{mikolov2013distributed}.  This applies either of two model architectures to  produce  word  vectors,  namely  continuous  bag-of-words  (CBOW) and  skip-gram (SG). Both methods are trained based on a neural prediction-based model. CBOW trains a model that aims to predict a word given its context, while SG does the inverse, namely to predict the context given its central word.
    
    \item \textbf{GloVe} \cite{pennington2014glove} learns efficient word representations by performing training on aggregated global word-word co-occurrence statistics from a corpus. It is known for learning good general language features by capturing words' co-occurrences globally across corpora.
    
     \item \textbf{FastText} \cite{bojanowski2017enriching} learns word representations through a neural LM. Unlike GloVe, it embeds words by treating each word as being composed of character n-grams instead of a word whole. This allows it to not only learn rare words but also out-of-vocabulary words.
\end{itemize}


\textbf{Contextualised embeddings}: Unlike classic embeddings, this type of embeddings
are known for capturing word semantics in context.

\begin{itemize}
    \item \textbf{ELMo} \cite{peters2018deep} learns contextualised word representations based on a neural LM with a character-based encoding layer and two BiLSTM layers (these are discussed later in Section~\ref{subsec:rnn}). The character-based layer encodes a sequence of characters of a word into the word's representation for the subsequent two BiLSTM layers that leverage hidden states to generate the word's final embedding.
    
    \textbf{Flair} \cite{akbik2018contextual}. This trains a model for producing contextualised word embeddings using neural character LM (1-layer BiLSTM), leading to lower computational resources and stronger character-level features. Although its effectiveness has been recognised in sequence labelling tasks, its performance in text classification remains unexplored.
    
    \textbf{BERT} \cite{devlin2018bert} is a recently proposed Transformer-based~\cite{vaswani2017attention} language representation model trained on a large cross-domain corpus. Unlike ELMO, which learns representations through bidirectional LMs (i.e., simply the combination of left-to-right and right-to-left representations), BERT applies a Masked LM to predict words that are randomly masked. Instead of the recurrent learning of a sequence, BERT uses the Multi-Head Cross-Attention mechanism~\cite{vaswani2017attention,bahdanau2015neural} to learn global dependencies between the words of a sequence, leading to it being more parallelisable and exhibiting superior performance (these are discussed later in Section~\ref{subsec:plms}). Through a process of fine-tuning,  BERT has been demonstrated to achieve state-of-the-art results for a range of NLP tasks. Other similar embeddings optimised for BERT include DistilBERT~\cite{sanh2019distilbert}, RoBERTa~\cite{liu2019roberta} and ALBERT~\cite{kobayashi2018contextual}.
\end{itemize}

Among so many choices of word embeddings, it is important to know how to select them for text classification. As an preliminary step before undertaking the work that constitutes the main contribution of the thesis, an initial study was conducted to investigate the factors influencing the choice of word embeddings for classification tasks~\cite{Wang2020a}. In this study, the aforementioned word embeddings are used with a CNN and a BiLSTM sequence encoder (see Section~\ref{subsec:cnn} and~\ref{subsec:rnn}) for four benchmarking text classification tasks. To summarise the major findings of this work, the results show that CNN as the sequence encoder outperforms BiLSTM in most situations, especially for document context-insensitive datasets. This study recommends choosing CNN over BiLSTM for document classification datasets where the context in sequence is not as indicative of class membership as sentence datasets. For word embeddings, concatenation of multiple classic embeddings or increasing their size does not lead to a statistically significant difference in performance despite a slight improvement in some cases. For context-based embeddings, the results show that BERT overall outperforms ELMo, especially for datasets consisting of long documents. Compared with classic embeddings, both achieve an improved performance for datasets with short texts while the improvement is not observed for longer texts. 

Assuming that a word embedding is selected and used to represent the input text at this stage, the next step of the process is to learn a sequence representation for text classification (see Figure~\ref{fig:arch-deep-for-tc}). The following sections introduce various neural network models that can be used for this purpose.

\subsection{Convolutional neural network}
\label{subsec:cnn}

\begin{figure}[!h]
    \centering
    \includegraphics[width=\linewidth]{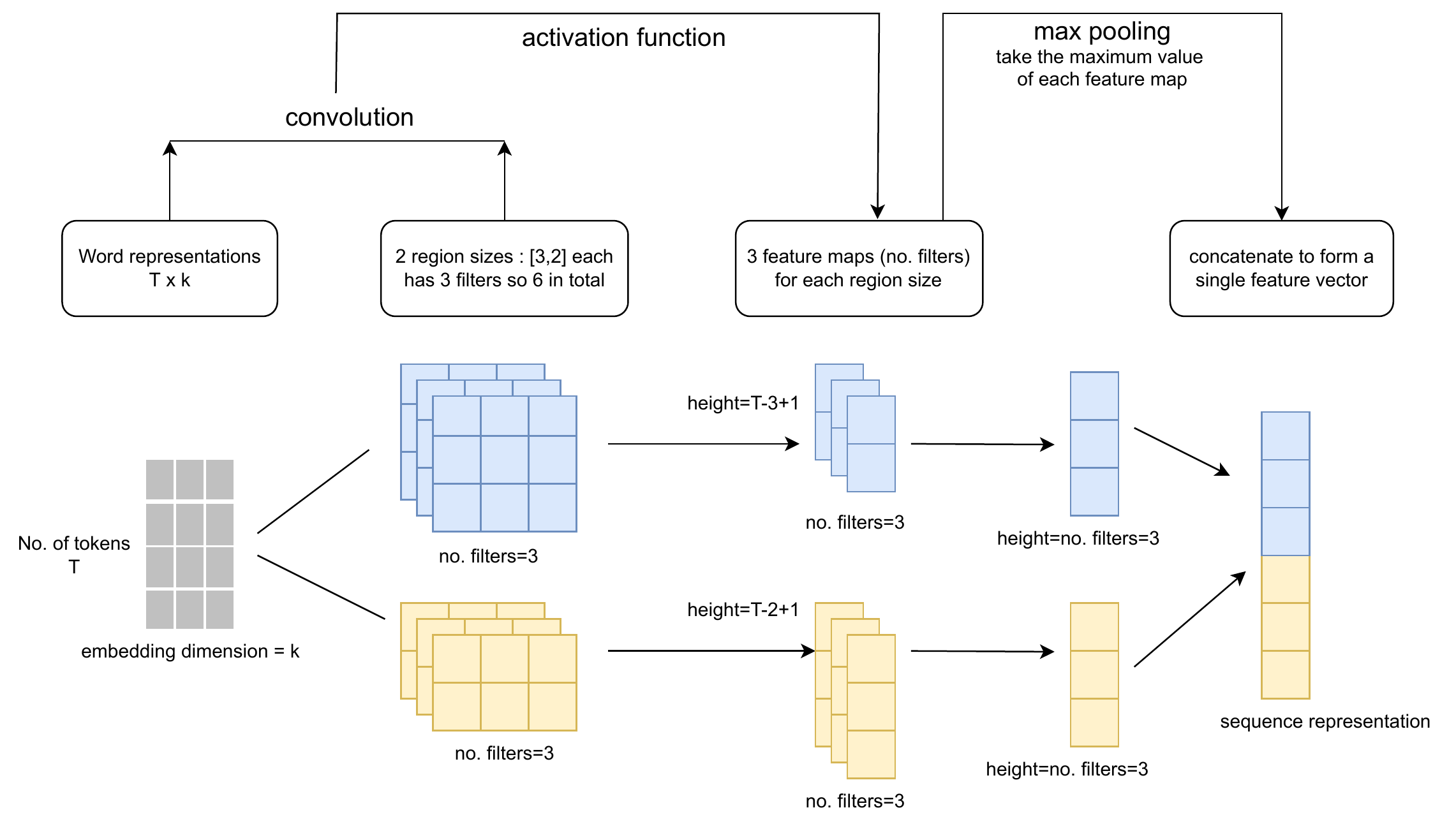}
    \caption{CNN as an example of sequence representation learning}
    \label{fig:cnn-seq-flow}
\end{figure}

Convolutional neural network (CNN)~\cite{zhang2018deep,zhang2017sensitivity} is a variant of feed-forward neural network originally employed in the field of computer vision. It consists of multiple convolutional layers, each of which acts as a local filter to extract features from the input by sliding (or convolving) it, like the cells in visual cortex of the human brain receiving local features (e.g. light or contour, known as receptive fields) of an input image. Figure~\ref{fig:cnn-seq-flow} shows the workflow of CNN as an example of sequence representation learning. In the first stage, the input matrix (i.e. the word representations/embeddings) is converted to feature/activation maps through the convolution layer. The convolution process consists of element-wise multiplications between the input matrix and the defined local filters which are analogous to windows. Mathematically, the local filters are simply arrays of weights, and their number and region size can be customised. In this example, there are two sizes of such filters: $2$ and $3$, each of which has $3$ filters~\footnote{Here the number of filters is set to be $3$ for demonstration simplicity and it is much more than this in practice.}. For the $3$ filters with region size $3$, each slides from the top to bottom of the input matrix (element-wise and sum if the sliding stride is 1) and output 3 feature maps (or representation) consisting of vectors with each size being $T-3+1$ (here $T$ stands for the number of input tokens). This is the same for the $3$ filters with region size $2$, outputting $3$ feature maps consisting of vectors with each size being $T-2+1$. 

Following the convolutional layer, a pooling (or sub-sampling) layer is applied to reduce the spatial size of the representation, while retaining the most salient information of the previous representation. For instance, max pooling in Figure~\ref{fig:cnn-seq-flow} refers to the sub-sampling operation that takes only the maximum number out of the vectors of 3 filters of each size and thus outputs a single vector with height being $3$ (i.e. the number of filters). In practice, the convolution and pooling processes are usually conducted multiple times to learn deeper representations. In this simple example, the sequence representation is eventually generated by concatenating the two vectors from the pooling layer. CNN is a widely-used model as a feature extractor for learning representations in text classification~\cite{conneau2016very,dos2014deep,kim-2014-convolutional,Zhang2015CharacterlevelCN} because of its focus on finding local clues within textual data. Taking a document as an example, there are often phrases or n-grams that are in different places but are very informative as to which topic the document relates to. CNN is good at finding such local indicators, irrespective of the positions.

\subsection{Recurrent neural network}

\label{subsec:rnn}

Recurrent neural networks (RNNs) are a class of neural networks that process data in a sequential way~\cite{sherstinsky2020fundamentals}. Its sequential attribute makes it well suited to NLP tasks such as text generation and text classification. In basic erms, given an input text sequence represented by $T$ word vectors, an unit of RNNs takes each word vector at a time step and process it to next unit. A time step refers to a specific point in the processing of the sequence, at which the RNN processes a single word vector. Hence, each unit not only takes the input of current time step but the output from previous time step. Depending on the number of tokens in the input sequence, the number of unit in RNNs is dynamic. In RNNs, the parameters and activation function are usually shared across all time steps, which is known as \textit{typing} or \textit{sharing} of parameters in neural networks. For a RNN, depending on the task, it usually has different output format. For a text generation task, a RNN model is used to train a language model where the model predicts the next word given the previous sequence of words. Hence the output in this case is a sequence of vectors where each vector indicates the word prediction at that time step. For text classification, RNNs simply output a sequence representation (a vector) summarising the whole sequence, which is then used to get the probability distribution across all classes.
\begin{figure}[!h]
    \centering
    \includegraphics[width=\linewidth]{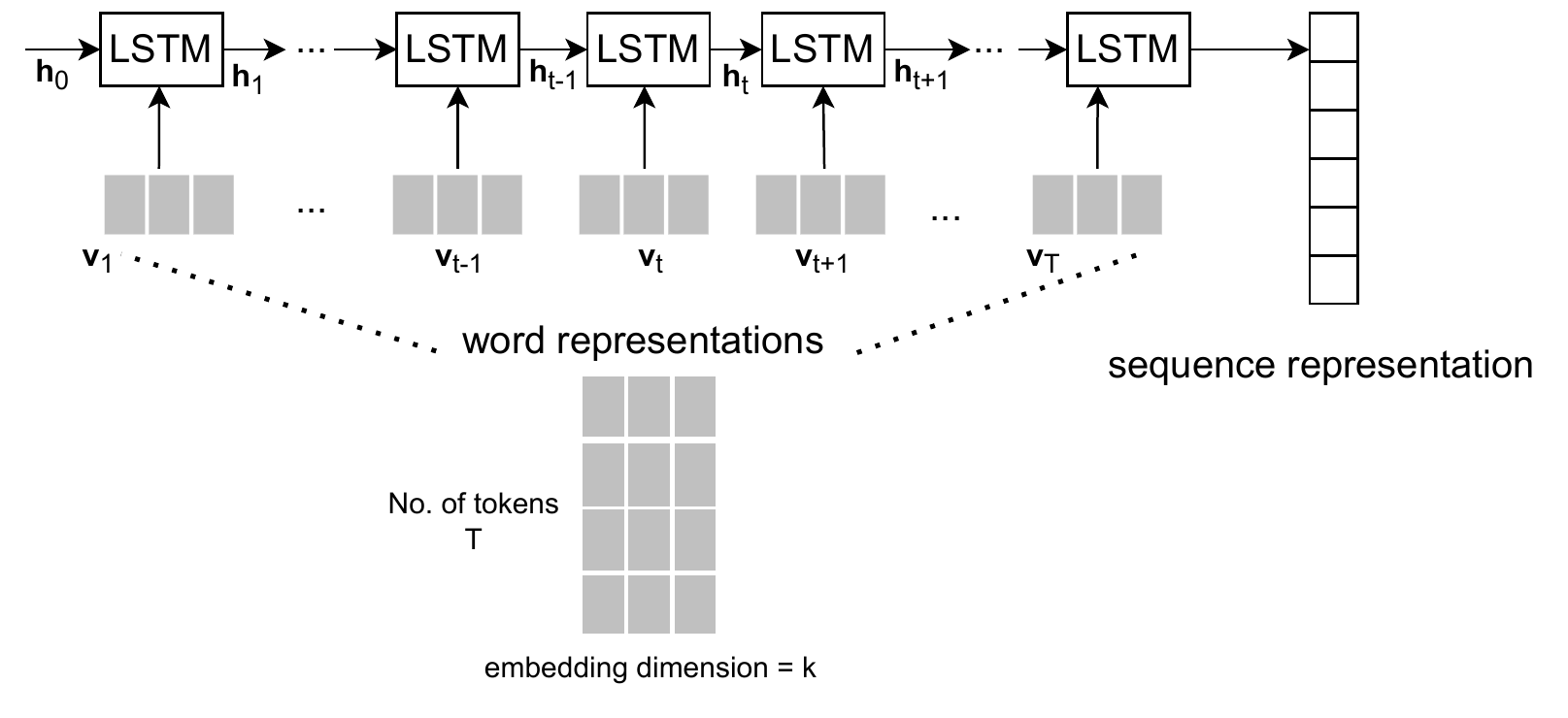}
    \caption{LSTM as memory cell of RNN for sequence representation learning}
    \label{fig:lstm-seq-flow}
\end{figure}

Regarding the unit of RNNs, long-short term memory (LSTM)~\cite{hochreiter1997long,Zhou2016TextCI} is a memory-based unit in a RNN. Although there are many other variations of RNN-based memory units, such as GRU~\cite{cho2014learning}, only LSTM is detailed below due to its popularity and ability to maintain information for long sequences~\footnote{Check~\cite{hochreiter1997long} for the gradient vanishing and exploding problem}. Figure~\ref{fig:lstm-seq-flow} shows the uni-directional (left-to-right) workflow of LSTM as memory cell of RNN for sequence representation learning. In a general sense, it encodes the input matrix to a sequence representation. At each time step $t$ in a token vector $\textbf{v}_t$, it summarises important information in the sequence as far as a hidden state $\textbf{h}_t$, using the previous hidden state $\textbf{h}_{t-1}$ as input. It does so by remembering or forgetting information through a set of transition functions (or so-called ``gates'') in the unit. Ultimately it summarises the sequence by outputting a single vector at the last time step. The LSTM transition functions are defined as follows~\cite{Zhou2016TextCI}:

\begin{equation}
\centering
\label{eq:lstm}
     \begin{array}{c@{\qquad}c}
         {\textbf{e}_t} = \sigma ({\textbf{W}_i} \cdot [{\textbf{h}_{t - 1}};{\textbf{v}_t}] + {\textbf{b}_i}) \\
    {\textbf{f}_t} = \sigma ({\textbf{W}_f} \cdot [{\textbf{h}_{t - 1}};{\textbf{v}_t}] +{\textbf{b}_f})\\
    {\textbf{q}_t} = \tanh ({\textbf{W}_q} \cdot [{\textbf{h}_{t - 1}};{\textbf{v}_t}] +{\textbf{b}_q})\\
    {\textbf{o}_t} = \sigma ({\textbf{W}_o} \cdot [{\textbf{h}_{t - 1}};{\textbf{v}_t}] +{\textbf{b}_o})\\
    {\textbf{c}_t} = {\textbf{f}_t} \odot {\textbf{c}_{t - 1}} + {\textbf{e}_t} \odot {\textbf{q}_t}\\
    {\textbf{h}_t} = {\textbf{o}_t} \odot \tanh \left( {{\textbf{c}_t}} \right)
     \end{array}
\end{equation}

where $\sigma$ and $tanh$ stand for sigmoid and hyperbolic tangent activation functions respectively. Element-wise multiplication is denoted by $\odot$ and $;$ refers to the concatenation of vectors. The parameters $\textbf{W}s$ and bias $\textbf{b}s$ are shared across all time steps and normally are randomly initialised before training (the initial hidden state $\textbf{h}_0$ is usually initialised with zeros). To better understand how the transition works, $\textbf{f}_t$ can be described as the forget gate that controls what information from the old memory cell should be forgotten. The input gate $\textbf{e}_t$ is used to control what new information is to be stored in the current memory cell and $\textbf{o}_t$ is described as an output gate to control what to output based on the memory cell $\textbf{c}_t$ (the initial memory state $\textbf{c}_0$ is usually initialised with zeros). In LSTM-based RNNs, the above process happens recurrently from time 0 to $T$. 

Unlike a LSTM, which summarises information in one direction from left to right, bidirectional LSTM (BiLSTM) sees the context of a token at time step $t$ in both left-to-right and right-to-left directions. In BiLSTM, the right-to-left process applies the same the transition functions as in the left-to-right process. In practice, there could be multiple layers of LSTM units stacked upon each other to learn deeper representations of the input data. In addition, depending on the problem domain, there exists different ways of generating the final sequence representation, such as concatenation of the mean of the outputs at each time step (bag-of-means)~\cite{Joulin2016BagOT} or only the output at the last time step as in the case of Figure~\ref{fig:lstm-seq-flow}.

Given the extensive studies in recent years on deep models for text classification, CNN and LSTM are just two examples of many in the literature. For example, \cite{Zhou2015ACN} combines CNN with LSTM for text classification; \cite{Yang2016HierarchicalAN} applies hierarchical attention networks;~\cite{Zhou2016TextCI}~improves text classification by integrating bidirectional LSTM with two-dimensional max pooling; and ~\cite{peters2018deep} applies a bi-attentive classification network (BCN) with pre-trained contextualised ELMo embedding, achieving strong performance in text classification.


\subsection{Pre-trained language models}

\label{subsec:plms}

More recently, transfer learning has gained great success in language understanding, which pre-trains on large textual corpora to gain a pool of general knowledge through unsupervised learning and then fine-tunes on the given training dataset (knowledge transfer) through supervised learning for a specific language task such as text classification~\cite{ruder2019neural}. The pre-training usually refers to the language modelling step and the fine-tuning refers to the downstream task training step. The major efforts into transfer learning include ULMFiT~\cite{howard2018universal}, ELMo \cite{peters2018deep} and transformer-based pre-trained language models (PLMs)~\cite{vaswani2017attention,devlin2018bert,radford2019language,raffel2020exploring}. The former two are pre-trained language models based on recurrent networks. One drawback of such language models is that the sequential nature of recurrent networks is not good for parallelisation within training sequences. This is critical when the length of any sequence is very long where batching across the training sequences is limited by memory constraints~\cite{vaswani2017attention}. As an important step towards tacking this issue, Ashish et al. 2017~\cite{vaswani2017attention} proposed the Transformer architecture that dispenses with recurrence and convolutions entirely and instead applies an attention mechanism for sequence representation learning. The Transformer mainly consists of an encoder and a decoder, which was originally proposed for machine translation. It is noted that in the context of machine translation, the input and output sequences are in different natural languages but share a common semantic representation. The encoder in the Transformer is used to learn the representation of a source sequence, while the decoder generates a target sequence based on the source representations from the encoder. Since the onset of Transformer, much work combining it with the idea of transfer learning has been introduced in recent years, achieving great success in various language processing tasks such as text classification and text generation. This line of work can be divided into three categories: encoder-based, decoder-based, encoder-decoder based. 




\subsubsection{Encoder models}
\label{subsubsec:encoder}

\begin{figure}[!h]
    \centering
    \includegraphics[width=\linewidth]{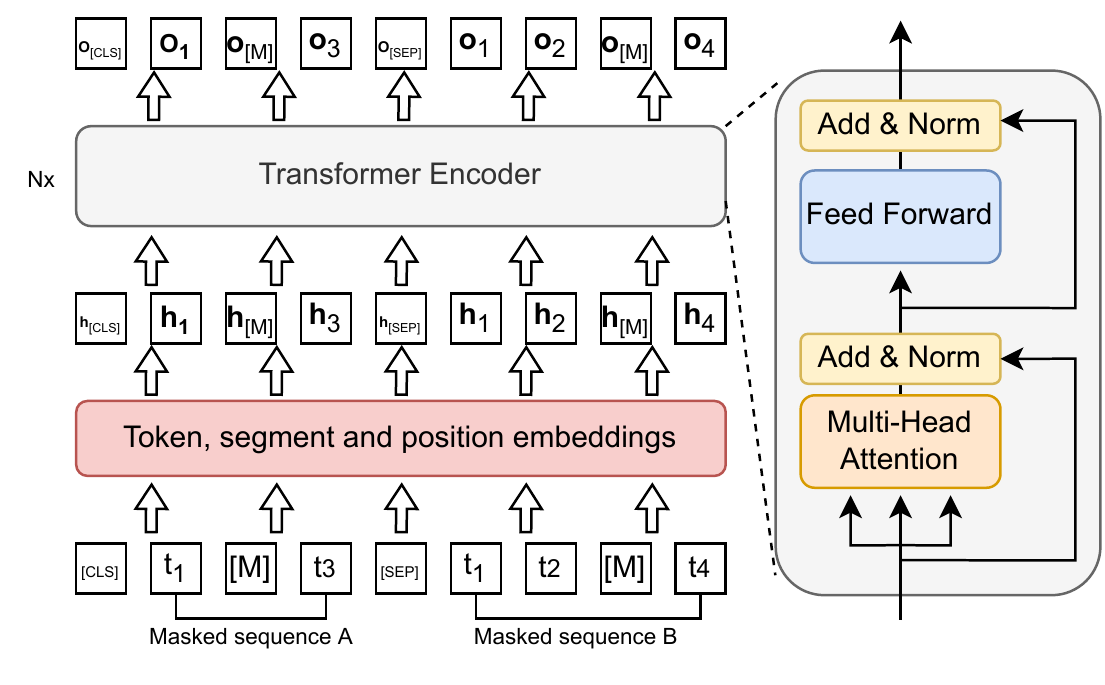}
    \caption{The workflow of transformer-encoder based BERT at pre-training}
    \label{fig:transformer-encoder-bert}
\end{figure}

The representative work in this category is BERT~\cite{BERT2018}. Depending on the transformer encoder, it is pre-trained using big unlabelled text data in an unsupervised (or usually referred to as self-training) manner on two tasks: masked language modeling (MLM) and next sentence prediction (NSP). To better describe BERT's pre-training, Figure~\ref{fig:transformer-encoder-bert} depicts the workflow of BERT at the pre-training stage. 

To achieve the MLM and NSP tasks, a pair of sequences A and B are used as the inputs and the pair are masked. The masking is conducted on the tokens of the input pair where a small portion of randomly-chosen tokens are masked (denoted as the symbol $[\text{M}]$ in the figure) while the remainder are unchanged (denoted as $t_1,t_2,t_3,t_4$). In addition to the masked tokens, two special tokens are used in BERT: $[\text{CLS}]$ and $[\text{SEP}]$. The former is the special classification token whose final output state is used as the aggregate sequence representation for classification tasks (the NSP task at pre-training and downstream classification tasks at fine-tuning). The latter is the separation token used to distinguish sequence A from B. In BERT, the input pair are tokenised by WordPiece~\cite{wu2016google}, which takes a word part instead of a whole word as a token. Once the input tokens are prepared, the next step is to represent them with vectors via the embedding layer. In BERT, each token $t_i$ is represented by the element-wise addition of three types of embeddings: token, segment and position embeddings.

\begin{equation}
    \textbf{h}_i = \text{Embed}_{\text{tok}}(t_i)+\text{Embed}_{\text{seg}}(t_i)+\text{Embed}_{\text{pos}}(t_i)
\end{equation}

where $\textbf{h}_i \in \mathbb{R}^{d_{\text{model}}}$ ($d_{\text{model}}$ is the embedding dimension or the model's hidden state size). The token embedding is used to represent the unique language information of each token, i.e., its content. The segment embedding consists of two vectors where the first one indicates that $t_i$ is from segment A and the second implies that it is from segment B. Hence, the purpose of segment embedding is again to enable the model to distinguish the two input sequences. The position embedding adds the positional information of each token to its representation so that the model is aware of the positions of each token. 

Having represented the input tokens with vectors, let $\textbf{H}$ be the hidden state that is a matrix representing all tokens. As introduced earlier, BERT is based on the transformer encoder relying on an attention mechanism to learn deep representations for the input sequences. The transformer encoder is a building block of BERT, which can be viewed as a layer of a neural network with two sub-layers. The first is known as the scaled dot-product cross-attention sub-layer and the second refers to a simple linear feed-forward sub-layer. In each sub-layer, there is a residual connection between the input and output, along with a normalisation on the output. The attention sub-layer plays a crucial role in allowing the representations of individual tokens to depend on one another. Before being fed to the attention sub-layer, the representation of each token is treated independently. The scaled dot-product attention mechanism is then used to allow a token to ``communicate" with other tokens, namely enabling the dependency between the token representations. The attention is said to be cross or bi-directional since a token is not only ``communicated'' with the tokens on the left but also the tokens on the right. Mathematically, it is achieved by introducing three matrices: the keys $\textbf{K}$, the queries $\textbf{Q}$ and the values $\textbf{V}$. To obtain them, they are simply linearly transformed from the hidden representations $\textbf{H}$, calculated as follows:


\begin{equation}
    \begin{aligned}
        \textbf{Q} = \textbf{H}\textbf{W}^Q \\
        \textbf{K} = \textbf{H}\textbf{W}^K \\ 
        \textbf{V} = \textbf{H}\textbf{W}^V
    \end{aligned}
\end{equation}

where $\textbf{H} \in \mathbb{R}^{L\times d_{\text{model}}}$ and $L$ refers to the number of input tokens or the sequence length. The weights $\textbf{W}^Q, \textbf{W}^K, \textbf{W}^V \in \mathbb{R}^{d_{\text{model}} \times d_{\text{model}}}$ are learnable parameters of the model. Using the keys, queries and values, the output of the attention function can be viewed as a weighted sum of the values. The weights are calculated by a softmax function on the scaled dot-product of the queries with the keys, presented as follows:



\begin{equation}
\label{eq:dot-scaled-product-attention}
\operatorname{Attention}(\textbf{Q}, \textbf{K}, \textbf{V})=\operatorname{softmax}\left(\frac{\textbf{Q} \textbf{K}^T}{\sqrt{d_k}}\right) \textbf{V}
\end{equation}

As the equation shows, the dot-product is scaled by being divided by the root square of the keys' dimension size $d_k$ (it is equal to $d_{\text{model}}$ in this case). To know the importance of other tokens to a token, the token attends to information of other tokens, which is quantified by the attention weights. The attention weights can be viewed as the similarity scores between the token's key and the queries of other tokens, computed by the scaled dot-product. As a result, the tokens with higher similarities pay attention to the token with greater importance (higher weight). In reality, instead of applying such a single attention function on the $d_{\text{model}}$-dimensional $\textbf{Q}$, $\textbf{K}$ and $\textbf{V}$, the multi-head attention approach is used to capture multiple aspects of a token's attention to other tokens, calculated as follows: 

\begin{equation}
    \begin{aligned}
\textbf{H}_{\text{att}} = \operatorname{MultiHead}(\textbf{Q}, \textbf{K}, \textbf{V})=\text {Concat}\left(\operatorname{head}_1, \ldots, \text {head}_{\mathrm{h}}\right) \textbf{W}^O \\
\text {where head}_{\mathrm{i}}=\operatorname{Attention}\left(\textbf{Q}\textbf{W}_i^Q, \textbf{K}\textbf{W}_i^K, \textbf{V}\textbf{W}_i^V\right)
    \end{aligned}
\end{equation}

where $\textbf{W}_i^Q \in \mathbb{R}^{d_{\text {model }} \times d_k}, \textbf{W}_i^K \in \mathbb{R}^{d_{\text {model }} \times d_k}, \textbf{W}_i^V \in \mathbb{R}^{d_{\text {model }} \times d_v}$ and $\textbf{W}^O \in \mathbb{R}^{h d_v \times d_{\text {model }}}$ are the learnable parameters of the model. In this case, $d_k=d_v=d_{\text {model }} / h$ where $h$ indicates the number of heads and $d_v$ is the values' dimension size. The multi-head attention concatenates individual attentions computed by the attention function on the projected keys, queries and values, which allows a token to jointly attend to information of other tokens in multiple aspects. Let the output of multi-head attention be denoted as $\textbf{H}_{\text{att}}$. It is added to the previous hidden representations $\textbf{H}$ known as the residual connection, followed by layer normalisation~\cite{ba2016layer}:
\begin{equation}
    \textbf{H}= \text {LayerNorm}(\textbf{H}+\textbf{H}_{\text{att}})
\end{equation}

Now the hidden state $\textbf{H}$ is re-assigned by the normalised addition of the previous hidden state and the attention output. Then this is fed to the position-wise feed-forward sub-layer that consists of two linear transformation with a ReLU activation in between. The first projects the hidden state to an intermediate dimension and the second projects it back to the original dimension $d_{\text{model}}$, presented as follows:


\begin{equation}
\textbf{H}_{\text{ff}}=\operatorname{FFN}(\textbf{H})=\text{ReLU} \left(\textbf{H} \textbf{W}_1+\textbf{b}_1\right) \textbf{W}_2+\textbf{b}_2
\end{equation}

where $\textbf{W}_1 \in \mathbb{R}^{d_\text{model}\times d_{\text{ff}}}, \textbf{W}_2 \in \mathbb{R}^{d_{\text{ff}} \times d_\text{model}}$ and $d_{\text{ff}}$ stands for the intermediate dimension that is usually greater than $d_{\text{model}}$. Similar to the attention sub-layer, the residual connection along with a layer normalisation is then conducted on the hidden state and the transformation output.
\begin{equation}
    \textbf{H}= \text {LayerNorm}(\textbf{H}+\textbf{H}_{\text{ff}})
\end{equation}

So far the transformer encoder of BERT has been described and it can be viewed as a layer of a neural network. In BERT, there are usually multiple such layers stacked together for the purpose of learning deep representations. In other words, the output of the first layer $\textbf{H}$ is used as the input for the next layer whose components remain the same as described previously: the attention sub-layer and transformation sub-layer. This repeats $N$ times where $N$ is the number of hidden layers\footnote{Depending on the versions of BERT, $N$ varies. The base version of BERT contains 12 such layers and the large version has 24 layers.}. For the output of the last layer, there are deep hidden representations for each input token, denoted as $\textbf{o}_i$ (see Figure~\ref{fig:transformer-encoder-bert}). There are two types of tokens whose output states require special attention. The output state of $[\text{CLS}]$ can be treated as the aggregate representation of the input sequences and hence it is used for the NSP task. During the pre-training stage, the task of Next Sentence Prediction (NSP) involves classifying whether a given sequence B follows a given sequence A in the unlabeled text corpus. If sequence B is the next sentence in the corpus after sequence A, it is considered a positive example for the NSP task. If sequence B is randomly sampled from the corpus and does not immediately follow sequence A, it is considered a negative example for the NSP task. The output states of the masked tokens $[\text{M}]$s go through a softmax layer for computing the loss of the MLM task where the true labels are the original tokens before being masked. To pre-train a BERT, the losses of the two tasks are joined together as the objective function to optimise the model's parameters via back-propagation.

When it comes to downstream tasks fine-tuning, the parameters of BERT are initialised with the pre-trained parameters so that it inherits the general language knowledge learnt from the pre-training stage. To fine-tune BERT for a sequence pair classification task such as textual entailment, the input remains the same as the input for pre-training except for the masking. For a sequence classification task such as crisis tweets categorisation, still without the masking, the input only contains a single sequence (i.e., only sequence A). To be consistent with the pre-training, the fine-tuning normally takes the output state of $[\text{CLS}]$ as the sequence representation for text classification. To fit BERT on various classification tasks via fine-tuning, a task head such as a linear layer is usually added on the top of BERT, which projects the sequence representation to the class space that estimates the probability distribution across all classes. Taking text classification as an example, the loss function can be written as follows where $ \textbf{W}_o$ and $\textbf{b}_0$ are the weights and bias of the last linear layer projecting the output hidden state of $[\text{CLS}]$ to class space.

\begin{equation}
\begin{aligned}
J(s) = -\sum_{i=1}^m p(y_i|s)\log \hat{p}(y_i|s) \\
\hat{p}(y|s) = \text{softmax}(\textbf{o}_{\text{[CLS]}}\textbf{W}_o+\textbf{b}_0)
\end{aligned}
\end{equation}

Based on the transformer encoder, BERT adopts the MLM and NSP tasks for pre-training, resulting in great success in a wide range of language tasks via fine-tuning. This has prompted much follow-up work to optimise it. The optimisation is mainly seen in the literature regarding the memory use, computational cost and design choices of BERT. Since BERT has been introduced, many variants have been proposed to address certain shortcomings and incorporate improvements. The following is a selection of the most prominent among these.

\begin{itemize}
    \item \textbf{DistilBERT}~\cite{sanh2019distilbert} is a distilled version of BERT. Compared to the original BERT, it has fewer trainable parameters and is thus lighter, cheaper and faster during training and inference. Given the size of the reduced model, the original paper reports that it still keeps comparative language understanding capabilities and performance on downstream tasks.
    \item \textbf{RoBERTa}~\cite{liu2019roberta} is an optimised variant of BERT with several changes made. First, it uses a byte-level BPE as the tokenizer~\cite{wang2020neural} and uses the MLM as the only pre-training task (NSP is removed). Besides this, some changes are made to the hyper-parameters for pre-training, including much larger mini-batches and learning rates, etc. The results show that these changes help boost the downstream performance, which indicates that the previously design choices for BERT are sub-optimal.
    
    \item \textbf{ALBERT}~\cite{lan2019albert} is a derivative of BERT that is mainly optimised for memory efficiency with two parameter-reduction techniques. First, it splits the embedding matrix in a BERT-like architecture into smaller matrices (separation of the embedding size and the hidden size), leading to reduced memory use while mathematically maintaining equivalent effect. In addition, it uses repeated layers, where the parameters are shared across different BERT hidden layers. This optimisation results in a much smaller memory footprint although the computational cost remains similar to the original BERT (the same iteration through all hidden layers is still required).
    
    \item \textbf{ELECTRA}~\cite{Clark2020ELECTRAPT} maintains essentially the same architecture and size as the original BERT except for a change in the embedding matrix as in ALBERT. What makes it stand out is that it adopts a different pre-training approach. Unlike the MLM pre-training objective used in BERT, it trains a generator using the MLM objective to replace tokens in a sequence and meanwhile it trains a discriminator with the objective of identifying which tokens were replaced by the generator in the sequence. It is shown to outperform other transformers on language understanding benchmarks when using the same amount of computational power.
    
     \item \textbf{DeBERTa}~\cite{he2020deberta} is optimised upon RoBERTa using fewer training data by introducing two novel techniques for pre-training: disentangled attention mechanism and an enhanced mask decoder. In DeBERTa, the representation for each token consists of two vectors that encode its content and position respectively. The disentangled attention calculates the attention weights between tokens by applying disentangled matrices on the representations of their contents and positions. Besides this, instead of using the output softmax as in BERT, DeBERTa applies an enhanced mask decoder to predict the masked tokens for pre-training. The results show that DeBERTa achieves improved performance on many downstream tasks as compared to RoBERTa even using half of its training data.
\end{itemize}


\subsubsection{Decoder models}
\label{subsubsec:decoder}
Unlike encoder-based PLMs, the decoder-based PLMs have been widely studied in the literature given their strong generalisation capability in text generation. The most well-known models of this type are the Generative Pre-Training (GPT) family~\cite{radford2018improving, radford2019language,brown2020language}. Compared to BERT-like models, GPTs are pre-trained on the traditional language modelling task with the architecture being similar to the transformer decoder. Regarding the architecture,  similar to BERT, the encoder of a GPT consists of two sub-layers: the attention sub-layer and the feed-forward layer. In contrast with the cross attention sub-layer in BERT, the attention sub-layer of GPT is called a ``masked self-attention layer''. The major difference is that the masked self-attention only allows the current token to pay attention to its previous tokens, i.e., it is only informed by the tokens on the left side instead of both sides as in the cross attention of BERT (unidirectional versus bidirectional). In mathematical terms, the attention weights for the current token are computed by the dot product between the keys and queries of only the tokens on the left side (see Equation~\ref{eq:dot-scaled-product-attention}).

Regarding the pre-training task of GPTs, the language modelling is referred to as the next token prediction task or casual language task (CLM) distinguished from MLM. In general terms, CLM takes a sequence of previous tokens as the input (known as the context window) and uses the context to predict the next token. At the pre-training stage, the loss function for pre-training a GPT is defined as follows:

\begin{equation}
J_1(s) =- \sum_{i=1}^{T} \log \hat{p}\left(t_i \mid t_{1}, \ldots, t_{i-1} ; \theta\right)
\end{equation}

The objective of CLM is to maximise the likelihood of current token $t_i$ given previous tokens $t_{1}, \ldots, t_{i-1}$. When adapting it to a downstream task via fine-tuning, the loss function can be defined as follows:

\begin{equation}
J_2(s) = -\sum_{i=1}^T \log \hat{p}\left(y \mid t_{1}, \ldots, t_{i-1}; \theta\right)
\end{equation}

This defines the objective of generating the target token $y$ (i.e., the next token $t_i$) given the context tokens $t_{1}, \ldots, t_{i-1}$, which applies to downstream tasks that are based on sequence generation, such as question answering. In particular for a text classification task, the equation can be simplified as follows:

\begin{equation}
J_2(s) = -\log \hat{p}\left(y \mid t_{1}, \ldots, t_{T}; \theta\right)
\end{equation}

where the target token $y$ refers to the label or class name of the input sequence $\{t_1,...,t_T\}$. To summarise GPTs in fine-tuning, the downstream task can be viewed as a conditional generation task that coincides with the pre-training CLM task.

The GPT family of language models includes GPT-1~\cite{radford2018improving}, GPT-2~\cite{radford2019language}, GPT-3~\cite{brown2020language}, and ChatGPT~\footnote{\url{https://openai.com/blog/chatgpt/}}, etc. These models differ in terms of both the amount of training data they use and their model size. For instance, GPT-3 has 175B parameters, which is 10 times more than GPT-2 (1.5B parameters) and GPT-2 has 10 times more parameters than GPT-1 (117M parameters). In terms of data size, GPT-3 also uses significantly more text data for pre-training than GPT-2, which itself uses more data than GPT-1. ChatGPT is fine-tuned from a large-scale GPT model (GPT-3.5) using reinforcement learning from human feedback, excelling at answering user questions in a dialogue setting. The development of GPTs has seen the trend of large language models for various downstream tasks driven by the finding in the literature that larger language models with more data brings better generalisation capability~\cite{kaplan2020scaling}.

\subsubsection{Encoder-Decoder models}
\label{sec:encoder-decoder-models}
The architecture of encoder-decoder PLMs is more like a standard Transformer comprising an encoder and a decoder~\cite{vaswani2017attention}. The encoder-decoder PLMs are usually referred to as sequence-to-sequence (seq2seq) models since they take a sequence as the input and generate a sequence as the output. In many cases, the input sequence is also called the source sequence and the output sequence is called the target sequence. At a high level, the encoder learns to encode the source sequence to a vector that can represent its contextualised linguistic features. Conditional on the source representation, the decoder then learns to generate the prediction words iteratively. Mathematically, given a source sequence $s^a\colon\{t_1^a,t_2^a,...,t_{T^a}^a\}$, the seq2seq model generates predictions, i.e., the target sequence $s^b\colon\{t_1^b,t_2^b,...,t_{T^b}^b\}$ through a parameterised estimation of conditional probability distribution as follows.

\small
\begin{equation}
\label{eq:seq2seq_overall_loss}
 J(s^a,s^b)= - \sum_{i=1}^{T^b} \hat{p}(t_i^b | t_{1:i-1}^b, f(t_{1:T^a}^a;\theta_{e});\theta_{d})
\end{equation}
\label{eq:seq2seq-general}
\normalsize

Where $f(t_{1:T^a}^a;\theta_{e})$ refers to the mapping function from the source sequence $s^a$ to its contextualised representation, learnt by the encoder with parameters $\theta_{e}$. Likewise, $\theta_{d}$ is the parameters for the decoder to learn the function of conditional generation: $\hat{p}(\cdot)$. From this equation, it is easy to know that the next token of the target sequence $t_i^b$ is generated conditional on both the source sequence and previous tokens of the target sequence.

To look into the inner parts of the architecture of decoder-encoder PLMs, the encoder consists of hidden layers with each layer containing a cross-attention sub-layer and a feed-forward sub-layer, which corresponds to the encoder layer described in Section~\ref{subsubsec:encoder}. The decoder consists of hidden layers with each layer containing a masked self-attention sub-layer, an encoder-decoder cross attention sub-layer and a feed-forward sub-layer where the self-attention sub-layer and feed-forward sub-layer coincide with the decoder layer as described in Section~\ref{subsubsec:decoder}. The self-attention sub-layer is the same as the masked self attention used in GPTs, ensuring the next token that is generated is conditional on the previous tokens of the target sequence. The encoder-decoder attention is similar to the cross attention used in BERT. The difference is that the attention weights in the encoder-decoder attention are calculated between the keys of the encoder's tokens and the queries of the decoder's tokens. This attention ensures that the target sequence is generated conditional on the source sequence.

Due to the seq2seq attribute, the encoder-decoder PLMs can be easily adapted to various language tasks that can be viewed as a text-to-text problem. For text classification, the document becomes the source textual sequence and the label becomes the textual target sequence. For question answering, the question can be the source sequence and the answer can be the target sequence. This also applies to other language tasks such as text generation, summarisation, etc. By converting the input examples of various tasks to the text-to-text format, the encoder-decoder PLMs hence can easily be trained on multiple tasks at the same time (i.e., multi-task learning) based on the Equation~\ref{eq:seq2seq_overall_loss}. One important such work is T5~\cite{raffel2019exploring}, which is pre-trained on a multi-task combination of unsupervised and supervised tasks where each task is converted into a text-to-text format. For unsupervised tasks, T5 adopts a denoising mechanism as the pre-training objective using big unlabelled text data where the source sequence is partially corrupted and the objective is to generate the target sequence that predicts the corrupted parts of the source sequence. For supervised tasks, T5 relies on existing annotated resources of various bench-marking language tasks and converts them into a text-to-text format for pre-training.

By pre-training in the manner of T5, it has been shown to exhibit strong performance in downstream tasks via fine-tuning. This has prompted many follow-ups of T5 for different use cases. To just name a few of many; mT5~\cite{xue2021mt5} is a multilingual T5 model pre-trained on big text data from more than 100 languages; byT5~\cite{xue2022byt5} is a T5 model pre-trained on byte sequences instead of sub-word token sequences, boosting performance in tasks that are sensitive to spelling and pronunciation; LongT5~\cite{guo2021longt5} is a variant of T5 designed especially for long sequences; BART~\cite{lewis2019bart} is pre-trained based on different training objectives, which introduces noise to the input by means of various operations including token deletion, order shuffling, text in-infilling, etc., exhibiting strong performance in downstream generation tasks.

\section{Conclusions}

This chapter has introduced the general background of this research. The background review put its focus on text classification as the crisis messages categorisation task is in essence a classification problem. Considering that the major contributions of this thesis utilise neural network based approaches, this chapter has covered the general ideas of machine learning and the basics of neural networks. In addition, much of the content of this chapter has focused on the use of various neural networks including CNN, RNN and PLMs for text classification. They are important models upon which the contributions of this research are built. In particular, the introduction to the Transformer-based PLMs (Section~\ref{subsec:plms}) provides essential background of the crisis messages categorisation problem in low-data settings, which will be reviewed in the next chapter.

\chapter{Crisis Messages Categorisation in Low-data Settings}
\label{ch:literature}

As mentioned in Chapter~\ref{ch:introduction}, it is difficult to obtain labelled data relating to emerging crisis events and it can take time to annotate data, so there is a need for methods that can be used to categorise social media messages related to crises in low-data settings. This chapter reviews existing methods that can be used for this purpose and thus be used as baselines for evaluating the work presented in the later chapters of this thesis. The review first covers a family of methods called ``domain adaptation'', which involves adapting a model trained on labelled data from one domain (the source domain, a distribution of data) to another domain (the target domain, a different distribution of data). When the methods are used for crisis message categorisation, known as ``crisis domain adaptation'' and discussed in Section~\ref{sec:cda}, it involves adapting a model trained on labelled data from past crisis events (i.e. the source domain) to categorise messages from a current crisis (i.e. the target domain). The review also covers existing methods for training a model on a small amount of labelled data from the target domain, known as ``few-shot learning''. These methods are relevant for crisis message categorisation when labelled data from the past events is not available and it is feasible to annotate a small dataset for emerging events so the model can be trained on the small labelled dataset. These methods are discussed in Section~\ref{sec:cfs}. To further reduce or eliminate the reliance on labelled target data, the review also covers existing methods for training a model without any labelled data from the target domain, known as ``zero-shot learning''. These methods are relevant for crisis message categorisation when no time and annotation effort is available for annotation so the model can be trained only on unlabelled data from emerging events. These methods are discussed in Section~\ref{sec:czs}. Finally, related datasets and evaluation metrics used by this research are covered in Section~\ref{sec:rds-ems}.


\section{Crisis domain adaptation}
\label{sec:cda}

In crisis response, it can be difficult to use computational models to categorise messages from a new crisis event when labelled data for that event is not immediately available. To overcome this challenge, crisis domain adaptation involves building a model for a categorisation task using data from one or more past events (the source domain) and adapting it to categorise messages from one or more current events (the target domain). In order for this to be effective, the categorisation task must be the same between the source and target domains, meaning that the set of categories used for categorisation must be the same. For example, if the model was trained to categorise messages from past flooding events by the type of aid requested, it would need to use the same set of aid categories to categorise messages from a current earthquake event. The following presents the details of crisis domain adaptation.



\textbf{Definition}: Given a categorisation (classification) task $\mathcal{T}$ with a set of pre-defined aid categories, a source domain is represented by the data of $n$ events $D_s:\{D_s^1,D_s^2,D_s^n\}$ and a target domain is represented by the data of $m$ events $D_t:\{D_t^1,D_t^2,D_t^m\}$. Domain adaptation trains a model on $D_s$ and tests the model on $D_s$ within the same task $\mathcal{T}$. For simplicity, $D_s$ is said to be the source data and $D_t$ is the target data. The categorisation task $\mathcal{T}$ may be composed of multiple sub-tasks, each with its own set of categories, and domain adaptation can still be applied as long as the sub-tasks and their corresponding categories remain the same between the source and target domains. This is known as multi-task adaptation. Depending on the difference between the source domain and target domain, domain adaptation can be divided into in-domain and cross-domain adaptation. In-domain adaptation implies that the $m$ events in the target domain are the same as the $n$ events in the source domain, which is equivalent to fully supervised learning (training and testing on the same domain). Cross-domain adaptation emphasises that the $m$ events in the target domain are different from the $n$ events in the source domain (i.e., different data distribution such as hurricane versus flooding), which is the focus of this research~\footnote{Unless stated, domain adaptation implies cross-domain adaptation in the remaining content of this thesis.}. There are several categories of domain adaptation, including many-to-many, many-to-one, and one-to-one adaptation. In many-to-many and many-to-one domain adaptation (also known as multi-source domain adaptation), the source domain has multiple events ($n>1$) and the target domain has one or more events ($m\geq 1$). One-to-one adaptation is similarly defined. 

There are various methods in the domain adaptation literature that can be used when there is a varying amount of data available, including labelled source data, limited target data, or no target data. These methods can be grouped into categories based on the amount of data they require, with methods that require more data being listed first.

\begin{itemize}
    \item \textbf{Semi-supervised domain adaptation}: In addition to the usage of labelled source data, semi-supervised domain adaptation uses a small amount of labelled target data plus a larger quantity of unlabelled target data. To differentiate it from semi-supervised learning, it additionally uses labelled source data for model building.
    \item \textbf{Supervised domain adaptation}: In contrast to semi-supervised adaptation, supervised domain adaptation uses labelled source data and only a small amount of labelled target data with no unlabelled target data. The difference with supervised learning is that it uses labelled source data and a small quantity of labelled target data. 
    \item \textbf{Unsupervised domain adaptation}: Unsupervised domain adaptation is a branch of domain adaptation research that focuses on using labelled data from the source domain and unlabelled data from the target domain to build a model. This is of particular interest in real-world situations where it is costly to obtain labelled data for the target domain. 
    \item \textbf{Target data independent domain adaptation}: This gives the strictest limitation to the availability of target data. It uses labelled source data for domain adaptation, which avoids using any form of data related to the target domain. Although limited work has been conducted for domain adaptation that requires no target data, it is deemed an important research problem in areas such as crisis message categorisation where the acquisition of unlabelled data of target crises is not only very expensive but also impractical given the inherent time constraints (i.e. because it is not practical to wait for enough data to be gathered as an event unfolds before starting to categorise messages). With this motivation, the major contributions of this research in domain adaptation fall in this category.
\end{itemize}

This section reviews major approaches that have been proposed for domain adaptation in these categories. Among the approaches, some are proposed for non-crisis related tasks (such as movie reviews sentiment classification), which can be easily transferred to crisis message categorisation, and others are directly proposed for crisis message categorisation. To emphasise the importance of target data independent domain adaptation (the major focus of this research), the former three categories are reviewed as a joint category: target data dependent domain adaptation.

\subsection{Target data dependent domain adaptation}

The central idea behind domain adaptation is to align the source and target domain distributions or reduce the shift between the source and target data distributions~\cite{weiss2016survey}. The alignment refers to the process of bringing the data from the two domains into a common space where they can be compared and used to train a model. This can involve techniques such as projecting the data onto a shared latent space or matching the distributions of the data in the two domains. The goal of aligning the data is to make it easier to transfer knowledge from the source domain to the target domain, so that a model trained on the source domain can be more effective at making predictions on the target domain.

In target data dependent domain adaptation, the goal is to make the source domain similar to the target domain using labelled data from the source domain, unlabelled data from the target domain, and/or labelled data from the target domain. There are two main approaches to achieving this in the literature: methods based on instance selection, and methods based on feature adjustment. These methods aim to bridge the gap between the source and target domains in order to improve the performance of a model trained on the source domain on the target domain.


\subsubsection{Instance selection methods}

Instance selection methods aim to adapt a classifier to a new domain by choosing a subset of instances from both the source and target domains for the adaptation process. When there are few labelled target data and unlabelled target data available, a common way to implement instance selection for domain adaptation is to use a two-step process~\cite{jiang2007instance,sugiyama2007direct,tsuboi2009direct,mazloom2019hybrid}. In the first step, a subset of instances is selected from both the source and the target domains. This can be done using various methods, such as selecting the most representative or informative instances, or using a search algorithm to find the subset of instances that maximises the performance of the classifier. In the second step, weights are assigned to the selected instances based on their relative importance or relevance for the adaptation process. This can be done using various techniques, such as assigning higher weights to instances from the target domain or to those that are more difficult to classify correctly. The final classifier is then trained on the combined set of selected and weighted instances.


 When only unlabelled target data is available, the target data can be selected by pseudo-labelling the unlabelled target data using a classifier trained on the source data, and then added with their pseudo labels to the training data~\cite{dai2007transferring,tan2009adapting,peddinti2011domain,herndon2015evaluation,li2015twitter,li2018disaster}. Here, pseudo labels refer to predicted labels by the classifier and these predicted labels are treated as if they were ground truth labels. There are two approaches to pseudo-labelling unlabelled target data: using hard labels, which are determined through self-training, or using soft labels, which are obtained through expectation-maximisation~\cite{li2021domain}. Hard labels are discrete class assignments, while soft labels are probabilities or confidence scores indicating the likelihood that an instance belongs to a particular class. 




The self-training method~\cite{chen2011co} can be described with two steps. The first step is to train a classifier on the labelled source data that assigns hard labels (e.g., 0 or 1) to the unlabelled target with high confidence predictions. The second step is to combine the hard-labelled data with the source data for training the classifier. This two-step process repeats until the classifier converges or reaches the designated maximum number of iterations. In the expectation-maximisation (EM) method, the process can be divided into an expectation step and a maximisation step. In the expectation step, the expected likelihood of the target data is evaluated using a classification model with parameters estimated from the labelled source data. In the maximisation step, the parameters are optimised by maximising the expected likelihood evaluated in the expectation step. The goal is to find the set of parameters that best explains the observations in the target data. To maximise the expected likelihood, there are soft labels (e.g., 0.6 or 0.4) assigned to the unlabelled instances. In a similar way to the self-training method, EM is also an iterative method for instance selection and the iterative process does not stop until convergence or reaching the maximum number of iterations. The following gives a review of important instance selection methods that relate not only to crisis messages categorisation but other non-crisis related tasks.

In the work by~\cite{dai2007transferring}, a EM-based domain adaptation approach was proposed for news topic classification using  Na\"ive Bayes classifiers~\cite{rish2001empirical}\footnote{This work studied three domains represented by news of three resources with unified labels: 20 Newsgroups, SRAA and Reuters-21578.}. Basically, this approach first estimates the priors (the parameters of a Na\"ive Bayes model) using the labelled source data and then an EM algorithm is applied to tune the model based on the distribution of unlabelled data from the target domain. Their experimental results found that the approach performs better than the traditional supervised algorithms including Support Vector Machine (SVM~\cite{noble2006support}) and Na\"ive Bayes classifiers, especially when the source domain and target domain are similar.

Another work by~\cite{jiang2007instance} proposed an instance selection (and weighting) framework for domain adaptation assuming the availability of labelled source data, a small set of labelled target data and a large amount of unlabelled target data. This work takes into account three major aspects to improve the adaptation performance, which is tested on several major NLP tasks including part of speech tagging, spam filtering, etc. First, it decides whether a training instance $x$ in the source domain is ``misleading'' or not by comparing its distribution for a class $y$ in the target domain $p_t(y|x)$ using labelled target data with its corresponding distribution in the source domain $p_s(y|x)$ labelled source data. Besides, when assigning higher weight to labelled target instances than labelled source instances (more information from the target domain), this is more effective than excluding "misleading" source instances. Last, it is necessary to augment the training data with confidently predicted target instances.

In addition, in this work~\cite{tan2009adapting}, they proposed a two-step EM based approach for domain adaptation for the task of sentiment analysis\footnote{They studied three domains represented by reviews of three resources with binary sentiment labels: Education Reviews, Stock Reviews and Computer Reviews.}. First, to better utilise the source domain knowledge, they proposed Frequently Co-occurring Entropy (FCE) to filter out non-generalisable (i.e., domain-specific) features and only keep the important features that have similar co-occurring probability in the source and target domains. Next, to better incorporate the target domain knowledge, they proposed Adapted Naïve Bayes (ANB), which is a weighted variant of the multinomial Na\"ive Bayes Classifier. The main idea behind ANB is to use a weighted EM algorithm to bring confidently pseudo-labelled target instances to the training set where the weight is shifted from the source domain to the target domain. 

In another work by~\cite{peddinti2011domain}, they studied domain adaptation in the context of sentiment analysis on Twitter from the perspective of finding good source instances that can inform the target domain. To achieve this, they proposed two iterative algorithms based on EM and Rocchio SVM to select good labelled source instances. Unlike previous works~\cite{dai2007transferring,jiang2007instance,tan2009adapting} requiring unlabelled target data, this work assumes the availability of source domain data and a small amount of labelled target data. In the EM iteration, they train a SVM classifier on the labelled target data and use it to pseudo-label source instances . Last, only the confidently-classified source instances are added back to the training set. 

Another work by~\cite{herndon2015evaluation} used a weighted Na\"ive Bayes classifier based on the EM algorithm, which is similar to the previous works~\cite{tan2009adapting, herndon2014empirical}, for domain adaptation in the task of splice site prediction. When it comes to the question of how to select the unlabelled target data (i.e., to assign pseudo labels to the unlabelled target instances), they comparatively studied three approaches: using hard labels only via self-training, using soft labels only via EM and using a combination of soft and hard labels. The experimental results in the task of splice site prediction indicate that to use soft labels is generally better than the other alternatives.

Inspired by the aforementioned works, some works have been found in the literature for the study of domain adaptation in the task of crisis messages categorisation. For example, the work by~\cite{li2015twitter} was considered to be the first work in adapting general domain adaptation approaches to the crisis domain. They studied domain adaptation in the task of classifying disaster-related tweets where an old disaster is viewed as the source domain and a new disaster is treated as the target domain. They make use of the unlabelled target data for crisis domain adaptation based on the iterative EM algorithm and a weighted Na\"ive Bayes classifier. In their proposed approach, they first train the Na\"ive Bayes classifier on the labelled source data (training data) and then use it to soft-label the unlabelled target data. To adapt the classifier from the source domain to the target domain, the soft-labelled instances are then added to the training set and assigned with more weight for the next iteration of training. The iterative process continues until the classifier converges or reaching a designated number of iterations. 

In another work by~\cite{li2018disaster}, the problem of crisis domain adaptation was studied similarly to their previous work~\cite{li2015twitter}. The difference is that they use unlabelled target data for crisis domain adaptation based on a weighted classifier and the iterative self-training algorithm instead of the EM algorithm as in~\cite{li2015twitter}. In this work, the unlabelled target instances are hard-labelled by the pre-trained classifier. When combining the hard-labelled instances with the original labelled source instances, only the most confidently classified instances are added to the training set. In experiments, they set up the classifier with different choices including Na\"ive Bayes, SVM, Random Forests and Logistic regression. The results show it gives overall better performance when the classifier is Na\"ive Bayes than other choices. In addition, they compared the NB-based self-training approach with the NB-based EM approach and supervised NB using labelled source data only. It is found that the NB-based self-training approach overall outperforms the other two approaches.

 Building upon previous works~\cite{li2018disaster,mazloom2018classification}, this work~\cite{mazloom2019hybrid} proposed a hybrid domain adaptation approach for crisis tweet classification. The hybrid approach adopts the iterative self-training algorithm as in~\cite{li2018disaster} for building the classifier. However, it takes two extra steps to select only a better-aligned subset of labelled source data before the self-training step. The first is to apply the Least Squares Non-Negative Matrix Factorization (LSNMF)~\cite{lee2000algorithms} to reduce the instance-feature data matrix representing the source-target pairs of events to a low-dimensional space. In selecting the subset of labelled source data, KNN~\cite{guo2003knn} is then applied on the reduced matrix and used to select only the top $k$ nearest instances for each instance of the unlabelled target data. They found that the NB self-training using unlabelled target data outperforms other classifier choices and NB using labelled source data only, which are findings similar to those of~\cite{li2018disaster}.

Instead of using traditional ML models as in~\cite{li2018disaster}, this work~\cite{li2021combining} combined self-training with deep learning models including a CNN model and pre-trained language models (a BERT and a BERTweet, whaich is a variant of BERT pre-trained on English tweets~\cite{nguyen2020bertweet}) for crisis domain adaptation. In experiments with self-training, it is found that BERTweet overall outperforms CNN and BERT. In experiments with and without self-training, the results show  that self-training with unlabelled target data helps improve the adaptation performance, especially in situations where a large amount of target unlabelled data is available. This indicates the benefit of using unlabelled target data for crisis domain adaptation as compared to without it, which coincides with previous works~\cite{li2018disaster,mazloom2019hybrid}.


\subsubsection{Feature adjustment methods}
\label{subsec:feature-adjustment-methods}
Another branch of target data dependent domain adaptation methods is based on feature adjustment. It looks into different approaches of aligning the source domain with the target domain; that is to change the feature representations. Given features such as bag-of-words fixed vocabulary representations or continuous representations, the core idea behind these methods is to keep as many cross-domain features (generalisable) as possible while simultaneously filtering out domain-specific features. Depending on whether the generalisable features are learnt through representation learning using deep learning models, the feature adjustment methods can be divided into two categories; traditional based and deep learning based. 

Among the traditional based methods, these works by~\cite{blitzer2006domain,blitzer2007biographies} proposed structural correspondence learning (SCL) for domain adaptation in the tasks of part of speech tagging and sentiment analysis. SCL is adopted to choose features (known as pivotal features) that can help the classifiers distinguish between domains. By using the labelled source data and unlabelled target data, pivotal features are chosen in two ways: based on their common frequencies in both domains and mutual information in the situations such as sentiment analysis where frequently occurring words may vary significantly across domains. When some labelled target data is available, SCL uses it to correct the misalignment of the pivot features. Similarly, with the use of few labelled target data, this work by~\cite{daume2007frustratingly} studied multi-source domain adaptation by adopting a simple method to represent the target and source data where the feature representations are divided into three types: source domain-specific, target domain-specific and general. Regarding finding generalisable features for domain adaptation, this work by~\cite{jiang2007two} proposed a simple approach for domain adaptation with two-steps. The first step is  to find a set of generalisable features across domains and the second step is to only keep the target domain-specific features. In a similar way to SCL, this work by~\cite{pan2010cross} introduced a spectral feature alignment (SFA) method to align domain-specific words across domains into unified clusters with the help of domain-independent words acting as a bridge. In SFA, the clusters are used to reduce the gap between domain-specific words, which help train the classifiers for the target domain. Another important traditional feature-based method was by~\cite{sun2016return} who proposed CORrelation Alignment (CORAL) for domain adaptation for object recognition and sentiment analysis tasks. CORAL is a simple unsupervised domain adaptation approach to align the data distribution of the source domain with that of target domain. It minimises domain shift by aligning the covariance of source and target distributions. In the CORAL approach, the covariance statistics are calculated for each domain and the distributions are aligned by adjusting the source features using the covariance of the target features. This method was then adapted and used with a self-training iterative algorithm for adaptation in a crisis domain.~\cite{li2018domain}. 

Unlike the traditional methods aiming to obtain the domain-invariant features from bag-of-words fixed vocabulary feature representations, deep learning based methods apply neural networks to learn continuous feature representations for both domains. In the literature relating to feature based target data dependent domain adaptation, this is usually achieved by the autoencoder networks or domain adversarial networks. 

An autoencoder is a type of neural network that learns a latent representation for an input via an encoder and then reconstructs the representation via an decoder in an unsupervised manner~\cite{Minmin2012Marginalized}. To apply it to the domain adaptation problem, the encoder can be trained on the source data and unlabelled target data and used to obtain the feature representations of both domains. Subsequently, a supervised classifier can be trained on the representations for the target domain classification. For example, this work~\cite{glorot2011domain} introduced Stacked Denoising Auto-encoders (SDA) for domain adaptation for the task of sentiment analysis. In SDA, multiple autoencoders are stacked together as a neural network model and the model is trained by denoising or reconstructing the latent representations from the encoder whose input is corrupted. Once the model is trained, the encoder is applied to obtain the latent representations of both domains. The latent representations are then used as the input representations for an SVM classifier. Other works using autoencoders in a similar way for sentiment classification domain adaptations can be found in~\cite{Minmin2012Marginalized,zhuang2015supervised,zhou2016bi,ziser2017neural,ziser2018pivot}. Another important autoencoder-based work is by~\cite{ghifary2016deep} who proposed deep reconstruction classification networks (DRCN) for domain adaptation. The DRCN consists of an autoencoder that is co-trained with a source classification network. The encoder of the autoencoder is shared with the source classification network and thus can learn information from both domains. Finally, the source classification network trained on source domain data is adapted to the target domain. Built on top of DRCN, this work~\cite{li2020domain} used a RNN (LSTM) network as the autoencoder for crisis domain adaptation relating to the task of disaster tweet classification. It was found that the reconstruction task can bring benefits to the adaptation performance as compared to the classification task alone.

As an alternative to autoencoder based methods, domain adversarial neural networks (DANN)~\cite{ganin2016domain} achieve domain adaptation by training a feature extraction network with the supervised source classification network at the same time. The feature extraction network aims to learn domain-invariant features by maximising the loss of a domain classifier that is used to distinguish between the source and target domains. To achieve the joint learning, the loss of the source classification network is combined with the the loss of the domain classifier as the objective function for optimising the networks' parameters via backpropagation. Based on this adversarial idea, some related works have been conducted in the literature. For example, this work~\cite{li2017end} proposed an end-to-end adversarial memory network for cross-domain sentiment classification where an attention mechanism is used to capture pivotal features. Another work~\cite{du2020adversarial} adapted BERT for domain adaptation, achieving state-of-the-art performance in cross-domain sentiment classification. To enable BERT to be domain-aware and distil domain-specific features, BERT is post-trained on a target domain masked language task and a novel domain-distinguishing pre-training task in a self-supervised manner. By doing this, the adversarial training is then conducted to derive the robust domain-invariant features. Given its strong performance, DANN has been also adapted and applied to the crisis domain in recent years. this work~\cite{li2019identifying} used VGG-19 as the backbone of DANN for disaster damage images identification where the adversarial training is used to find a transformation that makes the source and target data indistinguishable. In another work, this work~\cite{alam2018domain} proposed a CNN based domain adversarial network with graph-based semi-supervised learning for crisis domain adaptation in the task of crisis message relevance identification. The network applies a joint training of a supervised, a semi-supervised and a domain classifier on data of both domains to learn good domain-invariant representations. The semi-supervised classifier is trained to learn latent representations by predicting contextual nodes in a graph that encodes similarity between labelled and unlabelled training examples. The domain classifier is trained to distinguish the domains and the supervised classifier is trained for the main classification task. The experimental results show significant improvements of this work over baselines.



\subsection{Target data independent domain adaptation}

Target data independent domain adaptation methods assume no availability of any form of target data, including labelled and unlabelled data. Multiple previous studies have shown that the performance of domain adaptation with no target data is inferior to that with large amounts of unlabelled target data~\cite{alam2018domain,li2018disaster,mazloom2019hybrid,li2021combining}. This indicates that domain adaptation can be difficult without incorporating the information from the target domain. Despite this challenge, target data independent domain adaptation is considered to be an important research problem in the area of rapid crisis messages categorisation on social media. This is mainly because, in real-word crisis response, the need for categorising a new crisis is urgent and the target data is not readily available when the new crisis occurs. Target data independent domain adaptation is proposed to meet this use case. From the perspective of building a computational model, the research question becomes how to build the model trained on past events (source domains) and adapt it to a new event (target domain). 

Given its importance and that existing works in this direction are limited, this research mainly focuses on target data independent domain adaptation. Among the existing works~\cite{nguyen2016rapid,li2018comparison,li2021combining}, one common feature is that they use pre-trained word embeddings as generalised representations for crisis tweets, followed by a classification model or a sentence encoder for classification. Since the word embeddings are usually pre-trained on general texts such as Wikipedia pages, books or news, the general language information learnt from the pre-trained stage can be used to represent textual sequences in different domains~\footnote{Within the problem domain of this research, these textual sequences are messages relating to crisis events.}. In the literature, the commonly-used pre-trained word embeddings include Word2Vec, GloVe, and FastText (see Section~\ref{subsubsec:pre-trained word-embedding} for further discussion). Once the tweets are represented by the word embeddings, the subsequent classification can be achieved by a simple traditional ML model such as Na\"ive Bayes, which uses the representations as the input features. Another way to achieve classification is to apply a sentence encoder to learn a contextual sequence representation from the word representations. The sentence encoder normally refers to neural network models such as a CNN, LSTM or BERT. In such an encoder, a classification linear layer can be easily added on the top of the sentence encoder and hence the classification is done along with the sentence representation learning. Below gives a review of important existing works related to target data independent domain adaptation.




The work by~\cite{nguyen2016rapid} studied the problem of rapid classification of crisis-related tweets. This work is conducted to particularly address the real-world crisis situation where no target data is available when a new target crisis occurs. When there is no target data available at the beginning of the crisis occurrence, they found that a CNN model with crisis-specific pre-trained word embeddings outperforms traditional ML models and a CNN with general pre-trained word embeddings such as Word2Vec. However, to further improve the adaptation performance, they suggest using unlabelled target data by either regularised adaptation or instance selection as the target crisis unfolds over time.

Another work by~\cite{li2018comparison} identified the problem of no target data being available for domain adaptation in time-critical situations as disasters occur. They proposed using pre-trained word embeddings as the generalised representations for crisis tweets. They explored a wide range of both word embeddings and sentence encodings with traditional ML algorithms for crisis adaptation classification tasks. They found that general pre-trained GloVe embeddings overall outperform other embeddings and the GloVe embeddings trained on crisis data bring better results on more specific crisis tweet classification tasks.

Another work by~\cite{li2021combining} comparatively studied target data independent and target data dependent domain adaptations using CNN and BERT models with self-training. Another similar study of applying LSTM with various word embeddings for domain adaptation is found in~\cite{li2021domain}. The experimental results show that self-training with a large quantities of unlabelled target data boosts the performance as compared to without it. When there no target data is available, the results show that the pre-trained language model BERT performs much better than CNN and LSTM. Besides this, it further improves the performance when the BERT is pre-trained on tweets generally (i.e. not specifically on crisis-related tweets).

\section{Crisis few-shot learning}
\label{sec:cfs}

In rapid crisis response, domain adaptation achieves crisis message categorisation by adapting a model trained on past crisis events (source domains) to a current new crisis event (target domain). However, one premise of domain adaptation is that the categorisation task of target events has to be exactly the same as that of the source events. In other words, the pre-defined aid types of the source data need to be consistent with those of the target data so that the adaptation is possible. Although this requirement is satisfied in some real-world situations, it is insufficient in situations where the responders seek new aid types in an unfolding event that were not considered for past events, even though annotated data with different labels is available. For example, the responders may look for a new water-based rescue aid type in a storm while this aid type has never been recorded (or been relevant to) past events such as fire.

This marks the importance of crisis few-shot learning in rapid crisis response, which uses no labelled source data but a small quantity of labelled or unlabelled target data for crisis message categorisation. To achieve crisis few-shot learning, a small data set comprising a few equal number of instances per class from the target event and an unlabelled corpus of the target event are also used when available. The former is relatively inexpensive to obtain as the annotation for a few instances can be done quickly at the beginning of a new crisis with little human labour. The latter is easy to acquire as the new crisis unfolds. Crisis few-shot learning asks the question of how to build a classification model using the few labelled target instances as well as the unlabelled target instances for unseen crisis message categorisation.

By examining the literature, crisis few-shot learning has been rarely studied. However, many works on few-shot learning have been conducted for other NLP tasks and these works can be adapted to the crisis domain and used as strong baselines for the research into crisis few-shot learning. The works on few-shot learning can be grouped into two main categories: prompt-based learning methods and augmentation-based methods. This section gives a review of them separately.  



\subsection{Prompt-based learning methods}
\label{subsec:prompted-based-few-shot}

Prompt-based methods for few-shot learning borrows the idea of transfer learning using pre-trained language models (PLMs). In the standard method of fine-tuning PLMs for downstream classification tasks, the labelled instances of target tasks are fed to the models directly. However, when the labelled instances are few, the fine-tuning performance can be poor, due to having insufficient data to learn from. Although this problem can be alleviated by very big PLMs such as GPT-3~\cite{brown2020language}, which uses few instances as the input context to generate the classification results directly with no need for fine-tuning, they are too large to be deployed in domains that require efficiency such as rapid crisis response. Given small or medium-sized PLMs, prompt-based learning conducts few-shot fine-tuning for downstream tasks in a different way. Instead of feeding the raw instances to the models, prompt-based learning reconstructs them via a set of prompt templates before feeding them to the models. The idea behind this is to reconstruct the instances so that they can be fine-tuned in a way that is more similar to the pre-training, leading to better transfer learning from pre-training. Three works have been identified that are particularly important in the area of using prompt-based learning with PLMs for few-shot classification. These are strong benchmarks for use in this research, and are discussed as follows.



Pattern-Exploiting Training (PET) was introduced by~\cite{schick2021exploiting} --- a semi-supervised approach that trains language models on a given task by reformulating few instances as cloze-style questions. The cloze-style questions are natural-language sequences encoding some form of task description where the label field is masked for prediction similar to the pre-training MLM task in PLMs such as BERT and RoBERTa. In PET, a set of hand-created task-specific patterns (i.e., prompt templates) are used to convert the training instances so that the PLMs can be fine-tuned in a way similar to the pre-training~\footnote{When applying PET to crisis few-shot learning, a rescue-related message may appear as ``I am on a boat, rescue needed!''. Assuming a pattern designed in this form: ``$\langle$TEXT$\rangle$. It's about $\langle$LABEL$\rangle$'', the message is then converted by this pattern to a fill-in-the-blank style ``fill-in-the-blank'' question: ``I am on a boat, rescue needed! It's about \_\_\_'' where the underlined blank text ``\_\_\_'' implies the label field masked for prediction.}. To leverage the unlabelled data of a given task, PET trains multiple PLMs on the few training data converted by different patterns and then applies the ensemble of PLMs to soft-label the unlabelled data. This coincides with the previously-mentioned self-training in domain adaptation where unlabelled data are pseudo-labelled first. Similarly, PET subsequently uses the soft-labelled data to train a classifier in a supervised fashion. By testing PET in experiments, it exhibits state-of-the-art performance in many general language tasks and even better performance than supervised methods in several tasks. However, one drawback of PET is its reliance on the unlabelled data and the hand-designed task-specific patterns.


To address this limitation of PET, the work by~\cite{gao2021making} proposed a technique for better few-shot fine-tuning of medium or small-sized language models (LM-BFF). In reformulating an input sequence with a few task demonstrations are appended to the template-based prompt as the input context for language model training. These demonstrations comprise the instances sampled from the few shot data of classes other than the class of the current input instance. Additionally, the templates for constructing the input are not hand-designed but automatically generated by leveraging a seq2seq language model (T5). They only use a few annotated instances as supervision and also explore automatically generated prompts and fine-tuning with demonstrations, avoiding the use of unlabelled data and making it a task-agnostic method. In experiments, LM-BFF showed much better performance than standard fine-tuning on many tasks and improved performance especially with the demonstrations as compared to PET. Although LM-BFF outperforms standard fine-tuning on small datasets, there is still a gap to match its performance with that of standard fine-tuning on larger datasets. It is also demonstrated with high variance for the results of LM-BFF due to the randomness in the selection of a small amount of data for model training, which is a common issue of many existing few-shot methods. In addition, there is evidence to indicate that LM-BFF favors certain tasks, e.g., the task generalisation problem.

 Similarly, with focus on prompt design, the work by~\cite{zhang2022differentiable} proposed a different approach to make small language models for better few-shot learners. They proposed Differentiable prompt (DART) that uses a few unused tokens in the language model, serving as the differentiable template and label tokens so that they can be optimised with backpropagation. DART uses differentiable prompts for few-shot learning, aiming to generate more discriminative representations than the fixed prompts that are used in PET. In addition, DART also applies an auxiliary fluency constraint object to ensure the association among the prompts embeddings. An ablation study shows that the differentiable templates and labels and the fluency constraint object all help improve the performance. It achieves better performance than LM-BFF without demonstrations and comparable performance to LM-BFF plus demonstrations. However, the instability problem and task generalisation problem still exist as open research questions, as stressed in previous few-shot learning works.
 
\subsection{Augmentation based methods}
\label{subsec:aug-based-methods}

Given few-shot data, prompt-based learning methods reformulate the input by using prompts to improve transfer learning from PLMs. Augmentation based methods instead tackle the few-shot learning problem by augmenting the original few-shot data. Data augmentation is an idea to bring better performance by increasing the quantity of training data. It was originally used in computer vision to obtain synthetic training images by various ways of modifying the original images such as rotation, cropping, shearing or scaling, etc~\cite{shorten2019survey}. In contrast to images, data augmentation in NLP (i.e., text augmentation) is deemed to be a non-trivial research problem since minor modifications to a piece of text may lead to a drastic change in its semantic meaning. This brings two major challenges. First, to bring new knowledge to the training data, the augmented synthetic instances are expected to be as diverse as possible to the original training instances, known as the ``lexical diversity'' challenge. Apart from being diverse, the synthetic samples also have to be label-aligned regarding their semantics (i.e., their semantic meanings should coincide with the labels that are assigned to them, otherwise confusing the model), known as the ``label alignment'' or ``semantic fidelity'' challenge. 


Data augmentation for text classification has been widely developed in the literature. The work by~\cite{zhang2015character} demonstrated that replacing words or phrases with lexically similar words such as synonyms or hyponyms/hypernyms is an effective way to perform text augmentation with minimal loss of generality. The authors identify the target words according to a predefined geometric distribution and then replace words with their synonyms from a thesaurus. Similarly, the work by~\cite{wei2019eda} proposed Easy Data Augmentation (EDA) for text classification that generates new instances from the original training data with four simple operations; synonym replacement, random insertion, random swap, and random deletion, while another work by~\cite{feng2020genaug} further explore these substitution techniques, particularly for text generation. The work by~\cite{wang2015s} instead exploit the distributional knowledge from word embedding models to randomly replace words or phrases with other semantically similar concepts. The work by~\cite{kobayashi2018contextual} build upon this idea by replacing words based on the context of the sentence, which they achieve by sampling words from the probability distribution produced by a bi-directional LSTM-RNN language model at different word positions. Back translation (BT) is another method for text augmentation~\cite{sennrich2016improving,shleifer2019low}. Here, new samples are generated by translating the original sentence to an intermediate language, before it is eventually translated back to the original target language. Sometimes, back translation with hops (BT-hops) is used where the original sentence is translated to multiple intermediate languages before it is translated back to the original language. More recently, to bring better label alignment, most works have used pretrained transformer-based language models for text augmentation by performing conditional text augmentation to obtain new sentences from the original training data. For example, the work by~\cite{wu2019conditional} proposed Conditional Contextual Augmentation (CBERT) that leverages the masked language model of BERT conditioned on label surface names to replace words contextually. Similarly, another work by~\cite{kumar2020data} used a seq2seq model BART depending on the label names to replace text spans contextually (BART-Span). One common feature among these works is that they increase the quantity of training data by reformulating or translating the original data. The reformulation usually involves word substitutions with synonyms and the translation results in sentence rephrasing. This feature makes them good for obtaining new instances with good label alignment but difficult to bring diverse (new knowledge) instances to the original training data. 

To ensure diversity, recent years have seen another line of work in text augmentation: generation-based methods. Here, novel instances are generated by generative language models trained on the original few training instances. For example, the work by~\cite{kumar2020data} proposed fine-tuning GPT-2 for text augmentation. In fine-tuning GPT-2, the raw instances are concatenated with their corresponding class names as the input. At inference time, a new sequence of a class is generated by using the class name along with several seed words of a sampled original sequence as the input context. Although the GPT-2 in this work is able to generate instances that are diverse in terms of semantics, it is found that the generation that is conditional on labels and seed words still brings noise to the training data (the label misalignment problem). Taking the noise into account, another work by~\cite{anaby2020not} (GPT-2-$\lambda$) introduced a post-denoising step to filter the noisy generated sentences. In this work, a new sequence of a class is generated that only depends on the class name. The post-denoising step involves training a classifier on the original training data and then using the classifier to make predictions for the generated sentences. The generated sentences with high-confidence predictions are selected and added to the training data. By testing different classification models, they found that using BERT as the classifier results in the best performance overall. It is also found that the post-denoising step helps to alleviate the label misalignment problem, thus resulting in better final performance.

\section{Crisis zero-shot learning}
\label{sec:czs}

Crisis few-shot learning is introduced for the real-world use case where the target categorisation task is different from the source categorisation task and so domain adaptation is not possible. Crisis zero-shot learning is introduced for the same purpose but to extend the use cases of crisis few-shot learning. In few-shot learning, a small quantity of instances per class are assumed to have been annotated. Although this annotation burden is small compared to fully-supervised learning, it is still time-consuming in some specific domains such as crisis categorisation. Additionally, to complete the annotation the class distribution is usually assumed to be uniform, i.e., equal numbers of instances per class are annotated. This uniform class distribution assumption may deviate significantly from the actual distribution, leading to bias in the classification models. To replace crisis few-shot learning, crisis zero-shot learning is considered to be a more challenging research problem as it achieves crisis message categorisation when there is no annotated data (i.e., zero labelled instances) from the target domain.

Considering the importance of crisis zero-shot learning, this research further investigates the potential of performing crisis message categorisation without any labelled data. Similar to crisis few-shot learning, little work has been found in the literature for crisis zero-shot learning, but many zero-shot learning studies have been conducted for other text classification tasks. This research in crisis zero-shot learning is inspired by the general works on zero-shot learning and hence they are used as strong baselines in the later chapters of this thesis. This section reviews them by dividing them into two categories: Indirectly-supervised learning methods and Weakly-supervised learning methods.

\subsection{Indirectly-supervised learning methods}

Given a classification task of a target domain, zero-shot learning aims to build a classifier that predicts test instances from that domain without supervision (i.e. without learning from any previously-annotated training instances). The lack of labelled target data makes it impossible to train the classifier in a directly-supervised way. This has motivated much research to train classifiers in an indirectly or distantly-supervised way. Indirectly-supervised methods transfer classifiers trained on external resources (e.g., labelled or unlabelled data from other domains) to the target domain. Similar to few-shot learning, zero-shot learning addresses the issue of different tasks between domains. Hence, the tasks across domains have to be unified so that the transfer learning with indirect supervision is possible. This task unification is normally achieved by introducing a matching model between text pairs. Basically, the matching model is first trained on some external datasets comprising text pairs and its objective is to learn the semantic similarity between two pieces of text. When tested for the target classification task, the matching model assigns a class to a piece of text according to the similarity between the text and the class that is represented by its surface name or a description of its name. The following gives details of important such methods in the literature.


The work by~\cite{veeranna2016using} proposed a very simple method for zero-shot learning. They used pre-trained word embeddings to represent n-grams of testing instances (sentences or documents) and the classes' surface names. To assign one or more classes to a testing instance, the cosine similarity between the instance's n-grams and each class is calculated and used as the semantic similarity score for classification. Instead of relying on readily-available pre-trained embeddings, another work by~\cite{pushp2017train} proposed three different LSTM-based neural networks to learn the relationship between text pairs consisting of news headlines along with their Search Engine Optmisation (SEO) tags. In this work, millions of news headlines along with SEO tags are used as the external resources to train the matching model that can be applied to classification tasks where each testing instance with the classes comprise the text pair for similarity calculation. Another work is done by~\cite{yin2019benchmarking}, which proposed an entailment approach that applies pre-trained BERT to learn the relationship between text pairs that consist of the premises and hypotheses from three textual entailment datasets (RTE/MNLI/FEVER). The entailment datasets are external supervision resources with annotations associated with text pairs to indicate whether the premise entails the hypothesis (high semantic similarity) or contradicts the hypothesis (low semantic similarity). Hence, they are good resources for a model to learn semantic similarities between text pairs. Other similar works using indirect supervision resources for zero-shot learning can be found in~\cite{levy-etal-2017-zero} and~\cite{obamuyide2018zero}, who studied zero-shot relation extraction by transforming it into a machine comprehension and textual entailment problem respectively.

In the aforementioned works, to assign a class to a testing instance, the instance has to be paired with every class's description, which leads to linear computation complexity as the number of classes increases. For optimisation, the work by~\cite{muller2022few} applied siamese networks and label tuning to tackle this inefficiency issue of the entailment approach at inference time. Another work to address the inefficiency issue is found in~\cite{puri2019zero}, who achieved zero-shot model adaptation to new classification tasks via generative language modelling. They trained GPT-2 on text pairs consisting of Reddit posts and multiple titles including correct ones and incorrect ones. Here, the text pairs are formatted as multiple-choice questions and the model's prediction is to identify the correct class given a testing instance. In addition, the work by~\cite{joe2021} attempted to overcome the inference inefficiency issue by distilling the entailment matching-based models from~\cite{yin2019benchmarking,ye2020zero}. This method assumes the availability of unlabelled target data. The matching model is first used to pseudo-label the unlabelled target data with soft labels and then use it for target classification model training in a standard supervised way. This method is closely related to weakly-supervised learning, which is introduced next.

\subsection{Weakly-supervised learning methods}
\label{subsec: weakly-sup-czs}

Indirectly-supervised learning methods achieve zero-shot learning by building a matching model to learn semantic similarities between text pairs. Weak supervision based methods assume the availability of unlabelled target data and pseudo-label the unlabelled data as the training data for the target classification model. Hence, the quality of pseudo-labelled data usually determines the final performance of the classification model. The pseudo-labelled can be generated using class representations or obtained by a matching method between unlabelled instances and classes. The classes are typically expanded before they are matched with the unlabelled instances in order to represent them with richer meanings. Two methods of expansion are typically used. The classes can be expanded computationally by the addition of related words or phrases known as (extremely weak supervision~\cite{meng-etal-2020-text}) or by adding a small number of human-provided seed words (known as weak seed-word driven supervision~\cite{wang2021x}). A lot of research has been done in the literature to examine class expansion and produce better-quality pseudo-labeled data. The most significant of these publications are reviewed below; they will serve as the research's foundation in subsequent chapters.

The work by~\cite{meng2018weakly} proposed a framework for Weakly-Supervised Neural Text Classification (WeSTClass) with two stages. The first stage is to use seed information to obtain the pseudo-labelled data for model pre-training\footnote{To note, this pre-training is different to the pre-training in transfer learning.}. In WeSTClass, the seed information can be adaptive to either the raw class surface names or human-provided seed words. Instead of matching unlabelled instances with the classes, they model the class semantics as a spherical distribution and generate pseudo instances of each class where class-related terms are diversified and concatenated to form a pseudo instance. After the model is pre-trained on the pseudo instances, the second stage refines the model by training it on the real unlabelled data; the predictions by the model on the unlabelled data are normalised and used as the ground truths to tune the model itself, known as the self-training stage. The results show that the self-training stage helps to substantially improve the final performance.

In another work by~\cite{mekala2020contextualized}, they proposed a contextualised weakly-supervised (ConWea) approach for text classification. ConWea is a seed-word driven supervision method that converts user-provided seed words for each class into contextualised seed words (i.e., multiple interpretations of the same word) using their contextualised representations from pre-trained contextualised word embedding models such as ELMo and BERT. The contextualised seed words are used to create a contextualised corpus where the original seed words occurring in the unlabelled instances are replaced with their contextualised seed words. To refine and expand the contextualised seed words, a classifier is trained on the contextualised corpus, which works with a comparative ranking component to help filter out ambiguous words and add highly class-indicative keywords in an iterative manner. The training process stops after convergence has occurred, where no more ambiguous or class-indicative keywords remain. The final classifier is then used to make predictions for the test data of the target task.

Regarding extremely weak supervision that does not involve using human-provided seed words to enrich the classes, the work by~\cite{meng-etal-2020-text} proposed to use label names only for text classification (LOTClass). In this work, the pseudo-labelled data is obtained by a token-level matching between the unlabelled instances and the classes. Before matching, a class is represented by its vocabulary (i.e., expansion) instead of its surface name and the vocabulary consists of the top-50 predicted words at the positions of the class's surface name occurrences in the unlabelled instances by leveraging the masked language modelling (MLM) task of BERT. To expand the class effectively in this way, the exact class name should occur within the unlabelled instances, which is also essential to the subsequent matching step. In terms of matching, in a  similar way to MLM, they proposed a masked category prediction (MCP) head on BERT where a token is assigned to a class if its top predicted words highly overlap with the class's vocabulary. Here, the classification model comprising the MCP head and BERT is trained to predict the implied classes of the class-assigned tokens with them masked. In a similar way to~\cite{meng2018weakly}, the model is finally self-trained on real unlabelled data, leading to further improved performance.

X-Class is another extremely weak supervision approach to text classification, proposed by~\cite{wang2021x}. They used class names only for text classification by building class-oriented instance representations first and then using the Gaussian Mixture Model (GMM) clustering algorithm~\cite{duda1973pattern} to obtain the pseudo-labelled data. X-Class can be broken down into three components: class-oriented instance representation estimation, instance-class alignment through clustering and classifier training based on confident classes. The first component obtains class-oriented instance representations by applying a tailed attention mechanism on the representations of instances and the class names based on pre-trained language models (specifically BERT). The second refers to pseudo-labelling where instances are aligned with the classes with confidence scores by GMM clustering. Finally, BERT is trained on the pseudo-labelled instances with high confidence scores.

\section{Related datasets and evaluation metrics}
\label{sec:rds-ems}

In the domain of crisis message categorisation on social media, many benchmarking datasets have been created for research purposes. These datasets are important resources for testing the effectiveness of the methods proposed in this research. Referring back to the process of real-world crisis response introduced in Section~\ref{sec:intro-res-context}, the datasets are reviewed by organising them into two categories: informativeness and information types. The informativeness datasets are used for the initial filtering, which is a binary classification problem aiming at finding crisis messages that are informative to users' aid needs. The information types datasets are used for further classification, aiming to categorise the informative messages into specific aid types known as information types such as ``search and rescue'', ``volunteer'', ``donations'' etc. One goal of this research is to investigate to what extent the proposed methods can also generalise to text classification tasks that are not specific to crisis messages. Therefore, a list of benchmarking datasets relating to other domains, such as emotion classification and topic categorisation, are used for testing the generalisation ability of the proposed methods. Based on the evaluation of previous works that use these datasets, most use the standard classification metrics such as Accuracy or F1 and a few use a tailored set of metrics for performance reporting and comparison. Hence, the tailored metrics associated with the datasets are reviewed in this section as well.

\subsection{Informativeness}
\label{sec:rds-ems-infos}

This section describes two crisis datasets commonly used for informativeness identification in domain adaptation: \textbf{nepal\_queensland}~\cite{li2018disaster} and \textbf{CrisisT6}~\cite{olteanu2014crisislex}. Table~\ref{tab:basic-info-informativeness-datasets} presents basic information about the two datasets.

\begin{table*}[!h]
\centering
\small
\begin{tabular}{llllll}
\toprule
Dataset           & No. of events & No. of classes & Class dist. & Data size & Task type \\
\midrule
nepal\_queensland & 2             & 2              & balanced           & $\sim$21k  & single-label, binary\\
CrisisT6          & 6             & 2              & balanced           & $\sim$60k & single-label, binary \\
\bottomrule
\end{tabular}
\caption{Basic information of informativeness datasets}
\label{tab:basic-info-informativeness-datasets}
\end{table*}


\begin{figure*}[!h]
     \centering
     \begin{subfigure}[b]{0.45\textwidth}
         \centering
         \includegraphics[width=\textwidth]{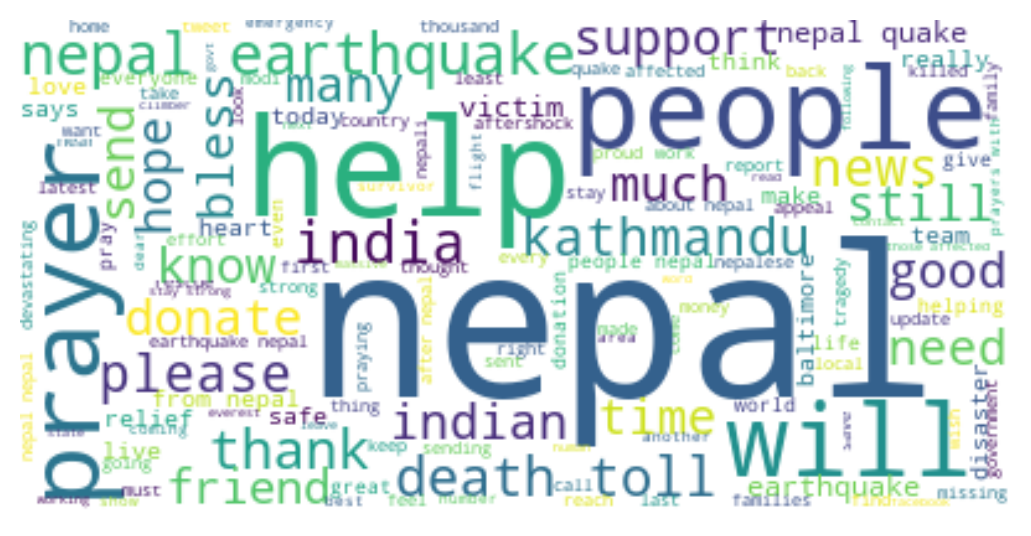}
         \caption{NE: Nepal Earthquake}
         \label{fig:nepal-earthquake}
     \end{subfigure}
     \begin{subfigure}[b]{0.45\textwidth}
         \centering
         \includegraphics[width=\textwidth]{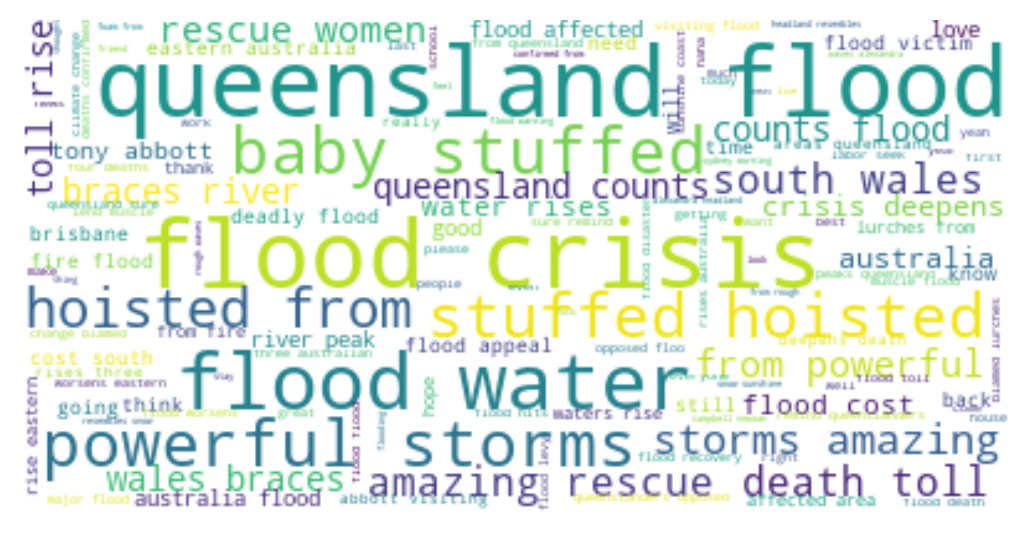}
         \caption{QQF: Queensland Floods}
         \label{fig:queensland-floods}
     \end{subfigure}
        \caption{Word clouds of two events from nepal\_queensland}
        \label{fig:nepal-queensland}
\end{figure*}

\begin{figure*}[!h]
     \centering
     \begin{subfigure}[b]{0.32\textwidth}
         \centering
         \includegraphics[width=\textwidth]{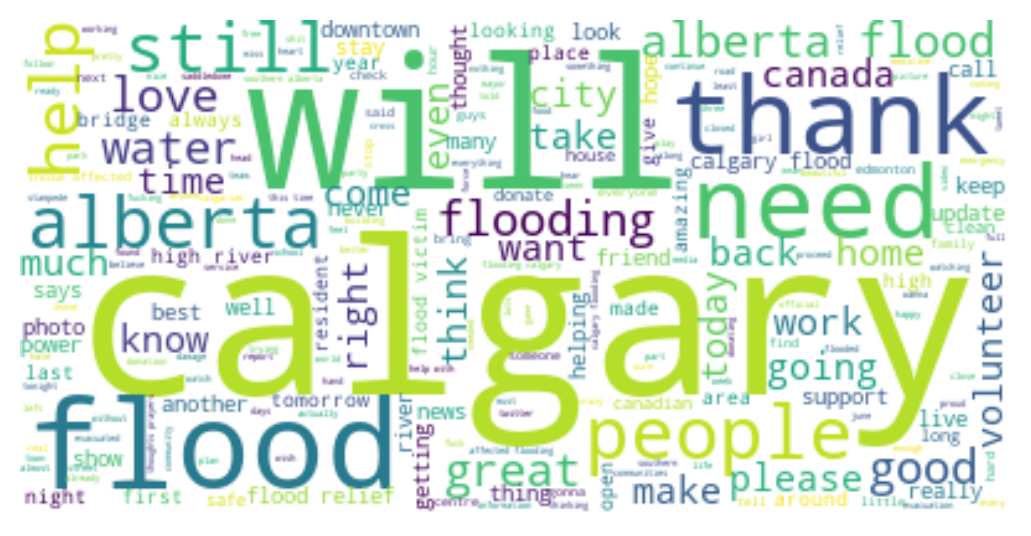}
         \caption{AF: Albert Floods}
     \end{subfigure}
     \begin{subfigure}[b]{0.32\textwidth}
         \centering
         \includegraphics[width=\textwidth]{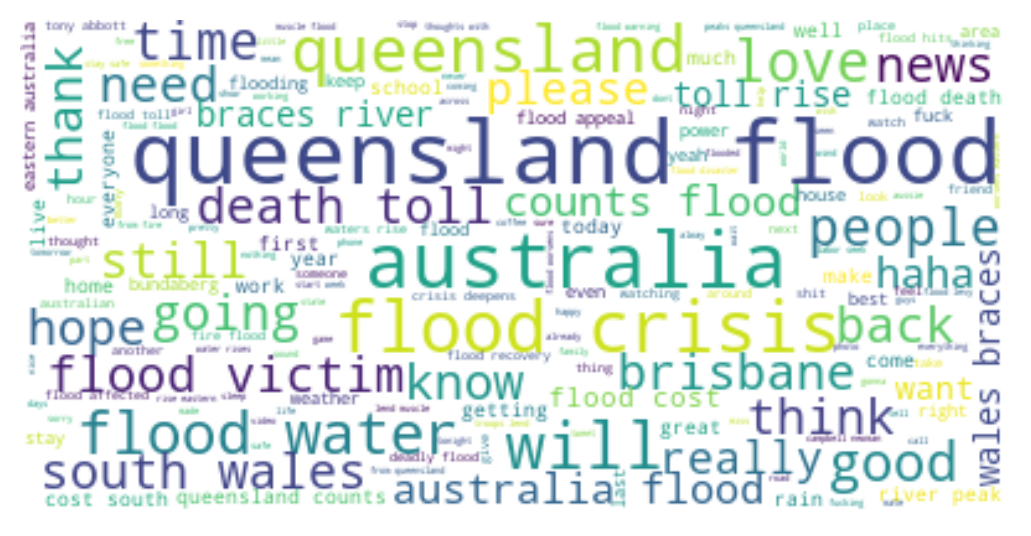}
         \caption{QF: Queensland Floods}
     \end{subfigure}
          \begin{subfigure}[b]{0.32\textwidth}
         \centering
         \includegraphics[width=\textwidth]{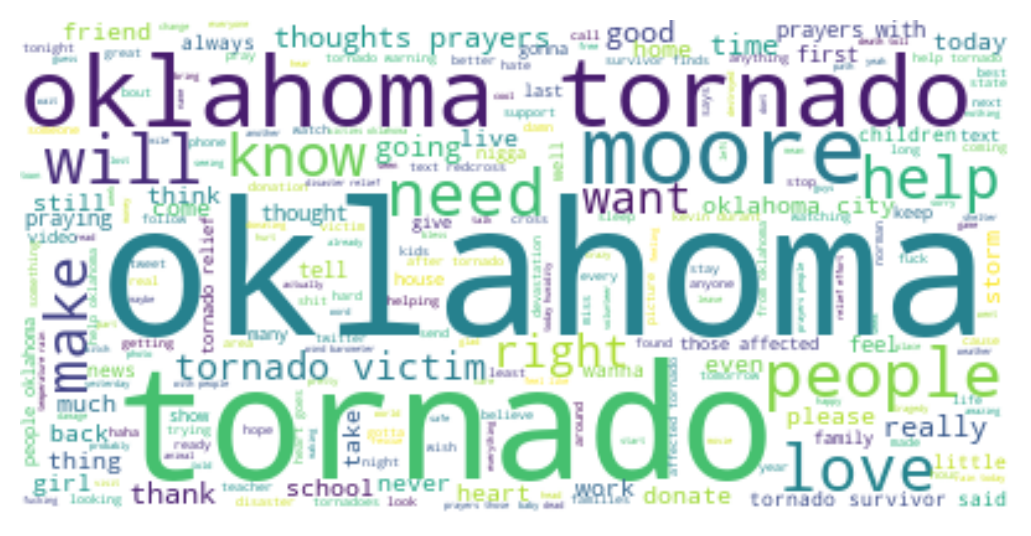}
         \caption{OT: Oklahoma Tornado}
     \end{subfigure}

         \begin{subfigure}[b]{0.32\textwidth}
         \centering
         \includegraphics[width=\textwidth]{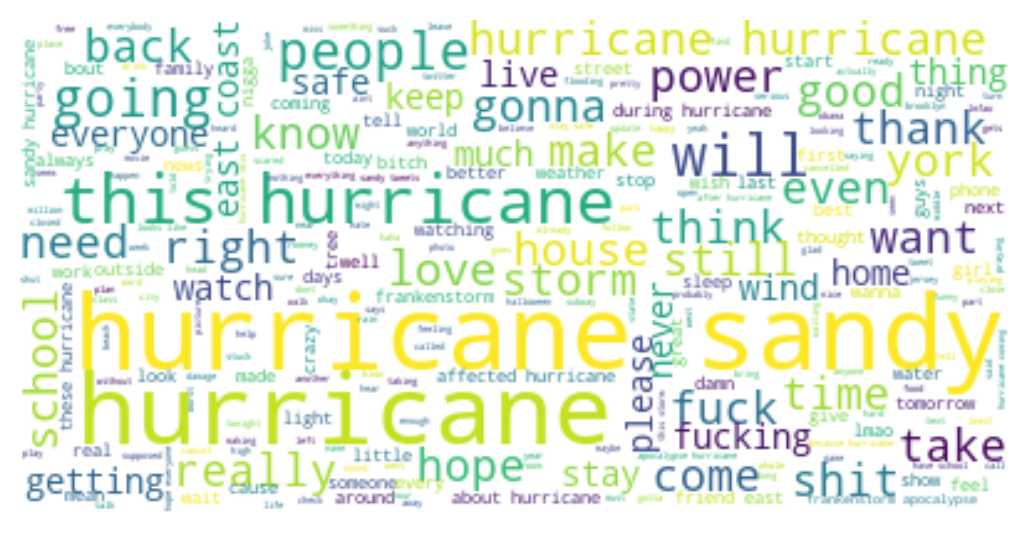}
         \caption{SH: Sandy Hurricane}
     \end{subfigure}
     \begin{subfigure}[b]{0.32\textwidth}
         \centering
         \includegraphics[width=\textwidth]{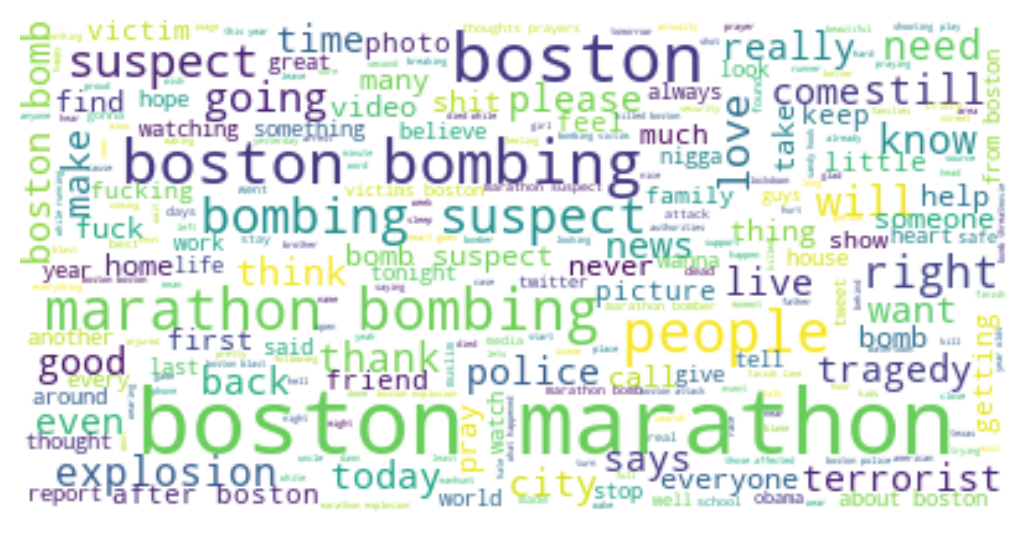}
         \caption{BB: Boston Bombings}
     \end{subfigure}
          \begin{subfigure}[b]{0.32\textwidth}
         \centering
         \includegraphics[width=\textwidth]{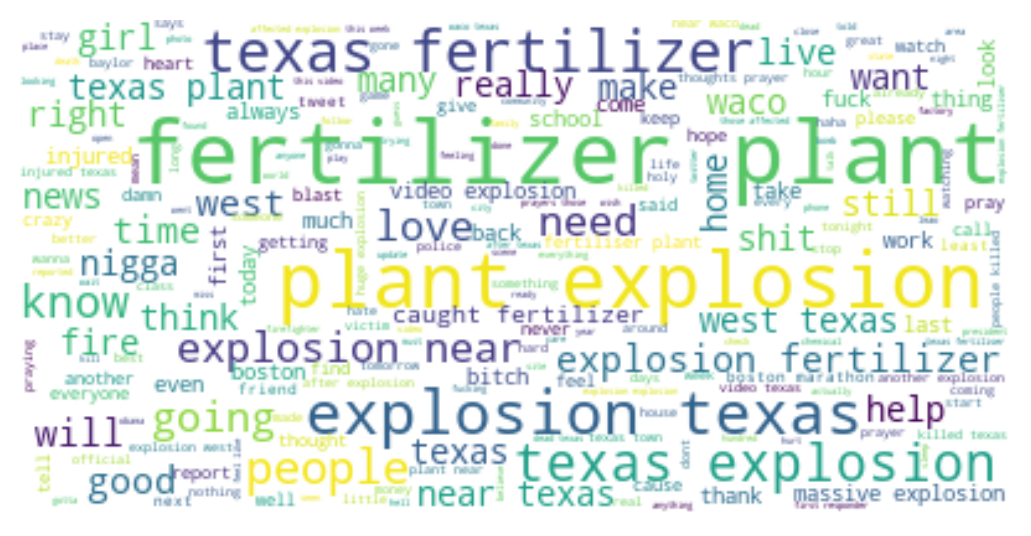}
         \caption{WTE: West Texas Explosion}
     \end{subfigure}
        \caption{Word clouds of six events from CrisisT6}
        \label{fig:wcs-t6-crisist6}
\end{figure*}

\begin{itemize}
    \item \textbf{nepal\_queensland} is a real-world crisis Twitter dataset collected during the 2015 Nepal earthquake (NE)\footnote{\url{https://en.wikipedia.org/wiki/April_2015_Nepal_earthquake}} and the 2013 Queensland floods (QQF)\footnote{\url{https://en.wikipedia.org/wiki/Cyclone_Oswald}}, originally proposed in~\cite{li2018disaster} for crisis domain adaptation. Figure~\ref{fig:nepal-queensland} depicts the word clouds of word distributions for the two events. This dataset consists of approximately 21,000 annotated tweets equally distributed across the two events, sampled from millions of tweets crawled through the Twitter API\footnote{\url{https://developer.twitter.com/en/docs/twitter-api}} using event-specific keywords and hashtags. In this dataset, each tweet is annotated by one of two classes (which are well balanced): \textit{relevant} if the tweet contains aid-related information such as donations or food requests and \textit{not\_relevant} if not.
    \item \textbf{CrisisT6} was originally released by~\cite{olteanu2014crisislex}. It is a collection of approximately 60,000 annotated tweets posted during six crisis events with approximately 10,000 per event. The word clouds of six events are presented in Figure~\ref{fig:wcs-t6-crisist6}. They are real-world crisis events occurred between 2012 and 2013, which can be summarised in three broad categories: Flood (AF: Albert Floods 2013\footnote{\url{https://en.wikipedia.org/wiki/2013_Alberta_floods}} and QF: Queensland Floods 2013\footnote{\url{https://en.wikipedia.org/wiki/Cyclone_Oswald}}), Hurricane (OT: Oklahoma Tornado 2013\footnote{\url{https://en.wikipedia.org/wiki/2013_Moore_tornado}}) and SH: Sandy Hurricane 2013\footnote{\url{https://en.wikipedia.org/wiki/Hurricane_Sandy}} and Explosion (BB: Boston Bombings 2012\footnote{\url{https://en.wikipedia.org/wiki/Boston_Marathon_bombing}} and WTE: West Texas Explosion 2013\footnote{\url{https://en.wikipedia.org/wiki/West_Fertilizer_Company_explosion}}). Similarly, the tweets are sampled before annotation from around 10 million tweets collected through the Twitter API using keywords and geographical regions or coordinates. Each tweet in this dataset is annotated by one of two classes according to relatedness (a similar concept to informativeness): \textit{on-topic} if the tweet is related to the event and \textit{off-topic} if not and the tweets are balanced in terms of class distribution. 
\end{itemize}

\subsection{Information types}
\label{sec:rds-ems-infotypes}

This section introduces three crisis datasets for information types classification: \textbf{TREC-IS}~\cite{mccreadie2019trec,mccreadie2020trec,buntain2021incident}, \textbf{HumAID}~\cite{alam2021humaid} and \textbf{Situation}~\cite{strassel2017situational,mayhew2018university}. These are suitable resources for the studies of crisis domain adaptation, few-shot and zero-shot learning. Table~\ref{tab:basic-info-infotypes-datasets} presents the basic information of these corpora, followed by a detailed description below.

\textbf{TREC-IS} is the dataset that has arisen as part of the challenges posed by the TREC Incident Streams (IS) track~\cite{mccreadie2019trec,mccreadie2020trec}. The IS track is an annual task run as part of the Text REtrieval Conference (TREC)\footnote{\url{https://trec.nist.gov/}}, aiming to encourage more mature social media-based emergency response technology. It was launched in 2018 and has run in many editions since then\footnote{The track ended in 2021 and the editions include 2018, 2019A, 2019B, 2020A, 2020B, 2021A and 2021B.}. In a each edition after the first, the data from past editions was used as the training data and a new unlabelled corpus was used as the test data (and the training data for following edition). Hence, the TREC-IS dataset are accumulated, and are divided into subsets based on the track's editions. By the time the track ended, it had accumulated four subsets of editions with annotations: 2018, 2019A, 2019B, and 2020A. Table~\ref{tab:trecis-stats-subsets} shows the important statistics of the four subsets.

\begin{table*}[h]
\centering
\begin{tabular}{llll}
\toprule
Subsets & \multicolumn{1}{l}{Data size} &No. events& Avg. length   \\
\midrule
2018    & 17581                  &15   & 35                 \\
2019A   & 7098                    &6 & 37                   \\
2019B   & 9122                     &6  & 48             \\
2020A   & 6658                     &15  & 42         \\
\midrule
Total   & 40459                  &42   & 40         \\
\bottomrule
\end{tabular}
\caption{Statistics of TREC-IS sub-datasets where \emph{No. events} means the number of events for a set. \emph{Avg. length} refers to the average length of instances.}
\label{tab:trecis-stats-subsets}
\end{table*}

It is notable that each subset consists of tweets relating to different crisis events and that the events do not overlap between the subsets, which makes the dataset a suitable resource for many-to-many crisis domain adaptation. Combining all the subsets, the TREC-IS dataset consists of approximately 40,000 annotated tweets across 42 crisis events. The events occurred between 2012 and 2020 including the COVID-19 pandemic~\cite{buntain2021incident}. For the full details of these events, the reader is referred to the home page of the IS track\footnote{\url{https://www.dcs.gla.ac.uk/~richardm/TREC_IS/}}. Regarding annotation, the dataset is annotated by two types of classes: ``information types'' and ``priority levels''.

For information types, each tweet is assigned with one or more information types (thus this is a multi-label classification task). There are 25 information types defined in this dataset where 6 are defined as ``actionable'' and the remaining 19 as ``non-actionable''. Actionable types are related to more urgent aid needs such as \textit{a request for search and rescue} or \textit{asking for a particular service or physical good}, etc. Non-actionable tweets tend to be less urgent including \textit{asking people to volunteer to help the
response effort}, \textit{reporting first party observation}, etc. For a complete description of the information types, the reader is referred to Appendix~\ref{appendix:trecis-its}. In addition to information types, each tweet is also labelled with one of four priority levels indicating the urgency level of the tweet: \textit{critical}, \textit{high}, \textit{medium} or \textit{low}. As indicated by the annotation in this dataset, participants in the IS track are asked to develop systems that are capable of categorising the information types and estimating the priority of tweets in a supplied test set.

Regarding evaluation metrics, given that the tweets are annotated by both information types and priority levels, the IS track proposed a set of tailored metrics to uniformly evaluate the performance of the participating systems on the tasks of information type classification and priority estimation. The metrics can be broadly divided into four categories: \textbf{Ranking, Alerting Worth, Information Feed Categorisation} \textbf{Prioritisation}, described as follows.

\begin{itemize}
    \item \textbf{Ranking} (range $0$ to $1$): In this category, NDCG~\cite{jarvelin2002cumulated} is the priority-centric metric used to evaluate the quality of submitted test tweets ranked by their priority scores. By default, it measures the top 100 submitted tweets per event and reports the average across all events as the final score. A high NDCG score implies that the system has achieved a good quality of priority-based ranking. It is noted that this metric was not used for the 2019 editions, but was later introduced from 2020.
    \item \textbf{Alerting Worth} (range $-1$ to $1$): This is inspired by the alerting use case in a real-world emergency response system (i.e, identifying aid needs that require urgency). It not only measures the effectiveness of a system in generating true alerts but also penalises the system in generating consecutive false alerts that would make end users lose trust in the system. Two components of Alerting Worth are AW-HC and AW-A, which measure the effectiveness of true alerts within the scope of tweets annotated to be critical or high, and within the scope of all priority-level tweets respectively. 
    \item \textbf{Information Feed Categorisation} (range $0$ to $1$): This is used to evaluate the aspect of information type classification performance by a system. To better reflect a system's utility to emergency response officers, it consists of three specific metrics, Actionable F1, All F1, and Accuracy, denoted by CF1-H, CF1-A, and Cacc respectively. CF1-A macro-averages the F1 scores across all information types, while CF1-H macro-averages the F1 scores only across the 6 actionable types as presented in Table~\ref{tab:actionable-versus-non}. The two metrics indicate the performance of information type categorisation by only taking the target class per information type into account. Cacc computes the categorisation accuracy micro-averaged across information types.
    \item \textbf{Prioritisation} (range $0$ to $1$): This is applied to measure the performance of priority level predictions, consisting of two specific metrics: PErr-H and PErr-A. In IS 2019, PErr-H and PErr-A were used to measure the root mean squared error (RMSE) between the predicted priority scores and actual priority levels for actionable and all information types respectively. However, since IS 2020, the priority predictions were changed to categorical priority levels instead of numeric scores as in 2019. The metrics were thus changed to Actionable F1 (PErr-H) and All F1 (PErr-A) respectively. Both are computed by averaging the macro-F1 scores on priority label predictions per information type. Unlike PErr-A, which averages the F1 scores for all information types, PErr-H averages the F1 scores for actionable types only.
\end{itemize}

As a general reference, the metrics in the category of information feed categorisation indicate the effectiveness of the information type classification task while the rest evaluate the effectiveness of the priority estimation task. Considering the characteristics of this dataset, this research uses it as the resource for the study of many-to-many crisis domain adaptation (Chapter~\ref{ch:adaptation}) where the subsets of previous years are used for model training and the subsets of late years are used for testing. To align with the IS track, the tailored metrics are used to compare the proposed methods in this research with baselines.

\begin{figure}[!h]
    \centering
    \includegraphics[scale=0.6]{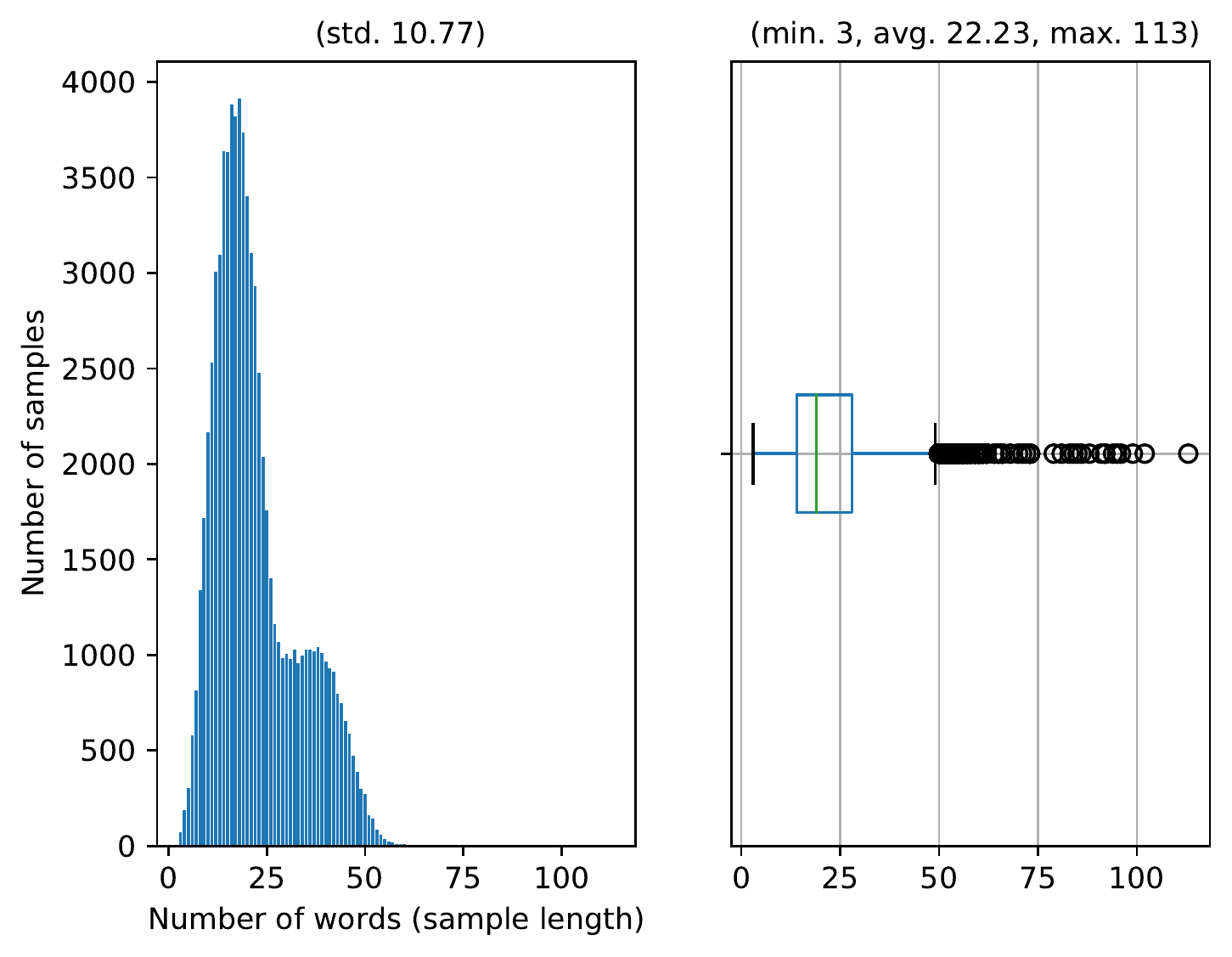}
    \caption{Word length statistics of HumAID}
    \label{fig:humaid-wl}
\end{figure}

\begin{figure*}[!h]
     \centering
     \begin{subfigure}[b]{0.24\textwidth}
         \centering
         \includegraphics[width=\textwidth]{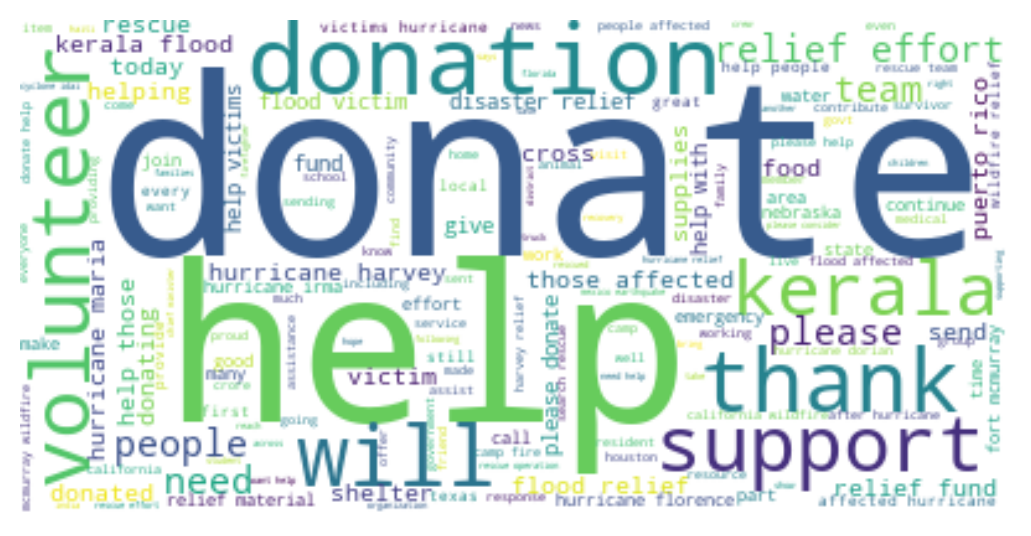}
         \caption{rescue volunteering or donation effort}
         \label{fig:humaid-rvde}
     \end{subfigure}
     \begin{subfigure}[b]{0.24\textwidth}
         \centering
         \includegraphics[width=\textwidth]{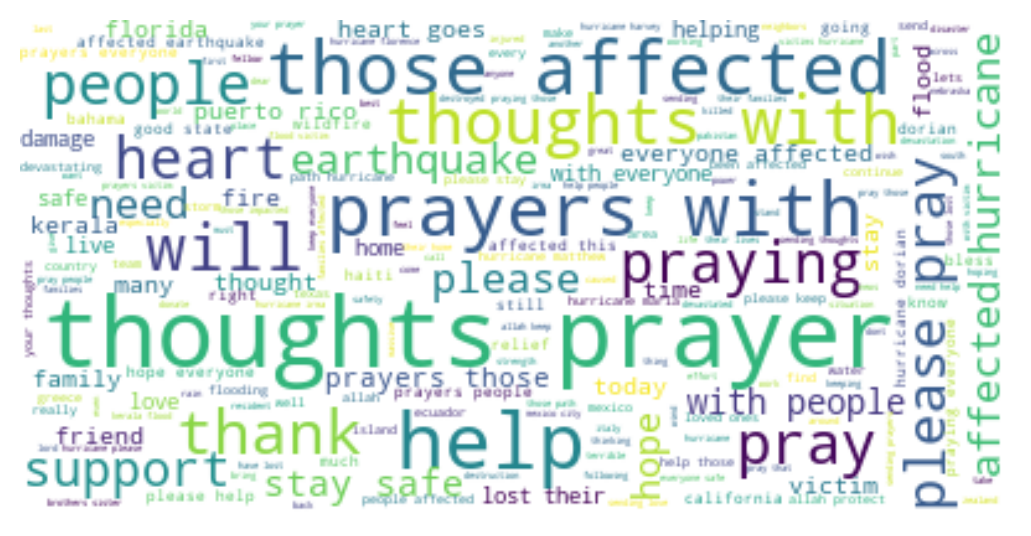}
         \caption{sympathy and support}
         \label{fig:humaid-ss}
     \end{subfigure}
        \begin{subfigure}[b]{0.24\textwidth}
         \centering
         \includegraphics[width=\textwidth]{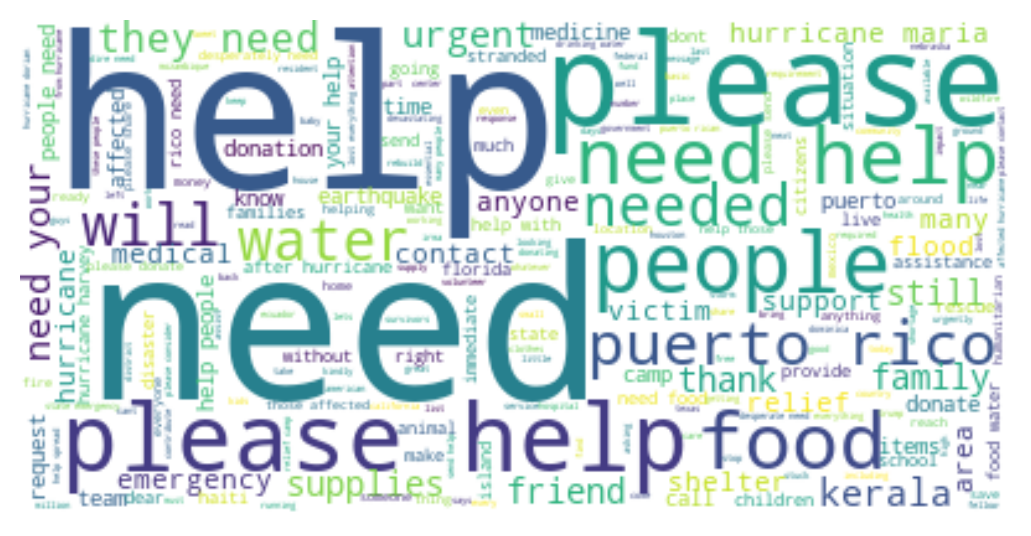}
         \caption{requests or urgent needs}
         \label{fig:humaid-run}
     \end{subfigure}
     \begin{subfigure}[b]{0.24\textwidth}
         \centering
         \includegraphics[width=\textwidth]{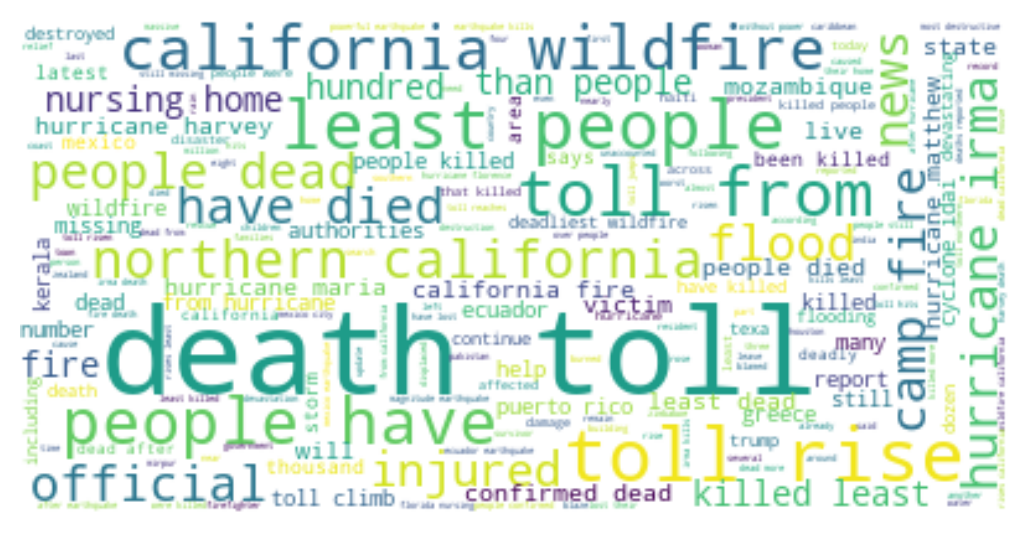}
         \caption{injured or dead people}
         \label{fig:humaid-idp}
     \end{subfigure}

          \begin{subfigure}[b]{0.24\textwidth}
         \centering
         \includegraphics[width=\textwidth]{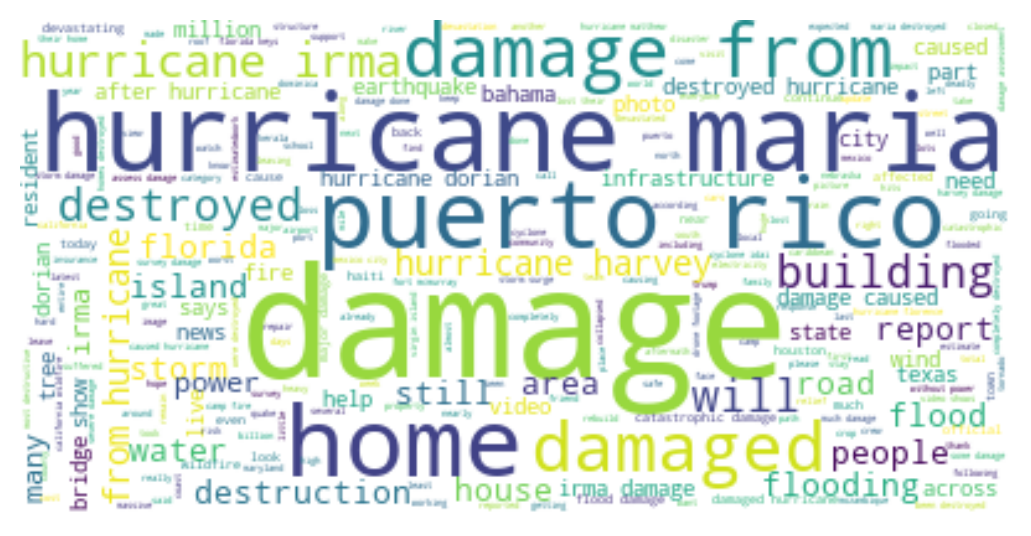}
         \caption{infrastructure and utility damage}
         \label{fig:humaid-iud}
     \end{subfigure}
     \begin{subfigure}[b]{0.24\textwidth}
         \centering
         \includegraphics[width=\textwidth]{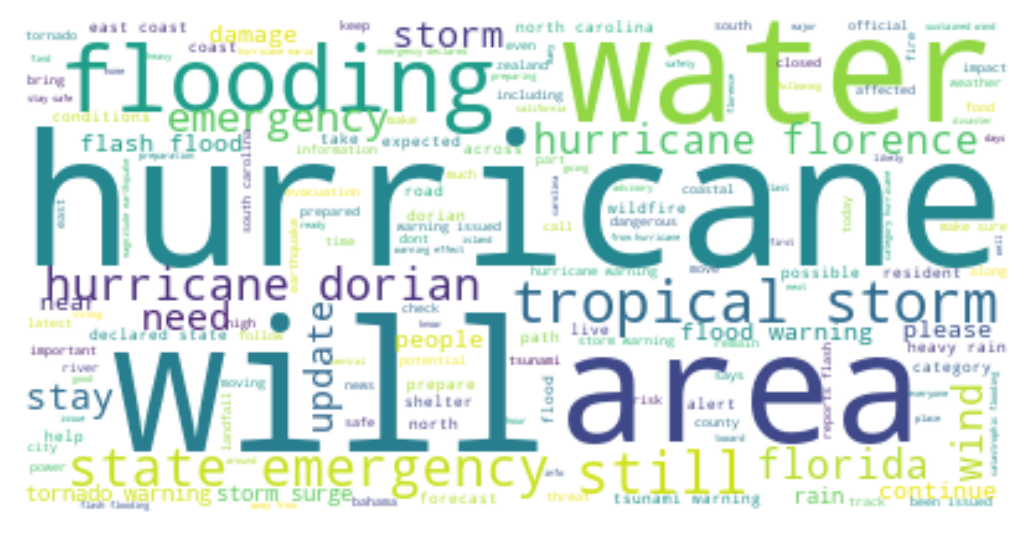}
         \caption{caution and advice \:\:\:\:\:}
         \label{fig:humaid-ca}
     \end{subfigure}
        \begin{subfigure}[b]{0.24\textwidth}
         \centering
         \includegraphics[width=\textwidth]{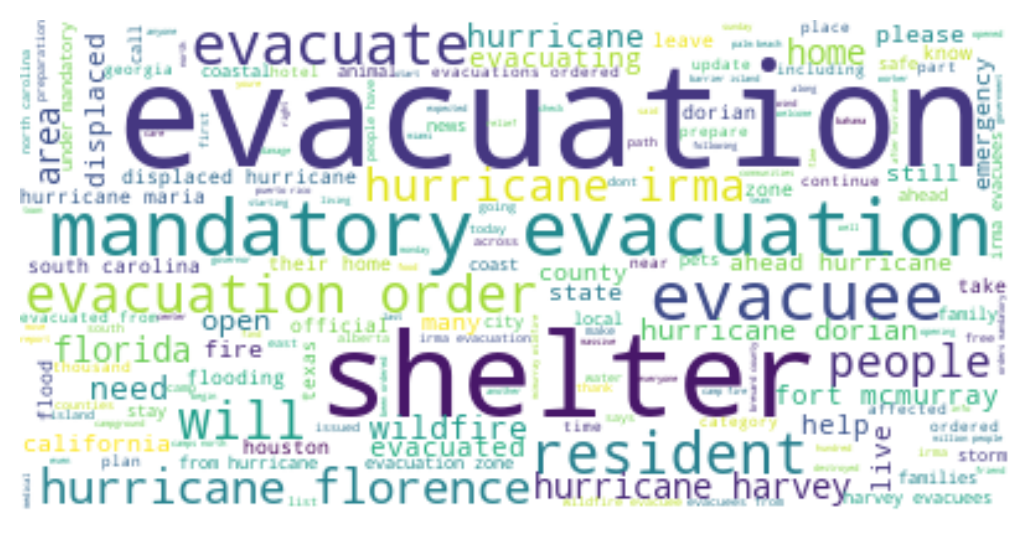}
         \caption{displaced people and evacuations}
         \label{fig:humaid-dpe}
     \end{subfigure}
     \begin{subfigure}[b]{0.24\textwidth}
         \centering
         \includegraphics[width=\textwidth]{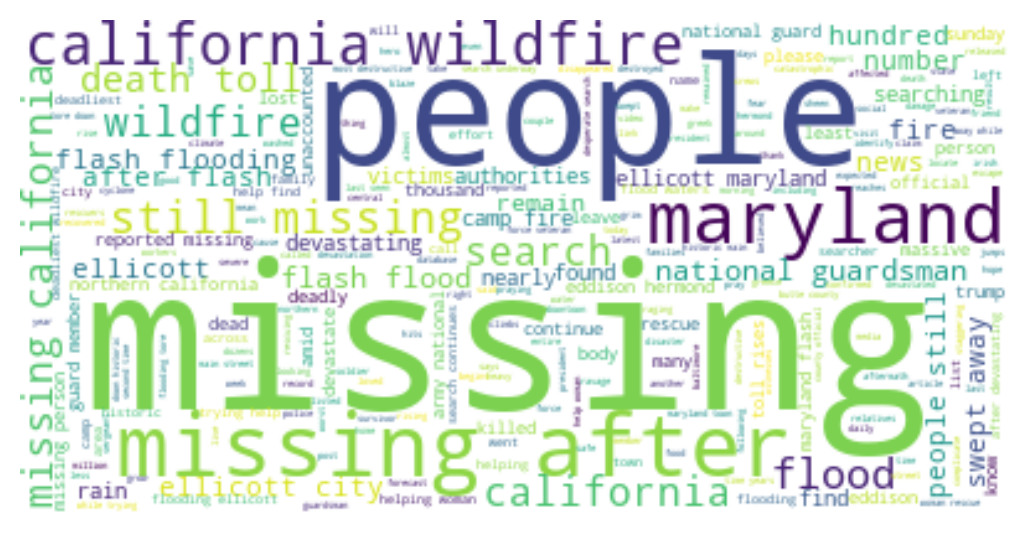}
         \caption{missing or found people}
         \label{fig:humaid-mfp}
     \end{subfigure}
        \caption{Word clouds of 8 informative humanitarian categories from HumAID}
        \label{fig:humaid-wc}
\end{figure*}

\textbf{HumAID}~\cite{alam2021humaid} is another dataset suitable for information type categorisation. This dataset consists of English tweets collected from 19 real-world disaster events. The events happened from 2016 to 2019, and various event types including wildfires, floods, hurricanes and so on\footnote{\url{https://crisisnlp.qcri.org/humaid_dataset}}. In contrast to TREC-IS, the dataset consists of approximately 77,000 tweets annotated by information type with splits of training, development and test set, described as ``humanitarian categories'' in the associated paper~\cite{alam2021humaid}. The data in Figure~\ref{fig:humaid-wl} illustrates the distribution of word lengths for all examples within the dataset. It is evident from the graph that a majority of the text samples are brief, containing words between 10 and 50 words in length. In addition, each instance of this dataset is annotated by one of 11 humanitarian categories (and thus represents a single-label, multi-class classification problem)\footnote{The complete categories are \textit{rescue volunteering or donation effort},\textit{sympathy and support}, \textit{requests or urgent needs}, \textit{infrastructure and utility damage}, \textit{injured or dead people}, \textit{caution and advice}, \textit{displaced people and evacuations}, \textit{missing or found people}, \textit{don't know can't judge}, \textit{not humanitarian}, \textit{other relevant information}}. The categories include highly urgent aid needs such as \textit{displaced people and evacuations}, \textit{missing or found people} as well as less urgent aid needs such as \textit{sympathy and support}, \textit{caution and advice}. Among the 11 humanitarian categories, there are three non-informative categories: \textit{don't know can't judge}, \textit{not humanitarian} and \textit{other relevant information}. In order to use this as a resource for information type categorisation after the informativeness identification, this research uses a modified version of the dataset where the instances of the three non-informative categories are removed. Figure~\ref{fig:humaid-wc} shows the word clouds of the 8 informative humanitarian categories of all examples within the dataset. This version only contains the informative categories whose surface names can effectively represent the meanings of the corresponding aid needs, which is well suited to the proposed methods of this research for crisis few-shot (Chapter~\ref{ch:sta-isa}) or zero-shot learning (Chapter~\ref{ch:pzsc}) as opposed to the version with the non-informative categories. 

\textbf{Situation}~\cite{strassel2017situational,mayhew2018university} is a multi-lingual dataset for detecting aid needs from crisis tweets written in a variety of natural languages. Although this dataset does not specify the crisis events from which the tweets are, the tweets are from a mixture of several crisis events. The dataset was originally designed for low-resource situation detection where annotated data is unavailable, which makes it a good resource to evaluate crisis zero-shot learning. Unlike TREC-IS and HumAID, this dataset is annotated by a more general set of information types, known as ``situation types''~\cite{strassel2017situational}. There are 11 situation types defined in this dataset: \textit{shelter}, \textit{search}, \textit{water}, \textit{utilities}, \textit{terrorism}, \textit{evacuation}, \textit{regime change}, \textit{food}, \textit{medical}, \textit{infrastructure}, and \textit{crime violence}. By their surface names, it can be seen that some of them represent a general description of an event such as \textit{crime violence} and \textit{terrorism}, which are different to the specific aid needs in TREC-IS and HumAID. In a similar way to TREC-IS, each instance of this dataset is annotated by one or more of the 11 situation types (i.e., representing a multi-label classification task). Given the scope of this research, the English version of the dataset comprising approximately 6,000 annotated instances in splits of training, development and test set released by~\cite{mayhew2018university} is used. In this research, this dataset is used for zero-shot learning, which will be introduced in Chapter~\ref{ch:pzsc}.


\begin{table*}[]
\centering
\begin{tabular}{llllll}
\toprule
Dataset   & No. of events               & No. of classes & Class dist. & Data size & Task type                 \\
\midrule
TREC-IS    & 42                          & 25             & imbalanced         & $\sim$40k & multi-label, multi-class  \\
HumAID    & 19                          & 11             & imbalanced         & $\sim$77k & single-label, multi-class \\
Situation & N/A & 11             & imbalanced         & $\sim$6k  & multi-label, multi-class  \\
\bottomrule
\end{tabular}
\caption{Basic information of information types datasets}
\label{tab:basic-info-infotypes-datasets}
\end{table*}

\subsection{Beyond crisis}
\label{sec:rds-beyond-crisis}

In crisis zero-shot and few-shot learning, the proposed methods are also evaluated on domains other than the crisis domain. Although the methods are proposed with the motivation of crisis message categorisation, the extension to other domains indicates that this research emphasises the generalisation capability of the methods across different domains. For this purpose, a list of benchmarking datasets beyond crises are used in this research. Table~\ref{tab:rds-beyond-crisis} presents the basic information of the datasets. As this shows, these datasets are varied in multiple aspects. For example, they are from different domains including question type identification, emotion detection, sentiment analysis and news topic categorisation. To be specific, this research uses \textbf{Topic}, \textbf{UnifyEmotion} and \textbf{Emotion} as extended datasets in the study of crisis zero-shot learning and \textbf{SST-2}, \textbf{TREC}, \textbf{Emotion}, \textbf{AgNews}, \textbf{TweetSentiment} in the study of crisis few-shot learning. The datasets are detailed as follows.

\begin{table*}[!h]
\footnotesize
\centering
\begin{tabular}{llrll}
\toprule
Dataset          & Domain                            & \multicolumn{1}{l}{No. of classes} & Class distribution & Data size                         \\
\midrule
TREC             & question types identification     & 6                                  & imbalanced         &$\sim$6k  \\
Emotion          & tweets emotion detection          & 6                                  & imbalanced         & $\sim$20k \\
UnifyEmotion     & mixed messages emotion detection  & 10                                  & imbalanced         & $\sim$48k                         \\
TwitterSentiment & tweets sentiment analysis         & 3                                  & imbalanced         & $\sim$60k                         \\
SST-2            & short messages sentiment analysis & 2                                  & balanced           & $\sim$70k                         \\
Topic            & news topic categorisation         & 10                                 & balanced           & $\sim$1.4m                        \\
AgNews           & news topic categorisation         & 4                                  & balanced           & $\sim$127k              \\
\bottomrule
\end{tabular}
\caption{Datasets beyond crisis domain}
\label{tab:rds-beyond-crisis}
\end{table*}
 
\begin{itemize}
    \item \textbf{TREC}~\cite{li-roth-2002-learning} is a dataset for question type classification. The instances of this dataset comprises approximately 6,000 questions in short sentences. Approximately 4,500 questions are collected by Hovy et al. 2001~\cite{hovy-etal-2001-toward} and the rest are from TREC 8, TREC 9 and TREC 10, which are annual task runs similar to TREC-IS proposed by the Text REtrieval Conference (TREC). The questions are originally annotated with 6 coarse class labels and 50 fine class labels. This research uses the version with 6 coarse class labels: \textit{Abbreviation}, \textit{Description}, \textit{Entities}, \textit{Human beings}, \textit{Locations} and \textit{Numeric values} and uses the original splits of this dataset.
    \item \textbf{Emotion}~\cite{saravia-etal-2018-carer} is a dataset for emotion detection comprising English Twitter messages that are pre-processed by removing hashtags and mentions. There are approximately 20,000 messages annotated with six emotion types: sadness, \textit{joy}, \textit{love}, \textit{anger}, \textit{fear} and \textit{surprise}. This research uses the original splits of the dataset published by~\cite{saravia-etal-2018-carer}.
    \item \textbf{UnifyEmotion }~\cite{klinger2018analysis} is another dataset for emotion detection. This dataset is constructed by combining multiple public emotion datasets from multiple domains including tweets, fairytales, artificial sentences, etc. Despite similar annotation, this dataset does not contain instances that overlap with Emotion. Instances are originally annotated by 9 emotion types plus a \textit{none} label if no emotion applies: \textit{surprise}, \textit{guilt}, \textit{fear}, \textit{anger}, \textit{shame}, \textit{love}, \textit{disgust} and \textit{sadness} and \textit{joy}. This research uses the version of the dataset from~\cite{yin2019benchmarking} without the ``none'' instances and uses the splits of the version released by~\cite{yin2019benchmarking}
    \item \textbf{TwitterSentiment }~\cite{rosenthal-etal-2017-semeval} is a sentiment classification dataset that consists of cleaned tweets, released by SemEval-2017 Task 4 Sentiment Analysis in Twitter. It consists of approximately 60k instances annotated with three class labels: \textit{positive}, \textit{neutral} and \textit{negative}. This research uses the original splits of the dataset published by~\cite{rosenthal-etal-2017-semeval}.
    \item \textbf{SST-2}~\cite{socher-etal-2013-recursive} is a benchmarking dataset for binary sentiment classification. It consists of approximately 70,000 short movie reviews annotated with \textit{positive} and \textit{negative} labels. This research uses the original splits of the dataset published by~\cite{socher-etal-2013-recursive}.
    \item \textbf{Topic} is a large-scale dataset for news topic categorisation. It refers to the Yahoo dataset released by~\cite{zhang2015character} and consists of approximately 1.4 million news articles annotated by 10 news topics (well-balanced): \textit{Education} \& \textit{Reference}, \textit{Society} \& \textit{Culture}, \textit{Sports}, \textit{Entertainment} \& \textit{Music}, \textit{Politics} \& \textit{Government}, \textit{Computers} \& \textit{Internet}, \textit{Family} \& \textit{Relationships}, \textit{Science} \& \textit{Mathematics}, \textit{Health} and \textit{Business} \& \textit{Finance}. This research uses the original splits of the dataset published by~\cite{zhang2015character}.
    \item \textbf{AgNews}~\cite{Zhang2015CharacterlevelCN} is a dataset comprising approximately 1 million news summaries (short documents) collected by more than 2,000 news sources, which can be used for multiple research purposes such as clustering, classification and  information retrieval. In its version for news topic categorisation, approximately 127,000 instances are annotated with four news topics: \textit{World}, \textit{Sports}, \textit{Business} and \textit{Sci/Tech}. This research uses the original splits of this version released by ~\cite{Zhang2015CharacterlevelCN}.
\end{itemize}



\section{Conclusions}

This chapter conducted a review of the specific domain of crisis message categorisation. Based on the proposals of this research, the review is conducted in three parts under the context of low-data availability for crisis messages categorisation: crisis domain adaptation, crisis few-shot learning and crisis zero-shot learning. For domain adaptation, both target data dependent and independent approaches are introduced. These approaches serve as important inspirations for the proposed methods in this research for crisis domain adaptation. In addition, important existing state-of-the-art zero-shot and few-shot methods are reviewed, which are closely related to the proposed zero-shot and few-shot methods proposed in this research. They also serve as strong baselines against which the proposed methods are compared. Finally, a list of benchmarking datasets within the crisis domain and beyond crises are reviewed. They are important resources for testing the effectiveness of the proposed methods in this research as comparing to baselines. Next, this thesis will introduce the details of the proposed methods in crisis domain adaptation (see Chapter~\ref{ch:adaptation} and Chapter~\ref{ch:adaptation2}), crisis few-shot learning (see Chapter~\ref{ch:sta-isa}) and crisis zero-shot learning (see Chapter~\ref{ch:pzsc}) respectively.


\chapter{Many-to-many Crisis Domain Adaptation}
\label{ch:adaptation}
\section{Introduction}

In real-world crisis message categorisation, labelled data relating to an emerging new event is not readily available. It is costly in terms of both time and human labour to annotate such a labelled data set, in a situation where timeliness is essential to emergency response. When there is a lack of labelled data for a new event, it is difficult to build a computational model that can learn the categorisation task for the event directly in a fully-supervised fashion. This highlights the necessity for domain adaptation to perform crisis message categorisation. Domain adaptation, which is a key aspect of transfer learning, involves learning to perform the same tasks in different domains where a model is trained on labelled data from a source domain plus unlabelled data from a target domain in some cases (check Section~\ref{sec:cda}). When applying domain adaptation in the context of crisis message categorisation, it is  known as Crisis Domain Adaptation. In that situation, one or more past crisis events are used as the source domain and one or more new emerging events are considered to be the target domain. Assuming that the categorisation task (i.e. the set of labels to be applied) is the same, crisis domain adaptation studies the problem of building a model using the labelled data of past events and adapting the model to perform categorisation on new text instances referring to a new event.

Based on this motivation, this chapter presents a series of studies for many-to-many crisis domain adaptation. It refers to the real-world use case in crisis domain adaptation where multiple source events are used to train a model that can be applied to many target events while the latter refers to the adaptation where one or more source events are used to train a model that is applied to only one target event. It first introduces several approaches for a crisis message categorisation task comprising two sub-tasks: information types classification and priority estimation for many crisis events, which can be viewed as a many-to-many crisis domain adaptation problem (Section~\ref{sec:many-to-many-adaptation}). This includes single-task learning with machine learning and neural networks (Section~\ref{subsec:word-embeddings-data-augmentations}) as well as a multi-task learning technique with pre-trained language models (Section~\ref{subsec:mtl-plms}). The former achieves the categorisation task by learning machine learning models and neural networks separately on the two sub-tasks and the latter achieves the categorisation task by learning pre-trained language models jointly on the two sub-tasks.

The work presented in this chapter has previously been presented in published works~\cite{congcong2020cls,Wang2021b,Wang2021c}.

\section{Many-to-many adaptation}
\label{sec:many-to-many-adaptation}

\begin{figure}[!h]
    \centering
    \includegraphics[width=\linewidth]{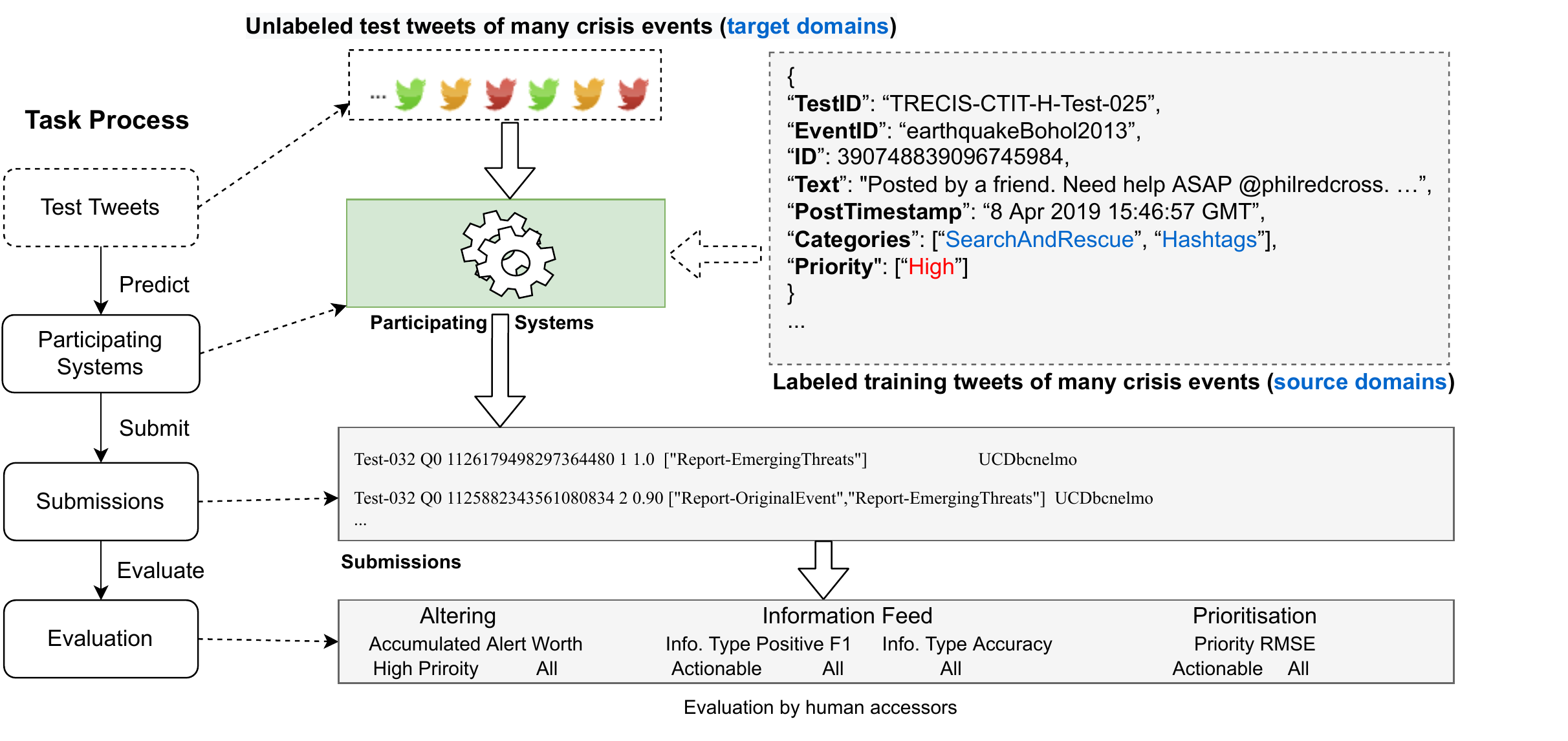}
    \caption{The TREC Incident Streams (IS) track as many-to-many adaptation}
    \label{fig:trec-is-task-process}
\end{figure}

This research conducts many-to-many crisis domain adaptation according to the timescales and tasks of the Incident Streams (IS) track. The IS track is a research initiative as part of the Text REtrieval Conference (TREC)\footnote{\url{https://trec.nist.gov/}}, which proposes a crisis message categorisation task consisting of two sub-tasks: information type classification and priority estimation. Its aim is for the community to explore more mature social media-based emergency response technology. The track ran annually from 2018 to 2021 inclusive and two editions were run in several of these years. In every edition, a training set and test set containing tweets from many crisis events are available. Figure~\ref{fig:trec-is-task-process} demonstrates the task process as many-to-many crisis domain adaptation. Overall, the TREC-IS dataset (from Section~\ref{sec:rds-ems-infotypes}) is the combination of the datasets used for these challenges. Here, unlabelled test tweets from many new events are viewed as the data of target domains and labelled training data from many past events are viewed as the data of source domains. For a participating system, they can develop different methods leveraging the source data and test their methods on the target data by submitting runs to the track. The submitted runs of participating systems are subsequently judged by human assessors using the tailored evaluation metrics as introduced in Section~\ref{sec:rds-ems-infotypes}.

Given the characteristics of this track, this research has explored a series of methods for many-to-many crisis domain adaptation by submitting runs to the track between 2019 and 2020. In the 2019 editions of this track, the methods mainly fall into the category of  single-task learning with traditional machine learning models such as Logistic Regression and simple neural network models such as LSTM. Having learnt experience from the 2019 editions, the methods evolved to a multi-task learning approach using pre-trained language models such as BERT, leading to state-of-the-art performance in the 2020 editions. This section presents the methods and major findings for crisis message categorisation in the context of many-to-many crisis domain adaptation.

\subsection{Single-task learning with machine learning and neural networks}
\label{subsec:word-embeddings-data-augmentations}

The TREC IS track ran twice in 2019 (Edition 2019-A and 2019-B). In both editions, the information type classification and priority estimation sub-tasks are learnt separately by two models. In Edition A, the models mainly consisted of traditional machine learning algorithms with hand-crafted features as input representations. After gaining some experience from the first participation, the participation at Edition B aimed to explore the potential of simple neural networks such as LSTM for the two sub-tasks. In 2019-A, the models used the subset 2018 of the TREC-IS dataset (see Table~\ref{tab:trecis-stats-subsets} in Section~\ref{sec:rds-ems-infotypes}) as the training set from which a 10\% portion is sampled as the validation set. In 2019-B, the subset 2018 is used as the training set and the subset 2019A as the validation set.


\begin{figure*}[!h]
  \centering
  \includegraphics[width=\linewidth]{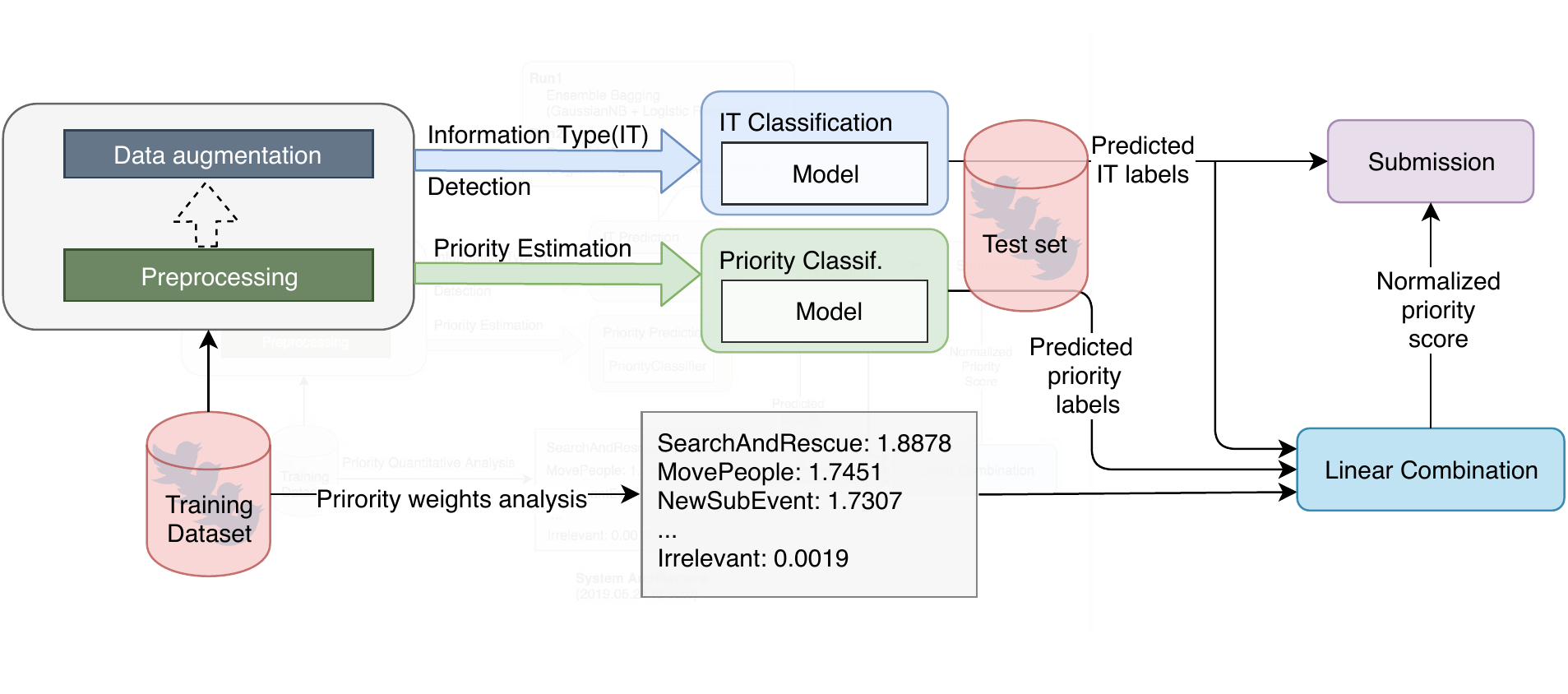}
  \caption{Single-task learning for information types classification and priority estimation}
  \label{fig:sys-arch}
\end{figure*}

Figure~\ref{fig:sys-arch} presents the overall system architecture of single-task learning for information type classification and priority estimation. The system began by preprocessing the tweets from a training set. The preprocessing was applied for all runs in both editions for data cleaning and refining. The preprocessing steps included removing URLs and punctuation (like \&, \text{@}, etc.). In addition, an in-domain out-of-vocabulary dictionary from \cite{imran2016twitter} was used to correct any misspelled words in tweets. 

Having being preprocessed, dataset analysis found that the classes of TREC-IS were severely imbalanced (it contains few tweets with more urgent information types and more critical priority levels), therefore data augmentation strategies are applied. After augmentation, the tweets were fed to train two models separately for information type classification and priority estimation. The architecture also allows for different models to be incorporated (here machine learning models and simple neural network models). Once the models were trained, they were used to make predictions for the unlabelled tweets in the test set. At this stage, each test tweet was predicted with one or more information types and one priority label. Instead of using the model predicted priority, it integrates with the predicted information types via a linear combination, defined as follows:

\begin{equation}
    {p_i} = (1 - \lambda ) \times {w_i} + \lambda  \times \delta({\hat{p}_i})
    \label{eq:trecis-2019a-linear-combination}
\end{equation}

where, given a tweet $i$, $p_i$ refers to its final predicted priority score and $w_i$ refers to the average priority weight based on the predicted information types for the tweet, which is obtained by looking up an information-type-to-weight table as presented in Figure~\ref{fig:sys-arch}. The look-up table is constructed by quantitatively analysing all tweets in the training set; the weight for an information type is averaged over the priority scores of the tweets that belong to that information type. In the equation, $\delta$ is the mapping function that maps the model predicted priority label $\hat{p}_i$ to the numeric priority score according to the mapping schema: low=0.25, medium=0.5, high=0.75 and critical=1.0. In the formula,  $\lambda$ is a parameter that can be adjusted to give more or less weight to the model predicted priority label and analysed priority weight and it is set to be 0.5 in practice, to give equal contribution. The intuition behind the linear combination is that the information type predictions are used to inform the priority prediction as the information types can implicitly reflect the priority levels such as a request of search and rescue implying a critical priority. Also, priority is quantified to a score between 0 and 1 with this combination.

\subsubsection{Traditional machine learning as the classifier}

In the TREC-IS 2019-A edition, four runs based on different variations of the aforementioned architecture were submitted, namely UCDrunEL1, UCDrunEL2, UCDrunELFB3 and UCDrunELFB4. All the runs were implemented based on machine learning models using feature representations. In all the runs, the up-sampling augmentation technique SMOTE~\cite{Chawla2002SMOTESM} was leveraged for mitigating the imbalanced dataset problem and binary relevance was used for information types (multi-label) classification. In addition, a Linear Regression model was trained for predicting priority levels. The runs differed in how they represented tweets and trained their information type classifiers with different scenarios. Table~\ref{tab:scenarios-trecis-2019a} summarises the combinations of four scenarios for the runs. Below are detailed the four scenarios. 

\begin{table}[h]
\centering
\renewcommand{\arraystretch}{1.5}

\begin{tabular}{lll}
\hline
                 & Simple ensemble          & Actionable-specific \\
\hline
Word2vec         & UCDrunEL1   & UCDrunEL2       \\
Feature boosting & UCDrunELFB3 & UCDrunELFB4 \\ 
\hline
\end{tabular}
\caption{Four runs at TREC-IS 2019-A}
\label{tab:scenarios-trecis-2019a}
\end{table}

\textbf{Word2vec}. For runs with this scenario, the tweets are represented by an in-domain word2vec-based world embeddings known as crisis word embeddings~\cite{imran2016twitter}, which are pre-trained on 52 million crisis-related tweets using a Continuous Bag Of Words (CBOW) architecture with negative sampling along with 300 word representation dimensionality. To generate a tweet representation, the bag-of-means strategy was used, which averaged the vectors of all individual words to output a single vector per tweet.

\textbf{Feature boosting}. This scenario combined the pre-trained word2vec representing a tweet with 21 other hand-crafted features. These features were extracted based on the commonly-applied practices in the literature of crisis tweet classification~\cite{beigi2016overview,neppalli2018deep}. They are based on the content of tweets, and include binary values (e.g. the presence of URLs, hashtags or emoticons), and numeric counts of important words in a tweet (e.g. the count of emotionally-positive or negative words, numbers, instructive words like ``donate'', ``call'', ``search''). The specific features chosen for a tweet were the number of hashtags (numeric), named entity count (numeric), digit count (numeric), number of important crisis-related verbs with reference to the crisis lexicon~\cite{olteanu2014crisislex} (e.g. trapped, stuck, move, etc., numeric), sentiment polarity (categorical, -1, 0 or 1), tweet length (word\_length, char\_length, numeric), ratio of uppercase letters (numeric) and retweet check (binary, 0 or 1).

\textbf{Simple ensemble}. In this scenario, two classifiers worked together to decide the information types to be assigned to a tweet: a Logistic Regression (LR) classifier and a Na\"ive Bayes classifier. Prior to deciding which machine learning models to employ in the ensemble approach, an initial experiment was conducted to evaluate a range of candidates, including Logistic Regression (LR), Na\"ive Bayes, Decision Tree, Random Forest, SVM, among others. The outcomes indicated that Logistic Regression (LR) and Na\"ive Bayes exhibited the highest performance among the candidates, and therefore, they were chosen for the ensemble approach. Additionally, the experiment revealed that the combined approach (simple ensemble) outperformed the individual models, which is why the two models were used together to make predictions for information type classification. The information types predicted for a tweet were determined by the averaged predicted probability of the two classifiers where an information type is assigned when the averaged probability is above 0.5.

\textbf{Actionable-specific}. Similar to the simple ensemble, this scenario also applied two classifiers to decide the information types to be assigned to a tweet. However, the difference is that the classifiers here were a LR classifier and an actionable classifier. The actionable classifier in the system was a linear Support Vector Machine (SVM)  model that was trained only on the training tweets that had actionable types (see Table~\ref{tab:actionable-versus-non} in Appendix~\ref{appendix:trecis-its}). Similar to the approach taken in the simple ensemble, an initial experiment was conducted to determine the optimal model for classifying actionable information types. SVM was ultimately selected as it achieved the highest performance among the candidate models. To assign information types for a tweet, the LR classifier first predicted a general list of information types and then the predicted list was forwarded to the actionable classifier so as to add any actionable information types not predicted by the LR classifier.

\subsubsection{Neural networks as the classifier}

In the TREC-IS 2019-B edition, four runs based on different approaches were submitted, namely UCDbaseline, UCDbilstmalpha, UCDbilstmbeta and UCDbcnelmo. Each used a neural network based model (except for UCDbaseline, which reused UCDrunELFB3, consisting of a simple ensemble with feature boosting, with the new data as this run was overall the best in terms of performance among the runs submitted to the 2019-A edition). For the remaining three runs, different models with data augmentation strategies were applied to explore the influence they have on the performance. The following describes the technical variations of each run. 

\textbf{UCDbilstmalpha}. To address the class imbalance problem, this run applied the text generation strategy GPT-2~\cite{radford2019language} for data augmentation. This extracts rare/less-represented tweet samples with respect to information types and priority in the training set and then uses the extracted samples as conditional samples for GPT-2 to generate synthetic samples. Table~\ref{tab:gen-gpt-2} gives some examples of generated samples by GPT-2 given raw critical tweets from the training set as the conditional samples. To represent a tweet, this run used GloVe in the embedding layer and a standard BiLSTM as the classification models. For information type classification, \textit{binary cross entropy} was used as the objective function. For priority level classification, \textit{categorical cross entropy} was used as the objective function.

\begin{table}[!h]
      \centering
\footnotesize
\renewcommand{\arraystretch}{1.5}
    \begin{tabular}[width=0.6\linewidth]{p{8cm}|p{7cm}}
    \toprule
    Conditional raw sample & Generated sample\\
    \hline
    LANDSLIDE!... road blocked in Costa Rica after M7.6 earthquake & 1 car submerged on a hillside and one car in water \\ 
\hline
More Than 1,000 People Are Now Missing In The California Wildfires & The fires are currently raging, causing extensive damage across vast areas of California \\ \hline
If you are interested in helping on the Earthquake relief & efforts, please call the local offices \\ \hline
Over 350 killed in Kerala floods & Thousands of people were trapped in the city \\ \hline
    
    \end{tabular}
      \caption{Generation of tweets with ``critical'' priority by GPT-2}
        \label{tab:gen-gpt-2}
\end{table}

\textbf{UCDbilstmbeta}. Unlike UCDbilstmalpha, this run did not apply data augmentation. Instead, a variant of the \textit{binary cross entropy} from Equation~\ref{eq:cross-entropy-lr} is applied $-w_i[y \log (\hat{y})+\left(1-y\right) \log \left(1-\hat{y}_{i}\right)]$\footnote{\url{https://pytorch.org/docs/stable/generated/torch.nn.BCELoss.html}} where $w_i$ as a weight factor for handling class imbalance. Unlike in other runs where it is set to be 1, in this run it is $w_{i}=\frac{\left|y_{\max }\right|}{|y_i|}$ where $|y_i|$ is the number of samples of information type $y_i$ appearing in the training data and $\left|y_{\max }\right|$ is the number of samples associated with the most common information type. Like UCDbilstmalpha, BiLSTM is used as the classification models for information type classification and priority estimation.

\textbf{UCDbcnelmo}. The last run used an attention-based neural network architecture as the classification models, i.e. the Bi-attentive classification (BCN) network by~\cite{mccann2017learned}.In~\cite{peters2018deep}, BCN was combined with ELMo embeddings to achieve strong performance in sentiment classification. BCN+ELMo applies the contextualised ELMo word representations in the embedding layer of BCN. Based on the promising performance that BCN+ELMo achieved in similar tasks, it was adapted to the crisis domain as the last run. Like UCDbilstmalpha, this run also leveraged the GPT-2 based data augmentation strategy for alleviating the class imbalance problem.

\subsubsection{Evaluation and discussion}

In evaluation, all participating groups submit their runs to the IS track whose organisers evaluate the runs of all participating groups uniformly using their proposed metrics (see Section~\ref{sec:rds-ems-infotypes}). The results of all the 8 runs described above are reported below. Results in bold are the best in their column. In addition to the presence of scores that the submitted runs achieved, the median scores of all participating groups for each metric are also included for comparison\footnote{The overview paper of this track in 2019~\cite{mccreadie2019trec} includes the full results of all participating groups.}. 

Table~\ref{tab:eval-report-trecis-2019a} presents the returned evaluation results for the four runs at TREC-IS 2019-A as well as the median and best scores in each metric from all participating runs. Generally, the runs in most metrics achieved performance above the median. In particular, actionable RMSEs (the lower score the better performance) for the runs (Act. (PErr-H)) are a lot lower than the median and the actionable F1 of UCDrunELFB3 (Act. (CF1-H)) is notably higher than the median (0.1180 vs 0.0459). The weakest aspect of the runs was in the information type (IT) accuracy, with no run above the median (although UCDrunELFB4 was not substantially below median performance). To compare UCDrunELFB3 with UCDrunEL1, and UCDrunELFB4 with UCDrunEL2, it is shown that the use of feature boosting in UCDrunELFB3 and UCDrunELFB4 did not help in AAW but did improv actionable F1 (0.1180 vs 0.0970 and 0.0918 vs 0.0884) and All F1 (0.1827 vs 0.1703 and 0.1668 vs 0.1505). Likewise, if comparing UCDrunEL2 with UCDrunEL1, the use of an actionable classifier in UCDrunEL2 did not help improve information type F1 but did improve both AAW and information type accuracy. In summary, the runs are some of the better performing ones in terms of information type classification, but are somewhat too conservative in terms of alerting. In contrast to a lot of other participating runs, the runs have a better distribution across classes (there are no complete class failures), likely due to the application of SMOTE for mitigating the class imbalance problem.

\begin{table}[!ht]
\centering
\scriptsize
\renewcommand{\arraystretch}{1.0}
\begin{tabular}[width=1.0\linewidth]{|l|c|c|c|c|c|c|c|}
\hline
Runs         & \multicolumn{2}{c|}{Alerting}                         & \multicolumn{3}{c|}{Information Feed}                                                                & \multicolumn{2}{c|}{Prioritisation}                            \\ \hline
            & \multicolumn{2}{c|}{AAW}          & \multicolumn{2}{c|}{IT Positive F1}                             & \multicolumn{1}{l|}{IT Acc.} & \multicolumn{2}{c|}{RMSE}                                      \\ \hline
            & \multicolumn{1}{l|}{High Pri. (AW-HC)} & All (AW-A)             & \multicolumn{1}{c|}{Act. (CF1-H)} & \multicolumn{1}{c|}{All (CF1-A)}     & \multicolumn{1}{c|}{All (Cacc)}            & \multicolumn{1}{c|}{Act. (PErr-H)} & All (PErr-A)                         \\ \hline
UCDrunEL1   & -0.8744                   & -0.4442 & 0.0970                 & 0.1703              & 0.7784                              & 0.1149                 & 0.0617              \\
UCDrunEL2   & \textbf{-0.8556}                   & \textbf{-0.4382} & 0.0884                 & 0.1505                       & 0.8324                              & \textbf{0.1144}                & 0.0633                       \\
UCDrunELFB3 & -0.9343                   & -0.4700 & \textbf{0.1180}                 & \textbf{0.1827}              & 0.7853                              & 0.1171                              &\textbf{ 0.0603  }                          \\
UCDrunELFB4 & -0.9268                  & -0.4676 & 0.0918                 & 0.1668              & 0.8519                              & 0.1207                 & 0.0623            \\ \hline\hline
Median      & -0.9493                            & -0.4855          & \multicolumn{1}{c|}{0.0459}     & \multicolumn{1}{c|}{0.15995} & \multicolumn{1}{c|}{\textbf{0.8539}}         & \multicolumn{1}{c|}{0.1615}     & \multicolumn{1}{c|}{0.07545} \\ 
Best &  0.2983                 & 0.2572  & 0.1695                 & 0.2825              & 0.9039                              & 0.1132                 & 0.0563 \\ \hline
\end{tabular}
\caption{Evaluation results of four runs at IS 2019-A based on the TREC-IS evaluation metrics described in Section~\ref{sec:rds-ems-infotypes}. Best of the four submitted runs in columns are \textbf{bolded}.}
 \label{tab:eval-report-trecis-2019a}
\end{table}

\begin{figure}[!h]
    \centering
    \includegraphics[width=\linewidth]{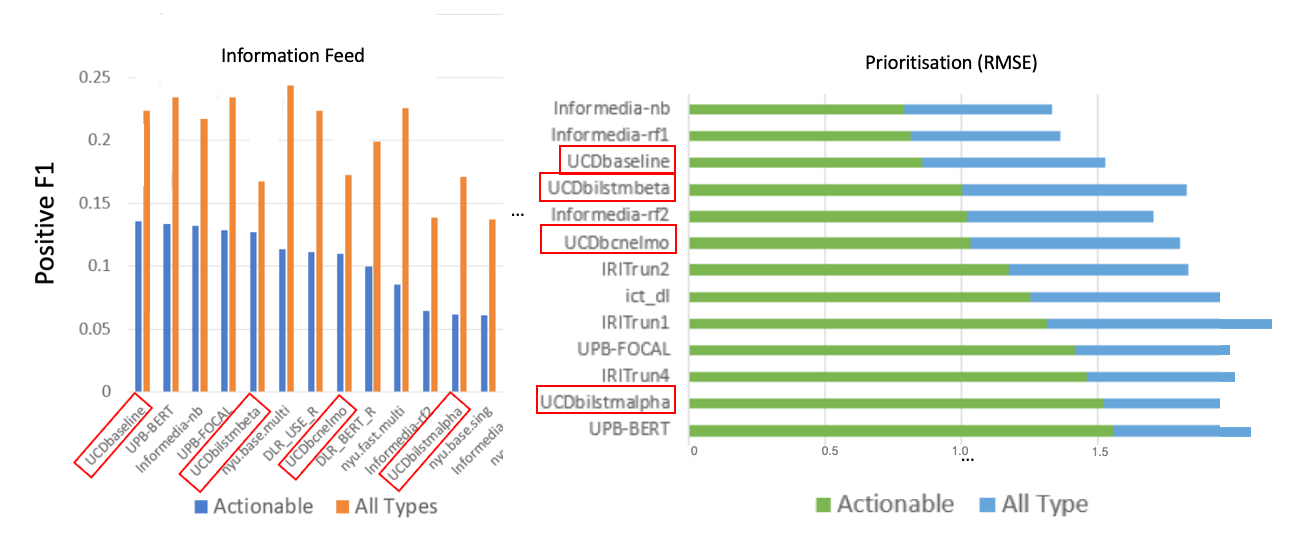}
    \caption{The runs at TREC-IS 2019-B among the top 13 participating runs out of 32 runs}
    \label{fig:compare-top13-trecis-2019b}
\end{figure}

Table~\ref{tab:eval-report-trecis-2019b} presents the returned evaluation results for the four runs at TREC-IS 2019-B. In general, the runs exhibit superior results compared to the median for the majority of metrics. In fact, for certain metrics like information feed F1 and Actionable RMSE, the runs demonstrate the best or nearly the best performance among all the participating runs. Figure~\ref{fig:compare-top13-trecis-2019b} compares the runs among the top 13 participating runs out of 32 runs in total. All the four runs appear in the top 13 in both information feed and prioritisation. In particular, UCDbaseline achieves the best (better than any other participating runs) performance in finding actionable information types as indicated by the actionable positive F1. Comparatively, UCDbilstmalpha is somewhat weak among the four runs. Although it achieves an information type accuracy of 0.86, which is marginally above the median of 0.8583, it does not do well in AAW, which indicates its weakness in correctly finding tweets for alerting. For UCDbilstmbeta, some of its improvements over UCDbilstmalpha are seen. In particular, UCDbilstmbeta achieves good performance in AAW, whose scores (-0.6047 and -0.3332) are increased from the median (-0.9197 and -0.4609) to some extent. It also outperforms UCDbilstmalpha in information type (IT) positive actionable F1. Its performance over UCDbilstmalpha implies that the use of loss weights helps to improve the overall performance. Compared to the other three runs, UCDbcnelmo obtained relatively even good performance across the metrics. It achieved better scores in almost every metric than the median to a good degree except that its information type accuracy is slightly lower than the median.



\begin{table}[!h]

\scriptsize 
\def\arraystretch{1.0}
\centering
 \begin{tabular}[width=1.0\linewidth]{|l|l|l|l|l|l|l|l|}
\hline
               & \multicolumn{2}{c|}{Alerting}                & \multicolumn{3}{c|}{Information Feed}                            & \multicolumn{2}{c|}{Prioritisation} \\ \hline
            Runs   & \multicolumn{2}{c|}{AAW} & \multicolumn{2}{c|}{IT Positive F1} & IT Acc. & \multicolumn{2}{c|}{Priority RMSE}  \\ \hline
           & High Pri. (AW-HC)         & All (AW-A)                  & Act. (CF1-H)           & All (CF1-H)                & All  (Caac)               & Act. (PErr-H)       & All (PErr-A)              \\  \hline
UCDbaseline    & -0.7856               & -0.4131              & \textbf{0.1355}      & \textbf{0.2232}     & 0.7495              & \textbf{0.0859}  & \textbf{0.0668}  \\ \hline
UCDbilstmalpha & -0.9287               & -0.4677              & 0.0614               & 0.171              & \textbf{0.86}       & 0.1521           & 0.0893           \\ \hline
UCDbilstmbeta  & \textbf{-0.6047}      & \textbf{-0.3332}     & 0.1269               & 0.1676              & 0.8378              & 0.1004           & 0.0822           \\ \hline
UCDbcnelmo     & -0.6961               & -0.3624              & 0.1099               & 0.1721              & 0.8452              & 0.1036           & 0.0769           \\ \hline
\hline
Median         & -0.9197               & -0.4609              & 0.0386               & 0.1361              & 0.8583              & 0.1767           & 0.1028           \\
Best & 0.2938                 & -0.1305 & 0.1355                 & 0.2343              & 0.8808                              & 0.0788                 & 0.0544 \\ \hline

\end{tabular}
  \centering
    \caption{Evaluation results of four runs at IS 2019-B based on the TREC-IS evaluation metrics described in Section~\ref{sec:rds-ems-infotypes}. Best of the four submitted runs in columns are \textbf{bolded}.}
  \label{tab:eval-report-trecis-2019b}
\end{table}

To summarise the runs in IS 2019, the runs in Edition B achieved overall better performance than the runs in Edition A as training data increased. It is also interesting to notice that UCDrunELFB3 from Edition A, rerun on Edition B denoted as UCDbaseline, achieved very strong performance in finding actionable information types and estimating priority comparing to other runs. This indicates the benefit of the linear combination (Equation~\ref{eq:trecis-2019a-linear-combination}, where information types are used to inform priority) as well as the simple ensemble strategy with hand-crafted features (multiple machine learning classifiers for information type predictions) applied in the UCDbaseline run. This motivated this research to propose a multi-task learning (MTL) approach followed by a simple ensemble combining multiple multi-task learners for information types classification and priority estimation in IS 2020 and 2021. For the runs in 2019 Edition B, it is also found that UCDbcnelmo achieves overall good performance, which implies the potential of attention-based neural networks for the IS task. This guided the MTL approach to be based on Transformer-encoder pre-trained language models such as BERT.   


\subsection{Multi-task learning with pre-trained language models}
\label{subsec:mtl-plms}

After reflecting on the experience from IS 2019, and the results of the runs submitted, improved solutions were sought for IS 2020. Here, this research proposes a multi-task learning (MTL) approach that learns the information type classification task and priority estimation task jointly by fine-tuning Transformer-encoder pre-trained language models such as BERT. Subsequently, an ensemble approach combining multiple such fine-tuned models is used for information type classification and priority estimation. In this section, the MTL approach followed by the ensemble approach is introduced and its effectiveness is discussed in comparison with single task learning approaches and other participating systems in IS 2020 and 2021. 


\subsubsection{Method}

\begin{figure*}[!h]
    \centering
       \includegraphics[scale=1.0]{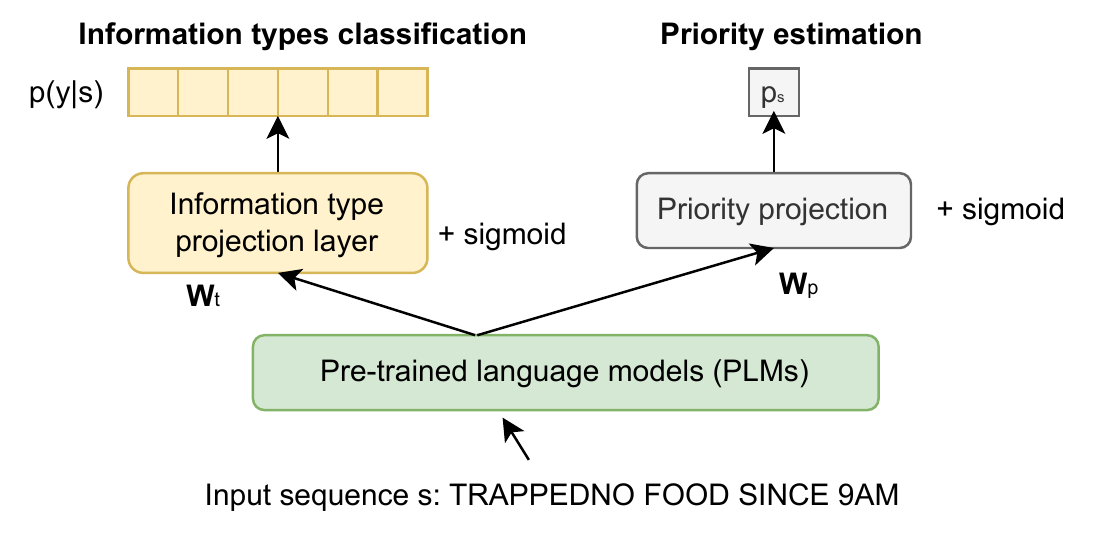}
    \caption{Overview architecture of Transformer-based multi-task learning (MTL) approach}
    \label{fig:trecis-mtl-method-overview}
\end{figure*}

 Figure~\ref{fig:trecis-mtl-method-overview} depicts the overview architecture of the MTL approach for the information type classification task and the priority estimation task. 

 \textbf{Information type classification}: The objective of information type classification is to predict the probability: $p_{(y_j|s)}$ referring to the likelihood of an input sequence $s$ (a tweet message) being assigned to the information type $y_j$. Since $s$ can be assigned to one or more information types, it is taken as a multi-label classification problem, estimated by the following equation.

\begin{equation}
    p_{(y|s)}=\sigma(f(s))
\end{equation}

where $\sigma$ is the sigmoid function (Equation~\ref{eq:sigmoid-function}) and $p_{(y|s)}$ is the estimated probability distribution across all information types $y:\{y_1,...y_j,...,y_m\}$ and $m$ is the number of information types. In the proposed method, an information type $y_j$ is assigned to $s$ when $p_{(y_j|s)}>0.5$.
 

To learn the function $f(\cdot)$, a neural network architecture incorporating a pre-trained language model (PLM) and an information type projection layer is deployed. Here, BERT is used as an example of a PLM, although the architecture supports the use of other PLMs as an alternative. BERT represents the sequence first by its output vector at the \texttt{[CLS]} token, denoted by $\mathrm{PLM_{CLS}}(s)$ and then this representation is used as the input of the projection layer. The projection process is formulated as follows.

\begin{equation}
    f(s)=\mathrm{PLM_{CLS}}(s)\times\textbf{W}_{t}
\end{equation}

Where $\textbf{W}_t \in \mathbb{R}^{d_{\mathrm{model}} \times m}$ is the learnable parameters of the projection linear layer and $d_{\mathrm{model}}$ is the hidden state dimension of the Transformer encoder based PLM (i.e., the dimensionality of the output vector at the \texttt{[CLS]} token).





\textbf{Priority estimation}: Similar to information type classification, the priority estimation task is treated as a regression task (due to the dependency between priority levels). In priority estimation, a numeric score $0\leq p \leq1$ is assigned to the input sequence $s$ to quantify its priority level ($p_s$), which is estimated as follows.

\begin{equation}
    p_s = \sigma(g(s))
\end{equation}

To learn the function $g(\cdot)$, a priority projection layer is added to the PLM, which transforms the \texttt{[CLS]} token output vector to the priority score that can be viewed as a numeric scalar, formulated as follows.

\begin{equation}
    g(s)=\mathrm{PLM_{CLS}}(s)\times\textbf{W}_{p}
\end{equation}

Where $\textbf{W}_p \in \mathbb{R}^{d_{\mathrm{model}} \times 1}$ is the learnable parameters of the priority projection layer and $d_{\mathrm{model}}$ is the hidden state dimension of the BERT-style Transformer encoder. 

\textbf{Joint learning}: As mentioned previously, the two tasks share the PLM, i.e., using the same layers of millions of parameters to represent the input sequence $s$ with the \texttt{[CLS]} output vector (Section~\ref{subsubsec:encoder}). The motivation behind this is twofold. Firstly, parameter sharing between multiple tasks is likely to enable one task to share its learnt knowledge with another~\cite{zhang2017survey}, which is desirable in solving the current problem: information types can inform priority levels and the other around. Secondly, it is faster to make predictions at inference time compared to training two separate models for every task.

To update the parameters of the projection layers as well as the shared parameters of the PLM, the joint loss $\mathcal{L}_{\mathrm{joint}}$ combining the information type prediction loss $\mathcal{L}_{\mathrm{it}}$ and priority prediction loss $\mathcal{L}_{\mathrm{pri}}$ is used as the objective function, defined as follows.

\begin{equation}
\mathcal{L}_{\mathrm{joint}}=\lambda \mathcal{L}_{\mathrm{it}}+(1-\lambda) \mathcal{L}_{\mathrm{pri}}
\label{eq:loss-all}
\end{equation}

Where $\lambda$ is a coefficient ($0\leq \lambda \leq1$) to adjust the relative weights of $\mathcal{L}_{\mathrm{it}}$ and $\mathcal{L}_{\mathrm{pri}}$. Considering information type prediction to be a multi-label classification problem, \textit{binary cross entropy} is used to define $\mathcal{L}_{\mathrm{it}}$. The following formulates the loss computation for one training sequence. For a mini-batch of sequences, the final loss is averaged over the sequences.


\begin{equation}
\begin{aligned}
    \mathcal{L}_{\mathrm{it}} = \sum_{y_j \in y}-b(y_j,s)\log(p(y_j|s)) - (1-b(y_j,s))\log(1-p(y_j|s))
\end{aligned}
\label{eq:loss-it}
\end{equation}

Where $b(y_j,s)$ is $0$ or $1$, indicating if the information type $y_j$ is assigned to the sequence $s$ (information type ground truth) and $p(y_j|x)$ is the probability score for predicting $s$ to be $y_j$. Considering priority estimation to be a regression problem, \textit{mean squared error} is used to define $\mathcal{L}_{\mathrm{pri}}$ as follows.

\begin{equation}
        \mathcal{L}_\mathrm{pri}=(m(r_s)-p_s)^2
        \label{eq:loss-pri}
\end{equation}

Where $r_s$ is the ground truth priority of $s$ and $m(\cdot)$ is the mapping function converting predicted categorical priority levels ($p_s$) to numeric scores according to the schema: $\{Critical:1.0, High:0.75, Medium:0.5, Low:0.25\}$. At prediction time, the mapping function is used in reverse to convert the predicted numeric score to a categorical priority level. For example, the priority of $s$ is predicted to be critical if $p_s$ lies between $1$ and $0.75$ inclusive.

\subsubsection{Experiments and results}

Training involves fine-tuning the PLMs to the joint downstream task of information type classification and priority estimation. For hyper-parameter selection, a grid search over learning rate: \textit{lr $\in$ \{5e-4,2e-4,1e-4, 5e-5, 2e-5, 1e-5\}} and mini-batch size: \textit{bs $\in$ \{8, 16, 32,64\}} is done with the development set. Finally, \textit{lr} and \textit{bs} are set to be 5e-5 and 32 respectively as they perform better empirically. Following the work in a similar domain~\cite{liu2020crisisbert}, the Adam optimiser~\cite{kingma2014adam} is used to update model parameters and a linear scheduler is applied for dynamically updating the learning rate, with 10\% warm-up ratio of the total training steps (12 epochs). Moreover, for the fine-tuning process, the value of $\lambda$ in Equation~\ref{eq:loss-all} was set to 0.5, which assigns equal weight to both the information type loss and priority loss. This decision was based on a preliminary study, which discovered that $\lambda$ of 0.5 yields satisfactory performance in both tasks. To arrive at this conclusion, a grid search was performed over $\lambda$ values ranging from 0 to 1.0 in increments of 0.1.


Since this approach was proposed in the context of IS 2020 (Edition A), in experiments, the 2020A subset was used as the test set that consists of tweets from 15 crisis events (see Section~\ref{sec:rds-ems-infotypes}). For the training and development sets, a data set combining the 2018, 2019A and 2019B subsets was formed first. Then a sampled 10\% of the combined set was used as the development set and the remaining 90\% as the training set. In evaluation, in addition to the tailored metrics (Section~\ref{sec:rds-ems-infotypes}), a harmonic mean metric (\textbf{HarM}) was added to the list as an indicator of overall performance across all metrics. In calculating this harmonic mean, AW-HC and AW-A were first normalised to lie between $0$ and $1$. 

In this work, three experiments were conducted to test the effectiveness of the proposed approach in different aspects. The first experiment (Ex1) was conducted to compare it to single-task learning baselines. The second experiment (Ex2) was conducted to explore the use of different PLMs with the approach, and examine their performance both individually and in ensembles. Finally, the approach was tested in the editions of IS 2020 (Ex3). The following presents each experiment along with the discussion of results.



    
    


\vspace{2px}

\textbf{Ex1: Single task learning baselines}

In the first experiment to explore the difference between the multi-task learning approach and single task learning approach, \texttt{BERT\_base}\footnote{\url{https://huggingface.co/bert-base-uncased}} is used as a baseline to fine-tune two models independently for the information type classification and priority estimation tasks in a single task learning (STL) scenario. Specifically, the loss functions of Equation~\ref{eq:loss-it} and Equation~\ref{eq:loss-pri} are used separately to train two single task models. In contrast, the proposed multi-task learning (MTL) scenario uses the joint loss function of Equation~\ref{eq:loss-all}. In addition to the Transformer-based deep learning baseline, traditional machine learning (ML) algorithms such as Logistic Regression (LR) are considered to be strong baselines in this problem domain, as evidenced in previous work~\cite{mccreadie2020trec,congcong2020cls}. Hence, another baseline that trains two separate LR-based classifiers for the two single tasks is implemented in the experiment also. To make this baseline as strong as possible, the development set is used to set up the hyper-parameters of the LR classifiers (using grid search) including \textit{C $\in$ \{0.01,0.1,1,1.0,10,100\}}, \textit{ngram\_range~$\in$~\{(1),(1,2),(1,3)\}} and \textit{weighting~$\in$~\{count,TF\text{-}IDF\}}. Following empirical study, the setups ultimately used \textit{C=10}, \textit{ngram\_range=(1,2)} and \textit{weighting=TF\text{-}IDF}. 


The performance of the BERT-based and LR-based STL runs compared to the BERT-based MTL run is reported in Table~\ref{tab:mtl-study}. It shows that the BERT-based runs perform better than the LR runs except for a marginal decrease in Cacc score. More importantly, the MTL run achieves substantial improvement in NDCG and AW scores over the BERT-based STL run. For example, the BERT\_base + MTL run achieves the best scores in NDCG, AW-HC, AW-A and CF1-H. Although it can be seen that the BERT-based STL runs perform the best in prioritisation, this is at the cost of a loss of NDCG and AW performance. As a whole, the MTL scenario gains an advantage over the STL scenario not only in the overall effectiveness (as illustrated by the HarM score in the last column) but also in avoiding the need to train separate models for separate tasks.

\begin{table*}[h]
\centering
\scriptsize
\renewcommand{\arraystretch}{1.2}
\begin{tabular}{llllllllll}
\toprule

                      & \multicolumn{5}{c}{Priority estimation and ranking}                                           & \multicolumn{3}{c}{Info. types   classification} &               \\
                  
    \cmidrule(lr){2-6}                  
    \cmidrule(lr){7-9}

                  & \textbf{NDCG} & \textbf{AW-H} & \textbf{AW-A} & \textbf{PErr-H} & \textbf{PErr-A} & \textbf{CF1-H}   & \textbf{CF1-A}   & \textbf{Cacc}   & \textbf{HarM} \\
LR + STL         & 0.4495        & -0.4856       & -0.2627       & 0.1718          & 0.2216          & 0.0898           & 0.1527           & \textbf{0.9113}          & 0.2109      \\
BERT\_base + STL & 0.4393        & -0.4057       & -0.2148       & \textbf{0.2402}$\star$          & \textbf{0.2758}$\star$          & 0.1084           & \textbf{0.1801}           & 0.8960          & 0.2510      \\
\midrule

BERT\_base + MTL      & \textbf{0.5101}        & \textbf{-0.2689}$\star$       & \textbf{-0.1569}$\star$       & 0.1923          & 0.2544          & \textbf{0.1382}$\star$           & 0.1638           & 0.8937          & \textbf{0.2609}  
\\
\bottomrule
\end{tabular}

\caption{Comparison between single task learning (STL) with Logistic Regression (LR) and BERT and the multi-task learning (MTL). The numbers in bold represent the highest performance in each column and those annotated with $\star$ indicates that the highest is ``confident'' compared to the next-highest in its column (Wilson Score Interval~\cite{wilson1927probable}, $p<0.05$). The difference between $c_1$ and $c_2$ is described as ``confident'' if their confidence intervals do not overlap.} 
\label{tab:mtl-study}
\end{table*}

\textbf{Ex2: Transformer selection and ensemble} 

In the architecture of the proposed approach (Figure~\ref{fig:trecis-mtl-method-overview}), one important component is the selection of Transformer encoder based PLMs. In the ``Method'' section above, BERT is used as an example of a PLM. However, since the introduction of BERT, the literature has seen many BERT variants being developed (Section~\ref{subsubsec:encoder}). Variants like DistilBERT~\cite{sanh2019distilbert}, ALBERT~\cite{sanh2019distilbert}, and ELECTRA~\cite{Clark2020ELECTRAPT} have been developed to optimise various aspects of BERT such as memory consumption, computation cost or pre-training representation learning. Studies have shown their promising performance through fine-tuning in downstream tasks such as text classification and reading comprehension. To examine their capabilities in the crisis domain, they are used along with BERT in the experiment.

The upper block of Table~\ref{tab:trans-study} presents the results of using different PLMs individually with the MTL approach, whereas the lower block reports the results of creating ensembles combining the predictions of multiple PLMs. The number appended to the individual runs refers to the trained model size and the number appended to the ensemble runs indicates the combination of corresponding individual run indices. As this shows, the BERT run has the same model size as the ELECTRA. However, there is no significant difference in performance between the runs when evaluated using the HarM score. Although BERT outperforms ELECTRA in AW, it loses this advantage in prioritisation. In comparison, DistilBERT and ALBERT are relatively smaller in size than BERT and ELECTRA. They still perform well overall, only slightly worse than the BERT and ELECTRA runs, which illustrates their effectiveness with much reduced model sizes (ALBERT in particular). 

\begin{table*}[hbt]
\centering
\scriptsize
\renewcommand{\arraystretch}{1.2}
\begin{tabular}{llllllllll}\toprule
                  & \multicolumn{5}{c}{Priority estimation and ranking}                                           & \multicolumn{3}{c}{Info. types   classification} &               \\
                  
    \cmidrule(lr){2-6}                  
    \cmidrule(lr){7-9}

Run   variants               & \textbf{NDCG} & \textbf{AW-HC} & \textbf{AW-A} & \textbf{PErr-H} & \textbf{PErr-A} & \textbf{CF1-H} & \textbf{CF1-A} & \textbf{Cacc} & \textbf{HarM} \\\hline

\multicolumn{10}{l}{\textit{Individual Transformer encoder based PLMs}}  \\
1. BERT\_base (110M)      & 0.5101        & -0.2689       & -0.1569   & 0.1923          & 0.2544     & 0.1382         & 0.1638         & 0.8937                 & 0.2609
 \\

2. DistilBERT\_base (66M)      & 0.4808        & -0.4533       & -0.2382   & 0.9004        & 0.1191    & 0.1376         & 0.1830                   & 0.2110          & 0.2264
           \\
3. ELECTRA\_base (110M)   & 0.5042        & -0.4011       & -0.2122  & 0.2059          & 0.2801     & 0.1514         & 0.1742         & 0.8958                  & 0.2689
           \\
4. ALBERT\_base\_v2 (11M) & 0.4669        & -0.4118       & -0.2190    & 0.1900          & 0.2720    & 0.0568         & 0.1707         & \textbf{0.9087}                 & 0.1923

\\\hline
\multicolumn{10}{l}{\textit{Ensemble runs}} \\
EnsembleA (1+3)     & \textbf{0.5207} & -0.2274 & -0.1406 & 0.1999 & 0.2560& 0.1738 & 0.1796 & 0.8722  & 0.2836
 \\
EnsembleB (2+4)     & 0.4848 & -0.3212 & -0.1823& 0.2081 & 0.2728 & 0.1407 & 0.2041 & 0.8844  & 0.2752
 \\
EnsembleC (1+2+3)   & 0.5206 & -0.1982 & -0.1282 & 0.2023 & 0.2589 & \textbf{0.1819} & 0.1909 & 0.8621 & 0.2919
 \\
EnsembleD (1+2+3+4) & 0.5176 & \textbf{-0.1613}$\star$ & \textbf{-0.1148}& \textbf{0.2594}$\star$ & \textbf{0.2966 } & 0.1754 & \textbf{0.2084} & 0.8545 & \textbf{0.3141}$\star$ \\\hline
\end{tabular}

\caption{Individual runs with different PLMs and ensemble runs that leverage the individual runs jointly making predictions for priority and information types.}
\label{tab:trans-study}
\end{table*}

Of the individual runs, each scores the highest performance in at least one metric, with none showing a significant overall performance improvement over the others. In order to exploit the power of these individual runs, this work proposes a simple ensemble approach that combines the individual runs to jointly make predictions for information types and priority. It is based on the hypothesis that such an ensemble may leverage the distinct benefits of these diverse PLMs to achieve greater overall performance across both tasks. The ensemble approach is described as follows.

\textbf{The ensemble approach}: Given a set of individual multi-task learners, $\{l_1,l_2,\ldots,l_n\}$, the final priority prediction for a tweet is made from the priority predictions by $\{l_1,l_2,\ldots,l_n\}$ according to a priority decision strategy, denoted by $P_{ds}$. Three options are set up to evaluate $P_{ds}$: \textit{\{Highest, Average, Lowest\}}. \textit{Highest} refers to always selecting the highest priority level given by any of the individual predictors. \textit{Lowest} represents the opposite strategy. The \textit{Average} scenario means taking an average score over all priority predictions in the union and then the final priority prediction is assigned based on this average score. The conversion between priority numeric score and level is applied via the mapping function, namely $m(\cdot)$ as introduced in Equation~\ref{eq:loss-pri}. Regarding information types, the final prediction for each tweet is made from the information type predictions by $\{l_1,l_2,\ldots,l_n\}$ according to an information type decision strategy, denoted by $I_{ds}$. Two options are available to determine $I_{ds}$: \textit{\{Union, Intersection\}}. \textit{Union} and \textit{Intersection} refers to always selecting the union and intersection respectively of information types assigned by the individual predictors.

\begin{table}[!h]
    \centering
    \footnotesize
\begin{tabular}{lrrrrrrrrr}
\toprule
                  & \multicolumn{5}{c}{Priority estimation and ranking}                                           & \multicolumn{3}{c}{Info. types   classification} &               \\
    \cmidrule(lr){2-6}                  
    \cmidrule(lr){7-9}
           & 
\multicolumn{1}{l}{\textbf{NDCG}} & \multicolumn{1}{l}{\textbf{AW-H}} & \multicolumn{1}{l}{\textbf{AW-A}} &
\multicolumn{1}{l}{\textbf{PErr-H}} & \multicolumn{1}{l}{\textbf{PErr-A}} &
\multicolumn{1}{l}{\textbf{CF1-H}} & \multicolumn{1}{l}{\textbf{CF1-A}} & \multicolumn{1}{l}{\textbf{Cacc}} &  \multicolumn{1}{l}{\textbf{HarM}} \\ 
Union-Highest        & 0.5170                             & -0.1613                            & -0.1148  & 0.2594                               & 0.2966                            & 0.1754                              & 0.2084                              & 0.8545                                                          & 0.3140                             \\ 
Union-Average        & 0.5066                             & -0.2489                            & -0.1491                         & 0.2475                               & 0.274   & 0.1754                              & 0.2084                              & 0.8545                                                             & 0.3036                             \\ 
Union-Lowest         & 0.4896                             & -0.5824                            & -0.2932                           & 0.1302                               & 0.2102  & 0.1754                              & 0.2084                              & 0.8545                                                            & 0.2369                             \\ 
Intersection-Highest & 0.5178                             & -0.1613                            & -0.1148     & 0.2342                               & 0.2715                       & 0.0303                              & 0.1105                              & 0.9291                                                            & 0.1387                             \\ 
Intersection-Average & 0.5061                             & -0.2489                            & -0.1491                            & 0.2184                               & 0.2485 & 0.0303                              & 0.1105                              & 0.9291                                                            & 0.1362                             \\ 
Intersection-Lowest  & 0.4888                             & -0.5824                            & -0.2932                           & 0.1321                               & 0.2215   & 0.0303                              & 0.1105                              & 0.9291                                                          & 0.1233                             \\ 
\bottomrule
\end{tabular}
    \caption{Evaluation results of the \texttt{EnsembleD} run with varying strategies for merging information types and priority levels. Each row is named as $x-y$ where $x$ is the information type strategy, i.e. $I_{ds}$ and $y$ is the priority level strategy, i.e. $P_{ds}$.}
    \label{tab:ensemble-decision}
\end{table}

In choosing $P_{ds}$ and $I_{ds}$, the ensemble run combining all four individual runs from the top of Table~\ref{tab:trans-study}, namely EnsembleD, is tested with different $P_{ds}$ and $I_{ds}$ and the results are reported in Table~\ref{tab:ensemble-decision}. The results show that the $Union$ runs substantially outperform the $Intersection$ runs in information feed categorisation while yielding the same scores in the remaining metrics. For $P_{ds}$, it shows an increased performance in ranking, alert worth, and prioritisation as it changes from $lowest$ to $highest$. Based on the results, in the subsequent experiments, $I_{ds}$ is set to be \textit{Union} and $P_{ds}$ is set to be \textit{Highest} as this combination gives the best performance across the metrics.

With this setup, the experiment is then conducted to test the ensemble approach with different sets of $\{l_1,l_2,\cdots,l_n\}$ and the lower block of Table~\ref{tab:trans-study} demonstrates the results. The EnsembleA run combines the two relatively large models, BERT and ELECTRA, while EnsembleB combines DistilBERT and ALBERT. It shows that each ensemble has overall performance superior to its component models. This indicates that the ensemble approach adds benefit to the performance by leveraging the predictions of individual runs. The best-performing run among all experimental runs so far reported in this work is EnsembleD, with the HarM score reaching $0.3141$ as well as achieving strong scores in individual metrics except for in Cacc. However, It is analysed that Cacc is arguably the least important metric due to the heavy imbalance in the TREC-IS information types and the usual problems with accuracy-based metrics in such a scenario. A na\"ive run that predicts all tweets to have all information types will achieve a Cacc score of approximately 0.94, while being useless in practical terms and performing extremely poorly in the other metrics. To examine the performance of the ensemble runs in separate tasks, it is found that the ensemble runs outperform the single model runs in both the priority estimation and information type classification tasks. Although the overall performance increases as more individual models combined, the experiment found that the increase becomes marginal when more models than in EnsembleD are combined.

\begin{figure*}
     \centering
     \footnotesize
    \begin{subfigure}[b]{0.4\textwidth}
         \centering
         \includegraphics[width=\textwidth]{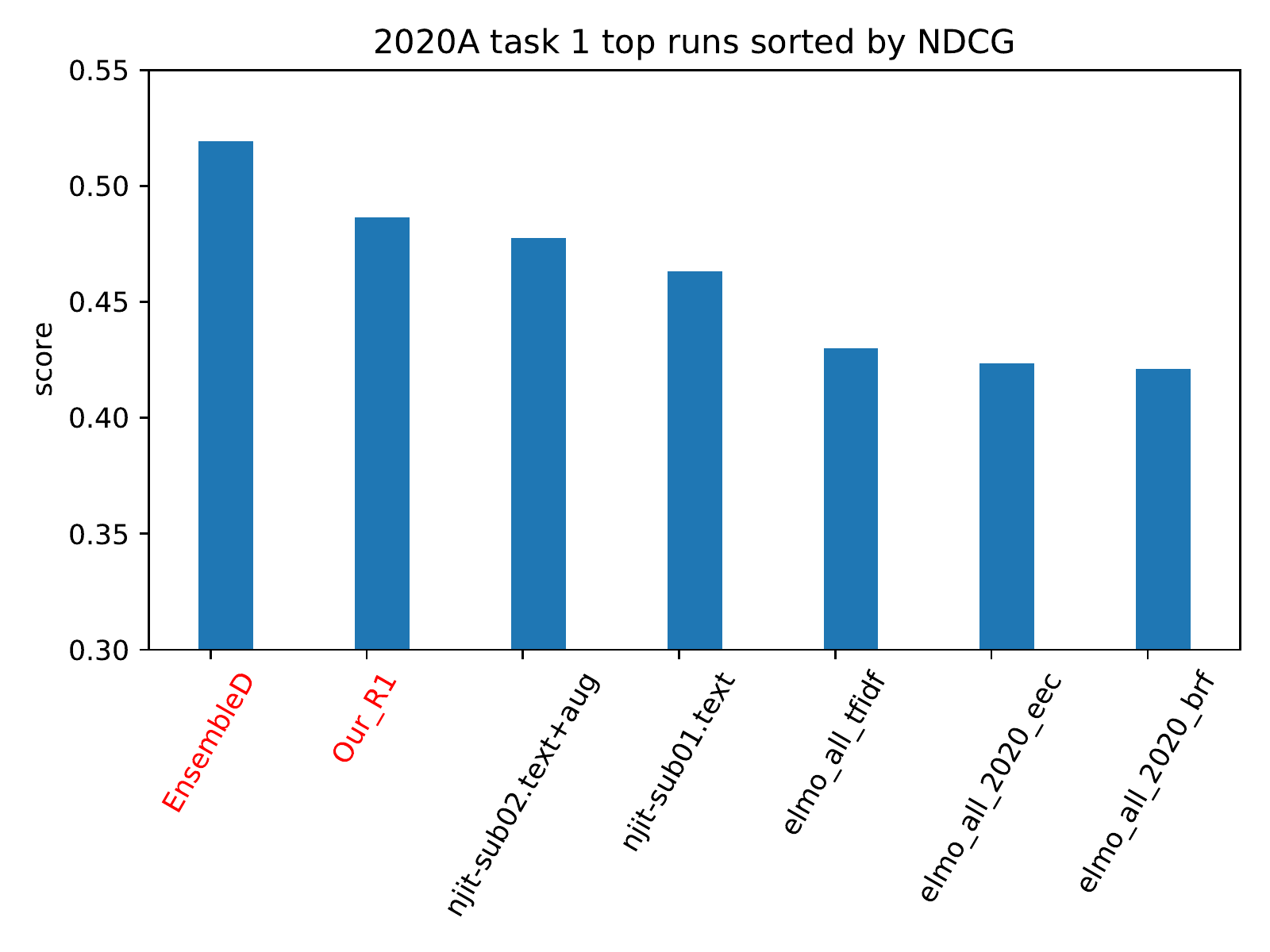}
         \caption{\textbf{Task 1 Ranking}}
         \label{fig:task1-ranking}
     \end{subfigure}
     \begin{subfigure}[b]{0.4\textwidth}
         \centering
         \includegraphics[width=\textwidth]{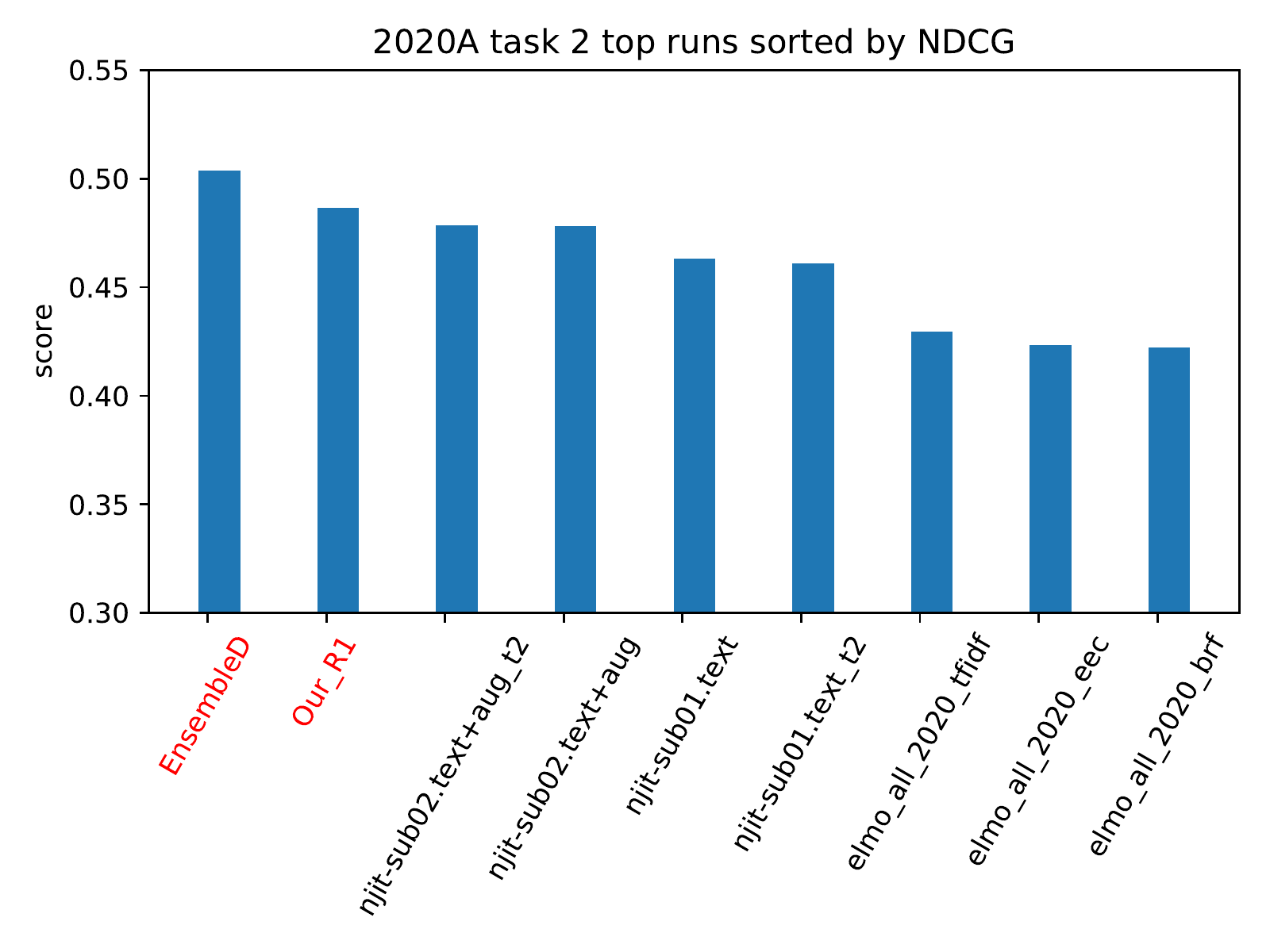}
         \caption{\textbf{Task 2 Ranking}}
         \label{fig:task2-ranking}
     \end{subfigure}
     \vspace{0.5cm}
     
      \begin{subfigure}[b]{0.4\textwidth}
         \centering
         \includegraphics[width=\textwidth]{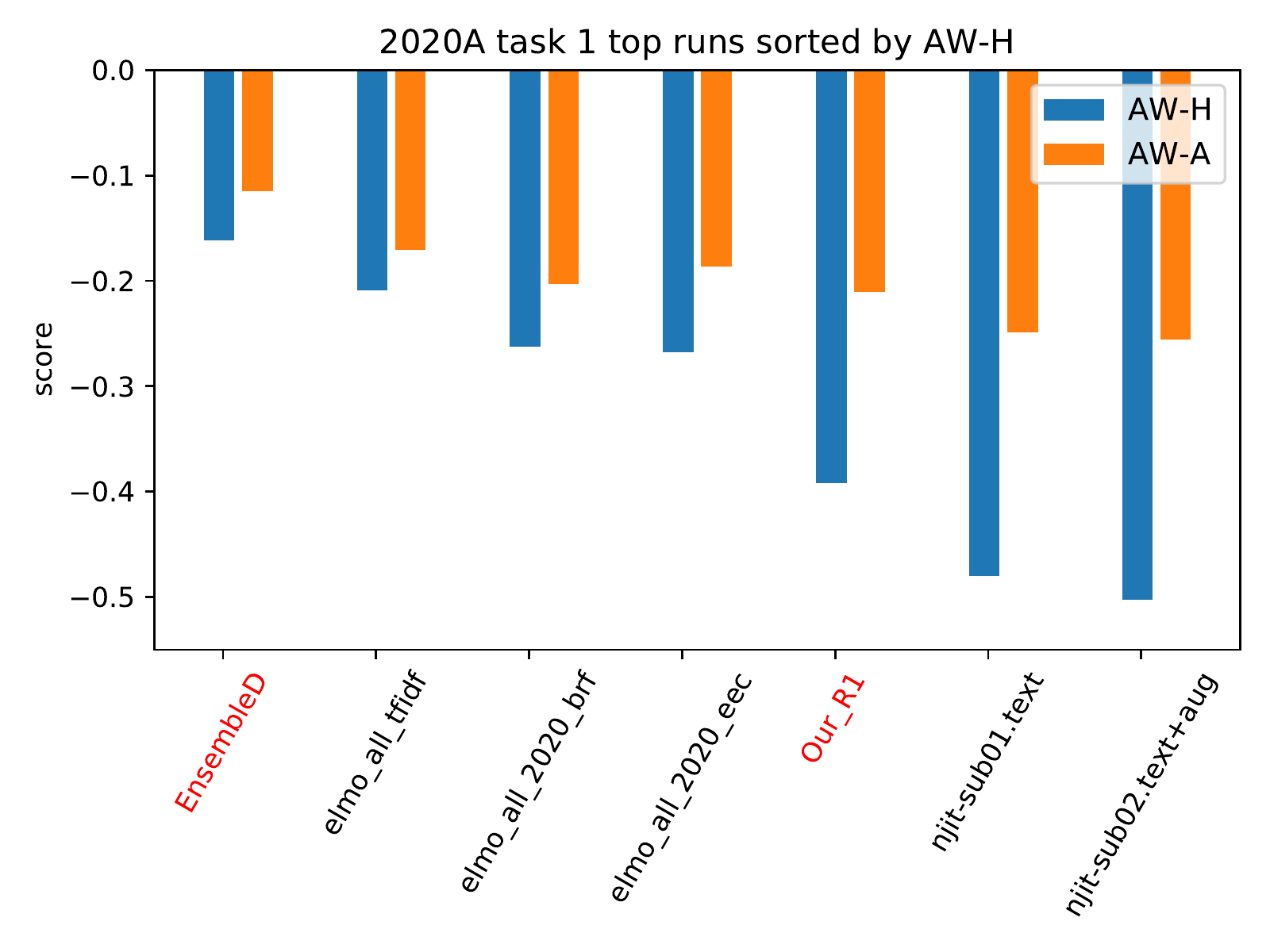}
         \caption{\textbf{Task 1 Alert Worth}}
         \label{fig:task1-aaw}
     \end{subfigure}
     \begin{subfigure}[b]{0.4\textwidth}
         \centering
         \includegraphics[width=\textwidth]{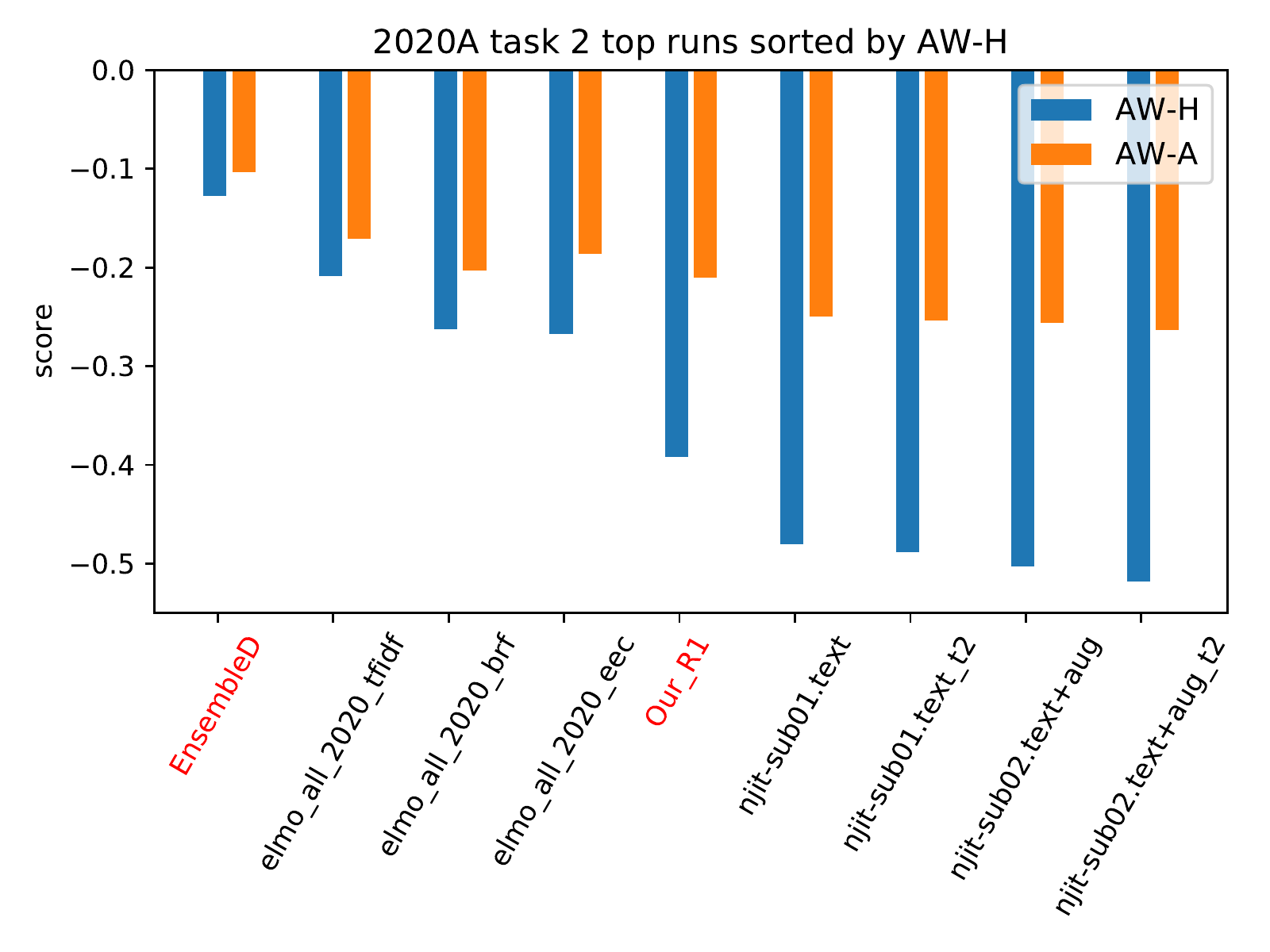}
         \caption{\textbf{Task 2 Alert Worth}}
         \label{fig:task2-aaw}
     \end{subfigure}
     \vspace{0.5cm}
     
      \begin{subfigure}[b]{0.4\textwidth}
         \centering
         \includegraphics[width=\textwidth]{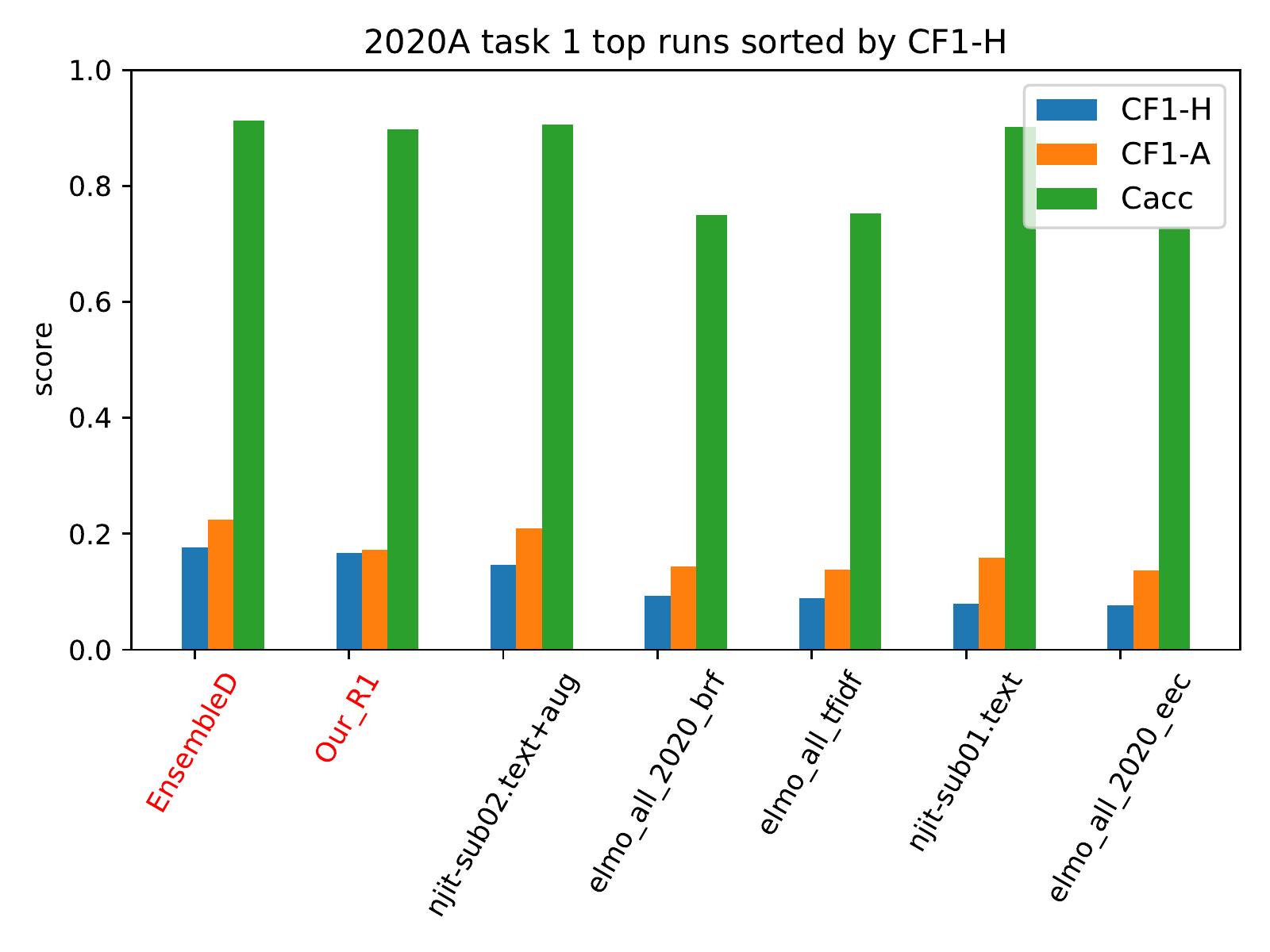}
         \caption{\textbf{Task 1 Information Feed Categorisation}}
         \label{fig:task1-infofeed}
     \end{subfigure}
     \begin{subfigure}[b]{0.4\textwidth}
         \centering
         \includegraphics[width=\textwidth]{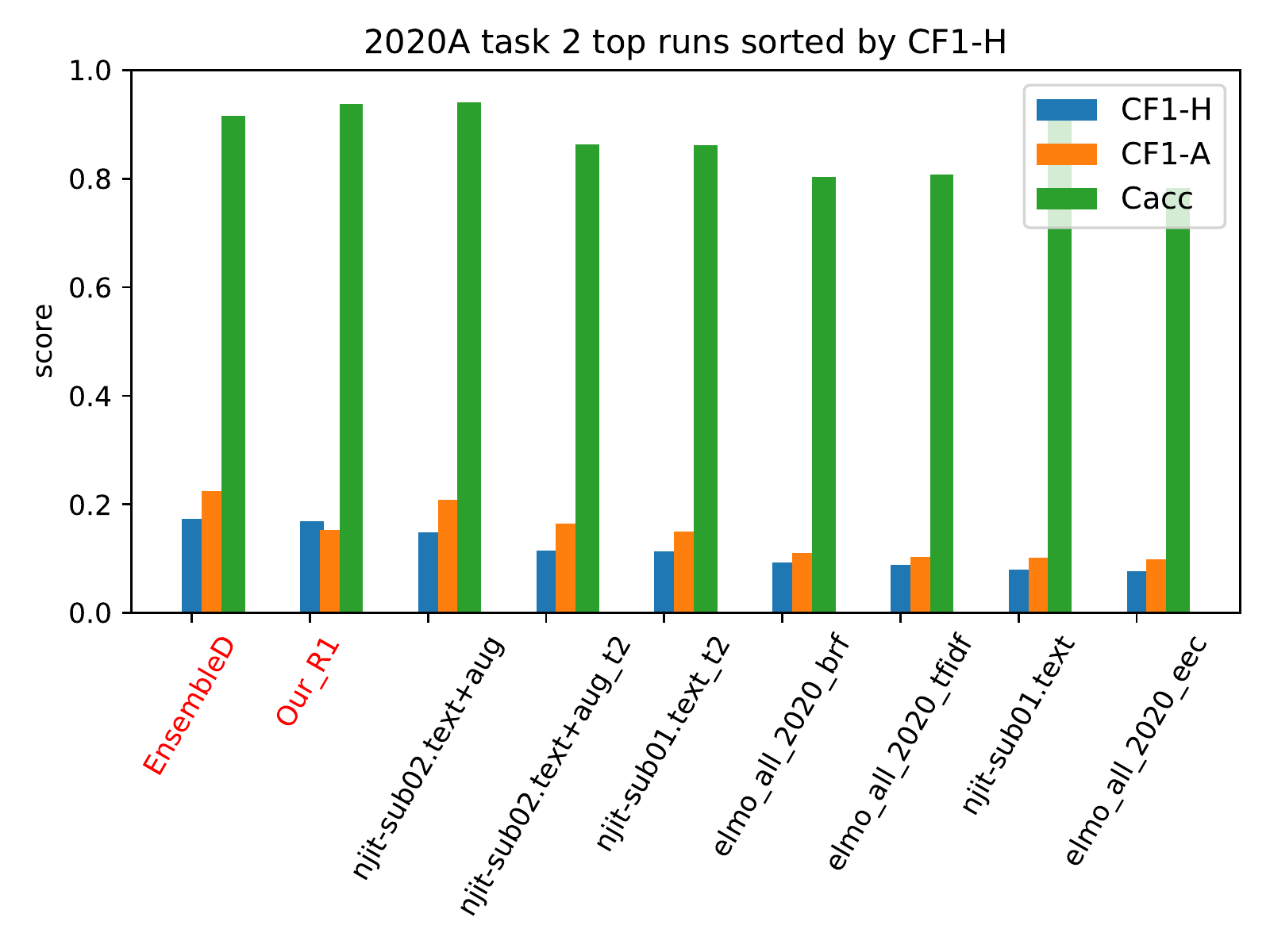}
         \caption{\textbf{Task 2 Information Feed Categorisation}}
         \label{fig:task2-infofeed}
     \end{subfigure}
     \vspace{0.5cm}
     
      \begin{subfigure}[b]{0.4\textwidth}
         \centering
         \includegraphics[width=\textwidth]{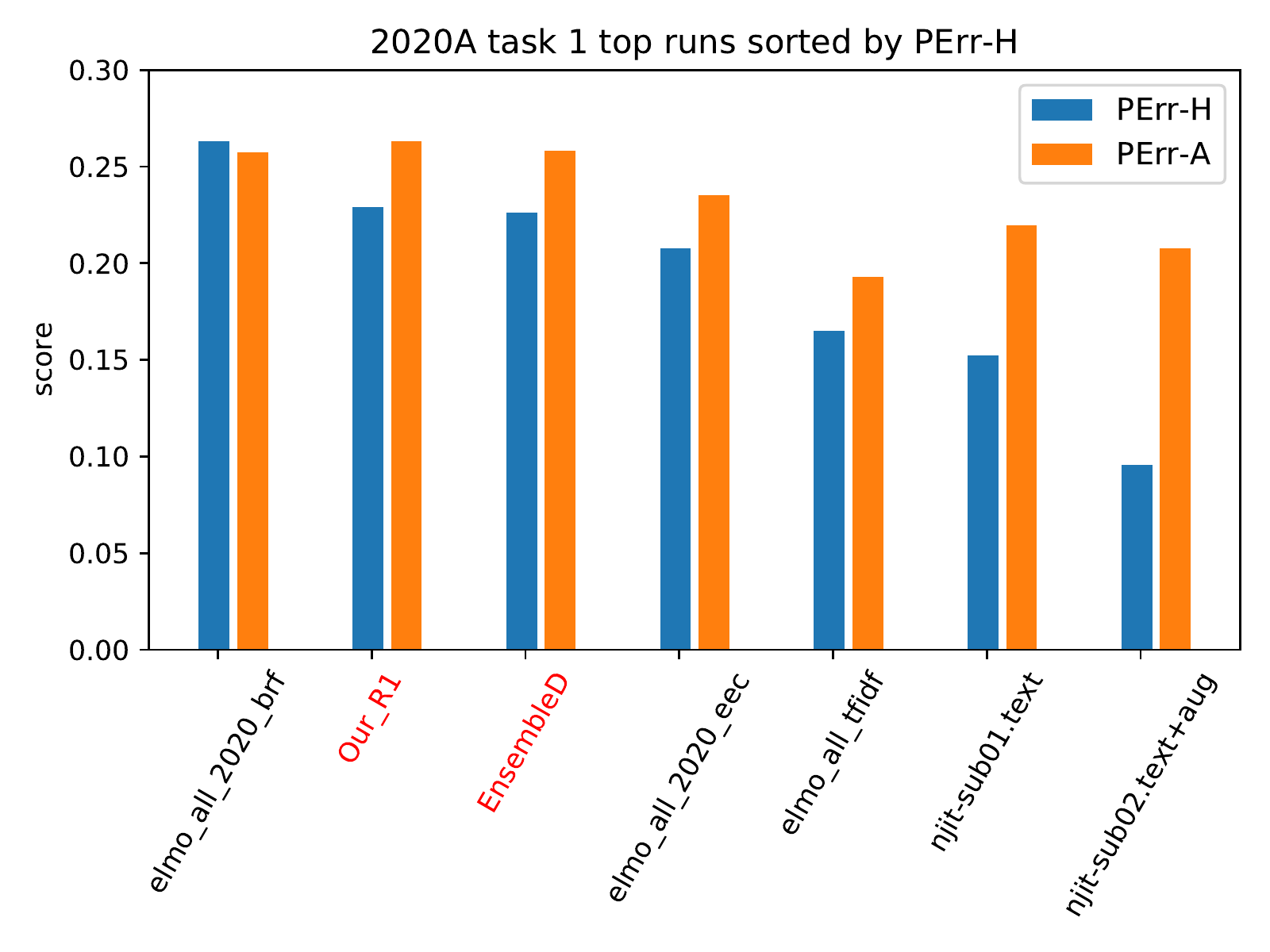}
         \caption{\textbf{Task 1 Prioritisation}}
         \label{fig:task1-priority}
     \end{subfigure}
     \begin{subfigure}[b]{0.4\textwidth}
         \centering
         \includegraphics[width=\textwidth]{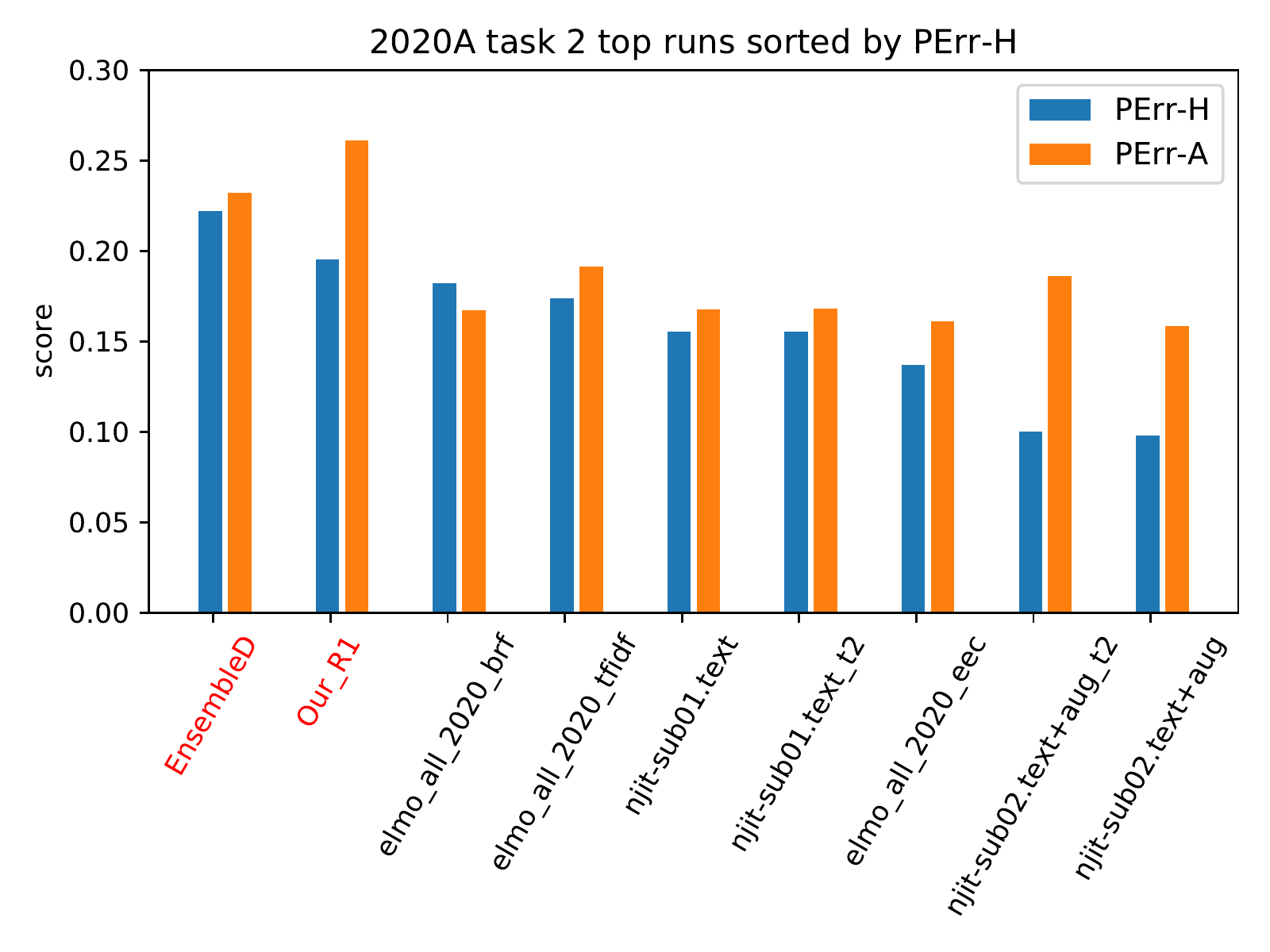}
         \caption{\textbf{Task 2 Prioritisation}}
         \label{fig:task2-priority}
     \end{subfigure}
     
    \caption{Performance comparison between TREC-IS 2020A top participating runs for both task 1 and 2. The runs based on the proposed MTL approach are highlighted in red.}
    \label{fig:treci2020a-results-plot}
\end{figure*}

\textbf{Ex3: Results at TREC-IS track}\label{subsec:is-perf}

The MTL-based runs have been found to have an advantage over BERT-based and LR-based STL baselines. A simple ensemble technique has also been proposed, which combines the individual MTL runs to make predictions for information types and priority levels. This ensemble technique improves performance compared to the individual runs. In this section, the performance of this approach is compared to state-of-the-art methods by comparing it to participating runs in the TREC IS track of 2020 (Edition A and B) and 2021 (Edition A and B).

%

In IS 2020 both Edition A and B, the track proposed two tasks: Task~1 and Task~2. The only difference between Task~1\footnote{The results reported so far are from Task~1.} and Task~2 is that Task~2 uses a reduced set of 12 information type labels, which includes 11 important information types\footnote{The information types include the 6 actionable information types in Table~\ref{tab:actionable-versus-non} plus Request-InformationWanted, CallToAction-Volunteer, Report-FirstPartyObservation, Report-Location and Report-MultimediaShare.} from the 25 used in Task~1 (see Table~\ref{tab:actionable-versus-non}) with the remaining 14 combined into the single category ``Other-Any''. Thus Task~2 emphasises a run's performance in identifying the information types that are most closely related to emergency response. Due to the common features shared between the two tasks, any runs submitted to Task~1 are also evaluated in Task~2.

In the 2020-A edition of TREC-IS, several runs were submitted, one of which was an MTL-based run called \texttt{Our\_Run1} that is similar to the \texttt{BERT\_base+MTL} run\footnote{The difference is that \texttt{Our\_Run1} is trained on the entire training set without first removing the development set used in the \texttt{BERT\_base+MTL}.}. Figure~\ref{fig:treci2020a-results-plot} plots the returned results of \texttt{Our\_Run1} and the top participating runs, which are evaluated for both Task~1 and Task~2. The performance of \texttt{EnsembleD} is also included for comparison to the submitted runs. Although \texttt{EnsembleD} was not officially submitted, it was subsequently evaluated using the official evaluation script provided by IS. The plotted results present the performance of the top participating runs from the perspective of four aspects: Ranking, Alert Worth, Information Feed Categorisation and Prioritisation. To view the full results of all participating runs, the reader is encouraged to refer to the track's overview paper~\cite{mccreadie2020trec}.

When examining the participating runs, it appears that they frequently achieve high scores in some metrics at the cost of lower scores in others. For example, the participating runs tagging with ``elmo'' relatively outperform \texttt{Our\_Run1} in Alert Worth (Figure~\ref{fig:task1-aaw} and \ref{fig:task2-aaw}) but fall far behind in Ranking and Information Feed Categorisation. In contrast, the participating runs tagged with ``sub'' achieve good scores that are near to \texttt{Our\_Run1} in Information Feed Categorisation (Figure~\ref{fig:task1-infofeed} and~\ref{fig:task2-infofeed}) but not in the Alert Worth metrics. Despite the loss in Alert Worth to the \texttt{elmo} runs, \texttt{Our\_run1} outperforms the top participating runs in the rest of metrics for both Task~1 and Task~2 with the only exception of a marginal loss to one \texttt{elmo} run in Task~1 Prioritisation (Figure~\ref{fig:task1-priority}). \texttt{Our\_run1} also achieves the highest HarM score, which implies overall best performance. However, there remains a strong argument that in practical terms different submitted runs are preferable in different situations, depending on the needs of the emergency responders, and that no overall best-performing system has been satisfactorily identified.

The \texttt{EnsembleD} run, however, achieves state-of-the-art performance in almost every metric, substantially outperforming the participating runs in most cases. Figure~\ref{fig:treci2020a-results-plot} indicates that there are only two evaluation figures (other than the less important Cacc discussed previously) where \texttt{EnsembleD} does not have the highest performance, with the difference being minor in both cases (the CF1-A from Figure~\ref{fig:task1-infofeed} and PErr-H from~\ref{fig:task1-priority}).

Given the tendency of participating runs to achieve imbalanced performance across the individual metrics, the \texttt{EnsembleD} stands out as a good choice with regards to its effectiveness in different aspects of emergency response, across the range of metrics. 

\begin{figure*}[!h]
     \centering
     \begin{subfigure}[b]{0.45\textwidth}
         \centering
         \includegraphics[width=\textwidth]{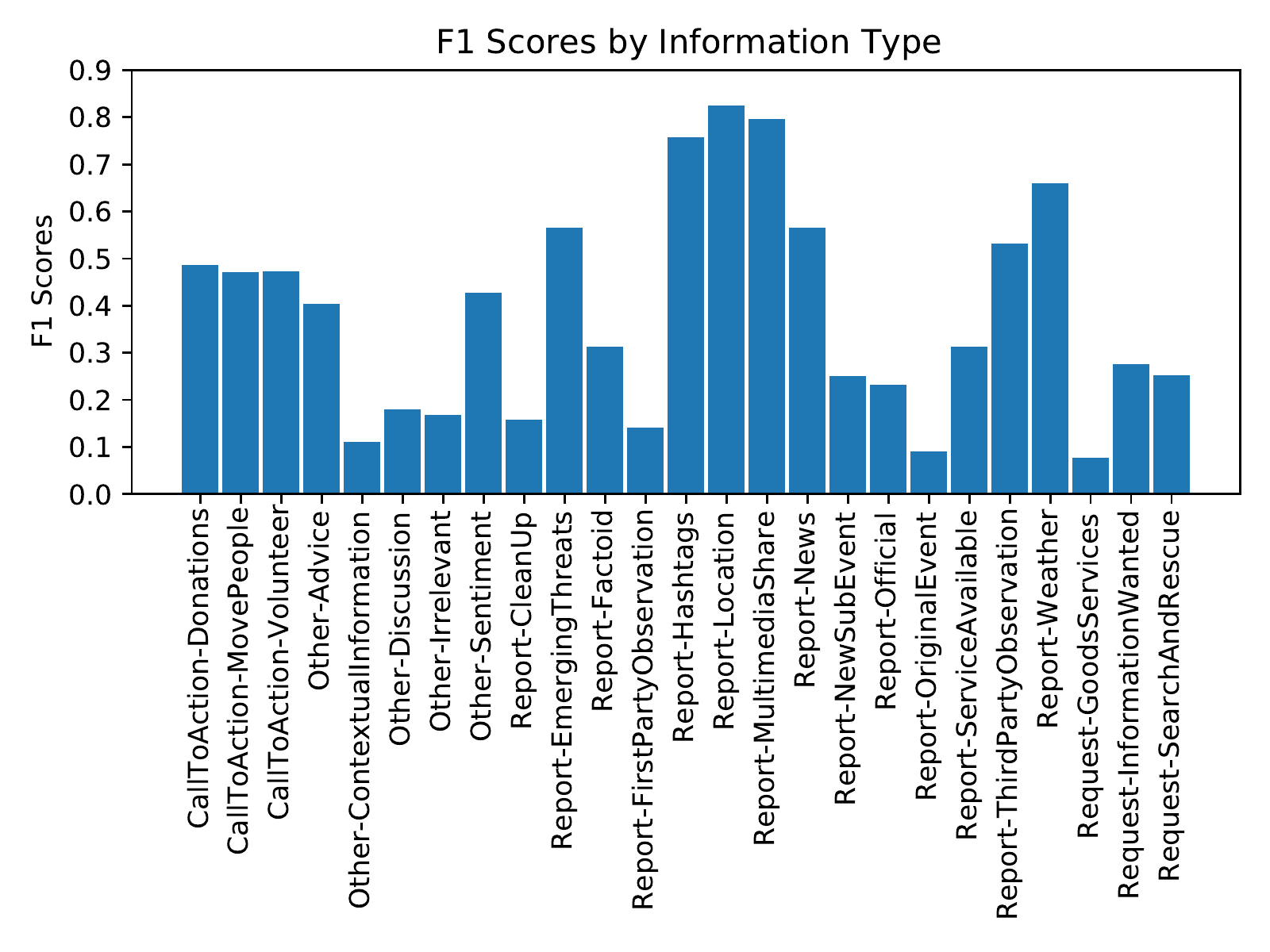}
         \caption{F1 Scores by IT}
         \label{fig:trecis-2020b-scores-per-it}
     \end{subfigure}
     \begin{subfigure}[b]{0.45\textwidth}
         \centering
         \includegraphics[width=\textwidth]{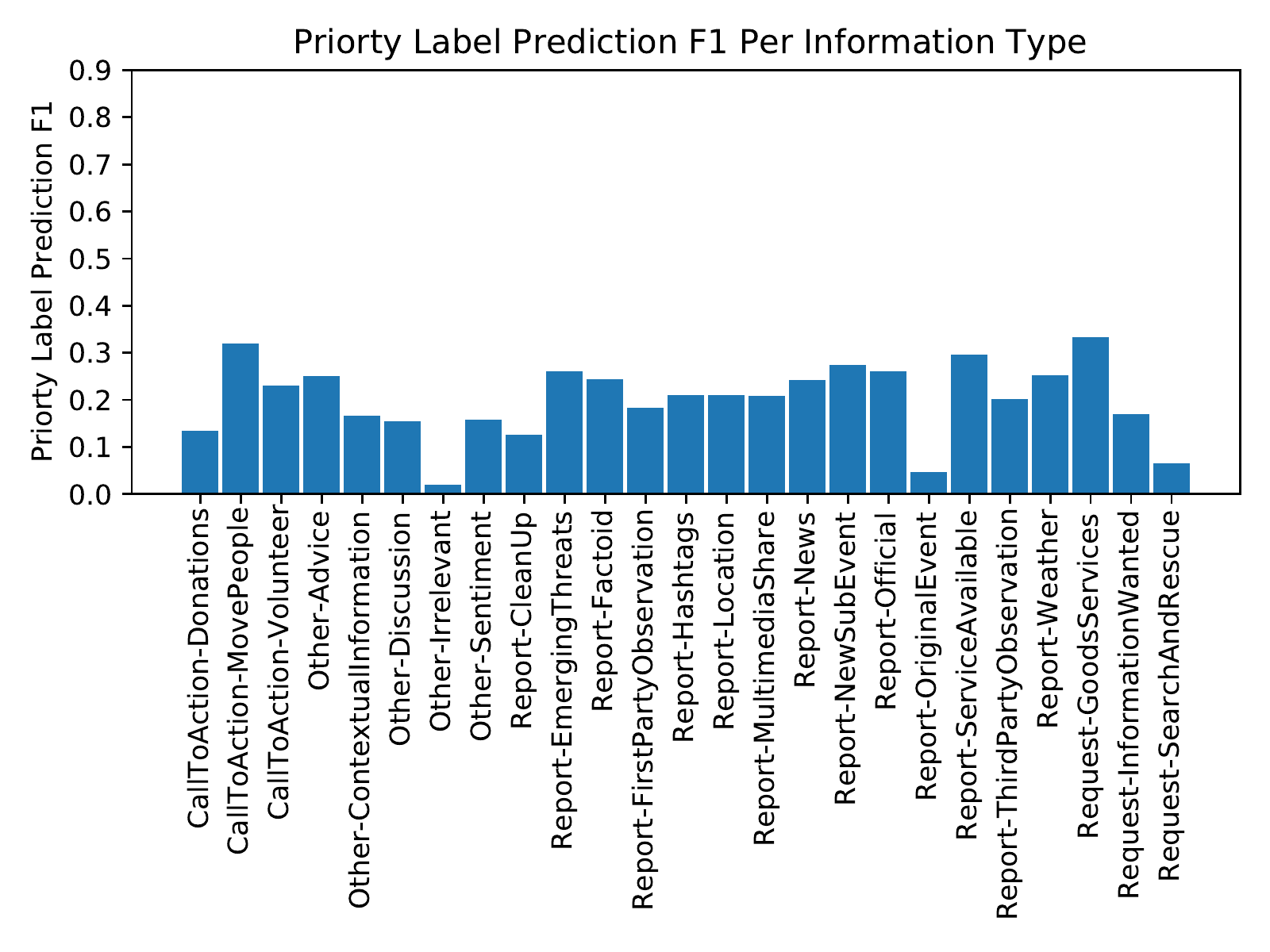}
         \caption{Priority level prediction F1 per IT}
         \label{fig:trecis-2020b-pri-per-it}
     \end{subfigure}
        \caption{Performance visualisation by information types (ITs) of \textbf{ucd-run1} in IS 2020B Task 1.}
        \label{fig:trecis-2020b-perf-vis-run1}
\end{figure*}

Having established the effectiveness of the approach in IS 2020A, it is further tested in further editions of the IS track. Table~\ref{tab:trecis-2020b-res-task1} and~\ref{tab:trecis2020b-res-task2} present the results of participating runs at IS 2020 Edition B Task 1 and 2 respectively, where \texttt{EnsemblD} (named as ``ucd-run1'' in the official submission of this edition) is compared to other participating runs. As can be seen from Table~\ref{tab:trecis-2020b-res-task1}, in Task~1, the MTL-based ensemble run substantially outperforms other participating runs in both information type classification and prioritisation~\footnote{The exception is accuracy, where only a small difference is observed across the participating runs: the MTL-based ensemble run is substantially higher than other participating runs in the remaining metrics.}. In particular, the run is effective in classifying actionable information types. For example, it achieves the top actionable information type F1 score of 0.3215 (CF1-H) and the best actionable priority F1 of 0.2009 (PErr-A). This is further evidenced by the run's performance in Task~2, as shown in Table~\ref{tab:trecis2020b-res-task2}, where the run outperforms the participating runs in every metric. To further examine the run's performance at the level of individual information types, the information type F1 scores and priority F1 scores per information type of the run are reported in Figure~\ref{fig:trecis-2020b-perf-vis-run1}. Figure~\ref{fig:trecis-2020b-scores-per-it} shows that the run performs well in categorising some actionable information types, such as ``CallToAction-MovePeople'' and ``Report-EmergingThreats'' but performs less well in other actionable information types such as ``Request-GoodsService'', as compared to the non-actionable information types. However, examining the priority F1s per information type in Figure~\ref{fig:trecis-2020b-pri-per-it}, it can be seen that the run performs relatively better in priority level prediction for actionable information types than non-actionable information types, where ``CallToAction-MovePeople'', ``Request-GoodsService'' and ``Report-ServiceAvailable'' are the top 3 that the run achieves in priority F1.

\begin{table*}[]
\scriptsize
\centering
\begin{tabular}{llp{2cm}p{1.5cm}p{1.5cm}p{1.9cm}p{1.5cm}}

\toprule
Run             & \textbf{nDCG}        & \textbf{CF1-H} & \textbf{CF1-A} & \textbf{Caac} & \textbf{PErr-H}  & \textbf{PErr-A} \\
\midrule
                  & \multicolumn{6}{c}{Participating runs} 
                  \\
BJUT-run        & 0.4346          & 0.0266          & 0.0581          & 0.8321          & 0.1744          & 0.0905          \\
njit.s1.aug     & 0.4480          & 0.2634          & 0.3103          & \textbf{0.8655} & 0.2029          & 0.1518          \\
njit.s2.cmmd.t1 & 0.4475          & 0.1879          & 0.2223          & 0.8475          & 0.2029          & 0.1518          \\
njit.s3.img.t1  & 0.4222          & 0.1879          & 0.2223          & 0.8475          & 0.1959          & 0.1417          \\
njit.s4.cml.t1  & 0.4164          & 0.1712          & 0.1465          & 0.8445          & 0.1054          & 0.1064          \\

ufmg-sars-test  & 0.3634          & 0.0001          & 0.0493          & 0.8337          & 0.1285          & 0.1378  \\
\midrule
EnsembleD (ucd-run1)        & \textbf{0.5033}          & \textbf{0.3215} & \textbf{0.3810} & 0.8520          &\textbf{0.2582}          & \textbf{0.2009}          \\
\bottomrule
\end{tabular}
\caption{Evaluation results of participating runs at TREC-IS 2020-B Task 1}
\label{tab:trecis-2020b-res-task1}
\end{table*}

\begin{table*}[]
\centering
\small
\begin{tabular}{lllll}
\toprule
Run               & \textbf{nDCG}        & \textbf{CF1-H} & \textbf{Caac} & \textbf{PErr-A} \\
\midrule\midrule
                  & \multicolumn{4}{c}{Participating runs} 
                  \\
BJUT-run        & 0.4350          & 0.0472          & 0.7977          & 0.1337          \\
njit.s1.aug     & 0.4487          & 0.3480          & 0.8846          & 0.1838          \\
njit.s2.cmmd.t1 & 0.4467          & 0.2494          & 0.8612          & 0.1838          \\
njit.s3.img.t1  & 0.4215          & 0.2494          & 0.8612          & 0.1708          \\
njit.s4.cml.t1  & 0.4176          & 0.1278          & 0.8360          & 0.1162          \\

ufmg-sars-test  & 0.3630          & 0.0127          & 0.8419          & 0.1480     \\
\midrule
EnsembleD (ucd-run1)        & \textbf{0.5020}          & \textbf{0.4036} & \textbf{0.8913   }       & \textbf{0.2320}          \\
\bottomrule
\end{tabular}

\caption{Evaluation results of participating runs at TREC-IS 2020-B Task 2}
\label{tab:trecis2020b-res-task2}
\end{table*}
\begin{table*}[]
\centering
\small
\begin{tabular}{lllllll}
\hline
                     & \textbf{nDCG} & \textbf{CF1-H} & \textbf{CF1-A} & \textbf{Caac} & \textbf{PErr-H} & \textbf{PErr-A}  \\
                     \hline
run1          & \textbf{0.6115} & 0.215  & 0.2951          & 0.8837          & 0.3032         & 0.3068         \\
mtl.ens.new & 0.5907          & 0.2579 & \textbf{0.3211} & 0.8646          & 0.3052         & 0.3125          \\
\hline
med                  & 0.5695          & 0.206  & 0.2823          & 0.8827          & 0.2113         & 0.2175          \\
max                  & \textbf{0.6115} & \textbf{0.2815} & \textbf{0.3211} & \textbf{0.8902} & \textbf{0.306} & \textbf{0.3211}  \\
\hline
\end{tabular}
\caption{The performance of the submitted runs at IS 2021A where the med and max rows present the median and maximum scores of each metric respectively across all participating runs.}
\label{tab:trecis-2021a-results}
\end{table*}
\begin{table*}[]
\centering
\small
\begin{tabular}{lllllll}
\hline
                     & \textbf{nDCG} & \textbf{CF1-H} & \textbf{CF1-A} & \textbf{Caac} & \textbf{PErr-H} & \textbf{PErr-A}  \\
                     \hline
run1          & 0.4499        & 0.2177                 & 0.247                    & 0.8966           & 0.2376                  & 0.2566                                   \\
mtl.ens.new    & 0.4555        & \textbf{0.251}         & \textbf{0.2623}          & 0.8753           & 0.2783                  & 0.2703                                       \\
\hline
med & 0.4272 & 0.1842 & 0.233 & 0.8947 & 0.2107 & 0.2031  \\
max         & \textbf{0.4791}       & \textbf{0.251}       & \textbf{0.2623}          & \textbf{0.9067}  &\textbf{ 0.2798}         & \textbf{0.2756 }                       \\
\hline
\end{tabular}
\caption{The performance of the submitted runs at IS 2021B where the med and max rows present the median and maximum scores of each metric respectively across all participating runs.}
\label{tab:trecis-2021b-results}
\end{table*}

In the last year of the TREC IS track, 2021, it changed from two tasks to only Task~1. In IS 2021, the MTL and ensemble methods with some variations are further tested over two editions of the track: 2021A and 2021B. Table~\ref{tab:trecis-2021a-results} and~\ref{tab:trecis-2021b-results} show the performance of the submitted runs compared to the median and maximum scores across all participating runs\footnote{Full results can be found in the overview paper~\cite{buntain2022incident}.}. The \texttt{run1} is simply the re-run version of the MTL method and \texttt{mtl.ens.new} is the re-run version of the MTL-based ensemble method. The difference of the two runs compared to \texttt{BERT\_base+MTL} and \texttt{EnsembleD} is that they use an optimised  BERT variant DeBERTa~\cite{he2020deberta} instead of the original BERT~\cite{BERT2018}. The performance of the two runs was consistent with previous editions, achieving the best or near-best scores in both the information type classification and priority estimation tasks. This offers strong evidence of the effectiveness of the MTL approach for crisis message categorisation in the context of many-to-to-many domain adaptation.

\subsubsection{Error analysis}

Even if EnsembleD and MTL-based runs perform well in many ways, it is still important to understand the many kinds of faults the system makes. Using the top EnsembleD run on the information type classification task, a qualitative error analysis was carried out to offer insights and aid the community in resolving issues in this area. Examples of incorrect information type predictions made by EnsembleD are shown in Table~\ref{tab:error-analysis-IT} where ``IT prediction'' denotes the set of information types assigned by EnsembleD and ``IT ground truth'' lists the information types assigned by human assessors as part of the IS track.

\begin{table*}[]
    \centering
    \footnotesize
\begin{tabular}{llp{4cm}p{3cm}p{3cm}}
\toprule
Id & Event                    & Text                                                                                                                                                                                               & IT prediction                                                                                               & IT ground truth                                                                                                       \\
\midrule
\# 1       & gilroygarlicShooting2020 & 6-year-old killed in Gilroy Garlic Festival Shooting https://t.co/MYgvoneYdC                                                                                                                       & Report-Factoid, Report-Location, Report-MultimediaShare, Report-News, Report-ThirdPartyObservation        & Report-Factoid, Report-Location, Report-MultimediaShare, Report-News                                           \\

\hline
\# 2       & gilroygarlicShooting2020 & Volunteers from Muslim-faith based charity @pennyappeal have turned up at Chapel school with loads of supplies for people evacuated from \#WhaleyBridge. Good on ya, lads. https://t.co/eBMM3evPAL & CallToAction-Donations, Report-Hashtags, Report-Location, Report-MultimediaShare, Report-ServiceAvailable & Report-Hashtags, Report-Location, Report-MultimediaShare, Report-ServiceAvailable, Report-ThirdPartyObservation \\
\hline
\# 3       & hurricaneBarry2020       & Rain water at Mississippi and Santa fe \#Denver @9NEWS https://t.co/a8eekFIODx                                                                                                                     & Report-Hashtags, Report-Location, Report-MultimediaShare, Report-News                                    & Other-Irrelevant                                                                                            \\
\hline
\# 4       & baltimoreFlashFlood2020  & I'm at Bronycon 2019 in Baltimore, MD https://t.co/oVXxmZ4JID                                                                                                                                      & Other-Irrelevant                                                                                      & Request-SearchAndRescue                                                                                     \\
\bottomrule
\end{tabular}
    \caption{Examples of error analysis for information types}
    \label{tab:error-analysis-IT}
\end{table*}

The first two examples show how the information type ``Report-ThirdPartyObservatio'' is handled. EnsembleD chose this information type in the first example, but it was not present in the ground truth information types. In the second example, EnsembleD did not select this information type, but the ground truth did. It is debatable whether all tweets containing crisis information should be labelled as first-party or third-party observations. That said, the image associated with \#2 suggests that this may be a first-party observation, with the author of the tweet posting an image of the volunteers mentioned. It's also worth noting that EnsembleD chose the ``CallToAction-Donation'' information type. Although this tweet does not explicitly call for action, the Twitter account referenced (@pennyappeal) is a charity appeal for crisis situations. Both of these observations indicate that context and subtlety make classification of information types difficult.


In example \#3, the geographical context is critical. The human assessor deemed it irrelevant because it relates to Denver, Colorado, whereas Hurricane Barry's main effects were felt primarily in Louisiana. The image attached depicts flood water during the same time period as the hurricane in question. As a result, many of the information types chosen by EnsembleD would have been correct if it had been within the geographical area in question. This example also highlights another challenge in terms of geographical context. Without context, ``Mississipp'' could refer to a US state or a river, and ``Santa Fe'' is the capital city of the US state of New Mexico. However, in this particular example, the names refer to street names in Denver, Colorado, where the image of floodwater was taken. Although this was not the reason for this particular tweet's misclassification, it does point to a new challenge, implying that language models alone will not be sufficient to solve this problem, and that the incorporation of knowledge maps and other ontologies may be required to provide the necessary context to information type classification systems.


 The last example appears to merely illustrate human inaccuracy in the classification of information types. This tweet refers to the user's attendance at a conference that was over two days before the flash flood being discussed. This highlights the drawbacks of using human assessors as comparison sources. Personal preference, external context, and obvious mistakes make it crucial to analyse systems qualitatively in addition to presenting evaluation metrics.
 

\section{Conclusions}

In this chapter, the research in many-to-many domain adaptation was reported, which was based on the TREC IS track that proposed an information type task and a priority estimation task for categorising crisis-related tweets from many unseen events. At the early stage of this track, the proposed methods were based on single-task learning (STL), which trained machine learning models or simple neural network models separately for the two tasks. Having realised the limited effectiveness of these methods, this research took a further step to explore better solutions as the track evolved. Ultimately, a multi-task learning (MTL) approach that trained a Transformer encoder-based pre-trained language model (PLM) jointly for the two tasks was proposed. By testing this approach in many subsequent IS editions, it was found to achieve strong performance compared to other participating runs. In particular, a simple ensemble technique that leveraged multiple multi-task learners was found to achieve the overall best performance among participating runs. In addition, a qualitative analysis of the predictions made by the MTL-based ensemble system revealed that language models alone may not be sufficient for categorising crisis tweets, as subtle factors such as implicit mentions of charity agencies or ambiguous place names can make the system challenging. 


So far, the methods have been developed to build models trained on the entire corpus of training tweets from various distinct events and test them on the test tweets of many new events, thus corresponding to many-to-many crisis domain adaptation. This adaptation work can be useful in addressing the needs of users who want to apply computational techniques for crisis message categorisation without needing to know the specific types of source and target events. However, in reality, it is also of interest to determine how to select source events for a new event based on their characteristics, specifically the one-to-one or many-to-one crisis domain adaptation problems. In the next chapter, the work of using event-aware sequence-to-sequence PLMs for this purpose of adaptation is presented.

\chapter{Many-to-one and one-to-one Crisis Domain Adaptations}
\label{ch:adaptation2}

\section{Introduction}

The work into this many-to-many adaptation helps to meet the user needs when building a computational model for crisis message categorisation without needing to know the specific types of the source and target events (Chapter~\ref{ch:adaptation}). However, in reality, given a choice of source events, it is also interesting to know how to select appropriate source events for a new event based on their characteristics: the one-to-one and many-to-one adaptation problems. For this purpose, this chapter introduces CAST, a method that utilises pre-trained sequence-to-sequence language models for crisis domain adaptation. The details of CAST are presented in Section~\ref{sec:cast}, which includes the methodology in Section~\ref{subsec:cast-method}, the experimental setup in Section~\ref{subsec:cast-exps-setup}, and discussions of its effectiveness in both many-to-one and one-to-one adaptation settings in Section~\ref{subsec:cast-exps-findings}.

The work presented in this chapter has previously been presented in published work~\cite{Wang2021a}.

\section{Many-to-one and one-to-one adaptations}

\label{sec:cast}

This section presents work using sequence-to-sequence (seq2seq) PLMs for one-to-one and many-to-one crisis domain adaptations, which is abbreviated to CAST. Firstly, this work aims to test the effectiveness of CAST by comparing it to existing adaptation approaches. Secondly, referring to a crisis event as a domain, another important objective of this work is to explore the selection of source domains for adaptation to a target domain, which offers insights on how to efficiently use source data for model training.

In comparison to related work for crisis domain adaptation~\cite{li2018disaster, alam2018domain,liu2020crisisbert} (Section~\ref{sec:cda}), CAST uses only labelled source data without any unlabelled target data, i.e., it is target data independent. This makes CAST more suited to real-world use cases. This is because a crisis usually focuses on a specific aid-related topic at a certain stage. For example, an earthquake is normally more about ``Emerging Threats'' than ``Donation'' at early stages. Hence, when a good-quality representation of the target event distribution is needed, the unlabelled target data needs to be collected as the target crisis unfolds, before training and classification can take place. Instead of using target data, CAST fine-tunes a seq2seq PLM for crisis domain adaptation by adding a task description and an event description to the input sequences that belong to the event, which makes the model aware of the event that the input sequences relate to. The details of CAST are described as follows.

\subsection{Event-aware sequence-to-sequence pre-trained language models}
\label{subsec:cast-method}

\begin{figure}[h!]
    \centering
    \includegraphics[scale=
    0.9]{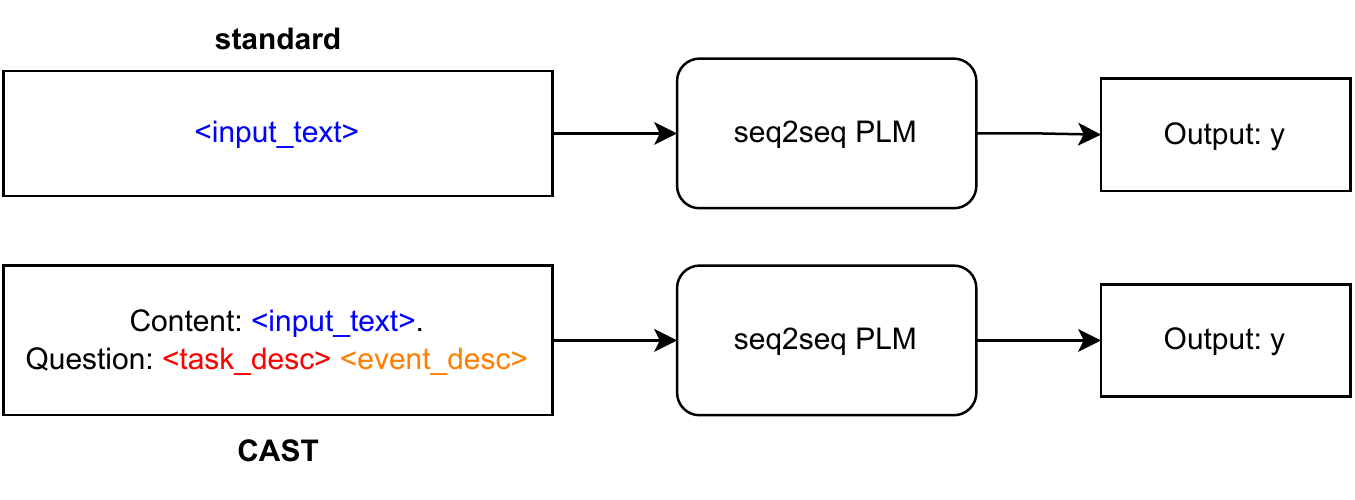}
    \caption{The standard and CAST approaches for crisis domain adaptation leveraging seq2seq PLMs.}
    \label{fig:cast_approach_overview}
\end{figure}

In this work, a Transformer decoder-encoder based model T5~\cite{raffel2019exploring} is used as an example of a seq2seq PLM. Unlike the standard way of fine-tuning a seq2seq model for a downstream task (Section~\ref{sec:encoder-decoder-models}), CAST is simple and trained by using a task description and/or an event description added to the input sequence (i.e., a crisis-related message), as illustrated in Figure~\ref{fig:cast_approach_overview}.  The task description essentially refers to a question-form natural language text specifying the target task and the event description refers to basic information of a crisis event including the event name and its occurrence location. Taking an input crisis message from 2015 Nepal earthquake in the task of informativeness classification as an example, the task description and the event description can be formed as ``Is this message informative to 2015 Nepal earthquake?''.

Mathematically, in the standard way of fine-tuning a seq2seq model for a downstream task, the input sequence $s\colon\{t_1,t_2,\ldots,t_T\}$ consists solely of the textual content itself. CAST is specifically proposed for crisis domain adaptation, taking into account both a task description $\mathrm{T_{desc}}$ and an event description $\mathrm{E_{desc}}$, leading to the new input $\hat{s}\colon\{\hat{t}_1,\hat{t}_2,\ldots,\hat{t}_{T'}\}$, formulated as follows.

\begin{equation}
    \label{eq:new_x}
\hat{s}_{1:T'} = \zeta_{\oplus}\left(s_{1:T}, \mathrm{T_{desc}}, \mathrm{E_{desc}}\right)
\end{equation}

Where $\zeta_{\oplus}$ is the reconstruction function that concatenates $s$ with $\mathrm{T_{desc}}$ and $\mathrm{E_{desc}}$ in natural language form. By this reconstruction, now the objective/loss function for training the seq2seq model is changed from Equation~\ref{eq:seq2seq_overall_loss} to be:


\small
\begin{equation}
\label{eq:seq2seq_overall_loss}
 J(\hat{s},y)= - \hat{p}(y | f(\hat{s}_{1:T'};\theta_{e});\theta_{d})
\end{equation}
\label{eq:seq2seq-general}
\normalsize

Here $y$ refers to the output label text in response to $\mathrm{T_{desc}}$ and $\mathrm{E_{desc}}$ in the input, e.g., ``yes'' or ``no'' to ``Is this message informative to 2015 Nepal earthquake''. As described, CAST differs from the standard approach in two main aspects. First, it considers $\mathrm{T_{desc}}$, which is inspired by prior work~\cite{wang-lillis-2020-ucd} using a task description in the input example for a COVID-related event extraction task. In addition, CAST considers $\mathrm{E_{desc}}$ for making the model domain-aware when tested on different events.

\subsection{Experiments setup}
\label{subsec:cast-exps-setup}

This section describes the experimental setup for the evaluation of CAST including benchmark datasets, adaptation scenarios and training details.



\subsubsection{Datasets}




%

 In this work, the datasets used in the experiments include \textbf{nepal\_queensland}~\cite{alam2018domain} and \textbf{CrisisT6}~\cite{olteanu2014crisislex}. These are two benchmark informativeness classification datasets consisting of crisis tweets from two and six events respectively as described in Section~\ref{sec:rds-ems-infos}. The two datasets are selected and used in this work since the binary informativeness classification task is compatible with the proposed method and they have equal sizes of instances across different crisis events. Considering that both datasets relate to the informativeness classification task, their labels are unified to the same target labels $y$, without altering the fundamental requirements of the task (i.e. to determine whether a message is informative with respect to a particular crisis event). In \textbf{nepal\_queensland}, \textit{relevant} is changed to \textit{yes} and  \textit{not\_relevant} is changed to \textit{no}, likewise for \textbf{CrisisT6}. Following this unification, the task description $\mathrm{T_{desc}}$ becomes the same for the two datasets. 

As \textbf{nepal\_queensland} releases splits of training, development and test sets, the same splits are used in the experiments for its in-domain adaptation. For cross-domain adaptation, the training and development sets of source events are used for model training and the test sets of target events are used for testing. As \textbf{CrisisT6} releases only a data set for each event, 5-fold cross validation is used in the experiments for its in-domain adaptation. For cross-domain adaptation, the data sets of source events are used for model training and the data sets of target events are used for testing.

\subsubsection{Adaptation scenarios}

Using the datasets, the experiments apply two adaptation scenarios for model training, as outlined in Figure~\ref{fig:cast_approach_overview}.

   \textbf{standard}: This scenario represents common practice in the literature for fine-tuning seq2seq Transformers on a downstream task, and is used as a baseline in the experiments. This scenario simply feeds the raw training tweets to the seq2seq model without any additional text being added. Here, a training tweet no matter what crisis event it belongs to is always constructed in the form: ``\textit{\{tweet\_text\}}''.
    
    \textbf{postQ}: This scenario represents a particular use case of the CAST method. Referring back to Equation~\ref{eq:new_x}, postQ takes into account $\mathrm{T_{desc}}$, $\mathrm{E_{desc}}$ and $\zeta_{\oplus}$. In this scenario, $\mathrm{T_{desc}}$ becomes ``\textit{Is this message relevant to}'' and $\mathrm{E_{desc}}$ becomes ``\textit{\{location\_name\}  \{crisis\_name\}}'', which are available in the selected datasets. For example, in \textbf{nepal\_queensland}, $\mathrm{E_{desc}}$ becomes ``\textit{Nepal Earthquake}'' or ``\textit{Queensland Floods}''. Finally, the function $\zeta_{\oplus}$ constructs the input sequence to be in the form of a question-answering sequence: ``\textit{Content: \{tweet\_text\}. Question: Is this message relevant to \{location\_name\}  \{crisis\_name\}?}''\footnote{For in-domain adaptation, the location name and crisis name are from the source event(s) at both training and testing time. For cross-domain adaptation, the location name and crisis name are from the source event(s) at training time and from the target event at testing time.}. This is determined by testing different variants of the task and event descriptions in a pilot study with CrisisT6, summarised as follows.
     


\begin{itemize}
    \item \textbf{variant 1}. This is similar to postQ except that \textit{\{location\_name\}} is removed from $\mathrm{E_{desc}}$, making the input location-agnostic to the model. The final input sequence is constructed like: ``\textit{Content: \{tweet\_text\}. Question: Is this message relevant to \{crisis\_name\}?}''. 
    \item \textbf{variant 2}. This is similar to postQ except that \textit{\{location\_name\}} and \textit{\{crisis\_name\}} are re-arranged such that the input is like: ``\textit{Content: \{tweet\_text\}. Question: Is this message relevant to a \{crisis\_name\} event that occurred in \{location\_name\}?}''.
    \item \textbf{variant 3}. This variant constructs the input sequence by setting  $\mathrm{T_{desc}}$ to be empty such that the input is like: ``\textit{Content: \{tweet\_text\}. Question: \{location\_name\} \{crisis\_name\}?}''.
\end{itemize}

The experimentation on these variants and postQ did not present any noticeable difference in performance. It is interesting to notice that there is no performance difference between variant 3 and postQ where variant 3 simply uses location and crisis name without including $\mathrm{T_{desc}}$ in the extended text. This is because given a specific classification task, $\mathrm{T_{desc}}$ will be the same for all training examples, thus leading to no difference to the model. However, ultimately the variant with $\mathrm{T_{desc}}$ (i.e., postQ) is chosen in the subsequent experiments mainly because this leaves room for expanding CAST to multi-task learning settings where $\mathrm{T_{desc}}$ becomes different for different tasks.

\subsubsection{Training details}

The experiments are conducted to examine the performance of the above two scenarios in many-to-one and one-to-one in-domain and cross-domain adaptations through fine-tuning seq2seq PLMs on the two benchmark datasets. Given a number of existing such seq2seq models (Section~\ref{sec:encoder-decoder-models}), the Transformer decoder-encoder based PLM T5~\cite{raffel2019exploring} is used as the target model in this study due to its availability of small and medium-sized pre-trained weights and its strong performance in various downstream language tasks. To be specific, the off-the-shelf \texttt{t5-small} and \texttt{t5-base} weights implying different model sizes are used in this study, which are abbreviated to \texttt{small} and \texttt{base}\footnote{\texttt{t5-small} and \texttt{t5-base} have around 60M and 220M parameters respectively. There are larger versions originally released by the authors, such as \texttt{t5-large}, \texttt{t5-3B} and \texttt{t5-11B}, these are not included given the scope of this research focuses on small and medium-sized PLMs.}. In fine-tuning, the learning rate is set to be $5e\text{-}05$ using Adam optimizer~\cite{kingma2014adam}, updated by a linear decay scheduler with warmup ratio $10\%$ of the total training steps (12 epochs)., which is based on the work in a similar domain~\cite{wang-lillis-2020-ucd}.



\subsection{Discussions and Findings}
\label{subsec:cast-exps-findings}

Having conducted extensive experiments with the two selected benchmark datasets, this section reports and discusses the results of CAST compared to baselines in many-to-one and one-to-one adaptations (including both in-domain and cross-domain). In addition, the insights on how to select source data for model training in many-to-one and one-to-one adaptations are presented. Specifically, in the context of real-world crisis domain adaptation, the study of one-to-one adaptation aims to identify the most suitable among a number of candidate source events (for which training sets are available) for adaptation to a new emerging target event (for which training data, either labelled or unlabelled, is not yet available). Similarly, the study of many-to-one adaptation investigates which combination of available source events is most suited to be adapted to an emerging target event.


    

\subsubsection{One-to-one adaptation}
\label{sec:one-to-one}

\textbf{Adaptation between two events}

Table~\ref{tab:nepal-queensland-in-domain} and~\ref{tab:nepal-queensland-cross-domain} present the in-domain and cross-domain performance respectively on the \textbf{nepal\_queensland} dataset across different runs. Regarding the in-domain performance, the standard and postQ runs are compared with the baseline CNN run~\cite{alam2018domain}. It is found that they substantially outperform this baseline for both the Nepal Earthquake and Queensland Floods events. For cross-domain adaptation, two baselines CNN+DA+GE~\cite{alam2018domain} and RNN+AE~\cite{li2020domain} that use unlabelled target data in training, i.e., target-data-dependent (TDD), are included for comparison. For target data independent (TDI) runs, apart from standard and postQ, two baselines CNN+DA~\cite{alam2018domain} and RNN~\cite{li2020domain} are reported. To compare standard and postQ with the TDD baselines, they substantially outperform the CNN+DA+GE run. When it comes to the state of the art TDD RNN+AE run, the postQ-base achieves competitive performance, with 87.06 versus 81.18 in NE$\boldsymbol{\rightarrow}$QQF and 64.12 versus 68.38 in QQF$\boldsymbol{\rightarrow}$NE (although the performance is less, it is important to note that the postQ as a TDI approach does not rely on any target data). However, the postQ-base outperforms the state of the art TDI RNN run in cross-domain adaptation with 87.06 versus 55.17 in NE$\boldsymbol{\rightarrow}$QQF and 64.12 versus 64.18 in QQF$\boldsymbol{\rightarrow}$NE. To compare the standard with the postQ in in-domain adaptation (see Table~\ref{tab:nepal-queensland-in-domain}), it is found that postQ performs very similarly to the standard. This makes sense since postQ is only different from standard in appending an extra task and event description. For in-domain adaptation, the appended text is the same for all training examples during training and at inference time, thus leading to no difference for the model when the extra text exists or not, which explains why they have the same level of performance. Interestingly for cross-domain adaptation, postQ does not achieve performance better than the standard method when using a small model. When using a bigger model (i.e., postQ-base), it seems that postQ performs much better than the standard method (see Table~\ref{tab:nepal-queensland-cross-domain}). This finding is further verified in the subsequent experiments on the \textbf{CrisisT6} dataset.

\begin{table}[!h]
\centering
\normalsize
\begin{tabular}{lrr}
\toprule
                        & \multicolumn{1}{l}{NE$\boldsymbol{\rightarrow}$NE} & \multicolumn{1}{l}{QQF$\boldsymbol{\rightarrow}$QQF} \\
                        \midrule
CNN~\cite{alam2018domain}  & 65.11                                   & 93.54                       \\
\textbf{standard-small} & 75.35                                   & 96.31                       \\
\textbf{standard-base}  & 74.40                                   & 96.83                       \\
\textbf{postQ-small}    & 79.25                                   & 96.34                       \\
\textbf{postQ-base}     & 77.57                                   & 96.81     \\
\bottomrule
\end{tabular}

 \caption{The in-domain adaptation weighted F1 scores for \textbf{nepal\_queensland} where \textbf{NE}: Nepal Earthquake and \textbf{QQF}: Queensland Floods. The CNN run refers to the supervised CNN run from~\cite{alam2018domain} and the runs in bold refer to the \textbf{standard} and \textbf{postQ} runs with \texttt{t5-small} and \texttt{t5-base}.}
 \label{tab:nepal-queensland-in-domain}
\end{table}

\begin{table}[!h]
\centering

\begin{tabular}{llrr}
\toprule                         
                      & source $\boldsymbol{\rightarrow}$ target      & \multicolumn{1}{l}{NE$\boldsymbol{\rightarrow}$QQF} & \multicolumn{1}{l}{QQF$\boldsymbol{\rightarrow}$NE} \\
                      \midrule
\multirow{2}{*}{TDD}  & CNN + DA + GE ~\cite{alam2018domain} & 65.92                                    & 59.05                      \\
                      & RNN + AE~\cite{li2020domain}        & 81.18                           & 68.38             \\
                      \midrule
\multirow{6}{*}{TDI} & CNN + DA ~\cite{alam2018domain}      & 60.94                                    & 57.79                      \\
                      & RNN~\cite{li2020domain}             & 55.17                                    & 64.18                      \\
                      & \textbf{standard-small}          & 82.43                                    & 58.99                      \\
                      & \textbf{standard-base}           & 77.39                                    & 60.25                      \\
                      & \textbf{postQ-small}             & 78.21                                    & 63.75                      \\
                      & \textbf{postQ-base}              & 87.06                         & 64.12   \\
                      \bottomrule
\end{tabular}
 \caption{The cross-domain adaptation weighted F1 scores for \textbf{nepal\_queensland}. The runs are presented in two categories: target data dependent (TDD) and target data independent (TDI). The CNN+DA+GE and CNN+DA refer to the CNN baseline runs in~\cite{alam2018domain} and RNN+AE and RNN refer to the RNN baseline runs in~\cite{li2020domain}.}
 \label{tab:nepal-queensland-cross-domain}
\end{table}


\textbf{Adaptation between more events}

Based on the \textbf{nepal\_queensland} dataset, some evidence of the effectiveness of the CAST-based runs (particularly for TDI cross-domain adaptation) has been identified as compared to the state-of-the-art. Next, the experiments are conducted upon the \textbf{CrisisT6} dataset that consists of six different crisis events representing a wide range of domains (details can be found in Section~\ref{sec:rds-ems-infos}).

\begin{figure*}[h!]
     \centering
     \begin{subfigure}[b]{0.4\textwidth}
         \centering
         \includegraphics[width=\textwidth]{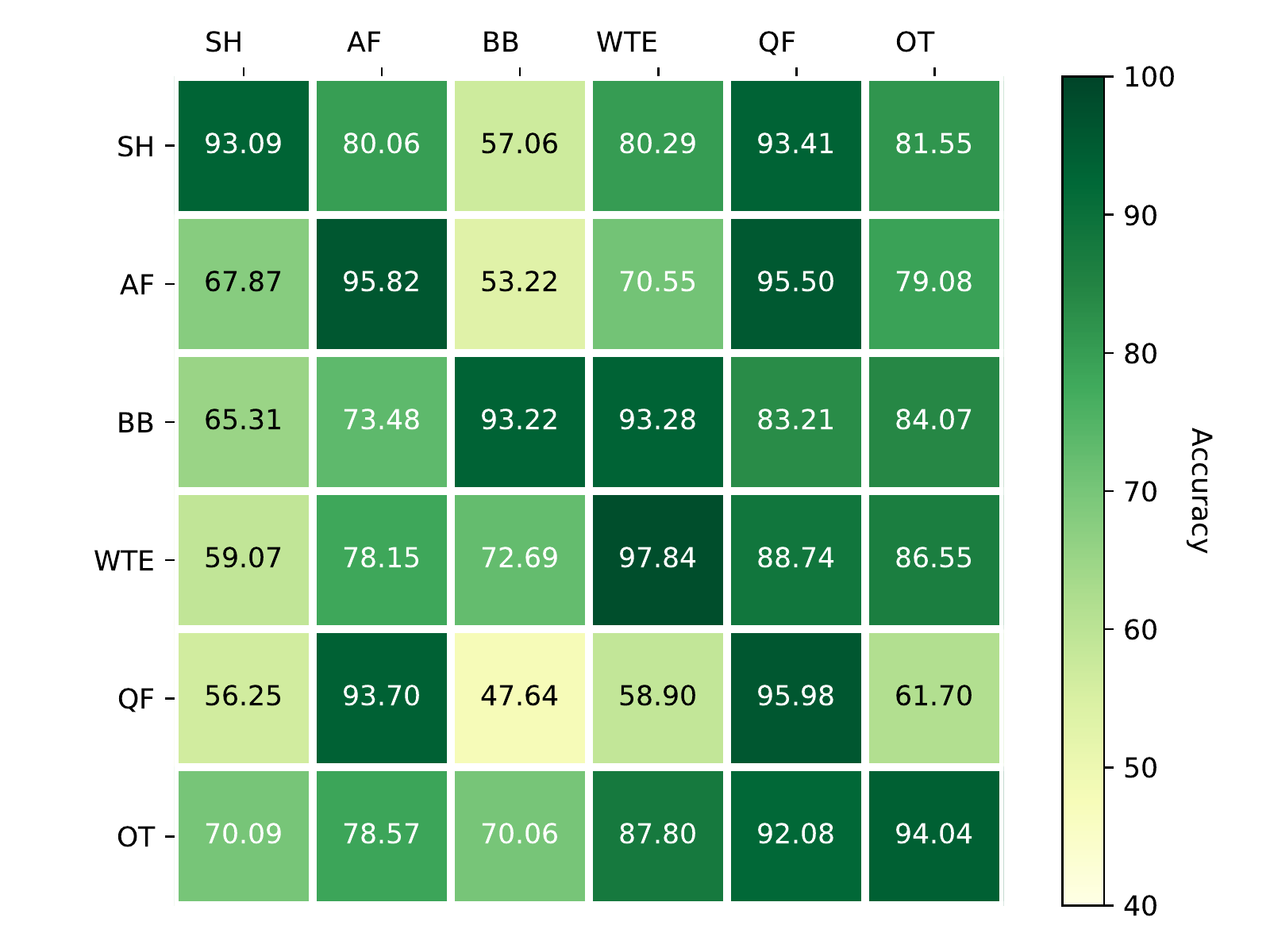}
         \caption{\textbf{small-standard}}
         \label{fig:small-crisist6-standard-acc}
     \end{subfigure}
     \begin{subfigure}[b]{0.4\textwidth}
         \centering
         \includegraphics[width=\textwidth]{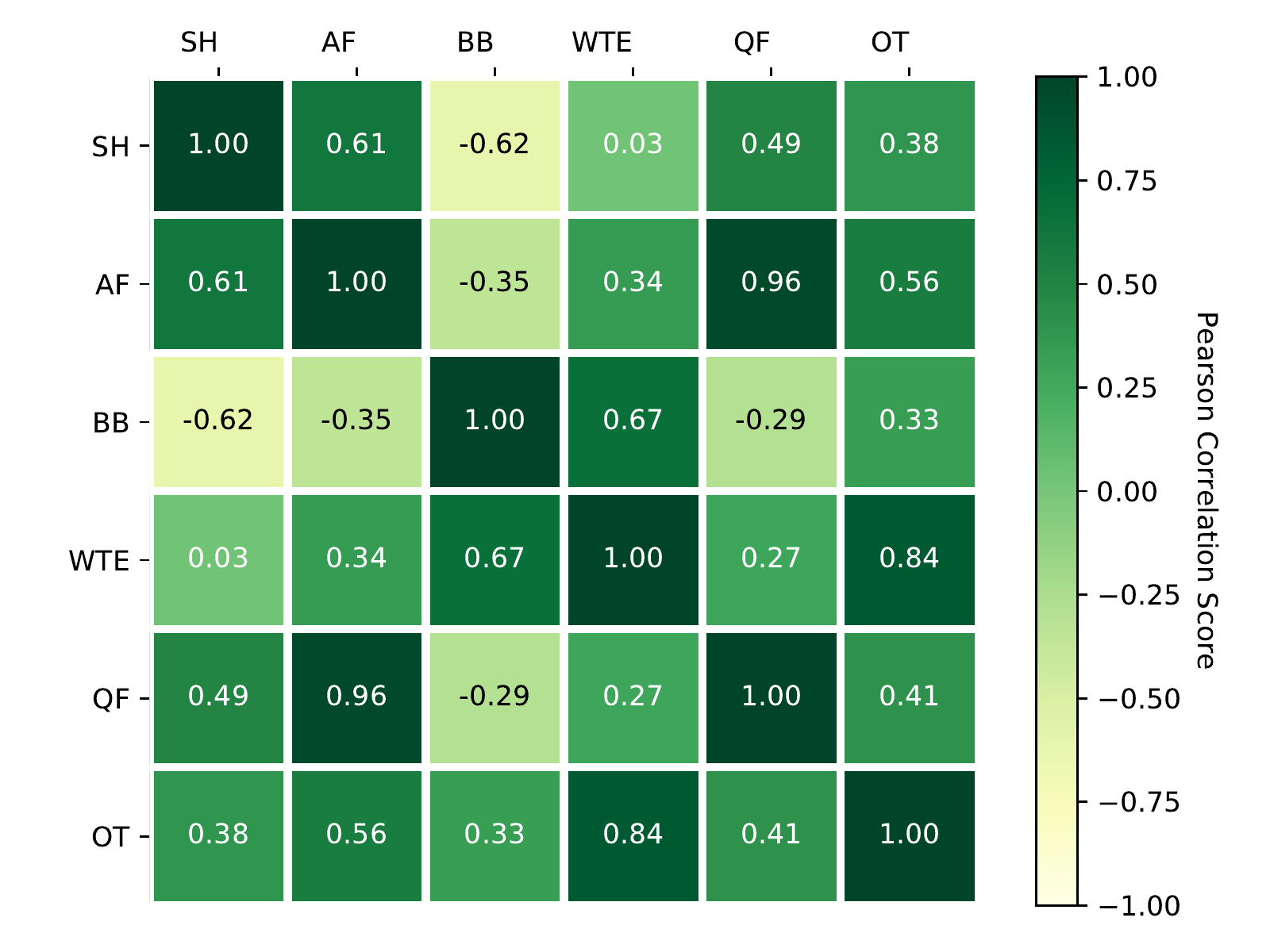}
         \caption{\textbf{small-standard} correlation}
         \label{fig:small-crisist6-standard-acc-correlation}
     \end{subfigure}

    \begin{subfigure}[b]{0.4\textwidth}
         \centering
         \includegraphics[width=\textwidth]{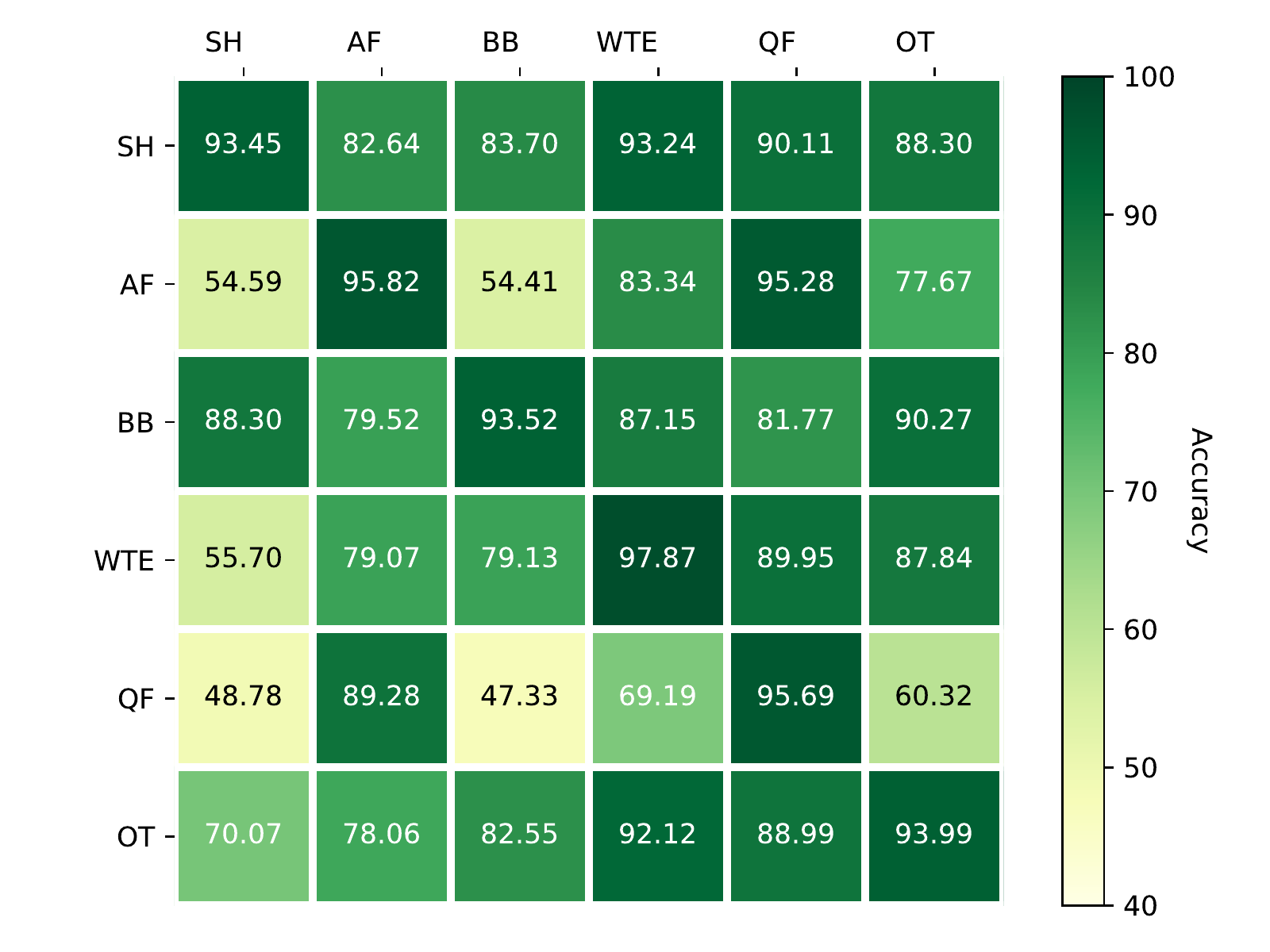}
         \caption{\textbf{small-postQ}}
         \label{fig:small-crisist6-postQ-acc}
     \end{subfigure}
     \begin{subfigure}[b]{0.4\textwidth}
         \centering
         \includegraphics[width=\textwidth]{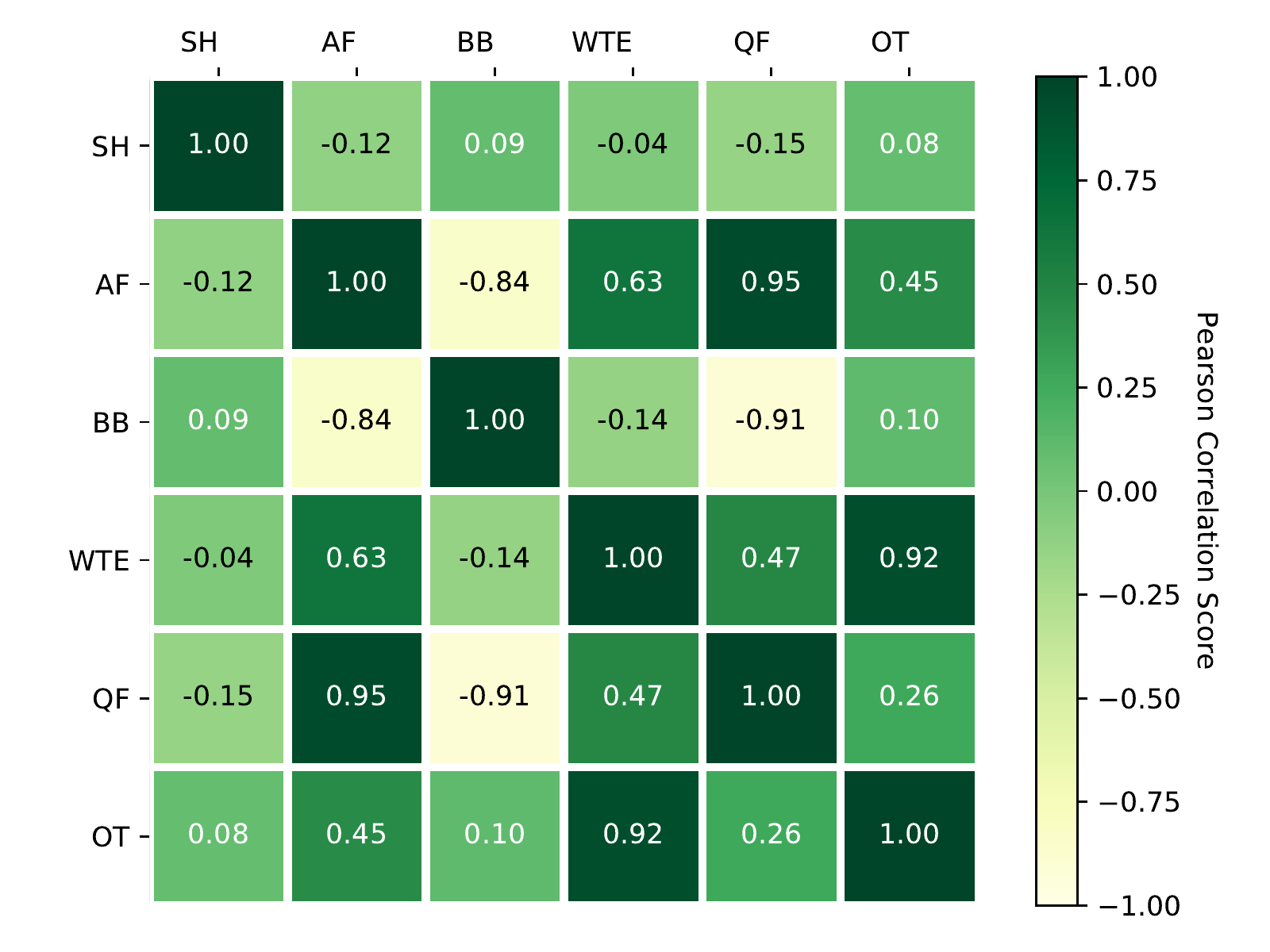}
         \caption{\textbf{small-postQ} correlation}
         \label{fig:small-crisist6-postQ-acc-correlation}
     \end{subfigure}
     
          \begin{subfigure}[b]{0.4\textwidth}
         \centering
         \includegraphics[width=\textwidth]{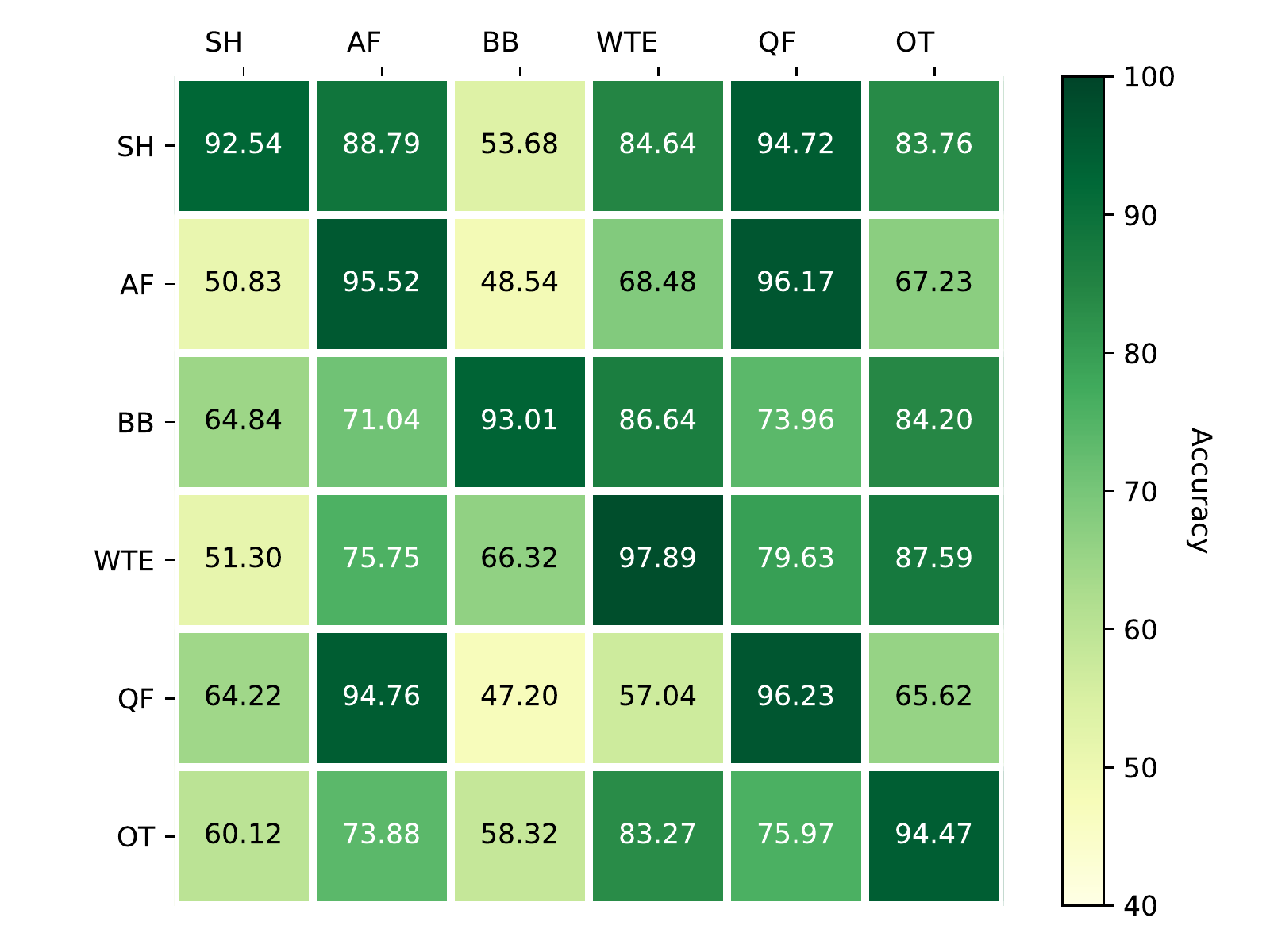}
         \caption{\textbf{base-standard}}
         \label{fig:crisist6-base-standard-acc}
     \end{subfigure}
     \begin{subfigure}[b]{0.4\textwidth}
         \centering
         \includegraphics[width=\textwidth]{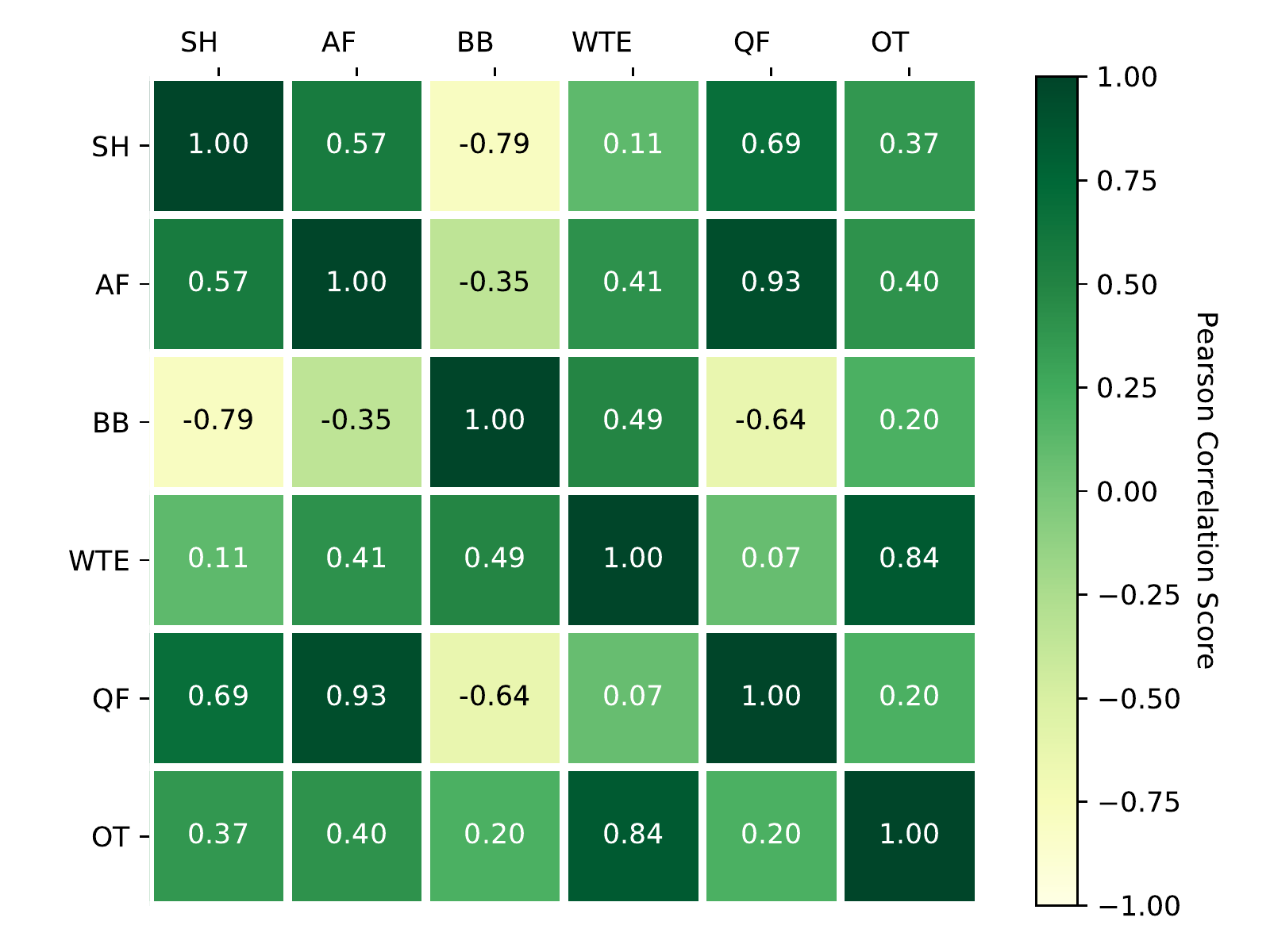}
         \caption{\textbf{base-standard} correlation}
         \label{fig:crisist6-base-standard-acc-correlation}
     \end{subfigure}

    \begin{subfigure}[b]{0.4\textwidth}
         \centering
         \includegraphics[width=\textwidth]{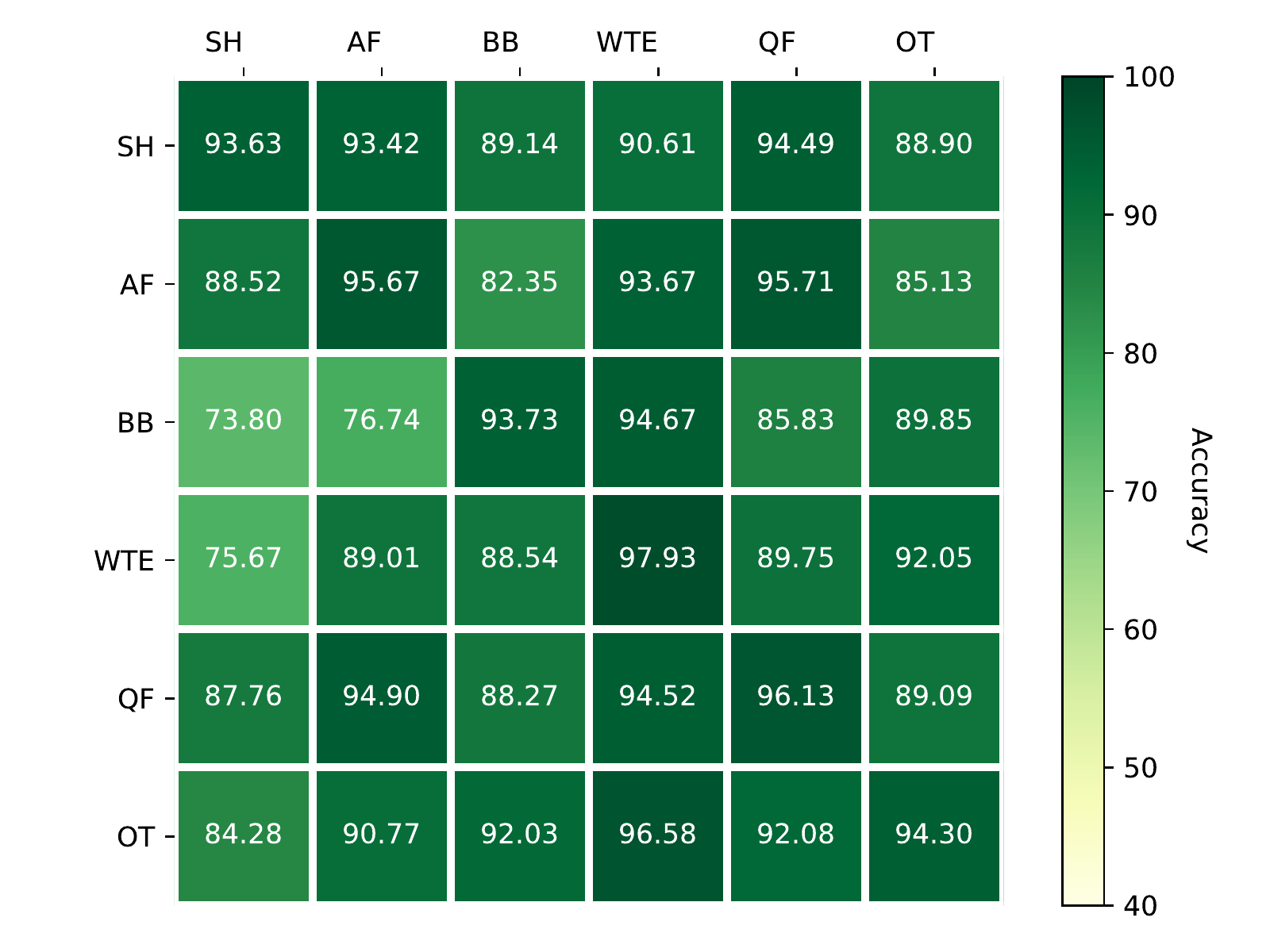}
         \caption{\textbf{base-postQ}}
         \label{fig:crisist6-base-postQ-acc}
     \end{subfigure}
     \begin{subfigure}[b]{0.4\textwidth}
         \centering
         \includegraphics[width=\textwidth]{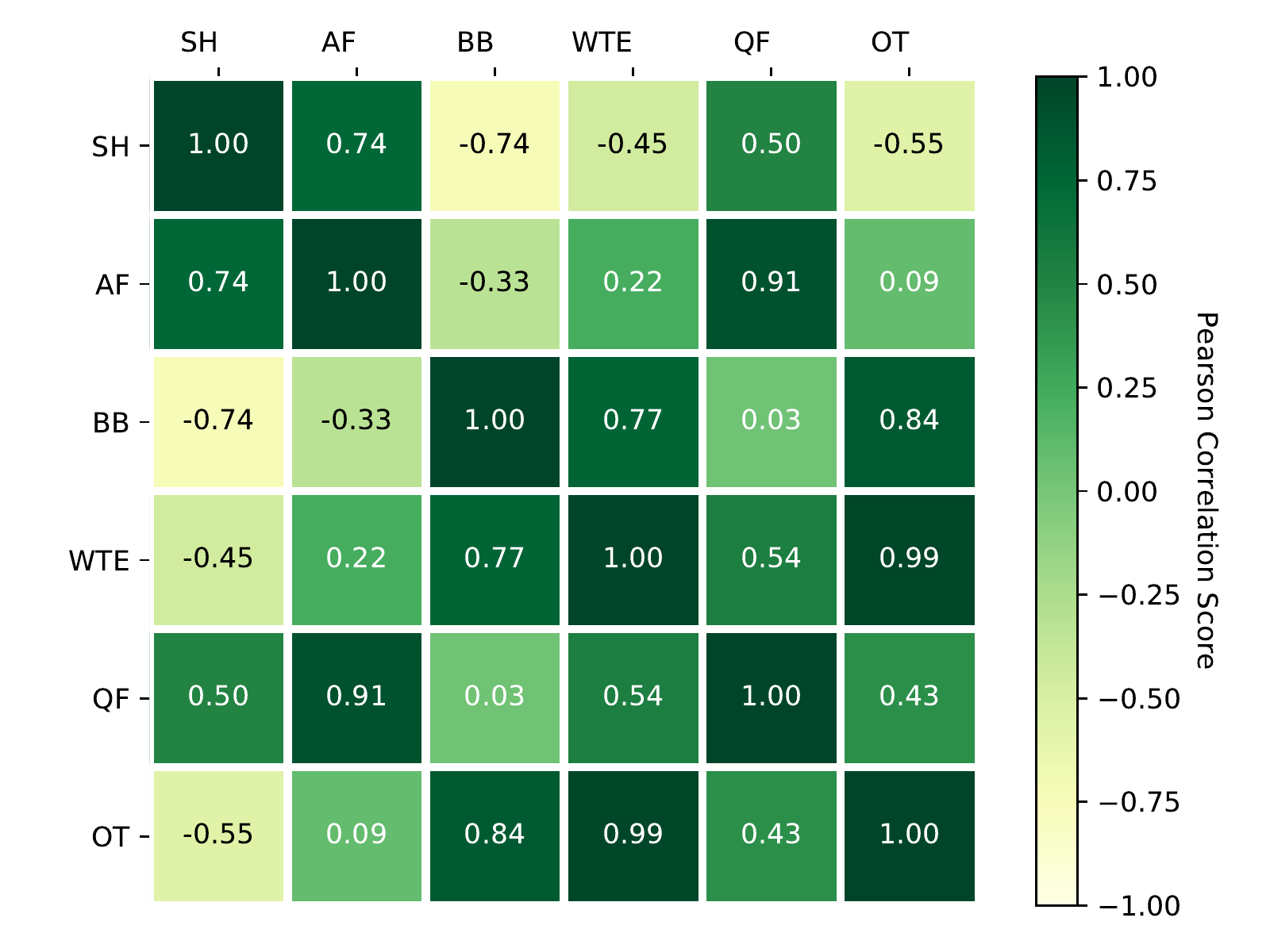}
         \caption{\textbf{base-postQ} correlation}
         \label{fig:crisist6-base-postQ-acc-correlation}
     \end{subfigure}
        \caption{Event adaptation accuracy and correlation for \textbf{CrisisT6} where SH: Sandy Hurricane, AF: Alberta Floods,
        BB: Boston Bombing, WTE: West Texas Explosion, QF: Queensland Floods, and OT: Oklahoma Tornado. Here QF is used to stand for Queensland Floods to distinguish it from QQF in \textbf{nepal\_queensland}.}
        \label{fig:small-base-standard-postQ-acc-cor}
\end{figure*}

Figure~\ref{fig:small-base-standard-postQ-acc-cor} reports the results of domain adaptation between the six events using both the standard method and postQ for fine-tuning the small and base models. The subfigures on the left side, i.e., (a), (c), (e), (g), present the accuracy scores in a matrix where the row represents the source events and the column stands for the target events. Hence, the diagonals refer to the in-domain performance and the rest are cross-domain scores. In addition to the adaptation matrices, a correlation matrix is also added to the right side of each of the adaptation matrices, i.e., (b), (d), (f), (h). The correlation matrices calculate the Pearson correlation between the rows of source events, which helps indicate how correlated two source events are in terms of their applicability to other target events.


Examining the matrices on the left, it is found that both the in-domain and cross-domain performance is consistent with \textbf{nepal\_queensland}. For example, in in-domain adaptation, postQ achieves similar performance to the standard with different model sizes\footnote{Per~\cite{liu2020crisisbert} that reported the average in-domain state-of-the-art accuracy for \textbf{CrisisT6} which is 95.6, the in-domain scores as seen from the diagonals approximate this state-of-the-art. Since this state-of-the-art was gained based on a random test leave-one-out evaluation which is different from 5-fold cross validation used in this work and thus this score is only for reference instead of direct comparison.}. Speaking of cross-domain adaptation, postQ performs competitively with the standard when using the \texttt{small} model (see Figure~\ref{fig:small-crisist6-standard-acc} and \ref{fig:small-crisist6-postQ-acc}). For the \texttt{base} model, postQ substantially outperforms the standard in cross-domain adaptation (see Figure~\ref{fig:crisist6-base-standard-acc} and \ref{fig:crisist6-base-postQ-acc}). As baselines, postQ-base is compared with two approaches presented in~\cite{li2018disaster}, using a Na\"ive Bayes classifier with and without self-training respectively (NB-ST and NN-S). Here, postQ-base is compared with their NN-S and NB-ST runs on 11 event pairs, as presented in Table~\ref{tab:t6-11-pairs}. This shows that the postQ-base substantially outperforms the target-data independent NB-S which is consistent with the results reported for \textbf{nepal\_queensland}. When compared to the TDD NB-ST, postQ-base achieves higher accuracy scores on almost all of the 11 event pairs, resulting in higher overall accuracy for postQ (90.02 versus 86.06 in average accuracy). 

\begin{table*}[]
\normalsize
\centering
\begin{tabular}{lrrrrrr}
\toprule
           & SH$\boldsymbol{\rightarrow}$QF & SH$\boldsymbol{\rightarrow}$BB & SH$\boldsymbol{\rightarrow}$WTE & SH$\boldsymbol{\rightarrow}$OT & SH$\boldsymbol{\rightarrow}$AF  & QF$\boldsymbol{\rightarrow}$BB   \\
NB-S & 76.84                     & 68.66                     & 77.21                      & 80.78                     & 71.06                      & 74.97                       \\
NB-ST      & 82.40                      & 84.06                     & 90.82                      & 87.76                     & 82.57                      & 81.86                       \\
\textbf{postQ-base} & 94.49                     & 89.14                     & 90.61                      & 94.49                     & 93.42                      & 88.27                       \\
\midrule
\midrule
           & QF$\boldsymbol{\rightarrow}$OT & QF$\boldsymbol{\rightarrow}$AF & BB$\boldsymbol{\rightarrow}$OT  & BB$\boldsymbol{\rightarrow}$AF & BB$\boldsymbol{\rightarrow}$WTE & Average \\
NB-S       & 84.13                     & 84.35                     & 73.81                      & 78.87                     & 94.77                      & 78.68                       \\
NB-ST      & 85.48                     & 86.91                     & 83.96                      & 86.01                     & 94.82                      & 86.06                      \\
\textbf{postQ-base} & 89.09                     & 94.90                      & 89.85                      & 76.74                     & 94.67                      & 90.52   \\
\bottomrule
\end{tabular}

\caption{Cross-domain accuracy comparison between postQ-base, NB-S (TDI) and NB-ST (TDD) from~\cite{li2018disaster} who studied the 11 event pairs of CrisisT6.}
\label{tab:t6-11-pairs}
\end{table*}

Regarding cross-domain adaptation, some interesting points are noticed. The results from~\cite{li2018disaster} present some evidence that similar event pairs (where the same or similar type of crises occurred at different locations) such as Queensland Floods (QF)$\boldsymbol{\rightarrow}$Alberta Floods (AF) and Boston Bombings (BB)$\boldsymbol{\rightarrow}$West Texas Explosion (WTE) are more likely to bring better scores than dissimilar pairs like Queensland Floods (QF)$\boldsymbol{\rightarrow}$Boston Bombings (BB) and Boston Bombings (BB)$\boldsymbol{\rightarrow}$Alberta Floods (AF) (see Table~\ref{tab:t6-11-pairs}). This evidence is enhanced in this study also. It is noted that the Alberta Floods (AF) and Queensland Floods (QF) relate to the same type of crisis (flooding) albeit in different locations at different times. It is interesting that in the standard and postQ runs these two events are \textbf{reciprocal}, indicating that either of them as the source event is well-suited to being adapted for the other as the target event. For example, the AF$\boldsymbol{\rightarrow}$QF adaptation always achieves accuracy around 95 and QF$\boldsymbol{\rightarrow}$AF adaptation achieves accuracy of 89.28 at the worst (Figure~\ref{fig:small-crisist6-postQ-acc}). Examining their correlation scores on the right, it is found that they are not only reciprocal, but also \textbf{highly correlated}, ranging from 0.91 to 0.96 (Figure~\ref{fig:crisist6-base-postQ-acc-correlation} and \ref{fig:small-crisist6-standard-acc-correlation}). This lends credence to the idea that similar event types have similar characteristics in terms of their applicability to cross-domain adaptation and could potentially be used interchangeably for a novel target event.

Another event pair with some similar characteristics are the Boston Bombing (BB) and the West Texas Explosion (WTE). These are perhaps less similar to the floods above in that one was an intentionally planted explosive device whereas the other was a factory fire that later resulted in an explosion. In this situation, it can be seen that whereas BB can be successfully adapted to WTE, the reverse is not the case. This may be related to the observation that BB is itself a difficult target event to adapt to, as evidenced by the fact that the cross-domain adaptation tends to be poorest in general when BB is the target.

Surprisingly, it is found that the Sandy Hurricane (SH) and Oklahoma Tornado (OT) datasets can not only be well adapted to AF and QF in most cases but also adapt well to WTE. This implies that there may be a certain degree of common linguistic features shared between the tornado/hurricane events and the explosion event. It seems that these findings indicate that the more similar a source event is to a target event, the more likely it is to exhibit better adaptation performance for that target event. Now such a question is naturally raised: does combining multiple similar source events add further benefit (many-to-one adaptation)?




\subsubsection{Many-to-one adaptation}


Considering the variety of crisis events and training efficiency, the many-to-one experimental runs are based on the \textbf{CrisisT6} dataset. The first experiment is leave-one-out cross-domain adaptation where one crisis event is chosen as the target domain and the union of the others as the source domain. Table~\ref{tab:one-out-t6-small} presents the results of fine-tuning using both the standard and postQ approaches. To compare the runs with the baseline by~\cite{li2018comparison}, it reveals that the CAST-based postQ run achieves 91.03 versus 89.6 in average accuracy. In addition, from this table it can be seen that there is no substantial difference between standard and postQ when AF, QF and OT are left out (they already achieve above 90\% accuracy). However, when leaving out SH and BB, standard performance is substantially lower than for postQ. This experiment indicates that postQ outperforms the standard approach when considering multiple events as the source domain. This is further justified by the next experiment.

\begin{table*}[]
    \centering
    
    \setlength{\tabcolsep}{10pt} 

    \begin{tabular}{lrlr}
     \toprule
\multicolumn{2}{c}{standard}              & \multicolumn{2}{c}{postQ}                 \\
\midrule
AF+BB+WTE+QF+OT$\boldsymbol{\boldsymbol{\rightarrow}}$SH & 74.35 & AF+BB+WTE+QF+OT$\boldsymbol{\rightarrow}$SH & 82.16 \\
SH+BB+WTE+QF+OT$\boldsymbol{\rightarrow}$AF & 95.22 & SH+BB+WTE+QF+OT$\boldsymbol{\rightarrow}$AF & 95.16 \\
SH+AF+WTE+QF+OT$\boldsymbol{\rightarrow}$BB & 70.18 & SH+AF+WTE+QF+OT$\boldsymbol{\rightarrow}$BB & 88.01 \\
SH+AF+BB+QF+OT$\boldsymbol{\rightarrow}$WTE & 89.39 & SH+AF+BB+QF+OT$\boldsymbol{\rightarrow}$WTE & 95.94 \\
SH+AF+BB+WTE+OT$\boldsymbol{\rightarrow}$QF & 95.59 & SH+AF+BB+WTE+OT$\boldsymbol{\rightarrow}$QF & 95.71  \\
SH+AF+BB+WTE+QF$\boldsymbol{\rightarrow}$OT & 90.52 & SH+AF+BB+WTE+QF$\boldsymbol{\rightarrow}$OT & 89.16\\
 \multicolumn{1}{c}{Average} & 85.88 &  \multicolumn{1}{c}{Average} & 91.03\\

\bottomrule
\end{tabular}
    \caption{Leave-one-out cross-domain adaptation accuracy using CrisisT6. The last row reports the average score. As a comparison, the best average score reported in the literature is 89.6~\cite{li2018comparison}.}
    \label{tab:one-out-t6-small}
\end{table*}

The next experiment is conducted to test if more source events bring better adaptation performance, whether they are similar or dissimilar. For this purpose, two event pairs: (QF, AF) and (BB, WTE) are selected. As indicated above, each pair contains two similar event types, while the two pairs themselves are dissimilar. The decision to choose these two pairs as similar event pairs is guided by existing work~\cite{li2018disaster} and the correlation scores that are presented in Figure~\ref{fig:small-base-standard-postQ-acc-cor}.

Table~\ref{tab:many-to-one-on-sim-pairs} presents the results of combinations of multiple source events adapted to the two pairs. First, it shows that postQ overall outperforms the standard in most situations (accuracy is much higher when BB and WTE are the target events and is at least the same level for QF and AF). However, the more interesting observation from this experiment is that simply increasing the number of source events does not guarantee benefits to the adaptation performance but it depends on what source events are be added. For example, when QF is the target event, only a trivial difference is observed between AF-to-QF and leaving QF out (thus combining all other crises as the source domain). A similar pattern is observed when AF is the target event.

\begin{table*}[h]

    \begin{subtable}[h]{0.45\textwidth}
        
        \scriptsize
  \begin{tabular}{lcc|cc}
        \toprule
                & \multicolumn{2}{c}{standard} & \multicolumn{2}{c}{postQ} \\
                & QF               & AF              & QF             & AF             \\
\midrule
AF              & 95.5             & -               & 95.28          & -              \\
QF              & -                & 93.7            & -              & 89.28          \\

\midrule
BB+WTE          & 83.79            & 74.69           & 87.27          & 76.18          \\

SH+OT        & 93.97            & 85.05               & 93.71          & 83.71              \\

AF+SH+OT        & 95.32            & -               & 95.76          & -              \\

QF+SH+OT        & -                & 95.45           & -              & 95.38          \\

SH+OT+BB+WTE        & 92.72               & 82.67           & 91.44              & 81.81         \\
\midrule
AF+SH+OT+BB+WTE & 95.59            & -               & 95.71          & -              \\
QF+SH+OT+BB+WTE & -                & 95.22           & -              & 95.16         \\
\bottomrule
\end{tabular}
       \caption{QF and AF as the target events}
       \label{tab:qf-af-many-to-one}
    \end{subtable}
    \hfill
    \begin{subtable}[h]{0.45\textwidth}
      
        \scriptsize
\begin{tabular}{lcc|cc}
\toprule
                & \multicolumn{2}{c}{standard} & \multicolumn{2}{c}{postQ} \\
                & BB               & WTE             & BB             & WTE            \\ \midrule
WTE             & 72.69            & -               & 79.13          & -              \\
BB              & -                & 93.28           & -              & 87.15          \\
\midrule
AF+QF           & 51.24            & 68.42           & 50.1           & 80.56          \\

SH+OT           & 75.71            & 89.32           & 82.14          & 92.01          \\

WTE+SH+OT           & 73.3            & -           & 85.3          & -          \\

BB+SH+OT           & -            & 92.44          & -          & 96.35          \\

AF+QF+SH+OT     & 63.94            & 79.6            & 78.42          & 86.29          \\
\midrule
WTE+AF+QF+SH+OT & 70.18            & -               & 88.01          & -              \\
BB+AF+QF+SH+OT  & -                & 89.39           & -              & 95.94          \\

\bottomrule
\end{tabular}
        \caption{BB and WTE as the target events}
        \label{tab:bb-wte-many-to-one}
     \end{subtable}
     \caption{Many-to-one cross-domain adaptation on similar event pairs}
     \label{tab:many-to-one-on-sim-pairs}
\end{table*}

Table~\ref{tab:qf-af-many-to-one} also indicates that adding BB+WTE seems not to add any benefit to the performance (indeed this reduces performance when compared with a source domain of SH+OT), and SH+OT can help to a degree (adding SH+OT to QF results in improved cross-domain performance when AF is the target).

Table~\ref{tab:bb-wte-many-to-one} demonstrates a similar outcome. When BB and WTE are the target events, AF+QF contributes little to the adaptation performance. Surprisingly, SH+OT not only helps QF and AF but also helps BB and WTE, which coincides with the one-to-one results as reported in the previous section. Hence, as a recommendation to maximise the adaptation performance to a target event (e.g., AF), it is good to combine its similar events (e.g., QF+SH+OT $\boldsymbol{\rightarrow}$ AF) as the source domain and exclude dissimilar events (e.g., BB+WTE)\footnote{Although there is no direct harm to the performance when including the dissimilar events (see the last two rows of Table~\ref{tab:qf-af-many-to-one} and~\ref{tab:bb-wte-many-to-one}), it is suggested to exclude them which can help reduce the training size and thus improve the training efficiency.}.



\section{Conclusions}

The chapter focused on developing adaptation models that learn from a previous crisis or crises in order to classify messages related to a new unfolding crisis situation, i.e., one-to-one and many-to-one crisis domain adaptations. To this end, this chapter presented CAST, which fine-tuned event-aware seq2seq Transformer-based PLMs for one-to-one and many-to-one domain adaptations between crisis events. To test the effectiveness of CAST, extensive experiments were conducted with two benchmark crisis informativeness classification datasets. In the one-to-one adaptation setting, CAST was demonstrated to be effective in cross-domain adaptation, outperforming the state of the art baselines without using any target data. Its advantage over the standard approach was more pronounced when training with a bigger model. In the many-to-one adaptation setting, CAST also outperformed the standard method and baselines in the literature. Interestingly, in the many-to-one adaptation setting, the results indicated that there was merit in choosing a source domain with similar characteristics (i.e. fine-tuning based on a similar type of crisis). If multiple existing similar events were available, these could be combined to form a larger source dataset to improve adaptation performance, while dissimilar events may have harmed classification performance.


Referring back to Chapter~\ref{ch:adaptation}, the work in the many-to-many adaptation helps guide the community in deploying a computational system for crisis message categorisation without the need for knowledge of specific types of source and target events. Here, the CAST work provides insights on how to select source events for crisis adaptation. However, in real-world crisis message categorisation, crisis domain adaptation only works when the categorisation task for target events is exactly the same as that of source events. In other words, the pre-defined aid types (classes) of the source data must be consistent with those of the target data in order for adaptation to be possible. In a real-world context, when there is a lack of annotated data for a new crisis event that emerges with new classes (due to the time-sensitive nature of such events), it is impractical to apply supervised approaches for categorizing the event. It is also very costly in terms of both human labour and time to annotate data for the new classes. This motivates the research to go beyond crisis domain adaptation and proposes crisis few-shot learning, which uses no labelled source data but a minimal quantity of labelled or unlabelled target data for crisis message categorisation. In the following chapter, the proposed augmentation approaches using PLMs for this purpose will be introduced.

\chapter{Augmentation for Crisis Few-shot Learning}
\label{ch:sta-isa}
\section{Introduction}

The previous two chapters discussed the use of supervised approaches, including domain adaptation, for crisis message categorisation. However, these approaches have limitations in meeting the real-world user needs for crisis response. Domain adaptation only works effectively when the categorisation task for the target event is the same as that of the source event, meaning that the pre-defined aid types (classes) must be the same for both. While this may meet some user needs, it has limited applicability to many real-world scenarios. In practice, responders may need to search for new types of information for emerging events. For instance, in a flood event, responders may request information about ``shelter'', but this type of information may not have been included in the annotations for previous events such as terrorist attacks. This makes it challenging to adapt the model trained on previous events. Moreover, the concept of domain adaptation relies on the availability of extensive training data related to past crises. However, in reality, this may not always be feasible, particularly when dealing with a novel crisis event, such as an epidemic with entirely new aid requirements or classifications, where no annotated training data is available due to the absence of similar prior incidents. Additionally, research from the previous chapter suggests that adapting the model using a similar previous event can be successful, but even when the events are similar, there is no guarantee that the results will be good as crises with similar characteristics may still have differences. These considerations have motivated this research to move towards building categorisation models that rely on minimal quantities of training data. The approaches outlined in this chapter aim to achieve ``crisis few-shot learning''. This refers to a situation where a very small quantity of labelled messages relating to the target event are required to bootstrap the system. This substantially reduces the burden of manually labelling data so as to be a manageable task in the early stages of an unfolding crisis. In addition to this, larger quantities of unlabelled data relating to the target event are also used, which do not require any manual annotations to be conducted during the crisis.

This chapter presents two augmentation-based approaches for crisis few-shot learning. The first approach, called Self-controlled Text Augmentation (STA), uses sequence-to-sequence (seq2seq) pre-trained language models (PLMs) to generate new crisis messages based on a small training set consisting of a few labelled messages for each class that the classifier is required to identify\footnote{A preprint of this STA work can be found in~\cite{wangsta2023}.}. To improve the quality of the generated messages in STA, the seq2seq model is used to evaluate the confidence of the generated messages. Only those with high confidence are selected and combined with the original labelled data to create the augmented data for training the categorisation model in a standard supervised manner (Section~\ref{sec:sta}). The second approach, called Iterative Text Augmentation (ISA), is an optimized version of STA that further improves the quality of the generated messages by incorporating an iterative mechanism and a de-duplication step into the STA pipeline (Section~\ref{sec:isa}).

\section{Self-Controlled Text Augmentation (STA)}
\label{sec:sta}


As discussed in Section~\ref{subsec:aug-based-methods}, the key to creating high-quality augmented data is to address the challenges of ``lexical diversity'' and ``semantic fidelity''. Lexical diversity refers to the diversity of the augmented samples in relation to the original samples, which helps to add knowledge to the training data. Semantic fidelity means that, despite being diverse, the augmented samples should have a reliable alignment between their semantic meanings and class labels, so as not to introduce noise into the training data. To tackle these challenges, this section introduces a novel generation strategy called Self-controlled Text Augmentation (STA). STA uses a seq2seq pre-trained language model with a set of task-specific prompts to generate new samples and then select the best ones. The prompts consist of classification and generation templates, allowing the model to learn both a classification task and a generation task jointly. The generation task is used to generate new samples intended to be related to particular classes by being dependent on the class surface name as an input and the classification task is used to choose only the high-confidence samples to be added to the training set, with the aim of including only samples with high semantic fidelity. The following section describes STA in the context of text classification.

\subsection{Self-controlled method}

\begin{figure*}[h!]
    \centering
    \includegraphics[scale=0.8]{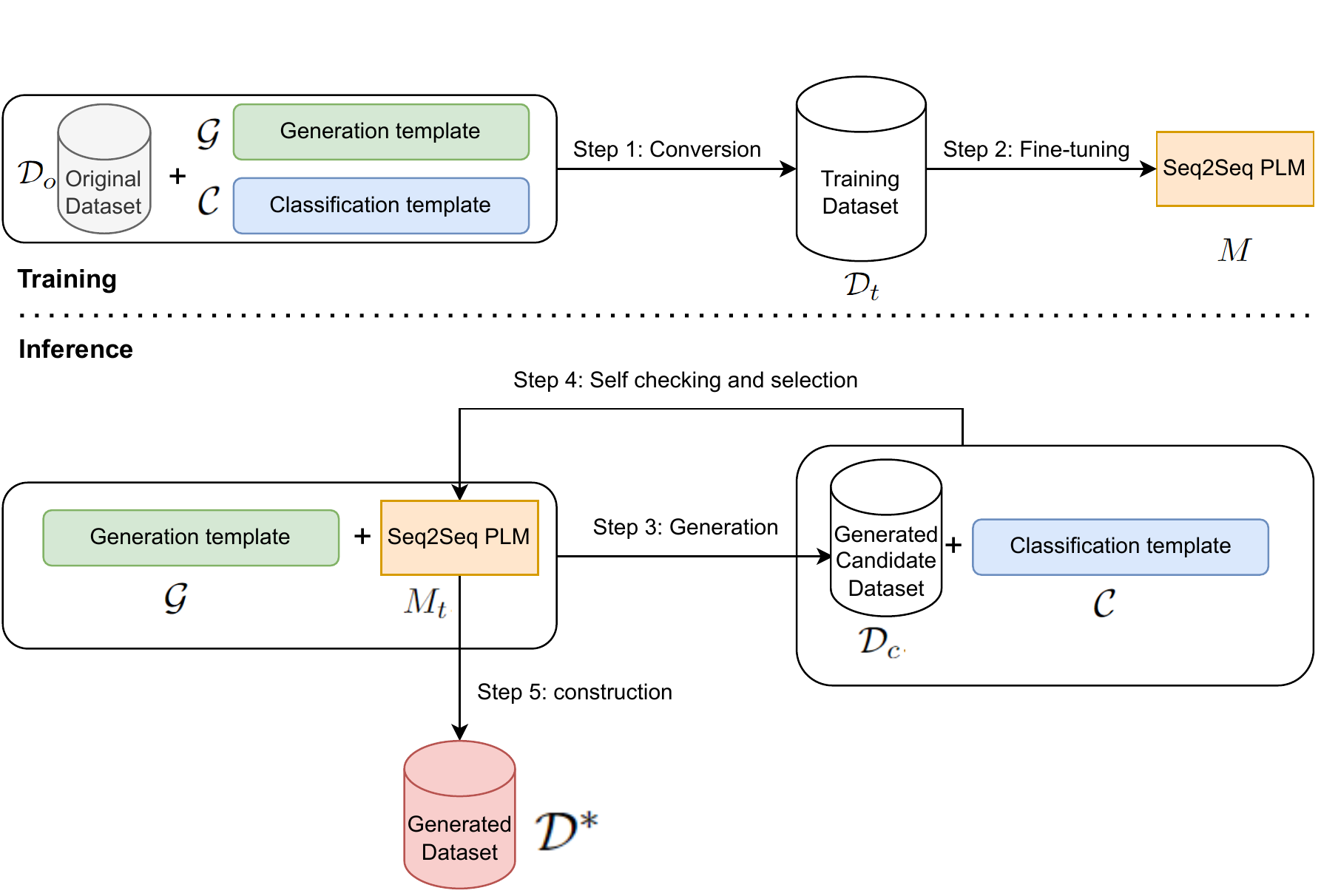}
    \caption{The architecture of the Self-controlled Text Augmentation approach (STA).}
    \label{fig:sta-method}
\end{figure*}



Figure~\ref{fig:sta-method} illustrates the workflow of STA and Algorithm~\ref{alg:sta-alg} states STA in simple terms. In Figure~\ref{fig:sta-method}, the upper portion outlines the finetuning component of the method (\textbf{Training}), whilst the lower portion demonstrates the procedure for generating novel data (\textbf{Inference}). In STA, a seq2seq language model is first fine-tuned using a small dataset (i.e. the original dataset) that is converted by two types of prompt templates: generation templates and classification templates. These templates recreate the original dataset in a way that allows the model to be trained on both a generation task and a classification task simultaneously. Once the model is trained, it can use the generation templates to generate new samples for a candidate dataset. However, not all of these candidates are used as augmentations, as some may be noisy. Instead, the model's predictions for each candidate using the classification templates are used to determine which candidates have high confidence and should be included in the generated dataset. This dataset, along with the original dataset, is then used to train a downstream classification model. The following provides a detailed description of the steps involved in STA, starting with prompts-based multi-task training and ending with data generation and self-checking.


\begin{algorithm}
\caption{: Self-Controlled Text Augmentation (STA)}\label{alg:sta-alg}
\begin{algorithmic}[1]

\Require Original dataset $\mathcal{D}_o$. Generation model $M$. Generation template $\mathcal{G}$. Classification template $\mathcal{C}$.

\State Convert $\mathcal{D}_o$ to training dataset $\mathcal{D}_t$ via $\mathcal{G}$ and $\mathcal{C}$.
\State Finetune $M$ on $\mathcal{D}_t$ in a generation task and a classification task jointly to obtain $M_t$.
\State Use $\mathcal{G}$ and $M_t$ to generate candidate dataset $\mathcal{D}_c$.
\State Apply $M_t$ to do classification inference on $\mathcal{D}_c$ with $\mathcal{C}$ to select the most confident examples.
\State Combine the final generated dataset $\mathcal{D}^*$ with the selected examples.
\State Use the combined data for downstream model training
\end{algorithmic}
\end{algorithm}

\begin{table*}[]
\centering
\footnotesize
\begin{tabular}{c|ll|c}
\toprule
\multicolumn{1}{l}{}        \textbf{Template}               &        & \textbf{Source sequence ($s$)}                                                              & \textbf{Target sequence ($t$)}                 \\
\midrule
\multirow{3}{*}{Classification }       & $c_1$ & Given \{Topic\}: \{$\mathcal{L}$\}. Classify: \{$x_i$\}                                    & \{$y_i$\}                           \\
                                           & $c_2$ & Text: \{$x_i$\}. Is this text about \{$y_i$\} \{Topic\}?                         & yes                                 \\
                                           & $c_3$ & Text: \{$x_i$\}. Is this text about \{$\overline{y}_i$\} \{Topic\}?              & no                                  \\
                                           \midrule
\multirow{2}{*}{Generation} & $g_1$ & Description: \{$y_i$\} \{Topic\}. Text:                                          & \{$x_i$\}                           \\
                                           & $g_2$ & Description: \{$y_i$\} \{Topic\}. Text: \{$x_j$\}. Another text: \{$x_i^{0\text{-}2}$\} & \{$x_i^{3...}$\} \\
                                           \bottomrule

\end{tabular}
\caption{Prompt templates of STA where $x_i$ refers to an input sequence and $y_i$ implies the label surface name of the sequence, ``Topic'' refers to a simple keyword describing the target task dataset e.g. ``disaster aid type'' and  $\mathcal{L}$ is the list of all class labels in the dataset. The symbol $\overline{y}_i$ in $c_{3}$ stands for any label in $\mathcal{L}\setminus\{y_i\}$, chosen randomly. In $g_{2}$, the $x_j$ denotes another sample from same class as $x_i$ (i.e. $y_j=y_i$) chosen randomly and $x_i^{0\text{-}2}$ refers to the first three words of $x_i$ as the context to generate the remaining words of $x_i$ i.e. $x_i^{3...}$.}
\label{tab:sta-templates}
\end{table*}    

\begin{table*}[h!]
\scriptsize
\centering
\begin{tabular}{p{25em}|p{20em}}
\toprule \specialrule{.1em}{.05em}{.05em}

\multicolumn{2}{p{45em}}{An example from a disaster aid type (Topic) classification dataset where the classes ($\mathcal{L}$): missing or found people, sympathy and support…}                                                                                                            \\
\hline
Text ($x$)                                                                                                                           & \textit{\textcolor{blue}{UPDATE: Body found of man who disappeared amid Maryland flooding}}

\\
 \hline  
Label ($y$)                                                                                                                       & \textit{\textcolor{red}{missing or found people}}                                                                            \\
\midrule \specialrule{.1em}{.05em}{.05em} 
\multicolumn{2}{l}{Converted  examples by classification templates: source($s$), target($t$)}                                                                                                                                                                                                                                             \\

\hline
\textit{Given disaster aid type: missing or found people,   sympathy and support…   Classify: \textcolor{blue}{UPDATE: Body found of man who disappeared amid Maryland flooding}}                                   & \textit{\textcolor{red}{missing or found people}}                                                                            \\
\hline

\textit{Text: \textcolor{blue}{UPDATE: Body found of man who disappeared amid Maryland flooding} Is this text about \textcolor{red}{missing or found people} disaster aid type?}                                       & \textit{yes}                                                                                 \\
\hline
\textit{Text: \textcolor{blue}{UPDATE: Body found of man who disappeared amid Maryland flooding} Is this text about negative disaster aid type?}                                       & \textit{no}                                                                                  \\

\midrule \specialrule{.1em}{.05em}{.05em}
\multicolumn{2}{l}{Converted  examples by generation templates: source($s$), target($t$)}                                                                                                                                                                                                              \\

\hline
\textit{Description: \textcolor{red}{missing or found people} disaster aid type. Text: }                                                                                            & \textit{\textcolor{blue}{UPDATE: Body found of man who disappeared amid Maryland flooding   }    }                             \\
\hline
\textit{Description: \textcolor{red}{missing or found people} disaster aid type. Text: \textcolor{blue}{UPDATE: Body found of man who disappeared amid Maryland flooding} Another text: Open missing people} & \textit{Search Database from Mati and Rafina areas \#Greecefires  \#PrayForGreece \#PrayForAthens}                                          \\
\hline
\textit{Description: \textcolor{red}{missing or found people} disaster aid type. Text: \textcolor{blue}{UPDATE: Body found of man who disappeared amid Maryland flooding} Another text: \#Idai victims buried}                    & \textit{in mass grave in Sussundenga, at least 60 missing -   \#Mozambique \#CycloneIdai \#Ci-cloneIdai} \\
\hline
\textit{Description: \textcolor{red}{missing or found people} disaster aid type. Text: \textcolor{blue}{top-notch action powers this romantic drama .} Another text: Rain on the} & \textit{way as search for missing continues after California   wildfires }  \\
\bottomrule
\end{tabular}
\caption{The demonstration of an example conversion by the prompt templates in Table~\ref{tab:sta-templates} where the input text is highlighted in \textcolor{blue}{blue} and label is highlighted in \textcolor{red}{red} for readability.}
\label{tab:sta-demo-ex}
\end{table*}

\subsubsection{Prompts-based multi-task training in seq2seq Models}
\label{subsubsec:sta_prompts}



Let $M$ be a pretrained seq2seq PLM. Such models consist of an encoder-decoder pair: the encoder takes a source sequence $s$ and produces a contextualised encoding sequence $ \overline{s}$. For each token $t_i$ that is to be generated for $\overline{s}$, the encoded input sequence and the sebsequence $t\colon\{t_1,t_2,..t_{i-1}\}$ (i.e. all tokens in the output sequence prior to $t_i$) are used as the input for the decoder to compute the conditional probability $p_{M}(t_{i} | t_{1: i-1},\overline{s})$ for $t_i$ and $p_{M}$ is estimated by trainable parameters of the model. The possible target output (a sequence) $t\colon\{t_1,t_2,...,t_m\}$ given $\overline{s}$ is generated via the factorisation:

\begin{equation}
\label{eq:conditional-generate}
    p_{M}(t_{1:m} | \overline{s}) = \prod_{i=1}^{m} p_{M}(t_i | t_{1: i-1},\overline{s})
\end{equation}

Let $\mathcal{D}_o= \{(x_i,y_i)\}|_{i=1}^n$ be a small corpus of few-shot data for text classification where $x_i \in \mathcal{X}$ and $y_i \in \mathcal{L}$ are an input text instance and its label respectively. The goal is to produce a training dataset $\mathcal{D}_t$ to finetune $M$ and ensure that it is primed for generating diverse examples that are faithful to the appropriate label.

Formally, a \emph{template} is a function $T: \mathcal{X} \times \mathcal{L} \to V^* \times V^*$ where $V$ is the vocabulary of $M$ and $V^*$ denotes the set of finite sequences of symbols in $V$. Given a set of templates $\mathcal{T}$, let $\mathcal{D}_t = \mathcal{T}(\mathcal{D}_o) = \bigcup_{T\in \mathcal{T}}T(\mathcal{D}_o)$. That is, each sample $(x_i,y_i) \in \mathcal{D}_o$ is converted to $|\mathcal{T}|$ samples in the training dataset $D_t$. Table~\ref{tab:sta-templates} lists all the templates that are specifically designed for classification and generation purposes and Table~\ref{tab:sta-demo-ex} demonstrates how this conversion is performed.

Crucially, two types of template families are constructed: classification templates $\mathcal{C}$ and generation templates $\mathcal{G}$ and the set $\mathcal{T}$ comprises both of these (i.e. $\mathcal{T} = \mathcal{C}\cup\mathcal{G}$).

\begin{itemize}
    \item \textbf{Classification templates} have the form $c(x,y) = (f_1(x),f_2(y))$ or $c(x,y) = (f_1(x,y),f_2(y))$ where $f_1$ and $f_2$ refer to functions that convert a piece of text to the source sequence and target sequence respectively. Here, the text $x \in \mathcal{X}$ is not a part of the target output.
    \item \textbf{Generation templates} have the form $g(x,y) = (f_1(y),f_2(x))$ or $g(x,y) = (f_1(x,y),f_2(x))$ where $f_1$ and $f_2$ refer to functions that convert a piece of text to the source sequence and target sequence respectively. Here, the label $y \in \mathcal{L}$ is not a part of the target output.
\end{itemize}

After obtaining the training dataset $D_t$ by the conversion, $M$ can then be finetuned to obtain $M_t$ (see \ref{subsec:sta_train_eval_details} for details on the training parameters used). The motivation here is that $D_t$ will ensure the model is finetuned to learn both how to generate a new piece of text of the domain basedon the label description as well as to classify a piece of text relating to the domain. For example, during the training stage, the model can use the first classification template $c_1$ in Table~\ref{subsubsec:sta_prompts}, to predict the label for an original sample $x_i$, which is a classification task. Meanwhile, using the first generation template $g_1$, the model can generate the original sample $x_i$ given its label $y_i$ as input, which is a generation task. At the inference stage, the model can use the generation template $g_i$ to generate a new sample given a label as input, and then use the classification template $c_1$ to classify the generated example and check its confidence, which is described next.



\subsubsection{Data generation, self-checking and selection}
\label{subsec:sta-self-check}

A two-step process is adopted here: first, candidates are generated and then a fraction of the candidates are selected to be included as augmentations. This process is conducted for each class separately so it is assumed that for the remainder of this section that a label $y \in \mathcal{L}$ has been fixed for both generation and selection (the same process is applied to other labels also).

The first objective is to generate $\alpha\times n_y$ samples where $n_y$ is the original number of samples in $\mathcal{D}_o$ for label $y$ and $\alpha$ is a multiplier hyperparameter  to control the ratio of generated samples to original samples. To perform this generation, Equation~\ref{eq:conditional-generate} is applied autoregressively to a chosen prefix or source sequence.

Referring back to Table~\ref{tab:sta-templates}, there are two choices of generation template sequence ($g_{1}$ and $g_{2}$) that can be used to construct $s$. Here $g_{1}$ is chosen over $g_{2}$ as the former only needs the label (the dataset description is viewed as a constant), i.e.  
$$g_1(x,y) = (f_1(y),f_2(x)).$$ 
which gives the model greater freedom to generate diverse examples. 

Thus $s$ is set to be $s=f_{1}(y)$ and $\alpha\times n_y$ samples are generated using the finetuned model $M_t$.

Now a synthetic candidate dataset for label $y$, $\mathcal{D}_c^y = \{(x_i,y)\}|_{i=1}^{\alpha \times n_y}$, is obtained, which will be refined using a self-checking strategy for selecting the generated samples based on the confidence estimated by the model $M_t$ itself. 

For each synthetic sample $(x,y)$, a source sequence is constructed using the template $c_1(x,y)= (f_1(x),f_2(y))=(f_1(x),\{y\})$, that is, $s$ is set to be $s=f_{1}(x)$. Given $s$, a score function $u$ is defined in the same way as in~\cite{schick2021exploiting}:
$$u(y|s) =  \log p_{M_t}(\{y\}|\overline{s})$$
Equivalently, this is the \emph{logit} computed by $M_t$ for the sequence $\{y\}$. Then the labels in $\mathcal{L}$ are normalised by applying a softmax over each of the scores $u(\cdot|s)$:

$$q(y|s) = \frac{e^{u(y|s)}}{\sum_{l\in \mathcal{L}}e^{u(l|s)}}$$

Finally, the elements of $\mathcal{D}_c^y$ are ranked by the value of $q$ and the top $\beta \times n_y$ samples ($\beta < \alpha$) are selected to form the final generated dataset for the specific label $y$: $D_*^{y}$. The overall $D_* = \bigcup_{y \in \mathcal{L}}D_*^{y}$ is the union of specific labels. Here, $\beta$ is called the \emph{augmentation factor} and $\alpha = 5 \times \beta$ is used. Thus, the self-checking technique selects the top $20\%$ of the candidate examples per class~\footnote{This is based on empirical experimental search over \{10\%, 20\%, 30\%, 40\%, 50\%\}.} to form the final generated $D^*$ that is combined with the original dataset $D_o$ for downstream model training.


\subsection{Experiments}

Next, extensive experiments are conducted to test the effectiveness of STA in low-data regimes. This section first describes the experimental setup including dataset and training details, and then outlines the baselines for comparison and discusses the results. Finally, an additional experiment is conducted to examine how STA can be generalised to other types of text classification tasks that are not specifically crisis-related.
 
\subsubsection{Dataset}

In this work, the \textbf{HumAID} dataset~\cite{alam2021humaid} is used as the benchmark dataset to test the effectiveness of STA. As discussed in Section~\ref{sec:rds-ems-infotypes}, this is a single-label multi-class dataset for information types classification, consisting of crisis tweets from 19 events between 2016 to 2019 annotated by information types including sympathy and support, missing or found people, etc. (These labels are easy to understand based on their surface names.). Considering at this stage STA supports single-label multi-class tasks and relies on good surface names, this justifies why this dataset is selected here.

\subsubsection{Training details}
\label{subsec:sta_train_eval_details}

When finetuning the seq2seq model, the pre-trained T5 base checkpoint is selected as the starting weights. For the downstream classification task, ``bert-base-uncased'' is fine-tuned on the original training data with the augmented (generated) samples. Regarding the pre-trained models, both are from the publicly-released version of the HuggingFace transformers library~\cite{wolf2019huggingface}\footnote{\url{https://github.com/huggingface/transformers}}. For the augmentation factor (i.e., $\beta$ in Section~\ref{subsec:sta-self-check}), the augmentation techniques including STA and the baselines are applied with $\beta$ between $1$ and $5$. Since this work focuses on text augmentation for classification in low-data settings, the experiments for both STA and the baselines are conducted on 5, 10, 20, 50 and 100 samples per class randomly sampled from the training set as per~\cite{anaby2020not}. Due to the randomness when sampling data of small size, all experiments are run $10$ times so that the average accuracy along with its standard deviation (std.) is reported on the full test set in the evaluation.


In finetuning T5, the learning rate is set to be $5 \times 10^{-5}$ using Adam~\cite{kingma2014adam} with linear scheduler ($10\%$ warmup steps), the training epochs to be $32$ and batch size to be $16$. At generation time, a top-k ($k=40$) and top-p ($p=1.0$) sampling technique is used for next token generation~\cite{holtzman2019curious}. In finetuning downstream BERT, the hyper-parameters are the same as those used in T5 finetuning, with the  exception that the training epoch is set to be $20$. 
The training epochs is set to be as large as possible with the aim of finding the best model when trained on a small dataset.

\subsubsection{Baselines}

In evaluation, STA is compared against a set of state-of-the-art techniques found within the literature of text augmentation. These approaches include a variety of augmentation procedures from easy reformulation to deep neural text generation. STA is compared to the augmentation techniques as they are directly related to STA in generating samples that can be used in the subsequent study for examining the quality of generated examples. 


\begin{itemize}
    \item \textbf{Baseline (No Aug.)} uses the original training data as the downstream model training data. The downstream classification model is directly trained on the original few-shot data with no augmentation applied.
    \item \textbf{EDA} refers to easy data augmentation in~\cite{wei2019eda}, which provides official implementations. Hence, their implementations are adapted to the selected dataset of this study for comparison.
    \item \textbf{BT} and \textbf{BT-Hops}~\cite{edunov2018understanding,shleifer2019low} refer to back-translation techniques. The former is implemented by taking one step back-translation from English to another language that is randomly sampled from the $12$ Romance languages provided by the ``opus-mt-en-ROMANCE'' model~\footnote{\url{https://huggingface.co/Helsinki-NLP/opus-mt-en-ROMANCE}} from the transformers library~\cite{wolf2019huggingface}. The latter adds random $1$ to $3$ extra languages to the back-translation using the same model.
    \item \textbf{GPT-2} is a deep learning method proposed in~\cite{kumar2020data} that uses GPT-2 for augmentation. Since no implementation is made available, the methodology described in the paper was followed. This was done by finetuning a GPT-2 base model on the small sampled training data sets (i.e., 5, 10, 20, 50, 100 samples per class to simulate the limited data scenario as in STA) and then the fine-tuned model is used to generate new samples that are conditional on both the label description and the first three words of an existing example.
    \item \textbf{GPT-2-$\lambda$} is similar to GPT-2 with the addition of the LAMBDA technique from~\cite{anaby2020not}, which also does not offer implementations. In a similar way to GPT-2, it is implemented by finetuning a GPT-2 base model on the small training data sets to generate new samples that are later confidence checked by a BERT base model (the LAMBDA component). 
    \item \textbf{CBERT}~\cite{wu2019conditional} is a strong word-replacement based method for text augmentation, for which an implementation is provided. Their implementation is applied to the selected dataset of this study for comparison.
    \item \textbf{BART-Span}~\cite{kumar2020data} uses the seq2seq BART model for text augmentation, and an implementation is not availalble. It is implemented as described in the paper, by finetuning the BART large model using the label names and the texts of 40\% consecutive masked words.
\end{itemize}

\subsection{Results and discussion}

\subsubsection{Classification performance}

Table~\ref{tab:sta-humaid-res} presents STA's accuracy on \textbf{HumAID}\footnote{The results are reported as average accuracy over $10$ random experimental runs, with the standard deviation in parentheses. Numbers in bold indicate the highest in each column}. It shows the effectiveness of STA in the crisis aid type classification task. It shows that STA outperforms a suite of state-of-the-art augmentation approaches, particularly when using fewer annotated training examples. When a higher number of samples ($50\text{-}100$) are used for training, it is seen that STA is slightly better than the baselines. However, STA is superior to other augmentation techniques when only a small number of examples are used to train the generator ($5\text{-}10\text{-}20$). In fact, STA demonstrates a difference of $+18.3$ and $+7.8$ compared to the best-performing augmentation baseline when trained on only $5$ and $10$ samples per class respectively, demonstrating its ability to generate salient and effective training examples from limited amounts of data in the crisis tweet classification task.
\label{subsec:sta-results}
\begin{table*}[!h]
\centering
\small
\begin{tabular}{l|lllll}
\toprule
Augmentation Method & 5                   & 10                   & 20                   & 50                   & 100                  \\\midrule \specialrule{.1em}{.05em}{.05em}
Baseline (No Aug.)  & 29.09(6.64)         & 37.08(6.42)          & 60.69(4.0)           & 80.01(0.93)          & 83.43(0.97)          \\ \hline    
EDA                & 49.49(4.48)         & 64.4(3.64)           & 74.69(1.53)          & 80.7(0.99)           & 83.53(0.56)          \\
BT                   & 45.76(5.67)         & 59.14(5.24)          & 73.47(2.05)          & 80.42(1.22)          & 83.1(0.68)           \\
BT-Hops             & 43.41(6.44)         & 57.52(5.21)          & 72.38(2.78)          & 80.12(1.11)          & 82.76(1.4)           \\
CBERT               & 44.8(7.56)          & 59.46(4.76)          & 73.43(1.73)          & 80.34(0.77)          & 82.69(1.16)          \\
GPT-2                 & 46.02(4.67)         & 55.72(5.72)          & 67.28(2.6)           & 77.82(1.61)          & 81.13(0.57)          \\
GPT-2-$\lambda$          & 50.73(8.62)         & 68.09(6.25)          & 78.51(1.32)          & 82.13(1.05)          & 84.2(0.79)           \\
BART-Span            & 42.4(7.31)          & 58.62(6.98)          & 70.04(3.73)          & 79.3(1.43)           & 83.33(0.93)          \\ \hline
\textbf{STA}          & \textbf{69.0(3.92)} & \textbf{75.84(3.34)}         & \textbf{80.17(1.59)} & \textbf{83.19(0.47)} & \textbf{84.52(1.14)} \\
\bottomrule
\end{tabular}
\caption{Results of STA and baselines on \textbf{HumAID} in $5, 10, 20, 50, 100$ examples per class. Accuracy is reported here and the numbers in \textbf{bold} indicate the highest in columns]}
\label{tab:sta-humaid-res}
\end{table*}

\subsubsection{Ablation Study}
\label{subsec:sta-ab-study}

To demonstrate the importance of the self-checking procedure, an ablation study is conducted to investigate STA with and without the self-checking step. The results without self-checking are shown in Table~\ref{tab:sta-humaid-abla-res} for \textbf{HumAID} (denoted as ``STA-noself''). As seen from this table, the approach demonstrates considerable improvements when the self-checking step is added across all tasks and training sample sizes, further supporting our augmentation technique. In fact, the difference between the two settings is considerable, with an average increase of $+5.7$ across all training samples sizes. It is hypothesised that the self-checking step more reliably controls the labels of the generated text, which greatly improves training stimulus and thus the performance on downstream tasks.

\begin{table*}
\centering
\small
\begin{tabular}{l|lllll}
\toprule
STA-noself          & 56.44(6.95)         & 70.23(4.27)          & 76.26(3.31)          & 79.35(4.45)          & 81.76(1.25)          \\
STA-twoprompts      & 68.71(10.92)        & \textbf{77.61(3.57)} & 80.06(1.71)          & 82.86(1.6)           & 84.31(0.69)          \\
\textbf{STA}          & \textbf{69.0(3.92)} & 75.84(3.34)          & \textbf{80.17(1.59)} & \textbf{83.19(0.47)} & \textbf{84.52(1.14)} \\
\bottomrule
\end{tabular}
\caption{Ablation study of STA on \textbf{HumAID} in $5, 10, 20, 50, 100$ examples per class. Accuracy is reported here and the numbers in \textbf{bold} indicate the highest in columns}
\label{tab:sta-humaid-abla-res}
\end{table*}

Of course, there are many possible choices for templates and permutations of template procedures. To further support the use of our multiple prompt templates used in STA (see Table~\ref{tab:sta-templates}), another ablation run is done for this purpose and its results are displayed in Table~\ref{tab:sta-humaid-abla-res} (denoted as ``STA-twoprompts''). These templates, one for classification ($c_1$) and one for generation ($g_1$), represent a minimalistic approach for performing generation-based augmentation with self-checking without the additional templates outlined in Table~\ref{tab:sta-templates}. The results show that the multiple templates used for STA provide slight improvements in all training sizes except for the 10-per-class case.


\subsubsection{Investigation of Lexical Diversity and Semantic Fidelity}
\label{subsec:sta-sfld-study}

As stated in Section~\ref{sec:sta}, one objective of this work is to generate data with high lexical diversity and semantic fidelity. To determine if STA is capable of producing new samples that outperform the baselines in these two aspects, additional experiments were conducted to analyse the samples produced by STA and the baselines. First, the following two measurements were used to quantify high lexical Diversity and semantic Fidelity.


\begin{itemize}
    \item \textbf{Generated Data Diversity.} The metric used for evaluating diversity is Unique Trigrams~\cite{feng2020genaug,kumar2020data}. This is determined by calculating the number of unique tri-grams as a proportion of the total number of tri-grams in a population. As the aim is to examine the difference between the generated data and the original data, the population consists of both the original and generated training data. For this metric, a higher score indicates better diversity.
    \item \textbf{Generated Data Fidelity.} The semantic fidelity is measured by evaluating how well the generated data retains the semantic meaning of its label. As per~\cite{kumar2020data}, it is measured by first finetuning a ``BERT-base-uncased '' on 100\% of the original training data of each classification task. The performance of the classifier on the test set of \textbf{HumAID} is $89.69$. After the finetuning, to measure the generated data fidelity, the finetuned classifier is used to predict the labels for the generated data and use the accuracy between its predicted labels and its associated labels (i.e., assigned labels by STA and baselines) as the metric for fidelity. Hence, a higher score indicates better fidelity.
\end{itemize}

\begin{figure}[h!]
    \centering
    \includegraphics[width=0.8\textwidth]{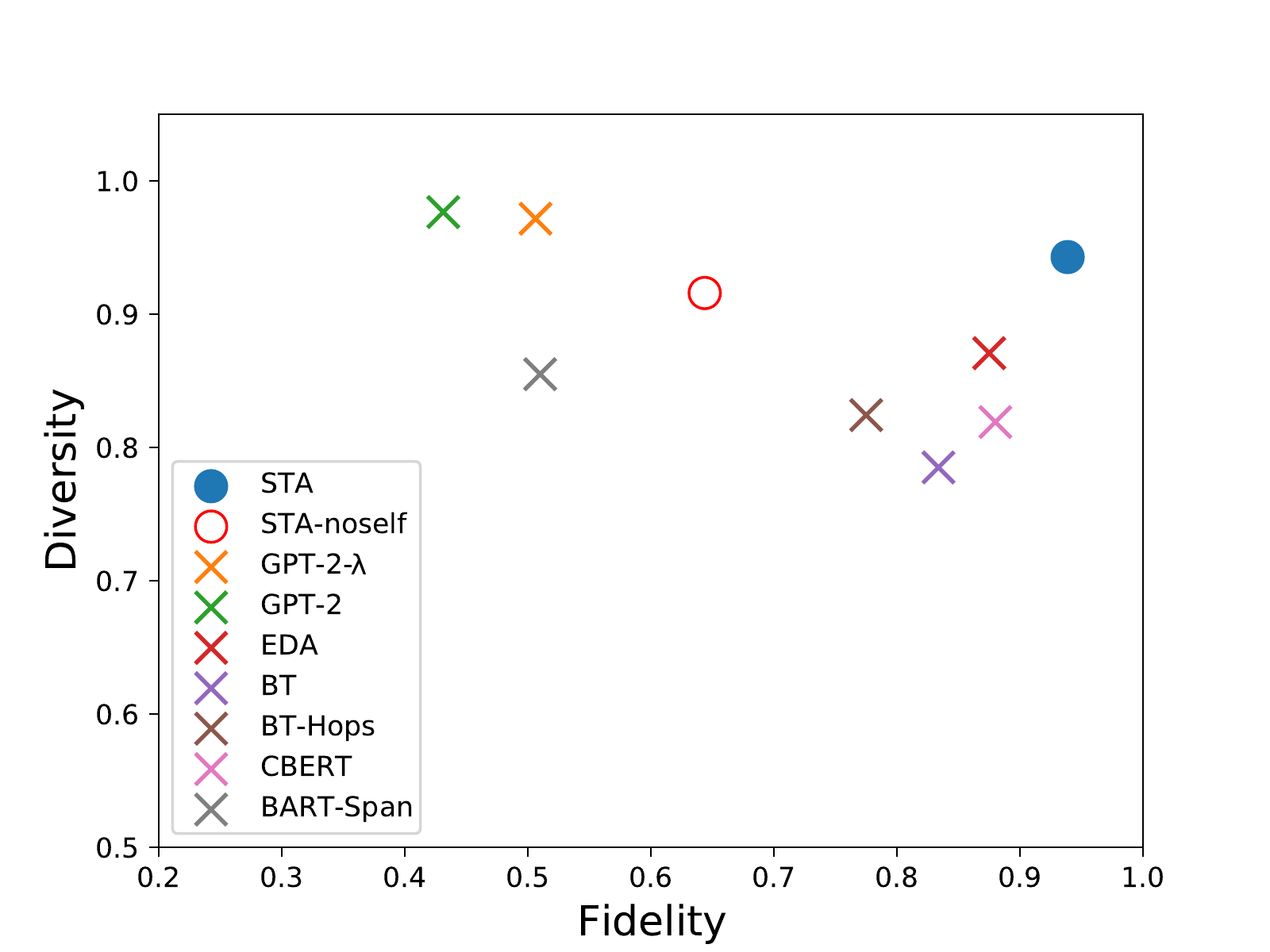}
           \caption{Diversity versus semantic fidelity of generated texts by various augmentation methods}
        \label{fig:sta-humaid-df}
\end{figure}

To present the quality of generated data in diversity and fidelity, the training data (10 samples per class) along with its augmented data ($\beta=1$) is used for investigation. Figure~\ref{fig:sta-humaid-df} depicts the diversity versus semantic fidelity of generated data by various augmentation methods for \textbf{HumAID}\footnote{The average scores over 10 runs are reported.}. It is found that generation-based approaches such as GPT-2 or GPT-2-$\lambda$, achieve strong diversity but less competitive fidelity (i.e. the generated samples are less representative of the labels they are intended to match). On the contrary, easy reformulation methods such as EDA perform well in retaining the semantic meaning but not in lexical diversity. The merit of STA is that it performs well in both diversity and fidelity, as can be seen from its position at the top-right of the figure. Finally, when comparing STA with and without self-checking, it can be seen that each approach produces highly diverse samples, although the self-checking step of STA results in a much higher level of semantic fidelity. This supports the notion that the generation-based approach (STA) is able to produce novel data that is lexically diverse, whilst the self-checking procedure can ensure consistent label retention, which produces a high semantic fidelity in the generated samples.



To conduct a qualitative analysis for the generated samples, Table~\ref{tab:sta-humaid-demo-aug-exs} demonstrates some original samples and augmented samples generated by different methods. It is noted that the $5$ augmented samples in each block are randomly selected instead of cherry-picked. Here an easy reformulation method EDA, a generation-based method GPT-2-$\lambda$ and STA-noself are used for comparison. As the results indicate, there is some difference between the original training samples and the augmented samples generated by STA and other methods. In comparison, the samples generated by STA tend to be not only diverse but also highly label-relevant (semantic fidelity). For example, despite high fidelity in EDA, it produces new samples by applying simple reformulation operations such as replacing ``flooding'' with ``Body'' for the first sample, leading to low diversity. Although GPT-2$\lambda$ looks good in terms of diversity, some of its generated samples intended for ``missing or found people'' are more suited to ``search and rescue'', resulting in unwanted noise. As a comparison, STA is superior in both aspects. Besides, without the self-checking component, (i.e., STA-noself) some noisy samples are also included (e.g., the second one in the demonstration), which reflects the effectiveness of the self-checking procedure in STA.

\begin{table*}[!h]
\scriptsize
\centering

\begin{tabular}{c|p{42em}}
\toprule
\multicolumn{2}{l}{Original   training examples and augmented examples for ``missing or found people"  of HumAID}                                                                      \\
\midrule

\multirow{5}{*}{Original}     & UPDATE: Body found of man who disappeared amid Maryland flooding                                                 \\
                              & Open Missing People Search Database from   Mati and Rafina areas  \#Greecefires   \#PrayForGreece \#PrayForAthens                             \\
                              & @ThinBlueLine614 @GaetaSusan   @DineshDSouza case in point, \#California Liberalism has created the hell  which has left 1000s missing 70 dead,...    \\
                              & Heres the latest in the California wildfires \#CampFire 1011 people are missing Death toll rises to 71 Trump blames fires on poor ...   \\
                              & \#Idai victims buried in mass grave in   Sussundenga, at least 60 missing - \#Mozambique  \#CycloneIdai \#CicloneIdai                                  \\
                              \midrule
\multirow{5}{*}{EDA}          & update flooding found of man who disappeared amid maryland Body                                                                                         \\
                              & open missing people search database from   mati escape and rafina areas greecefires prayforgreece prayforathens                                        \\
                              & created gaetasusan dineshdsouza hell in   point california missing has thinblueline the case which has left s   liberalism dead an countless people... \\
                              & heres blames latest in the california   wildfires campfire people are missing death toll rises to trump more fires on poor...            \\
                              & idai victims buried in mass grave in   sussundenga at mozambique missing least cycloneidai cicloneidai                                                 \\
                              \midrule
\multirow{5}{*}{GPT-2-lambda} & @KezorNews - Search remains in \#Morocco after @deweathersamp; there has   been no confirmed death in \#Kerala                                         \\
                              & \#Cambodia - Search \& Rescue is   assisting Search \& Rescue officials in locating the missing 27 year old   woman who disappeared in ...             \\
                              & @JHodgeEagle Rescue Injured After Missing   Two Children In Fresno County                                                                              \\
                              & \#Florence \#Florence Missing On-Rescue   Teams Searching For Search and Rescue Members \#Florence \#Florence   \#DisasterInformer \#E                 \\
                              & RT @LATTAODAYOUT: RT @HannahDorian:   Search Continues After Disappearance of Missing People in Florida                                                \\
                              \midrule
\multirow{5}{*}{STA-noself}   & Search Database from Matias, Malaysia, missing after \#Maria, \#Kerala,   \#Bangladesh \#KeralaKerala, \#KeralaFloods, ...                             \\
                              & RT @hubarak: Yes, I can guarantee you   that our country is safe from flooding during the upcoming weekend!  Previous story Time Out! 2 Comments       \\
                              & The missing persons who disappeared amid   Maryland flooding are still at large. More on this in the next article.                                     \\
                              & the number of missing after \#CycloneIdai   has reached more than 1,000, reports CNN.                                                                  \\
                              & RT @adriane@przkniewskiZeitecki  1 person missing, police confirm  \#CycloneIdai. \#CicloneIdai                                                        \\
                              \midrule
\multirow{5}{*}{STA}          & The missing persons who disappeared amid Maryland flooding are still at   large. More on this in the next article.                                     \\
                              & Search Triangle County for missing and   missing after \#Maria floods \#DisasterFire                                                                   \\
                              & Just arrived at San Diego International   Airport after \#Atlantic Storm. More than 200 people were missing, including   13 helicopters ...            \\
                              & Search Database contains information on   missing and found people \#HurricaneMaria, hashtag \#Firefighter                                             \\
                              & Were told all too often that Californians   are missing in Mexico City, where a massive flood was devastating. ...             \\
                              \bottomrule
\end{tabular}
\caption{The demonstration of original training samples and augmented samples for ``missing or found people'' of \textbf{HumAID}}
\label{tab:sta-humaid-demo-aug-exs}
\end{table*}

\subsection{Generalisation to domains beyond crisis}

Having studied the effectiveness of STA for crisis tweet classification, it is interesting to examine the extent to which it can generalise to other domains that are not crisis-related. Three more datasets beyond crisis are used for this purpose.

Following previous augmentation work~\cite{kumar2020data,anaby2020not}, two bench-marking datasets are used in the experiments: \textbf{SST-2}~\cite{socher-etal-2013-recursive} and \textbf{TREC}~\cite{li-roth-2002-learning}. The \textbf{Emotion}~\cite{saravia-etal-2018-carer} dataset is also included to extend the domains of testing STA's effectiveness. The details of each of these datasets can be found in Section~\ref{tab:rds-beyond-crisis}.

Following the same experimental procedures as for \textbf{HumAID}, STA is run on the three datasets (tasks) and its results compared to the baselines are reported, in addition to the results of an ablation study and an analysis of lexical diversity and semantic fidelity.

By comparison to the baselines, the results on the \textbf{SST-2} (Table~\ref{sta-sst2-res}), \textbf{Emotion} (Table~\ref{sta-emotion-res}) and \textbf{TREC} (Table~\ref{sta-trec-res}) classification tasks all demonstrate the effectiveness of the augmentation strategy. Similar to \textbf{HumAID}, in all cases STA achieves state-of-the-art performance for text augmentation across all low-resource settings. When a higher number of samples ($50\text{-}100$) are used for training, STA is better, as in the cases of \textbf{SST-2} and EMOTION tasks, or competitive, as in the case of \textbf{TREC}. Besides this, the performance of STA is superior to that of the other augmentation techniques in very low data regimes ($5\text{-}10\text{-}20$). In fact, across the three tasks, STA on average demonstrates a difference of $+6.4$ and $+3.7$ when trained on only $5$ and $10$ samples per class respectively. To summarise, the results demonstrate the ability of STA to generate salient and effective training samples from limited amounts of data for text classification tasks other than crisis-based.

\begin{figure*}[h!]
\centering
\subfloat[\textbf{SST-2}]{\label{fig:sst2_df}
    \includegraphics[scale=0.45]{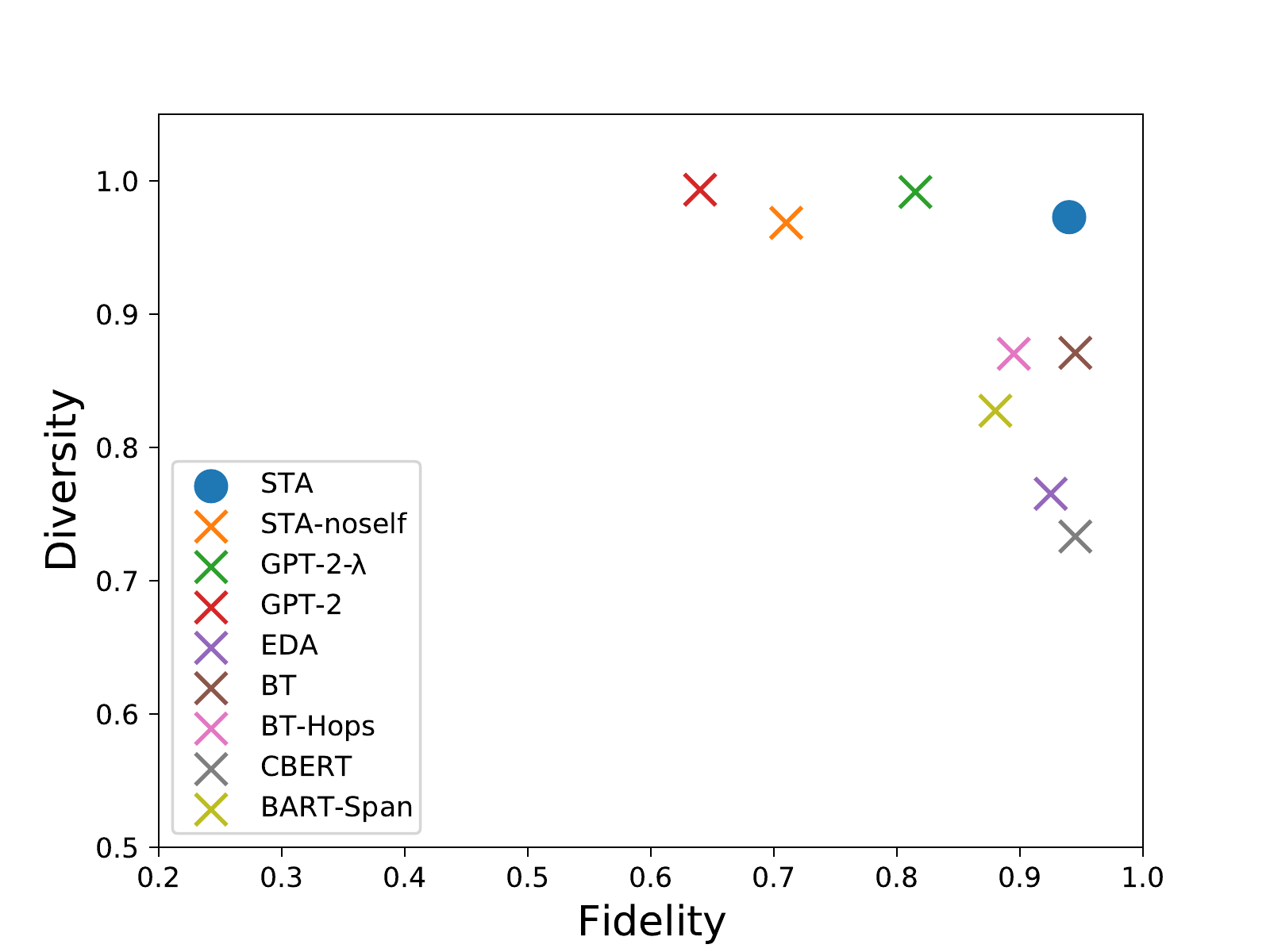}}
\subfloat[\textbf{Emotion}]{\label{fig:emotion_df}
    \includegraphics[scale=0.45]{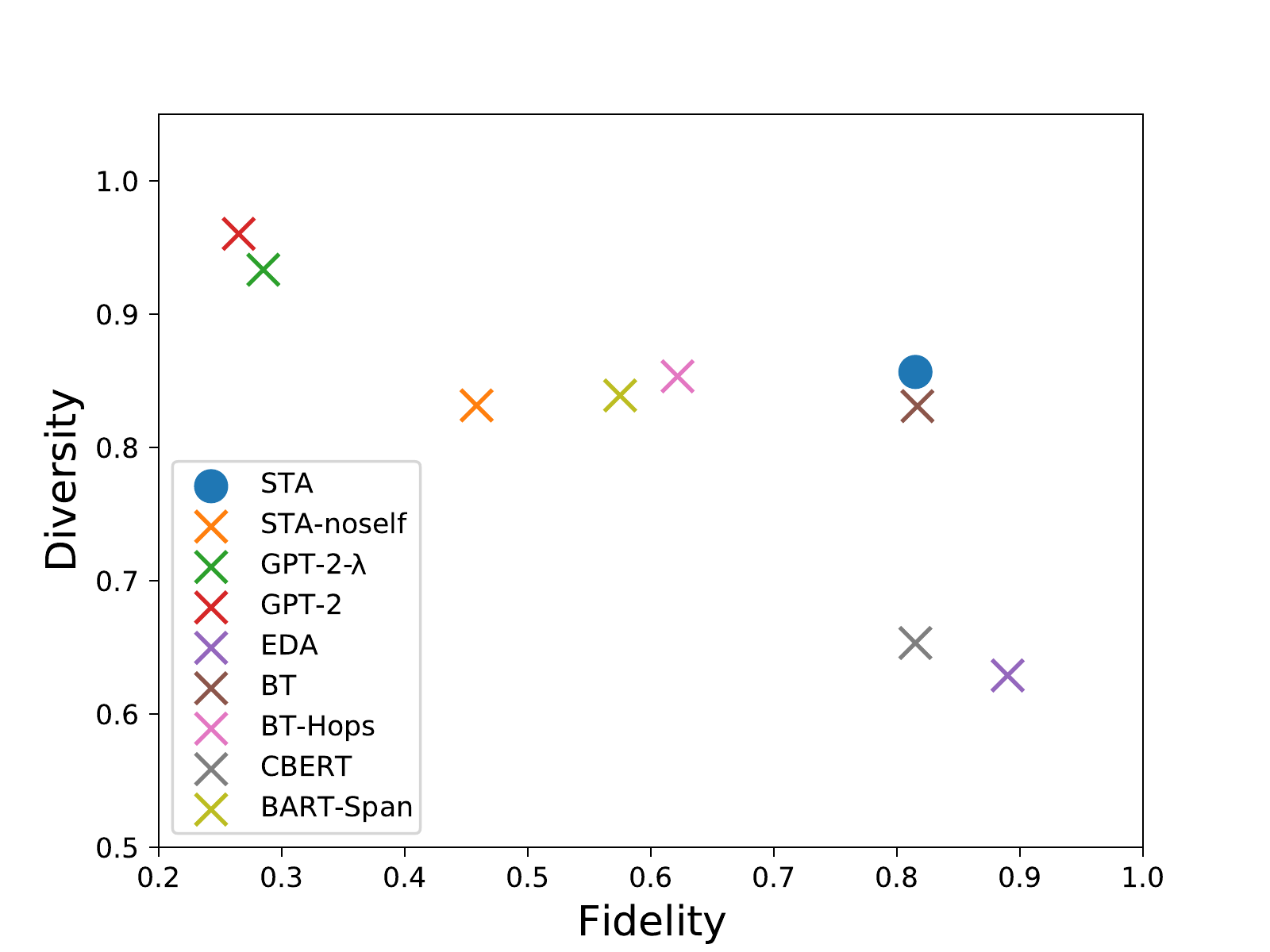}}
    
\subfloat[\textbf{TREC}]{\label{fig:trec_df}
\includegraphics[scale=0.45]{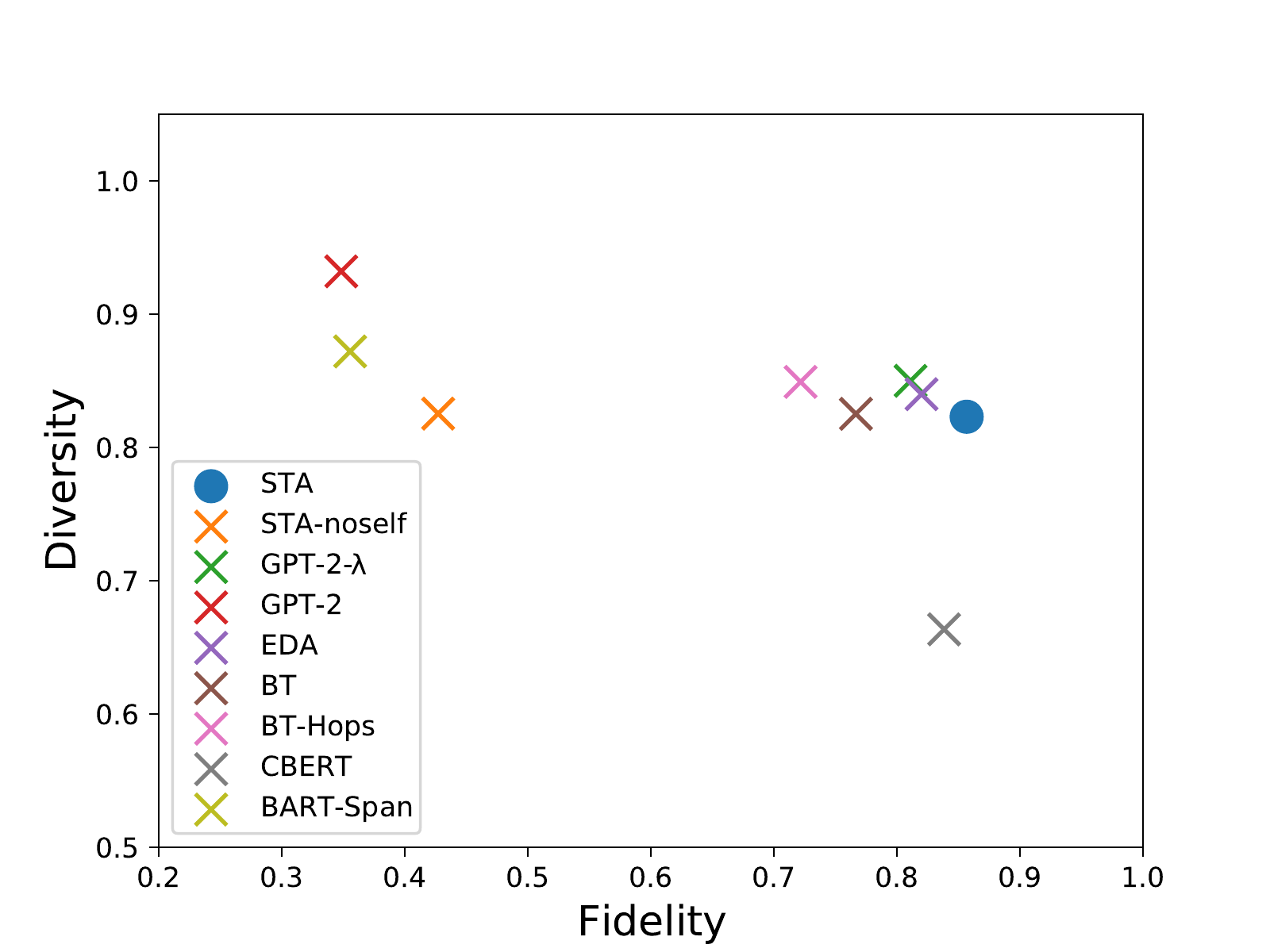}}
        \caption{Diversity versus semantic fidelity of generated texts by various augmentation methods}
        \label{fig:sta-general-df}
\end{figure*}

In the ablation study, it is noticed that the self-checking step and templates used for STA help to improve the downstream classification performance (rows ``STA-noself'' and ``STA-twoprompts''). This is consistent with the results for \textbf{HumAID} (see Section~\ref{subsec:sta-ab-study}). In the investigation of semantic fidelity and lexical diversity, Table~\ref{tab:sta_general_fidelity_clas_perf} shows the performance of the fully-trained classifier on the test set predicted by BERT that is trained on the whole training data for measuring semantic fidelity. Figure~\ref{fig:sta-general-df} depicts the diversity versus semantic fidelity of generated data by various augmentation methods and  STA across the three datasets. As can be seen from its position at the top-right of Figure~\ref{fig:sst2_df},~\ref{fig:emotion_df} and~\ref{fig:trec_df}, STA again achieves the best combination of semantic fidelity and lexical diversity compared to alternative approaches.

\begin{table*}[h!]
\centering
\small
\begin{tabular}{l|lllll}
\toprule
    \textbf{Augmentation Method}                          &  \textbf{5}                    & \textbf{10}                   & \textbf{20}                   & \textbf{50}                   & \textbf{100}                  \\
    
                              \midrule \specialrule{.1em}{.05em}{.05em} 
                              Baseline (No   Aug.) & 56.5\, (3.8)           & 63.1\, (4.1)           & 68.7\, (5.1)          & 81.9\, (2.9)          & 85.8\, (0.8)          \\
                             \midrule 
EDA~\cite{wei2019eda}                  & 59.7\, (4.1)          & 66.6\, (4.7)          & 73.7\, (5.6)          & 83.2\, (1.5)          & 86.0\, (1.4)           \\
BT~\cite{edunov2018understanding}                   & 59.6\, (4.2)          & 67.9\, (5.3)          & 73.7\, (5.8)          & 82.9\, (1.9)          & 86.0\, (1.2)          \\
BT-Hops~\cite{shleifer2019low}              & 59.1\, (4.6)          & 67.1\, (5.2)           & 73.4\, (5.2)          & 82.4\, (2.0)          & 85.8\, (1.1)          \\
CBERT~\cite{wu2019conditional}                & 59.8\, (3.7)            & 66.3\, (6.8)          & 72.9\, (4.9)          & 82.5\, (2.5)          & 85.6\, (1.2)          \\

GPT-2~\cite{kumar2020data}                 & 53.9\, (2.8)          & 62.5\, (3.8)           & 69.4\, (4.6)          & 82.4\, (1.7)           & 85.0\, (1.7)          \\
GPT-2-$\lambda$~\cite{anaby2020not}          & 55.4\, (4.8)& 	65.9\, (4.3)& 	76.2\, (5.6)& 	84.5\, (1.4)	& 86.4\, (0.6)
          \\
BART-Span~\cite{kumar2020data}            & 60.0\, (3.7)          & 69.0\, (4.7)          & 78.4\, (5.0)          & 83.8\, (2.0)           & 85.8\, (1.0)          \\
\midrule
STA-noself   & 66.7\, (5.0)   &	77.1\, (4.7)	  & 81.8\, (2.1)   &	84.8\, (1.0)   &	85.7\, (1.0)

\\
STA-twoprompts   & 69.8\, (4.9) &	 79.1\, (3.4)	 & 81.7\, (4.5) &	\textbf{86.0\, (0.8)} &	\textbf{87.5\, (0.6)}
\\

\textbf{STA}       & \textbf{72.8\, (6.2)} & \textbf{81.4\, (2.6)} & \textbf{84.2\, (1.8)} & \textbf{86.0\, (0.8)} & 87.2\, (0.6) \\
\bottomrule
\end{tabular}

\caption{STA on \textbf{SST-2} in $5, 10, 20, 50, 100$ examples per class}
\label{sta-sst2-res}
\end{table*}

\begin{table*}[h!]
\centering
\small
\begin{tabular}{l|lllll}
\toprule
    \textbf{Augmentation Method}                          & \textbf{5}                    & \textbf{10}                   & \textbf{20}                   & \textbf{50}                   & \textbf{100}                  \\
\midrule \specialrule{.1em}{.05em}{.05em}
Baseline (No   Aug.) & 26.7 \, (8.5)          & 28.5 \, (6.3)          & 32.4 \, (3.9)          & 59.0 \, (2.6)          & 74.7 \, (1.7)           \\
\midrule
EDA                  & 30.1 \, (6.2)          & 33.1 \, (4.3)          & 47.5 \, (5.0)          & 66.7 \, (2.7)          & 77.4 \, (1.8)          \\
BT                   & 32.0 \, (3.0)          & 37.4 \, (3.0)          & 48.5 \, (5.1)          & 65.5 \, (2.0)           & 75.6 \, (1.6)          \\
BT-Hops              & 31.3 \, (2.6)          & 37.1 \, (4.6)          & 49.1 \, (3.5)          & 65.0 \, (2.3)          & 75.0 \, (1.5)          \\
CBERT                & 29.2 \, (6.5)          & 32.6 \, (3.9)          & 44.1 \, (5.2)          & 62.1 \, (2.0)          & 75.5 \, (2.2)          \\
GPT-2                 & 28.4 \, (8.5)          & 31.3 \, (3.5)          & 39.0 \, (4.1)          & 57.1 \, (3.1)          & 69.9 \, (1.3)           \\
GPT-2-$\lambda$          & 28.6 \, (5.1)          & 30.8 \, (3.1)          & 43.3 \, (7.5)          & 71.6 \, (1.5)           & 80.7 \, (0.4)           \\
BART-Span             & 29.9 \, (4.5)           & 35.4 \, (5.7)          & 46.4 \, (3.9)          & 70.9 \, (1.5)          & 77.8 \, (1.0)          \\
\midrule
STA-noself & 34.0 \, (4.0)& 	41.4 \, (5.5)& 	53.3 \, (2.2)& 	65.1 \, (2.3)	& 74.0 \, (1.1)

\\
STA-twoprompts   & 41.8 \, (6.1) &	56.2 \, (3.0) &\textbf{64.9 \, (3.3)} &	75.1 \, (1.5)	 &81.3 \, (0.7)
\\

\textbf{STA}           & \textbf{43.8 \, (6.9)} & \textbf{57.8 \, (3.7)} & 64.1 \, (2.1)& \textbf{75.3 \, (1.8)} & \textbf{81.5 \, (1.1)} \\
\bottomrule

\end{tabular}
\caption{STA on \textbf{Emotion} in $5, 10, 20, 50, 100$ examples per class}
\label{sta-emotion-res}
\end{table*}

\begin{table*}[h!]
\centering
\small
\begin{tabular}{l|lllll}
\toprule
    \textbf{Augmentation Method}                          & \textbf{5}                    & \textbf{10}                   & \textbf{20}                   & \textbf{50}                   & \textbf{100}                  \\
\midrule \specialrule{.1em}{.05em}{.05em} 
Baseline (No   Aug.) & 33.9 \, (10.4)         & 55.8 \, (6.2)          & 71.3 \, (6.3)          & 87.9 \, (3.1)          & 93.2  \, (0.7) \\
\midrule

EDA                  & 54.1 \,  (7.7)         & 70.6 \, (5.7)          & 79.5 \, (3.4)          & 89.3  \, (1.9) & 92.3 \, (1.1)          \\
BT       &           56.0 \, (8.7) &	67.0 \, (4.1)	&79.4 \, (4.8)&	89.0 \, (2.4)	& 92.7 \, (0.8)                     \\
BT-Hops      &    53.8 \, (8.2)&	67.7 \, (5.1)&	78.7  \, (5.6)&	88.0  \, (2.3)	& 91.8 \, (0.9) \\
CBERT                &    52.2 \, (9.8)	&67.0 \, (7.1)&	78.0 \, (5.3)	&89.1 \, (2.5)&	92.6 \, (1.1) \\
GPT-2                &   47.6 \, (7.9) &	67.7 \, (4.9) &	76.9 \, (5.6)	 & 87.8 \, (2.4) &	91.6 \, (1.1) \\
GPT-2-$\lambda$           &    49.6 \, (11.0)&	70.2 \, (5.8)&	80.9 \, (4.4)&	\textbf{89.6 \, (2.2)} &\textbf{93.5 \, (0.8)}           \\

BART-Span            &   55.0 \, (9.9)	&65.9 \, (6.7)&	77.1 \, (5.5)&	88.38 \, (3.4)&	92.7 \, (1.6)        \\
\midrule 
STA-noself   & 45.4 \, (3.2)  & 	61.9 \, (10.2)  & 	77.2 \, (5.5)  & 	88.3 \, (1.2)	  & 91.7 \, (0.8)
\\
STA-twoprompts   & 49.6 \, (9.0)  & 	69.1 \, (8.0)  & 	81.0 \, (5.9)  & 	89.4 \, (3.0)  & 	93.1 \, (0.9)
\\
\textbf{STA}        & \textbf{59.6 \, (7.4)}    & \textbf{70.9 \, (6.6)} & \textbf{81.1 \, (3.9)} & 89.1 \, (2.7) & 93.2 \, (0.8) \\
\bottomrule
\end{tabular}

\caption{STA on \textbf{TREC} in $5, 10, 20, 50, 100$ examples per class}
\label{sta-trec-res}
\end{table*}

\begin{table}[h!]
    \centering
    \begin{tabular}{cccc}
    \toprule
         & SST-2 & Emotion & TREC \\
       Test  & 91.8	& 93.5&	96.6 \\
       \bottomrule
    \end{tabular}
    \caption{Accuracy (in \%) on test set predicted by BERT that is trained on the whole training data for measuring semantic fidelity.}
    \label{tab:sta_general_fidelity_clas_perf}
\end{table}

\subsection{Summary}

This section introduced a novel strategy (named STA) for text-based data augmentation that used pattern-exploiting training to generate training samples and achieve better label alignment. STA substantially outperformed the previous state-of-the-art augmentation approaches on the crisis tweet classification task across a range of low-resource scenarios. In addition, experiments also showed that STA generalised well to other domains such as sentiment, emotion, and topic classifications. Furthermore, an analysis of the lexical diversity and label consistency of generated samples was provided, demonstrating that the approach produces uniquely varied training samples with more consistent label alignment than previous work. 

This work explored the possibility of using data augmentation for boosting text classification performance when the downstream model was finetuned using pre-trained language models. The results showed that STA consistently performed well across different domains using the same experimental setup, which addressed the limitation stated in the previous work~\cite{kumar2020data} calling for a unified data augmentation technique. However, there remains room to improve STA. In STA, new samples are generated and self-checked using a one-time fine-tuned seq2seq model. This raises the question of whether iteratively fine-tuning the seq2seq model would be beneficial. Additionally, the samples generated by STA are not checked for duplicates among themselves. This prompts the question of whether adding a deduplication component would be useful. The iterative self-controlled augmentation (ISA) method is introduced next as an optimised work of STA.



\section{Iterative self-controlled augmentation}
\label{sec:isa}

\begin{figure*}
    \centering
    \includegraphics{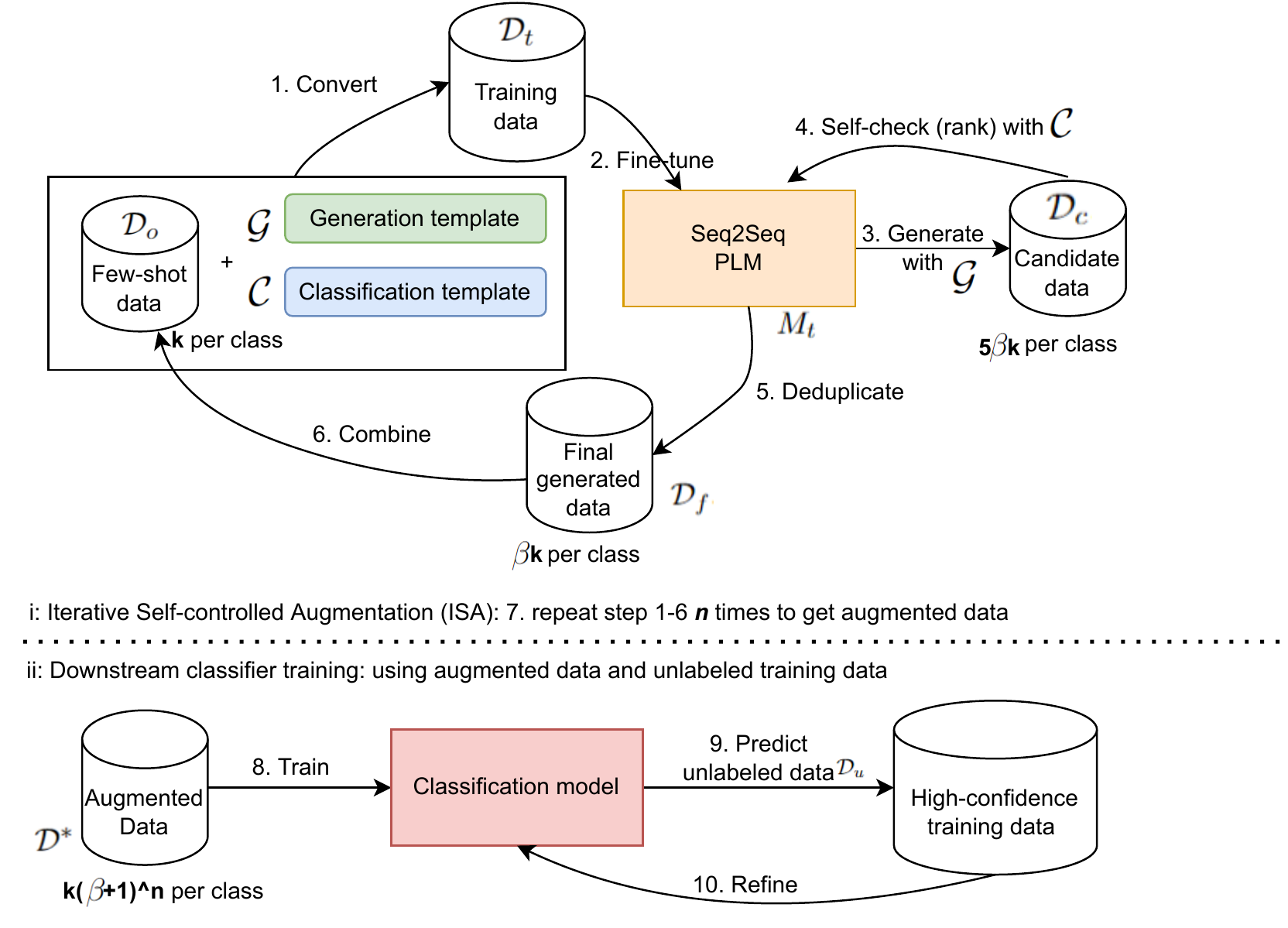}
    \caption{The overview of Iterative Self-controlled Augmentation (ISA) approach for few-shot text classification}
    \label{fig:isa-overview}
\end{figure*}

STA is demonstrated to be effective in the crisis message categorisation domain as well as for other text classification tasks. In particular, the self-checking mechanism helps to improve the quality of the generated texts by STA in terms of semantic fidelity. To ensure better quality of generated texts for both lexical diversity and semantic fidelity, iterative self-controlled augmentation (ISA) is introduced in this section. ISA is an optimised version of STA that adds two extra steps to the text generation process: an iterative step and a de-duplication step. Figure~\ref{fig:isa-overview} depicts the overview of ISA and Algorithm~\ref{alg:isa-alg} describes the overall process of using ISA for few-shot text classification in simple mathematical terms. The text in bold indicates the difference introduced by ISA as compared to STA (see Algorithm~\ref{alg:sta-alg}). 


\begin{algorithm}
\caption{: Iterative Self-Controlled Text Augmentation (ISA)}\label{alg:isa-alg}
\begin{algorithmic}[1]

\Require Few-shot labelled data $\mathcal{D}_o$. Generation model $M$. Generation template $\mathcal{G}$. Classification template $\mathcal{C}$. \textbf{Unlabelled target data $\mathcal{D}_u$.}

\State Convert $\mathcal{D}_{o}$ to training dataset $\mathcal{D}_{t}$ via $\mathcal{G}$ and $\mathcal{C}$.
\State Finetune $M$ on $\mathcal{D}_{t}$ in a generation task and a classification task jointly to obtain $M_{t}$.
\State Use $\mathcal{G}$ and $M_{t}$ to generate candidate dataset $\mathcal{D}_{c}$.
\State Apply $M_t$ to do classification inference on $\mathcal{D}_{c}$ with $\mathcal{C}$ to \textbf{rank $\mathcal{D}_{c}$ by prediction confidence}.
\textbf{\State De-duplicate the confidence-ranked samples through semantic similarity checking to get final generated dataset  $\mathcal{D}_{f}$.}
\State Combine the final generated dataset with the selected samples.
\textbf{\State Repeat step 1 to 6 \textit{n} times to get the final augmented training data $\mathcal{D}^*$.}
\State Use the augmented data for downstream classification model training
\textbf{\State Use the trained downstream model to make predictions for the unlabelled target data $\mathcal{D}_u$.
\State Select high-confidence predicted samples for model refinement.}
\end{algorithmic}
\end{algorithm}

For ISA, to get the augmented data (step $1-7$), first a seq2seq transformer (generation model) is fine-tuned on the training data that is converted from the original few-shot data by the generation templates and classification templates. The fine-tuned seq2seq model is then used for generating new samples and selecting samples with high confidence of being faithful to the appropriate label. Next, to increase the diversity of the selected samples, a de-duplication step is taken to check their semantic similarities. The selected data is then combined with the few-shot data for next round of model training and new samples are generated. This iterative process is repeated $n$ times to generate the augmented data for downstream classification training. When training the downstream classifier, it is first trained with the augmented data and then refined on the unlabelled data with high-confidence predictions by the classifier itself (step $8-10$). The following section introduces the details of ISA.

\subsection{Iterative self-controlled method}

Given the difference of ISA compared to STA, this section is divided into three components describing ISA for crisis few-shot text classification: iterative augmentation, generation de-duplication, and model refinement. 

\subsubsection{Iterative augmentation}
In a round of iterative augmentation, let $\mathcal{D}_{o}$ be the initial dataset for few-shot text classification, known as the seed or few-shot data. It consists of $k$ samples per class. Similar to STA, $\mathcal{D}_{o}$ is first converted to the training data $\mathcal{D}_{t}$ by the generation templates $\mathcal{G}$ and classification templates $\mathcal{C}$ (see Table~\ref{tab:sta-templates}). The seq2seq model is then fine-tuned on this training data and used to generate a candidate dataset $\mathcal{D}_{c}$ that consists of $5\beta k$ samples per class where $\beta$ refers to the augmentation factor (i.e., how many times more samples are generated for the original $k$ samples per class). By applying the self-checking mechanism of STA on $\mathcal{D}_{c}$, the $5\beta k$ samples per class of $\mathcal{D}_{c}$ are ranked by the confidence scores predicted by the seq2seq model (See Section~\ref{subsec:sta-self-check}), represented by $\mathcal{D}_{\hat{c}}$. Then $\mathcal{D}_{\hat{c}}$ is de-duplicated by semantic similarity checking, forming the final generated data of this round $\mathcal{D}_{f}$, consisting of $\beta k$ samples per class (introduced later in Section~\ref{subsec:isa-dedup}). This indicates that the de-duplication step is used to select only the top $20\%$ dissimilar samples from $\mathcal{D}_{\hat{c}}$ and add them to $\mathcal{D}_{f}$. After obtaining the selected augmented data $\mathcal{D}_{f}$ in one round, the union of $\mathcal{D}_{f}$ and $\mathcal{D}_{0}$ containing $k(\beta+1)$ samples per class becomes the training set for the next round of iteration. In ISA, the iteration continues until $n$ rounds have been completed. After $n$ rounds of iterations, mathematically, this leads to the final augmented data  $\mathcal{D}^*$ that comprises $k(\beta+1)^n$ samples per class. After the iterative augmentation obtaining $\mathcal{D}^*$, the next step is to use it along with an unlabelled dataset $\mathcal{D}_u$ for downstream model training, introduced in Section~\ref{subsec:isa-model-refinement}.






\subsubsection{Generation de-duplication}
\label{subsec:isa-dedup}

The purpose of de-duplication is to select samples from the confidence-ranked dataset $\mathcal{D}_{\hat{c}}$ that are as diverse as possible (taking a single iterative round as an example). After the de-duplication, only $\beta$ samples per class are obtained from the $5\beta k$ samples per class of $\mathcal{D}_{\hat{c}}$, forming the final selected augmented data $\mathcal{D}_{f}$. Algorithm~\ref{alg:isa-dedup} presents the details of the de-duplication process. In this de-duplication, $\mathcal{D}_{r}$ is a copy of the seed data $\mathcal{D}_{o}$ used as the reference data for checking the semantic similarity between each sample $x_i$ from $\mathcal{D}_{\hat{c}}$ and each sample $x_j$ of $\mathcal{D}_{r}$\footnote{Making $\mathcal{D}_{r}$ a copy of $\mathcal{D}_{o}$ is because $\mathcal{D}_{r}$ is dynamically updated in checking against $\mathcal{D}_{\hat{c}}$ and $\mathcal{D}_{o}$ must remain unchanged in order to be combined with $\mathcal{D}_{f}$ at the end of one round.}. To measure the similarity, cosine similarity is used and calculated as follows.

\begin{equation}
    s_{i,j} = \text{cosine}(\mathcal{V}(x_i), \mathcal{V}(x_j)), \text{where } x_i \in \mathcal{D}_{\hat{c}} \text{ and } x_j \in \mathcal{D}_{r}
\end{equation}

where $\mathcal{V}$ is a sentence embedding model for vectorising $x_i$ and $x_j$. To determine if $x_i$ is added to $\mathcal{D}_{f}$, $x_i$ is matched with all samples of $\mathcal{D}_{r}$ to get the similarity collection:  $S_i=\{s_{i,j}\}|_{j=1}^{\text{len}(\mathcal{D}_{r})}$. To de-duplicate, $x_i$ is added to $\mathcal{D}_{f}$ only when the maximum similarity score is less than or equal to a threshold $\epsilon$, formulated as follows.

\begin{equation}
  \mathcal{D}_{f} = \mathcal{D}_{f} \cup \{x_i\}, \text{if } \text{max}(S_i) \leq \epsilon
\end{equation}

Once $x_i$ is selected based on this check, it is also added to the reference data $\mathcal{D}_{r}$ for next candidate checking. With this de-duplication, the selected samples in $\mathcal{D}_{f}$ are diversified between themselves and also become dissimilar to the seed samples in terms of semantics.

\algrenewcommand\algorithmicrequire{\textbf{Input:}}
\algrenewcommand\algorithmicensure{\textbf{Output:}}
\begin{algorithm}
\caption{Generation De-duplication}
\label{alg:isa-dedup}
\begin{algorithmic}

\Require Original few-shot data: $\mathcal{D}_{o}$, Confidence-ranked data: $\mathcal{D}_{\hat{c}}$, Embedding model: $\mathcal{V}$
\Ensure Final generated data: $\mathcal{D}_{f}$. No. of candidates for deduplication: $5\beta k$, No. after deduplication ($20\%$): $\beta k$

\State $\mathcal{D}_{r} \gets \mathcal{D}_{o}$
\State $i \gets 0$
\State $\mathcal{D}_{f} \gets \emptyset$
\While{$i \neq 5\beta k$}
\State $j \gets 0$
\State $S_i \gets \emptyset$

\While{$j \neq \text{len}(\mathcal{D}_{r})$}
\State $s_{i,j} \gets \text{cosine}(\mathcal{V}(x_i), \mathcal{V}(x_j))$, where $x_i \in \mathcal{D}_{\hat{c}}$ and $x_j \in \mathcal{D}_{r}$
\State $S_i \gets S_i \cup \{s_{i,j}\}$

\State $j \gets j + 1$
\EndWhile

\If{$\text{max}(S_i) \leq \epsilon$}
    \State $\mathcal{D}_{f} \gets \mathcal{D}_{f} \cup \{x_i\}$ \# $x_i$ is selected and added to $\mathcal{D}_{f}$ if it hits a low similarity score.
    \State $\mathcal{D}_{r} \gets \mathcal{D}_{r} \cup \{x_i\}$ \# $x_i$ is selected and it has to be added to $\mathcal{D}_{r}$ for duplication checking to avoid selecting samples similar to it in the reaming steps of loop. 
\EndIf
\If{$\text{len}(\mathcal{D}_{f}) \geq \beta k$}
    \State break \# this indicates 20\% out of candidates are selected after de-duplication.
\EndIf

\State $i \gets i + 1$
\EndWhile
\end{algorithmic}
\end{algorithm}

\subsubsection{Model refinement using unlabelled data}
\label{subsec:isa-model-refinement}
After the iterative augmentation with de-duplication, the final augmented dataset
$\mathcal{D}^*$ is obtained for downstream model training. Apart from being trained on $\mathcal{D}^*$, the downstream model is also refined on an unlabelled corpus from the target task $\mathcal{D}^u$. The lower part of Figure~\ref{fig:isa-overview} shows that the process of downstream model refinement is iterative. Given a downstream model, it is first trained on $\mathcal{D}^*$. Then the model is used to make predictions for the unlabelled samples of $\mathcal{D}^u$. Among the predicted samples, only the samples with high confidence are selected for refinement (i.e., as the training data for the model in next iteration). The confidence of a sample is determined by its predicted probability by the model and the sample is selected when its confidence is above a threshold $\sigma$. For ISA, the refinement continues to iterate while high-confidence samples continue to be identified. The idea behind refinement is to enable the model to self-learn from its own predictions on the unlabelled samples (also known as ``self-training'' in the literature~\cite{meng2018weakly,meng-etal-2020-text}).

\subsection{Experiments}



The purpose of the experiments in STA was to compare its performance to augmentation-based few-shot baselines, using synthetic data generated by STA and the baselines to see how it impacted downstream classification performance. However, it is not yet clear how STA performs compared to prompt-based approaches (Section~\ref{subsec:prompted-based-few-shot}). Therefore in the following experiments both STA and ISA are compared to these prompt-based approaches. To ensure a fair and parallel comparison, STA and ISA are run using the same experimental setup as the prompt-based approaches, including the use of certain language models and the number of random sampling. In addition, ISA is also compared to the state-of-the-art augmentation-based STA to determine if the synthetic data generated by ISA is an improvement over STA. The following sections provide more details on the experimental setup.



\subsubsection{Experimental Setup}


For the generation seq2seq model, instead of the base version for STA, the large version of T5,  \texttt{t5-large} is chosen in the experiments for ISA. The hyperparameters of fine-tuning \texttt{t5-large} remain the same as for STA. To decide the iteration number $n$, a grid search over $n \in \{1,2,3,4,5\}$ is conducted and finally it is set to be $3$ based on its overall good performance with this setup. The augmentation factor $m$ is set to be $5$ based on a search over $m \in \{1,2,3,4,5,6,7\}$. Since ISA is proposed for few-shot text classification, the experiments are run in very low-data regimes: the seed data consists of $5$ samples per class ($k_1=5$). The seed data is sampled from the training set of the target task and the rest is treated as the unlabelled data for model refinement (i.e., $D_u$).

The de-duplication threshold $\epsilon$ is set to be $0.9$ and the threshold $\sigma$ for downstream model refinement is set to be $0.95$ given their overall good performance in initial experiments. For downstream model training, the hyperparameters remain the same as for STA but the initial model checkpoint is changed from \texttt{bert-base} to \texttt{roberta-large}\footnote{\url{https://huggingface.co/roberta-large}}, so as to be consistent with the few-shot literature~\cite{zhang2022differentiable,schick2021exploiting,gao2021making}. In the experiments, the parameters are validated based on the final refined model on a development set. It can be challenging to decide the development set as a small such set with great randomness makes it difficult to find the best model checkpoint while a large such set leads to deviation from the real-world few-shot use cases~\cite{gao2021making}. To overcome this problem $50$ examples per class are sampled from the original development set to form the few-shot development set for parameter validation. Additionally, to alleviate the instability of training on small datasets, the experiments are run on $5$ splits of the seed data sampled with different random seeds and hence the performance is reported as the average accuracy score with standard deviation over the $5$ runs.



\subsubsection{Baselines}

Apart from the baseline without augmentation (No Aug.), STA and three state-of-the-art prompt-based few-shot learning approaches are selected from the literature (Section~\ref{subsec:prompted-based-few-shot}) to which ISA is compared in the experiments:
PET~\cite{schick2021exploiting}, LM-BFF~\cite{gao2021making}, and DART~\cite{zhang2022differentiable}.

\begin{itemize}
    \item \textbf{PET} is a semi-supervised approach that fine-tunes language models on a task using cloze-style questions. Hand-crafted prompt templates convert the training data, and multiple models are trained to leverage unlabelled data. The ensemble of models is then used to soft-label unlabelled data.
    \item \textbf{LM-BFF} is a technique for improved few-shot fine-tuning of medium or small-sized language models. Task demonstrations are added to template-based prompts for language model training. Unlike PET, LM-BFF uses automatically generated templates from a seq2seq language model (T5), instead of hand-designed templates.
    \item \textbf{DART} employs unused tokens as differentiable template and label tokens for backpropagation optimization. Its differentiable prompts for few-shot learning aim to generate more discriminative representations than PET's fixed prompts.
\end{itemize}

STA is implemented following the same experimental setup as ISA, namely, 5 random runs in 5 examples per class setting using the large T5 and RoBERTa instead of the base T5 and BERT (as were used in the previous experiments). The three prompt-based baselines are common in making relatively smaller language models for better few-shot learners\footnote{This is opposed to large-scale language models such as GPT-3~\cite{brown2020language}. Although they exhibit state-of-the-art few-shot classification performance, considering fair comparison and the difficulty in re-implementing them, they are not used as baselines in the experiments.}. To adapt them to the crisis categorisation domain, they are implemented by
running the official code released by PET\footnote{\url{https://github.com/timoschick/pet}}, LM-BFF\footnote{\url{https://github.com/princeton-nlp/LM-BFF}}. and DART\footnote{\url{https://github.com/zjunlp/DART}} on target datasets in the few-shot setting of $5$ examples per class, as for STA and ISA.

\subsection{Results and discussion}

In a similar way to STA, ISA is first tested on the crisis message categorisation task represented by the \textbf{HumAID} dataset~\cite{alam2021humaid} and then on other domain tasks such as topic classification or sentiment classification represented by \textbf{Emotion}~\cite{saravia-etal-2018-carer}, \textbf{AgNews}~\cite{Zhang2015CharacterlevelCN}, \textbf{TweetSentiment}~\cite{rosenthal-etal-2017-semeval} and \textbf{TREC}~\cite{li-roth-2002-learning}. This section reports ISA's performance on these datasets as compared to the baselines and discusses the quality of generated texts by ISA as compared to STA. 

\subsubsection{Downstream classification}

Table~\ref{tab:isa-humaid-res} presents the accuracy and standard deviation (in parentheses) of ISA compared to the baselines on \textbf{HumAID} based on 5 runs with 5 examples per class using \texttt{roberta-large}. The first row denoted as Baseline (No Aug.) is a simple baseline that uses no augmented data but the original seed data for downstream model training. The table shows that the iterative self-controlled method with de-duplication to augment the seed data in ISA helps improve the downstream performance substantially as compared to the baseline without augmentation and STA with only self-controlled augmentation (78.0 versus 57.0 and 78.0 versus 72.2 respectively). In addition, although STA hits a comparable score with the best-performing prompt-based approach PET (72.2 vs 73.8), it is also found that ISA outperforms PET by a large margin, as indicated by the $+4.2$ difference (78 vs 73.8). To examine the effect of model refinement on ISA (Section~\ref{subsec:isa-model-refinement}), a run denoted as ``ISA-refine'' that does not use the unlabelled corpus for model refinement is included also. The results show that the refinement helps improve the downstream performance by a small margin, indicated by 78.0 versus 76.1.

\begin{table}[h!]
\centering
\begin{tabular}{ll}
\toprule
                   & HumAID               \\
                   \midrule
Baseline (No Aug.) & 57.0 (7.0)           \\

DART~\cite{zhang2022differentiable}               & 61.2 (5.1)          \\
PET~\cite{schick2021exploiting}                & 73.8 (1.5)         \\
LM-BFF~\cite{gao2021making}             & 57.1 (14.0)          \\

STA                & 72.2 (5.1)
                    \\
ISA-refine                & 76.1 (3.3)          \\
ISA    & \textbf{78.0 (2.6)} \\
\bottomrule
\end{tabular}
\caption{Accuracy of ISA comparing to baselines on \textbf{HumAID} based on 5 runs}
\label{tab:isa-humaid-res}
\end{table}

\subsubsection{Quality of generated texts}

ISA achieves strong downstream classification performance. It is interesting to know whether there is a correlation between the downstream performance and the quality of the generated texts. Hence, the investigation of the quality of generated texts by ISA is conducted by experimentation. The investigation serves to answer two major questions. First, how does the quality of texts generated by ISA comparing to those of STA? Second, what effect does the de-duplication have on ISA regarding the quality of generated texts?  

\begin{figure*}[h!]
\centering
\subfloat[ISA versus STA]{\label{fig:isa-sta-study-humaid-generated-texts-df}
    \includegraphics[scale=0.45]{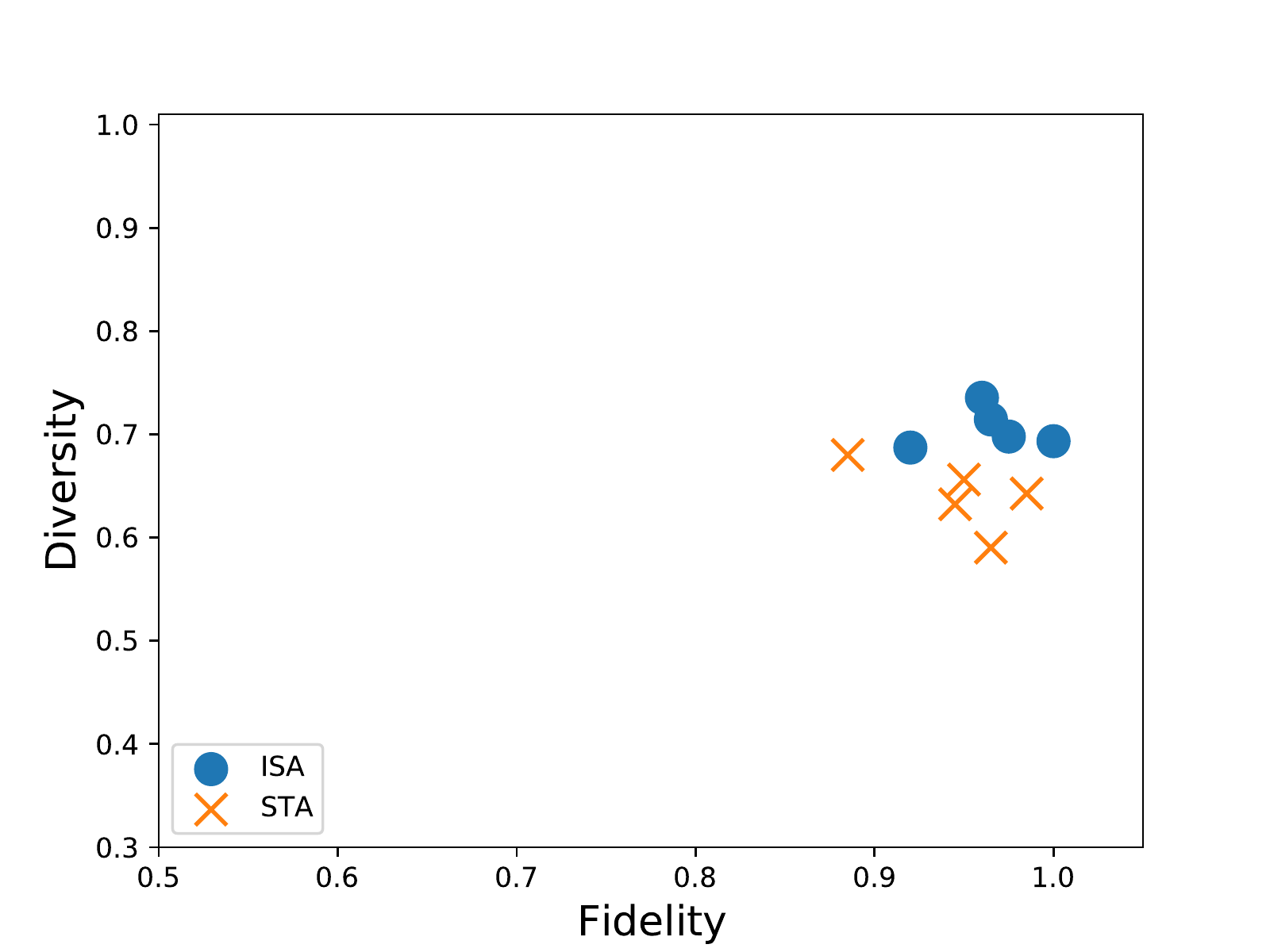}}
\subfloat[Deduplication versus No-Deduplication]{\label{fig:isa-sta-study-humaid-generated-texts-dedup}
    \includegraphics[scale=0.45]{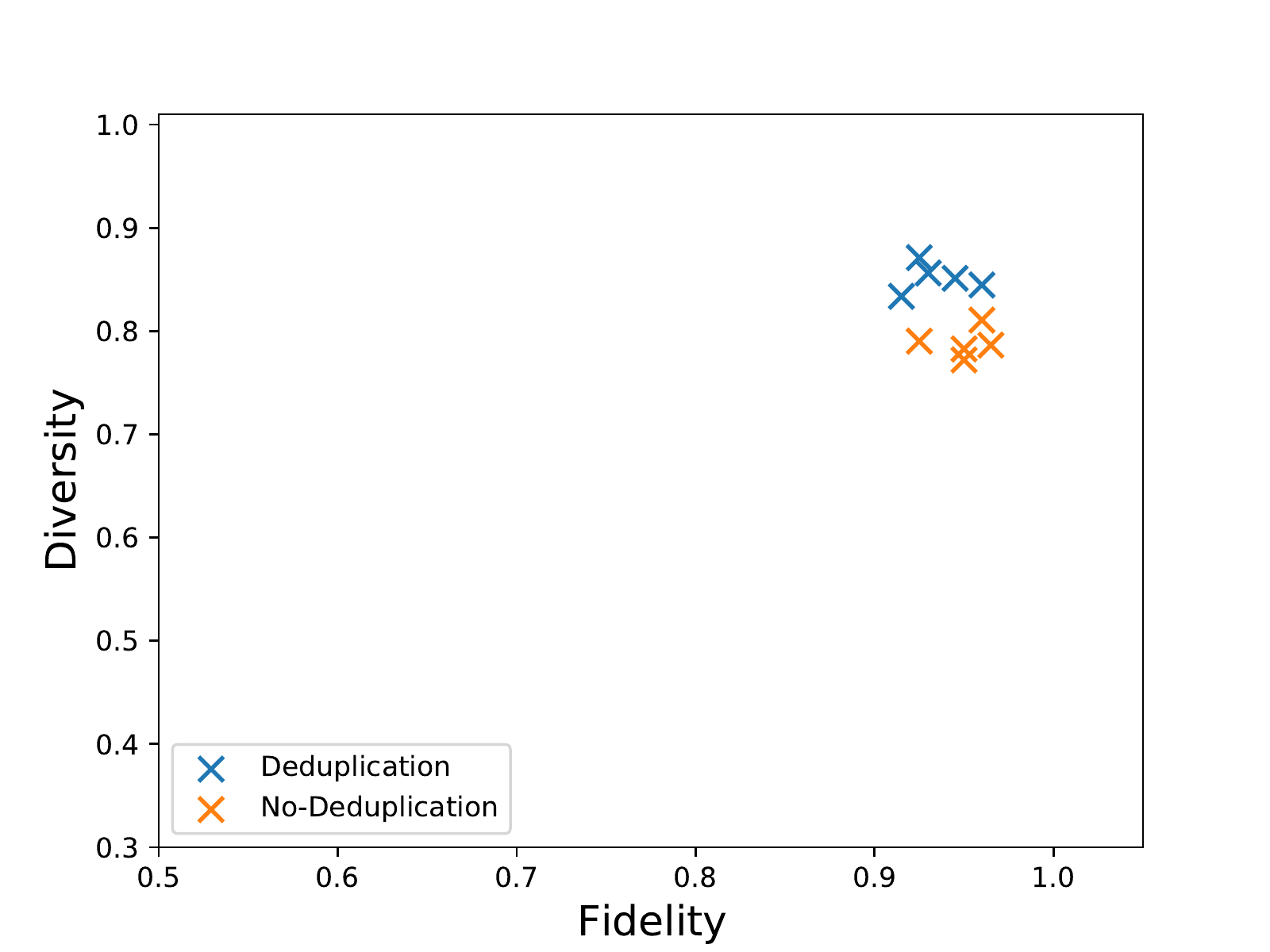}}
        \caption{The investigation of the quality of generated texts by ISA based on 5 random runs}
        \label{fig:isa-study-humaid-generated-texts}
\end{figure*}

The experiment to answer the first question follows the same methodology as was used for STA (Section~\ref{subsec:sta-sfld-study}), i.e, to measure the lexical diversity and semantic fidelity of the generated texts. Likewise, Unique Trigrams is used to measure diversity and fully-trained BERT base is used to measure fidelity. Figure~\ref{fig:isa-sta-study-humaid-generated-texts-df} plots the scores of lexical diversity and semantic fidelity of texts generated by ISA and STA based on 5 runs. It can be seen that ISA is positioned at the right-top of the graph, indicating its superiority in generating texts with both good lexical diversity and semantic fidelity as compared to STA. Furthermore, to conduct a qualitative analysis of the texts generated by the two methods, Table~\ref{tab:sta-demo-ex} lists five randomly-picked samples generated by STA and ISA and five original samples from the label ``missing or found people'' of \textbf{HumAID}. It reveals that despite good diversity of the generated samples by STA as compared to the original samples, there are some near-duplicates among them, such as \#1 and \#3 (the second block) both describing ``Missing in MD After Earthquake ''. By contrast, the generated samples by ISA are not only mutually diverse but also different from the original samples.

To study the effect of de-duplication, the experiment is designed simply to run ISA by setting the augmentation factor $m$ to be 1 and the iteration number $n$ to be 5. At the end of each iteration, the average diversity and fidelity scores of generated texts by ISA with and without de-duplication are recorded, which are plotted in Figure~\ref{fig:isa-sta-study-humaid-generated-texts-dedup}. The results show that ISA with de-duplication achieves better diversity scores while retaining competitive fidelity scores as compared to ISA without de-duplication. This indicates the positive effect of the de-duplication mechanism used in ISA.

\begin{table*}[h!]
\centering
\small
\begin{tabular}{cp{35em}}
\toprule
\multirow{5}{*}{Original} & 1: Maryland county grapples with devastation   left behind by flash flood One man remained missing after flash flooding tore   down historic Main Street in Ellicott City, Maryland.                                                                                                \\
                          & 2: \#California: 200 \#missing in wildfires,   @BBCWorld reports. \#disappeared \#MissingPersons                                                                                                                                                                                    \\
                          & 3: Catastrophic Floods: 1 Missing in MD   From The Weather Channel iPhone App                                                                                                                                                                                                       \\
                          & 4: TODAY clean up continues in Maryland   after the devastating floods.. while the search continues for a missing man.   We’re tracking the latest on @WMTWTV                                                                                                                       \\
                          & 5: One man is missing after devastating   flash flooding swept through Ellicott City, Maryland, turning streets into   rivers and carrying away cars.                                                                                                                               \\
                          \midrule
\multirow{5}{*}{STA}      & 1: Missing in MD After Earthquake | NBC News | 9to5Mac                                                                                                                                                                                                                              \\
                          & 2: : Said one person is missing after   devastating flash flooding swept through Ellicott City, Maryland, turning   streets into rivers and carrying away cars.                                                                                                                     \\
                          & 3: Disaster Relief: Missing in MD From The   Weather Channel iPhone App                                                                                                                                                                                                             \\
                          &4: For information on missing persons in   the area of the 2010 earthquake and tsunami click here. \#Earthquake                                                                                                                                                                     \\
                          &5: A large number of people are missing   after devastating flash flooding swept through Ellicott City, Maryland,   turning streets into rivers and carrying away cars.                                                                                                             \\
                          \midrule
\multirow{5}{*}{ISA}      &1: In addition to the search for bodies, the FBI launched an extensive   manhunt for possible missing people in connection with the devastating 2011   earthquake in Mexico. We are still searching for three missing persons. We   will continue to monitor the situation closely. \\
                          &2: RCMP continues search for a missing man   after devastating flash flooding swept through Ellicott City, Maryland,   turning streets into rivers and carrying away cars.                                                                                                          \\
                          &3: Hikers in Mexico City unaccounted for   after flash flooding swept through town on Sunday afternoon, leaving behind   cars and power lines.                                                                                                                                      \\
                          &4: In the aftermath of Hurricane Florence,   more than 2,000 people have been missing since the beginning of the year.                                                                                                                                                              \\
                          &5: \#Fire and Ice @ChickenFest @CycloneIdai   @RachelJ. @CycloneIdai @RachelJ. @CycloneIdai @WWII @WWII @MissingPersons       \\
                          \bottomrule
\end{tabular}
\caption{Original training examples and augmented examples generated by STA and ISA for ``missing or found people'' of \textbf{HumAID} using \texttt{t5-large}}
\label{tab:sta-isa-humaid-demo-aug-exs}
\end{table*}

\subsubsection{Generalisation to domains beyond crisis}

ISA is demonstrated to be very effective in the crisis message categorisation task, similar to STA, it is then tested on more classification tasks. Apart from \textbf{Emotion}~\cite{saravia-etal-2018-carer} and \textbf{TREC}~\cite{li-roth-2002-learning}, which are also used in STA, two extra datasets \textbf{AgNews}~\cite{Zhang2015CharacterlevelCN} and \textbf{TweetSentiment}~\cite{rosenthal-etal-2017-semeval} are added to test ISA in other domains. The former is a news categorisation task and the latter is a ternary sentiment classification task. More details are described in Section~\ref{tab:rds-beyond-crisis}.

\begin{table}[h!]
\centering
\begin{tabular}{lllll}
\toprule
                   & Emotion              & AgNews              & TweetSentiment      & TREC                \\
                   \midrule
Baseline (No Aug.) & 30.1 (4.4)           & 80.3 (3.2)          & 43.8 (7.2)          & 65.8 (3.8)          \\
DART~\cite{zhang2022differentiable}               & 48.1 (2.1)           & 82.0 (0.9)          & 54.2 (1.4)          & 71.3 (2.3)          \\
PET~\cite{schick2021exploiting}                & 46.9 (1.2)          & \textbf{86.7 (0.4)}          & 46.0 (0.7)          & 77.1 (1.7)          \\
LM-BFF~\cite{gao2021making}             & 52.9 (7.3)           & 60.9 (19.8)         & 51.5 (6.7)          & 40.2 (7.3)          \\
STA         &       52.2 (1.4)&	83.1 (3.0)&	51.7 (5.3)&	71.8 (1.6)
                   \\
ISA-refine                & 58.8 (2.0)           & 83.7 (0.8)          & \textbf{57.6 (7.3)} & 82.6 (3.2)          \\
ISA    & \textbf{61.2 (2.1)} & 86.1 (0.8) & 56.7 (8.8)          & \textbf{84.6 (3.1)} \\
\bottomrule
\end{tabular}
\caption{Accuracy of ISA comparing to baselines on \textbf{Emotion}, \textbf{AgNews}, \textbf{TweetSentiment} and \textbf{TREC} based on 5 runs}
\label{tab:isa-general-res}
\end{table}

Table~\ref{tab:isa-general-res} shows the results of ISA compared to baselines on the four datasets. ISA consistently outperforms the baseline without augmentation as well as STA with only self-controlled augmentation, implied by the average difference $+17.2$ and $+7.5$ respectively. It is also seen that ISA with the refinement outperforms ISA without it (``ISA-refine'') by a small margin, on average $+1.5$. For the prompt-based few-shot baselines, it is difficult to find one method that performs well across all datasets and instead they each tend to succeed on some datasets while producing poorer results in others. For example, DART performs well in \textbf{TweetSentiment} but not for \textbf{Emotion}, LM-BFF succeeds in \textbf{Emotion} but not for \textbf{TREC} and PET is good in \textbf{AgNews} and \textbf{TREC} but not for the others. However, ISA achieves overall strong performance across all of these datasets, outperforming the prompt-based few-shot baselines substantially in all situations except for a slight loss to PET in AgNews  (86.1 versus 86.7). This suggests that ISA is a generalisable approach for text classification of different domains in very low-data regimes.

\begin{figure*}[h!]
\centering
\subfloat[\textbf{Emotion}]{\label{fig:sta_isa_emotion_df}
    \includegraphics[scale=0.45]{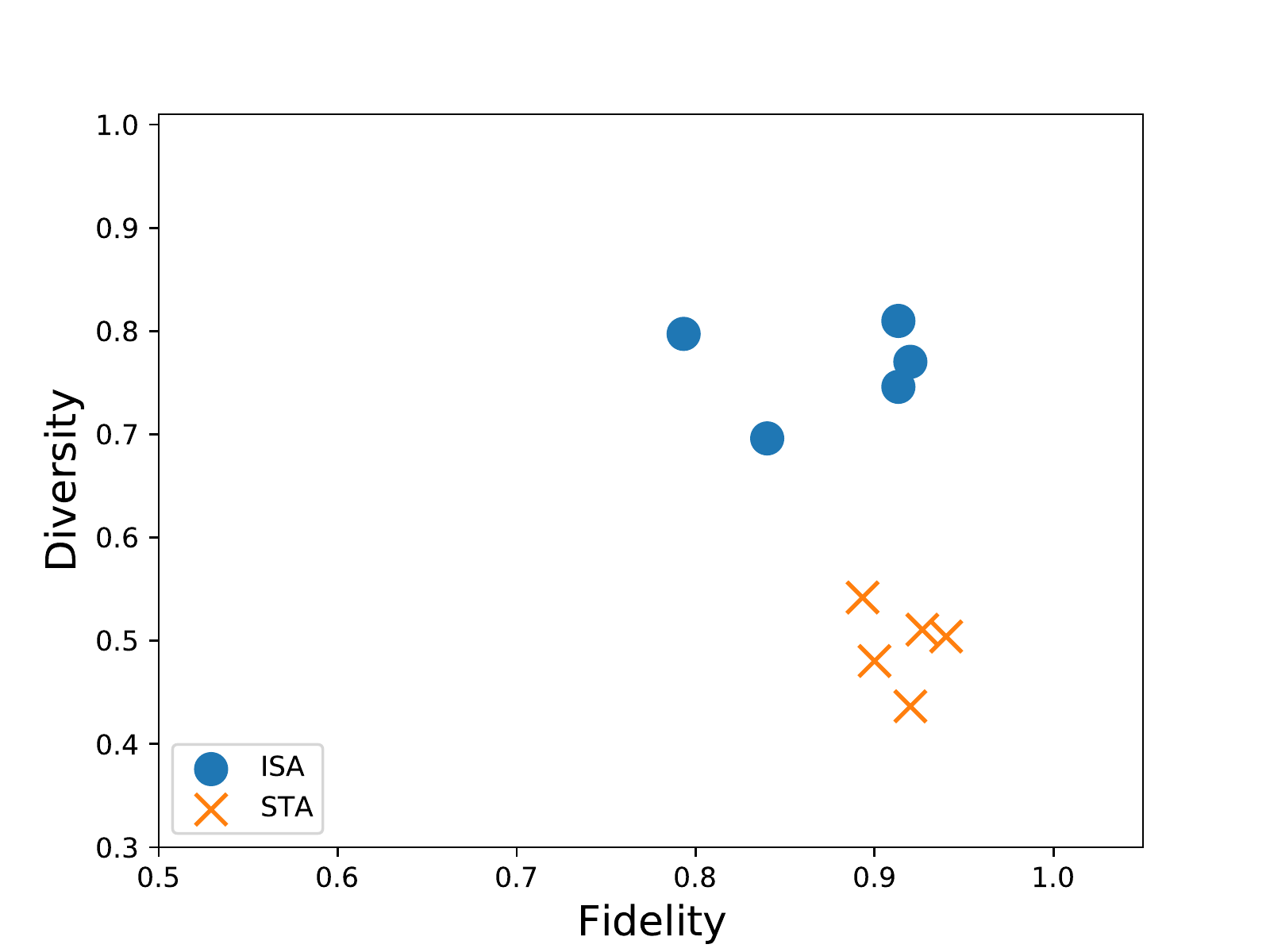}}
\subfloat[\textbf{AgNews}]{\label{fig:sta_isa_agnews_df}
    \includegraphics[scale=0.45]{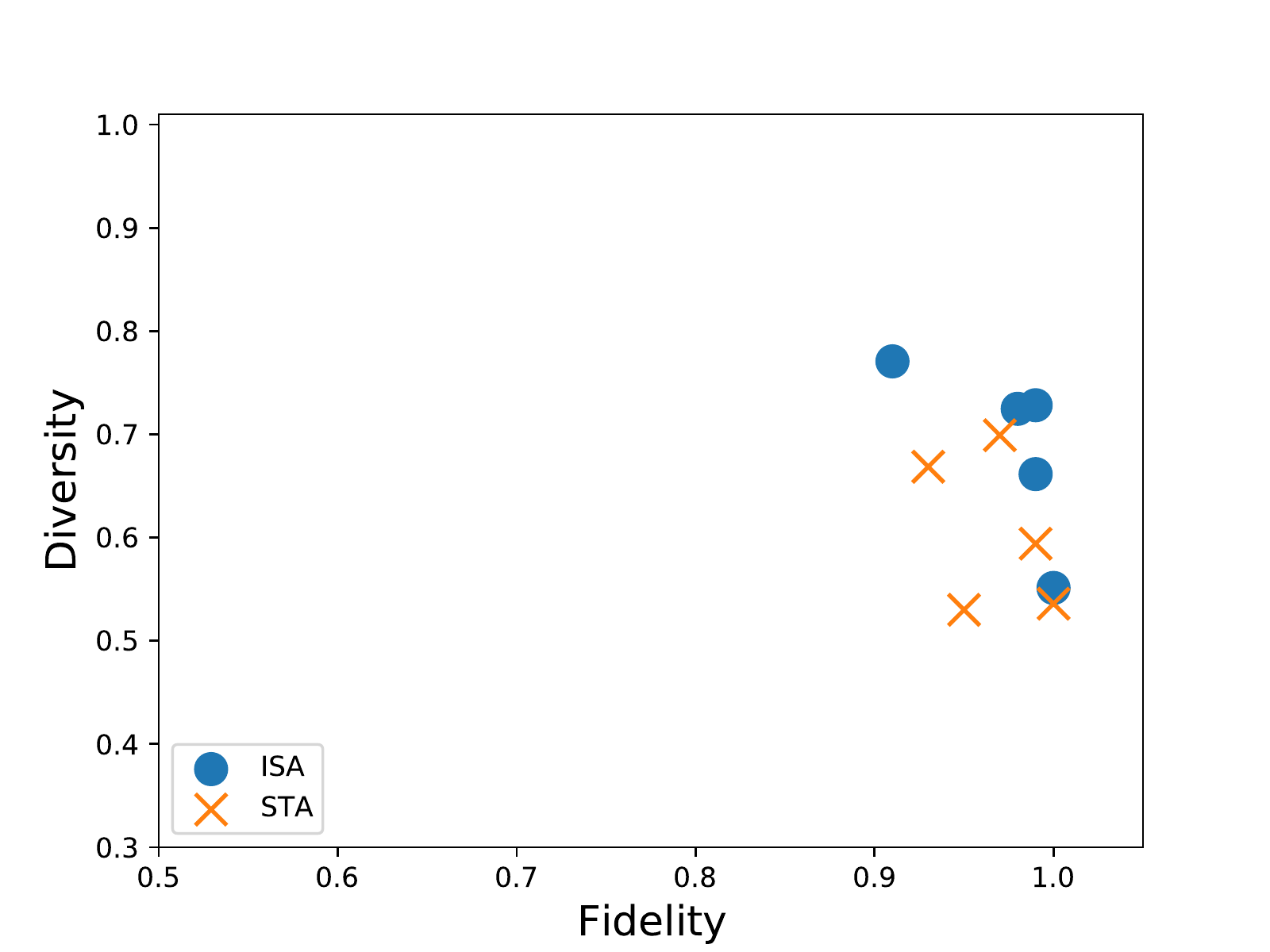}}
    
\subfloat[\textbf{TweetSentiment}]{\label{fig:sta_isa_tweet_df}
\includegraphics[scale=0.45]{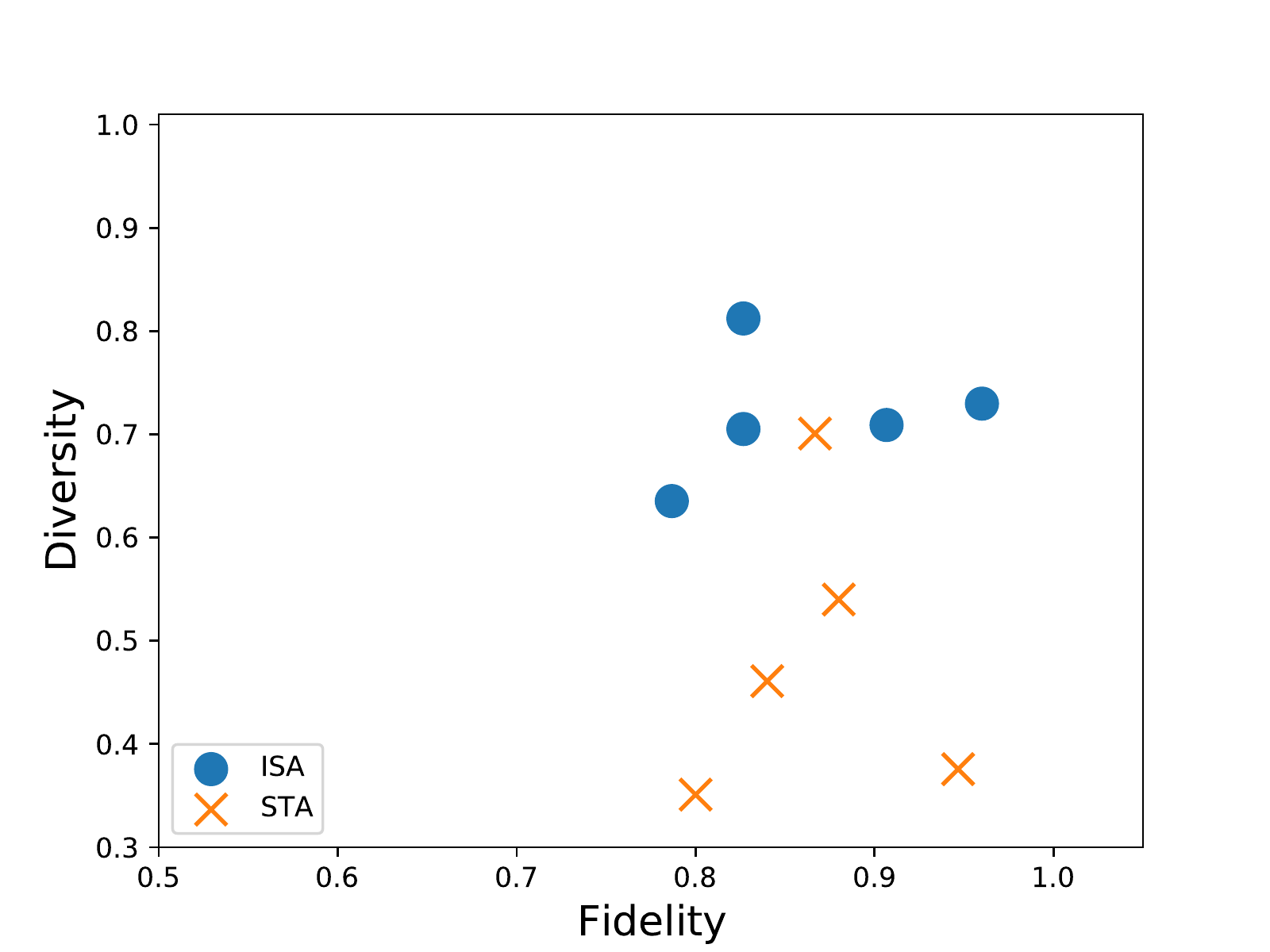}}
\subfloat[\textbf{TREC}]{\label{fig:sta_isa_trec_df}
    \includegraphics[scale=0.45]{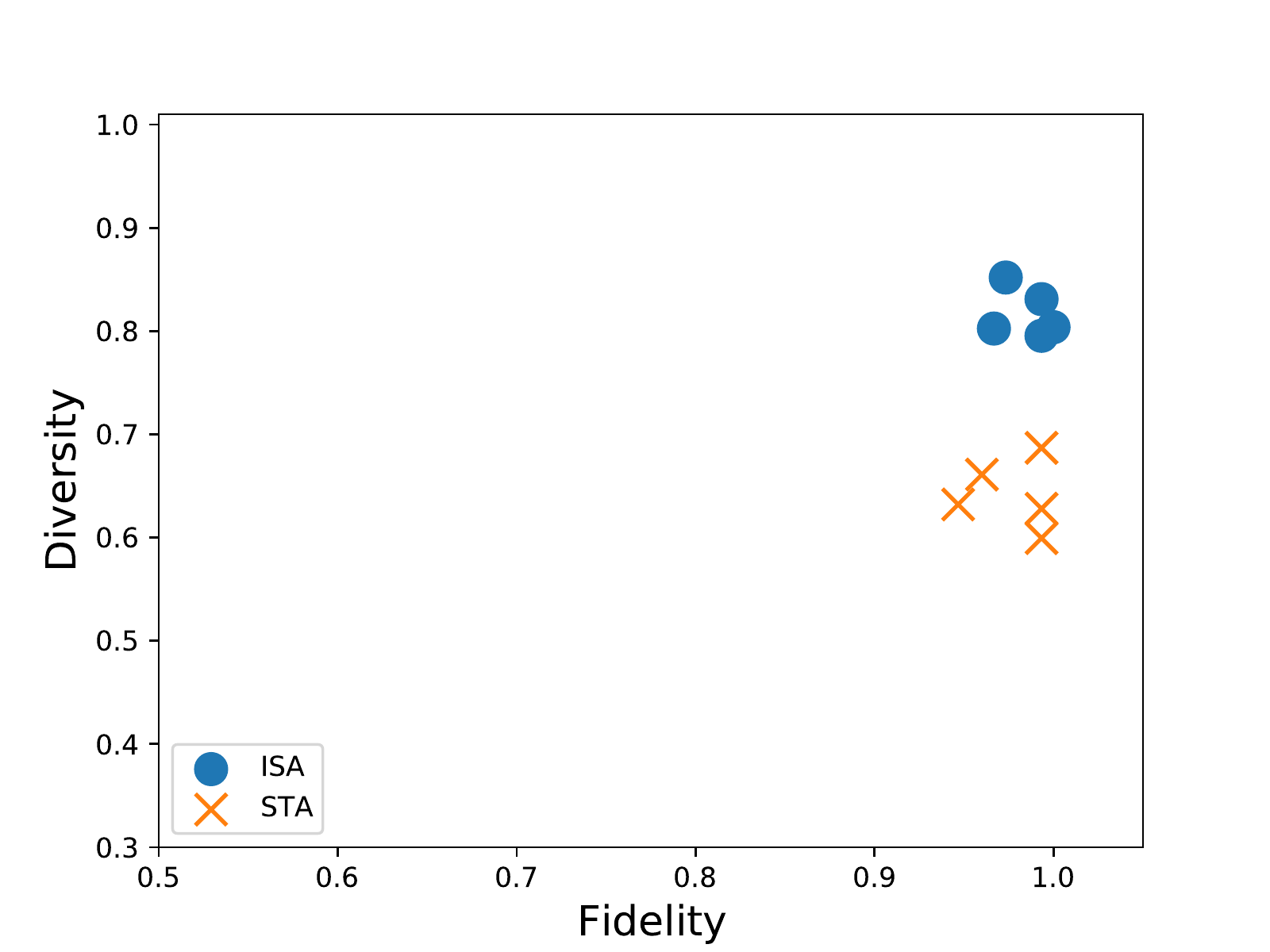}}
        \caption{Diversity versus semantic fidelity based on 5 runs}
        \label{fig:sta-isa-general-df}
\end{figure*}

\begin{figure*}[h!]
\centering
\subfloat[\textbf{Emotion}]{\label{fig:isa_iterno_dup_emotion_df_avg}
    \includegraphics[scale=0.45]{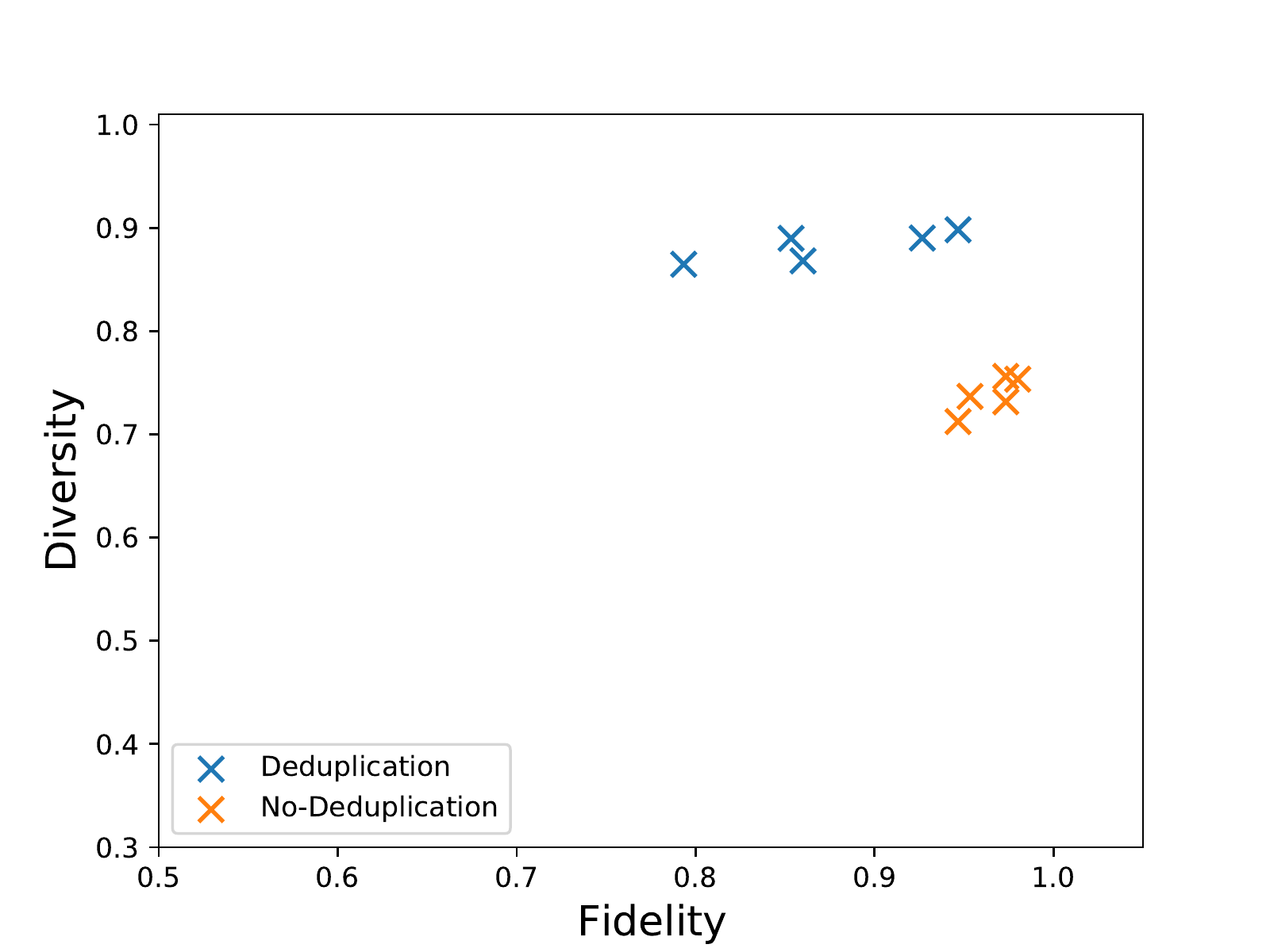}}
\subfloat[\textbf{AgNews}]{\label{fig:isa_iterno_dup_agnews_df_avg}
    \includegraphics[scale=0.45]{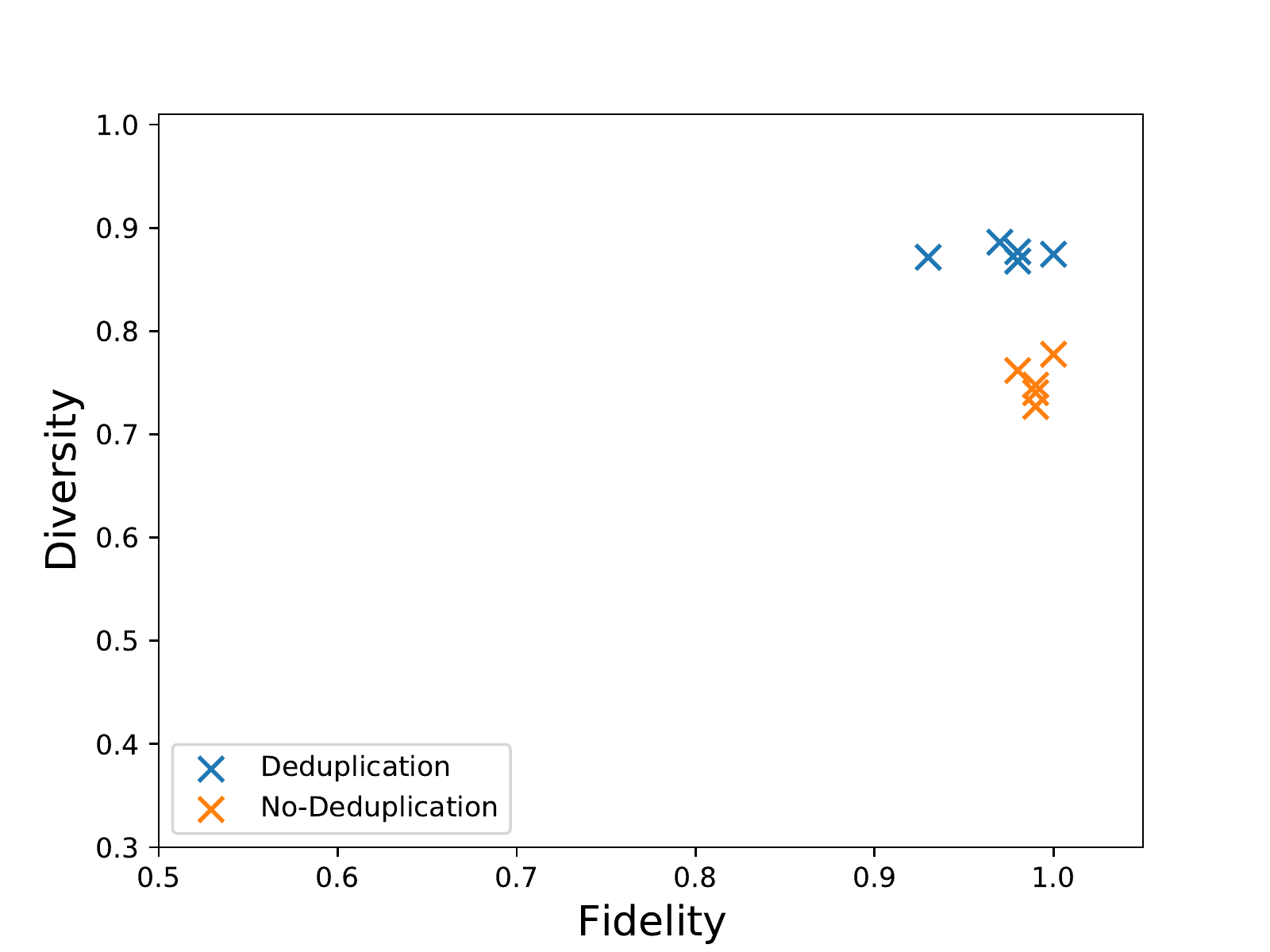}}
    
\subfloat[\textbf{TweetSentiment}]{\label{fig:isa_iterno_dup_tweet_df_avg}
\includegraphics[scale=0.45]{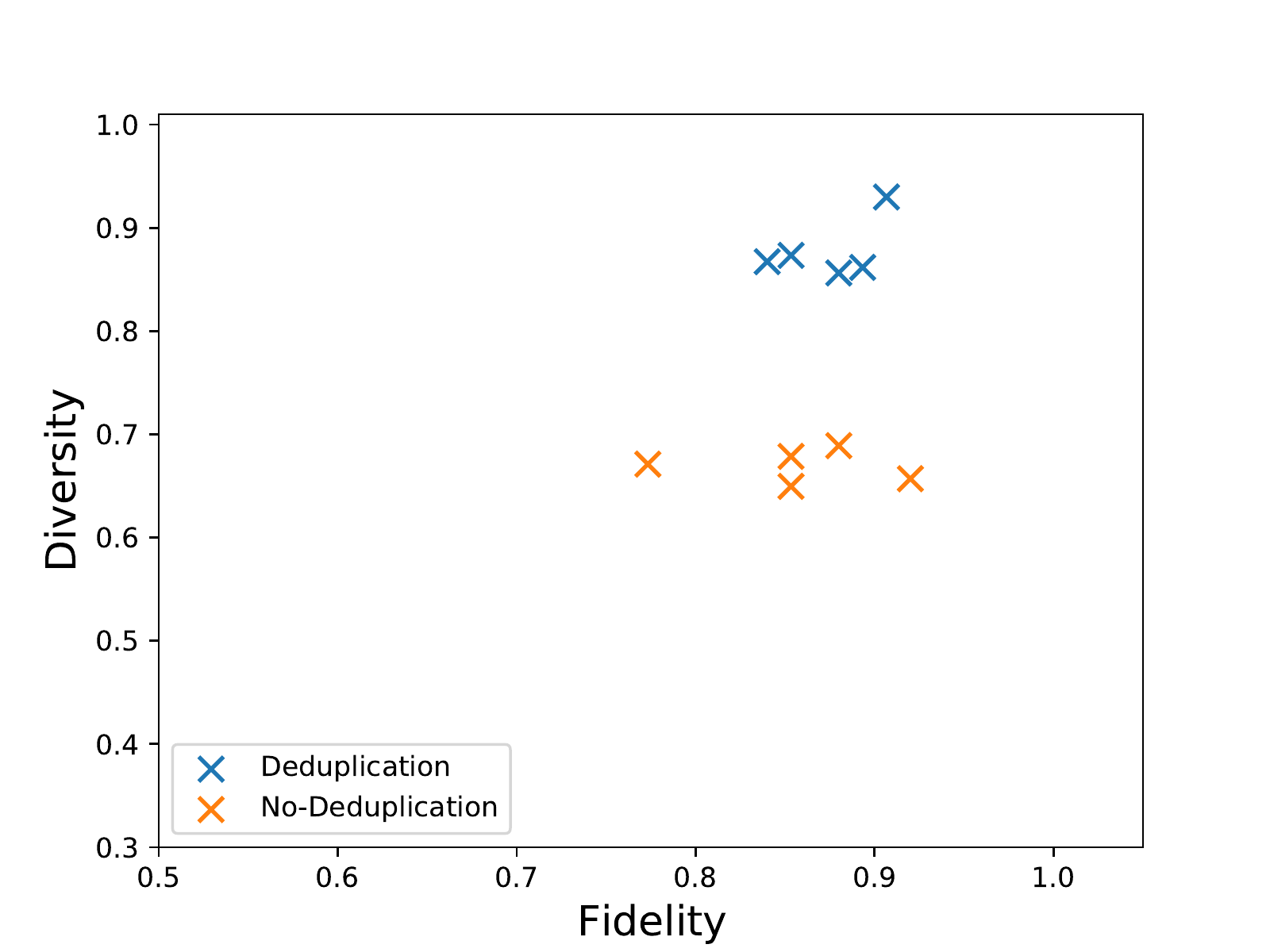}}
\subfloat[\textbf{TREC}]{\label{fig:isa_iterno_dup_trec_df_avg}
    \includegraphics[scale=0.45]{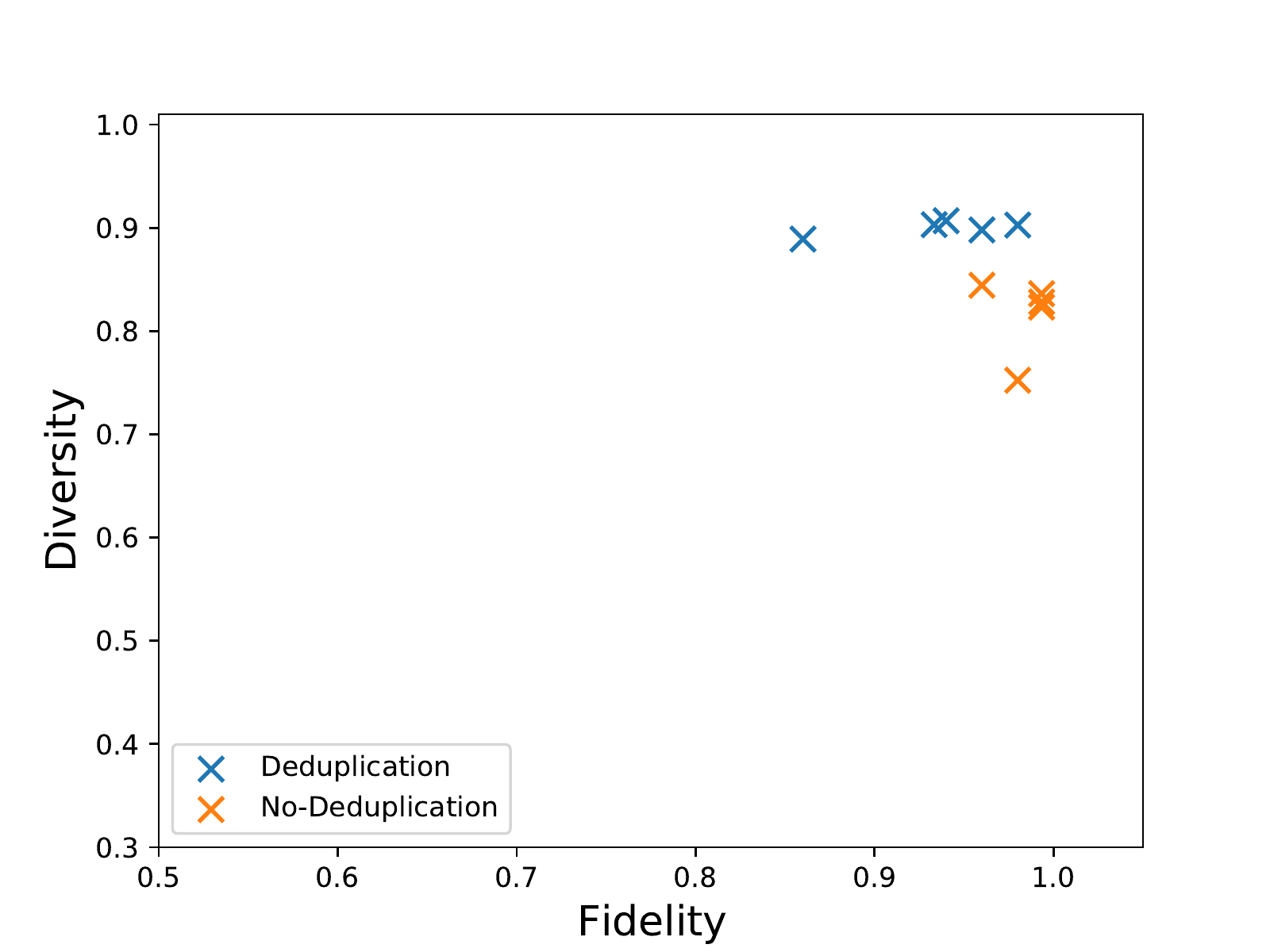}}

                \caption{Deduplication versus No-Deduplication for ISA based on 5 runs}
        \label{fig:sta-isa-general-dedup}
\end{figure*}

Regarding the study of the quality of generated texts, the experimental methodology remains the same as for \textbf{HumAID}. In measuring semantic fidelity, the accuracy of fully-trained BERT base on the test sets of \textbf{AgNews} and \textbf{TweetSentiment} is 94.78 and 74.9 respectively. Figure~\ref{fig:isa-sta-study-humaid-generated-texts-df} shows the results of diversity versus fidelity of texts generated by ISA and STA on datasets beyond crisis. It can be seen that ISA overall achieves better scores in both diversity and fidelity than STA. Figure~\ref{fig:isa-sta-study-humaid-generated-texts-dedup} depicts the effect of de-duplication on ISA. It shows that ISA with de-duplication achieves better diversity scores than without de-duplication. This reflects the correlation between the quality of generated texts measured by diversity and fidelity and the downstream performance: the better the quality of generated texts is, the better performance it brings to the downstream model performance.




\subsection{Summary}

In this section, ISA is introduced for few-shot text classification. It is built upon STA,  adopting an iterative self-controlled method with de-duplication to generate higher-quality samples for downstream model training. The results show that it outperforms STA as well as several state-of-the-art prompt-based few-shot baselines in the crisis message categorisation domain as well as for other domains such as sentiment and news classification. The investigation of the quality of texts generated by ISA reveals that ISA is superior in generating texts with good lexical diversity and semantic fidelity compared to the other approaches. The correlation between the quality of generated texts and downstream performance indicates that, for a generation-based augmentation approach for few-shot text classification it is important to ensure good lexical diversity and semantic fidelity for the generated texts. Hence, the future work can aim to further optimise the approach by controlling the quality of generated texts both in terms of lexical diversity and semantic fidelity. 

\section{Conclusions}

Various approaches have been proposed for few-shot text classifications within the NLP community and these can be directly adapted to the crisis message categorisation domain~\cite{zhang2022differentiable,schick2021exploiting,gao2021making}. This chapter introduced two novel augmentation approaches for crisis few-shot classification. STA is a self-controlled augmentation method that uses seq2seq pre-trained language models to generate new crisis messages based on only a small quantity of initial seed messages. It is motivated by and developed with the objective of optimising the quality of generated messages, i.e., the lexical diversity and semantic fidelity (label alignment). Eperiments comparing STA to existing augmentation baselines showed that STA is capable of generating new crisis messages with superior diversity and fidelity compared to the baselines. 

Following this, ISA was proposed, which introduces an iterative mechanism and a de-duplication mechanism to STA, aiming to further improve the quality of generated messages. The results indicate that ISA outperforms STA in terms of both diversity and fidelity, leading to improved downstream classification performance. STA was also shown to achieve comparable performance with the best prompt-based method, while ISA demonstrated the best performance among all the approaches evaluated for crisis message categorisation in few-shot settings. Not only were STA and ISA empirically effective in few-shot crisis message categorisation, further experiments also showed that they had strong generalisation capability to other text classification domains such as emotion or topic classifications. 

For crisis message categorisation, STA and ISA use a small quantity of annotated samples (performing well with as few as 5 labelled examples per class) and greater quantities of unlabelled data from the target crisis to generate more data for downstream model training. Although the few-shot seed data is small and can be annotated within a short time in real-world crisis response, it becomes time expensive when the number of pre-defined crisis-related aid classes become large. This is because STA and ISA requires a few annotated messages for every class in order to ensure that new messages can be generated for every class. Additionally, crisis few-short learning assumes equal class distribution, i.e, $k$ examples per class, which usually does not reflect the real-world distribution.

This raises the question of whether it is possible to completely eliminate the need for annotated target data. In the next chapter, the research will introduce a study on crisis zero-shot learning, which aims to categorise crisis messages without any annotated data. To address this challenge, the study will propose a novel approach called P-ZSC that uses pseudo-labelled data for zero-shot crisis message categorisation.
\chapter{Using Pseudo-labelled Data For Crisis Zero-shot Learning}
\label{ch:pzsc}

The previous chapter explored the use of two augmentation approaches for categorising crisis messages that require only a few annotated messages per aid type from emerging events. The annotation is relatively easy and efficient when there are only a few aid types, but becomes expensive when many aid types are of interest. Additionally, the methods assume that crisis messages are evenly distributed across aid types, which is rarely the case in real-world scenarios where the class distribution is often influenced by the characteristics of the crisis. For example, during a terrorist attack, requests for blood donations may be more common than requests for shelter or food, whereas the opposite may be true during a storm or flood. These considerations have motivated the current research to focus on crisis message categorisation using zero-shot learning, which involves using only easy-to-obtain unlabelled target data from emerging events.

This chapter first introduces the method for crisis zero-shot learning called P-ZSC (Pseudo-labelled Zero-Shot Categorization) (Section~\ref{sec:pzsc-intro}). The approach involves using pseudo-labelled data to train a model followed by a refinement process for crisis message categorisation (Section~\ref{sec:pzsc-method}). To obtain the pseudo-labelled data, each sample of the unlabelled data is matched with the descriptions of aid types represented by their surface names such as ``search and rescue''. After this matching, a label expansion component is then applied to generate a label vocabulary for each class in order to enhance the representation of the aid classes (Section~\ref{subsec:label-expansion-pzsc}). Matching is then conducted between the unlabelled samples and the label vocabularies, rather than the raw label surface names. Samples that achieve a high matching score with a label are assigned that label (Section~\ref{subsec:pla-pzsc}). The pseudo-labelled set is then used to pre-train a downstream classifier, which is finally self-trained on the unlabelled data (Section~\ref{subsec:pre-train-self-train-pzsc}). Finally, the results of P-ZSC compared to baselines are reported and discussed in experiments (Section~\ref{sec:pzsc-exps}).

The work presented in this chapter has previously been the subject of a peer-reviewed publication~\cite{wang2022using}.

\section{Introduction}
 \label{sec:pzsc-intro}

This section introduces P-ZSC, a weakly-supervised method for performing zero-shot learning during a crisis. Building upon previous work (Section~\ref{subsec: weakly-sup-czs}), P-ZSC uses a simple and effective algorithm to match unlabelled samples with classes to obtain pseudo-labelled data that can then be used to train the classification model. To avoid exact matching, P-ZSC obtains related phrases from a domain-specific unlabelled corpus to create a vocabulary for each class based on their embeddings. A sample is assigned with a particular class when it overlaps significantly with the vocabulary of that class. P-ZSC also uses self-training, a method that has been demonstrated to be effective~\cite{meng-etal-2020-text} to refine the categorisation model on the domain-specific unlabelled data after it has been trained on the pseudo-labelled data. The details of P-ZSC are provided below.

\section{Method}
 \label{sec:pzsc-method}

 Formally, the problem that the system aims to solve is defined as follows: for a classification task $\mathcal{T}$ (which can be either a single-label or multi-label task), given $n$ label names\footnote{In the context of a crisis message categorisation tasks, the label names are aid types for emergency response. In this description, more general terms are used to indicate its generalisabilty to other text classification tasks.} $\mathcal{Y}:\{y_1,y_2,\ldots,y_n\}$ and an unlabelled corpus $\mathcal{D}:\{d_1,d_2,\ldots,d_m\}$ containing $m$ examples from this task domain, the objective is to train a model $f$ that can assign one or more labels (depending on whether the task is single-label or multi-label classification) from $\mathcal{Y}$ to an example $x$ based on its probability estimation over $\mathcal{Y}$, namely, $f(x): p(y_1|x), p(y_2|x),..., p(y_n|x)$.

\begin{figure*}[!h]
    \centering
    \includegraphics[scale=
    0.8]{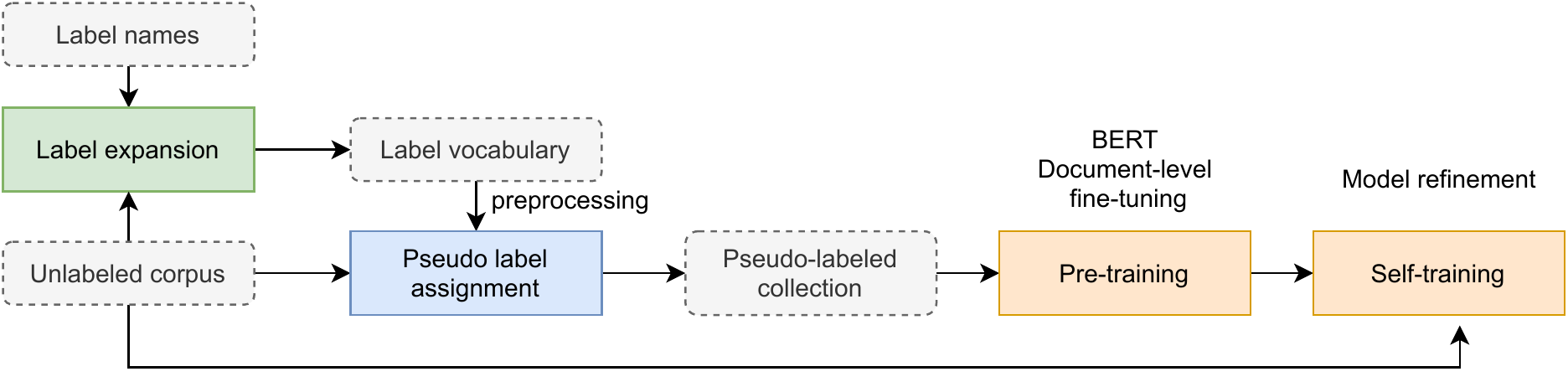}
    \caption{The architecture of P-ZSC}
    \label{fig:arch}
\end{figure*}

As illustrated in Figure~\ref{fig:arch}, the system has three stages: label expansion produces a label vocabulary to represent raw labels, pseudo label assignment matches the unlabelled corpus with label vocabularies to obtain a pseudo-labelled dataset, pre-training and self-training train a classification model on the pseudo-labelled dataset and then refines the model on the unlabelled corpus. These stages are described in detail in the following sections.
\subsection{Label expansion}
\label{subsec:label-expansion-pzsc}


In a classification task such as crisis message categorisation, the label names tend to consist of only one or two words such as ``missing or found'', ``goods services'' and ``donations'', which are insufficient to convey their potentially broad meaning. To enrich a label's meaning, a simple label expansion (LE) algorithm is proposed to find the most semantically-similar words or phrases to the label from the unlabelled in-domain corpus $\mathcal{D}$.

Given $\mathcal{Y}$ and $\mathcal{D}$, first all examples from $\mathcal{D}$ are combined and split into n-gram phrases consisting of all 1-grams, 2-grams and 3-grams that are found in any examples within $\mathcal{D}$ (including overlaps): $\mathcal{G}:\{g_1,g_2,g_3,\ldots,g_L\}$. Then, the similarity between $y_i \in \mathcal{Y}$ and $g_i \in \mathcal{G}$ is calculated via a sentence embedding model. Given a sentence embedding model $E$,\footnote{As an example of sentence embeddings models, the widely-applied \texttt{deepset/sentence\_bert} from the Huggingface model hub~\cite{wolf2019huggingface} is used in this work.} the matching score $\hat{s}_{i,j}$ between a label $y_{i}$ and an n-gram $g_{j}$ is calculated as the cosine similarity between the embeddings of the label and the n-gram respectively:

\begin{equation}
    \hat{s}_{i,j} = \text{cosine}(E_{avg}(y_{i}), E_{avg}(g_{j}))
\end{equation}

For any $y_{i}$ or $g_{j}$ with more than one token, $E_{avg}$ takes the average pooling as the output embedding. After this, each label $y_i \in \mathcal{Y}$ has a list of n-grams (i.e. a subset of $\mathcal{G}$) that are ranked by the similarity scores. The vocabulary for each label $y_i$ is denoted by $\hat{\mathcal{V}}_i: \{(\hat{v}_{i,k},\hat{s}_{i,k})\}|_{k=1}^{\hat{K}}$ where $\hat{v}_{i,k} \in \mathcal{G}$ represents the $k$th expanded n-gram in the vocabulary of label $y_i$ and $\hat{s}_{i,k}$ is the corresponding matching score.

\textbf{Label vocabulary pre-processing}: To maintain the quality of the label vocabulary, a further pre-processing step is taken to optimise the original vocabulary $\hat{\mathcal{V}}_i$ and a label's pre-processed vocabulary is denoted as: $\mathcal{V}_i: \{(v_{i,k},s_{i,k})\}|_{k=1}^K$. At this step, only those n-grams in $\hat{\mathcal{V}}_i$ where $\hat{s}_{i,k} \geq 0.7$ are selected, maintaining a minimum of 2 n-grams and a maximum of 100 n-grams per label\footnote{The parameters are fixed for all tasks to ensure actual zero-shot learning.}. Then a discounting is applied on the n-grams that co-occur across different labels, which is calculated by

\begin{equation}
\label{eq:discounting}
    s_{i,k} = \hat{s}_{i,k}*log_{e} \left( \frac{n}{LF(v_{i,k})} \right)
\end{equation}

where $n$ is the number of labels and $LF(v_{i,k})$ is the frequency of n-gram $v_{i,k}$ across the vocabularies of all labels. To know the effect of the label expansion approach on raw surface names, Table~\ref{tab:crisis-label-vocab} shows the label vocabularies for the labels from two crisis classification datasets \textbf{HumAID} and \textbf{Situation}. After this process, the labels are represented by their corresponding vocabularies that convey a broader semantic meaning of their raw surface names. For example, instead of just ``infrastructure and utility damage'', it is enriched with new terms such as ``property damage'', ``power lines'', etc. Notably, the top-ranked expanded phrases mostly consist of 1 and 2-grams, as matching high semantic scores with 3-grams can be challenging due to the complexity of semantic matching. Nevertheless, this work still leverages 3-grams for label expansion, as they may include crucial phrases such as ``people evaluated safet'', which can help expand the``evaluation'' label well (as shown in Table~\ref{tab:crisis-label-vocab}).

\begin{table*}[h!]
\centering
    \begin{subtable}[]{\textwidth}
        \centering
          \scriptsize
\begin{tabular}{lp{8cm}}
\toprule
Label                                  & Vocab                                                                                                                                                           \\
\midrule
rescue volunteering or donation effort & donation help, donate help, help donate, donating help, volunteers   helping, volunteering, volunteers help, rescue effort, relief donations,   rescue people   \\
sympathy and support                   & sympathy, condolences, compassion, support please, condolences families,   deepest condolences, support relief, please support, provide relief, kindness        \\
requests or urgent needs               & urgent need, urgent, needs, need, needing, require, urgently, demand,   needed, really need                                                                     \\
infrastructure and utility damage      & infrastructure damage, infrastructure, power lines, areas impacted,   damage caused, damage, property damage, areas affected, power outages, damage   amp       \\
injured or dead people                 & dead people, dead injured, people died, people dead, killed injured,   people dying, people killed, died injured, killed people, deaths                         \\
caution and advice                     & advice, careful, warns, warned, safety tips, advised, heed warnings, pray   safety, advisory, take care                                                         \\
displaced people and evacuations       & displaced people, people displaced, refugees, families displaced,   evacuations, evacuation orders, evacuation zones, displaced, evacuation, need   evacuate    \\
missing or found people                & people missing, missing people, people still missing, reported missing,   people lost, number missing, people unaccounted, lost lives, missing, still   missing \\
\bottomrule
\end{tabular}
\caption{\centering Label vocabularies for HumAID dataset.}
\label{lv:topic}
    \end{subtable}
    \vspace{1.5cm}
    \begin{subtable}[]{\textwidth}
        \centering
        \scriptsize
\begin{tabular}{lp{11cm}}
\toprule
Label          & Vocab                                                                                                                                             \\
\midrule
shelter        & shelter, temporary shelter, shelters, shelter food, refuge, food shelter,   protection, temporary shelters, makeshift shelters, emergency shelter \\
search         & search, searching, find, search operations, finding, seek, seeking                                                                                \\
water          & water, water borne, waters, water system, water drinking, water sources,   waterborne, water resources, water supplies, water supply              \\
utilities      & electricity supply, power lines, power supply, electricity water,   electricity, supply systems                                                   \\
terrorism      & terrorism, anti terrorism, terrorist groups, terrorist, terrorists,   terrorist attack                                                            \\
evacuation     & evacuation, evacuate, evacuated, people evacuated, people evacuated   safety, evacuated safety, evacuating                                        \\
regime change  & change, changed                                                                      \\
food           & food, including food, food clothing, food items, food clothes, food work,   foods, food supplies, food program, food programme                    \\
medical        & medical, medical services, medical team, medicine, doctors, medical care,   healthcare, medical attention, doctor, hospital                       \\
infrastructure & infrastructure, infrastructures, infrastructure including, structures                                                                             \\
crime violence & crime, crimes, violence   \\
\bottomrule
\end{tabular}
\caption{\centering Label vocabularies for Situation dataset.}
\label{lv:situation}
     \end{subtable}
     \caption{Top-ranked expanded phrases for crisis-related labels from \textbf{HumAID} and \textbf{Situation}}
     \label{tab:crisis-label-vocab}
\end{table*}

\subsection{Pseudo label assignment}
\label{subsec:pla-pzsc}


After expansion, a labelled collection (pseudo-labelled data) is next constructed for model training. Due to the lack of annotated data, the alternative is to construct the collection via the process of pseudo label assignment (PLA). To achieve this, a simple approach for PLA is adopted, which is described as follows:

A document $d_j \in \mathcal{D}$ is matched with a label's vocabulary $\mathcal{V}_i: \{(v_{i,k},s_{i,k})\}|_{k=1}^K$ (from the previous section) through a cumulative scoring mechanism. To assign a label $y_i$ to a document $d_j$, a  matching score between them is first calculated by

\begin{equation}
     s_{j,i}^* = \sum_{k=0}^{K}{s_{i,k}}\mathbb{I}(v_{i,k} \in \mathcal{G}_j)
\end{equation}

where $\mathbb{I}(\cdot)$ is the indicator function (evaluating to 1 if $v_{i,k} \in \mathcal{G}_j$ or 0 otherwise) and $\mathcal{G}_j$ is the set of n-grams that are contained in document $d_j$. For consistency with what was used in label expansion, here the n-grams also range from $n=1$ to $n=3$. Now $s_j^*:\{s_{j,i}^*\}|_{i=0}^n$ refers to the matching score of $d_j$ with every label from $\mathcal{Y}$. To decide if $d_j$ is assigned one or more labels, a threshold $\epsilon$ is defined. For single-label tasks, if the maximum value of $s_j^*$ at index $i$ is greater than $\epsilon$, then $y_i$ is assigned to $d_j$. For multi-label tasks, if the value of $s_j^*$ at any index is larger than $\epsilon$, then the label at that index is assigned to $d_j$. Thus only the examples achieving high matching scores (high-confidence) with the labels are likely to be pseudo-labelled. Examples for which no label exceeds the threshold are left unlabelled and are not used used in the next stage for pre-training, but can be used for self-training.

At this point, a pseudo-labelled collection is obtained, denoted as $\hat{\mathcal{D}}: \{(x_i,\mathcal{Y}_i)\}|_{i=1}^{N}$ where $N$ is the number of pseudo-labelled examples, and $\mathcal{Y}_i$ refers to the pseudo labels assigned to $x_i$ and it is a subset of $\mathcal{Y}$ consisting of a label for single-label tasks and multiple labels for multi-label tasks. To ensure the quality of $\hat{\mathcal{D}}$, the threshold $\epsilon$ should be chosen carefully. It will result in poor-quality pseudo-labelled data being generated if it is too low, but will result in zero examples for some labels if it is too high. Since the matching score $s_{i,k}$ is normalised by Equation~\ref{eq:discounting}, $\epsilon$ is set to be $log_{e}{n}$ where $n$ is the number of labels. However, this value can lead to zero pseudo-labelled examples for insufficiently-expanded labels (e.g., their vocabularies contain few phrases: particularly common in class-imbalanced datasets). Hence, $\epsilon$ is reduced by half when the label with fewest expanded phrases has fewer than $10$\footnote{This setup is fixed for all tasks and not tuned to different tasks to ensure actual zero-shot learning.}.

\subsection{Pre-training and self-training}
\label{subsec:pre-train-self-train-pzsc}

With $\hat{\mathcal{D}}$ as the supervision data, the next step fits the model $f$ to the target classification task by learning from the pseudo-labelled data. Instead of training from scratch, the pre-trained \texttt{bert-base-uncased}~\cite{BERT2018} is used as the base model ($f_{\theta}^*$) and this can easily be replaced by other pre-trained language models~\cite{wolf2019huggingface}. To be specific, for the task $\mathcal{T}$, a classification head is added to the base model. The classification head takes the \texttt{[CLS]} output (denoted as $\boldsymbol{h_{\text{[CLS]}}}$) of the base model as input and outputs the probability distribution over all classes:

\begin{equation}
\begin{aligned}
\boldsymbol{h} = f_{\boldsymbol{\theta}}^*(x) \\ p(\mathcal{Y}\mid x) = \sigma(\boldsymbol{W} \boldsymbol{h}_{\text{[CLS]}}+\boldsymbol{b})
\end{aligned}
\end{equation}

where $\boldsymbol{W} \in \mathbb{R}^{h \times n}$ and $\boldsymbol{b} \in \mathbb{R}^{n}$ are the task-specific trainable parameters and bias respectively and $\sigma$ is the activation function (softmax if $\mathcal{T}$ is a single-label task, or sigmoid for a multi-label task). In model training, the base model parameters $\boldsymbol{\theta}$ are optimised along with $\boldsymbol{W}$ and $\boldsymbol{b}$ with respect to the following cross entropy loss (on the pseudo-labelled data):

\begin{equation}
   \mathcal{L}_{pt}=-\sum_{i=0}^n y_{i}\log p(y_{i} \mid x)
\end{equation} 

Since $\hat{\mathcal{D}}$ is a subset of $\mathcal{D}$, there are many unlabelled examples not seen in the model's pre-training. As indicated in LOTClass~\cite{meng-etal-2020-text}, the unlabelled examples can be leveraged to refine the model for better generalisation via self-training. Hence, the unlabelled corpus $\mathcal{D}$ is then used for model self-training. This is done by splitting it into equal-sized portions (assuming each portion has $M$ examples) and then each of the examples in a portion is predicted by the model in an iteration with the predictions denoted as the target distribution $Q$. In each iteration, the model is trained on batches of the portion with the current distribution as $P$. The model is updated with respect to the following Kullback–Leibler (KL) divergence loss function~\cite{joyce2011kullback}:

\begin{equation}
    \mathcal{L}_{st}=\mathrm{KL}(Q \| P)=\sum_{i=1}^{M} \sum_{j=1}^{n} q_{i,j} \log \frac{q_{i,j}}{p_{i,j}}
\end{equation}

where $q_{i,j}$ and $p_{i,j}$ refer to the predicted probability of an sample $d_i$ to an label $y_j$ of $Q$ and $P$ respectively. In deriving the target distribution $Q$, it can be applied with either soft labelling~\cite{xie2016unsupervised} or hard labelling~\cite{lee2013pseudo}. Hard labeling involves using high-confidence predictions as the ground truth labels, which can be prone to error propagation. Soft labeling, on the other hand, calculates the target distribution for each instance, and does not require the use of preset confidence thresholds. As a result, soft labelling overall tends to bring better results~\cite{meng-etal-2020-text}, therefore the soft labelling strategy is used to derive $Q$ in this work, formulated as follows:

\begin{equation}
\begin{aligned}
q_{i,j}=\frac{p_{i,j}^{2} / p_{j}^*}{\sum_{j^{\prime}}\left(p_{i j^{\prime}}^{2} / p_{j^{\prime}}^*\right)}, p_{j}^*=\sum_{i} p_{i,j} \\
p_{i,j} = p(y_j \mid x_i) = \sigma(\boldsymbol{W}(f_{\boldsymbol{\theta}}^*(x_i))_{\text{[CLS]}}+\boldsymbol{b})
\end{aligned}
\end{equation}

This strategy derives $Q$ by squaring and normalising the current predictions $P$, which helps boost high-confidence predictions while reducing low-confidence predictions. 

\section{Experiments}
 \label{sec:pzsc-exps}
 
Having described the system components, the following sections describe extensive experiments to demonstrate the effectiveness of the proposed approach.

\subsection{Datasets}


In testing P-ZSC for crisis message categorisation, \textbf{Situation}~\cite{mayhew2018university, yin2019benchmarking} is selected in addition to \textbf{HumAID}~\cite{alam2021humaid}. In contrast with \textbf{HumAID}, \textbf{Situation} is a multi-label dataset where a crisis tweet is assigned to one or more aid types such as ``shelter'', ``search'', etc. Details of the datasets can be found in Section~\ref{sec:rds-ems-infotypes}. In evaluation, the \textit{label-wise weighted F1} metric is used for \textbf{Situation} considering it is a multi-label class-imbalanced dataset while accuracy is used for \textbf{HumAID}, which are also aligned with~\cite{yin2019benchmarking} and~\cite{alam2021humaid} respectively. To use them in zero-shot learning settings, the training and development sets of each dataset excluding labels are combined and used as an unlabeled corpus for model training, while the test set is used for testing.


\subsection{Experimental details}


In the experiments, to simulate a simple zero-shot setting, the training set of each dataset (pre-split training sets that come with the datasets) is used as the unlabelled corpus $\mathcal{D}$ and the surface names (e.g. ``missing or found'', ``shelter'', ``search'' etc) as the label set $\mathcal{Y}$. For model pre-training on the pseudo-labelled data, the \texttt{BERT-base-uncased} pre-trained checkpoint is fine-tuned on a batch size of 16 using Adam~\cite{kingma2014adam} as the optimiser with a linear warm-up scheduler for increasing the learning rate from $0$ to $5e-5$ at the first $0.1$ of total training steps and then decreasing to 0 for the remaining steps. For self-training, the hyper-parameters remain the same as in~\cite{meng-etal-2020-text}, with batch size $128$ and update interval $50$, which results in the number of training examples in each iteration (namely, $M$) being $50 \times 128$. 


\subsection{Baselines}

In these experiments, related existing methods (Section~\ref{sec:czs}) are selected as baselines to be compared with P-ZSC. The baselines include indirectly-supervised methods \textbf{Label similarity}~\cite{veeranna2016using}, \textbf{Entail-single}~\cite{yin2019benchmarking}, \textbf{Entail-ensemble}~\cite{yin2019benchmarking}, \textbf{Entail-Distil}~\cite{joe2021} and weakly-supervised methods \textbf{ConWea}~\cite{mekala2020contextualized}, \textbf{X-Class}~\cite{wang2021x}, \textbf{WeSTClass}~\cite{meng2018weakly} and \textbf{LOTClass}~\cite{meng-etal-2020-text}. In addition, the fully \textbf{Supervised BERT} is added to the list for comparison. The experimental setup for them are described as follows.

\begin{itemize}
    \item \textbf{Label similarity} uses pre-trained word embeddings to compute the cosine similarity between the class label and every 1-gram to 3-gram of the example. In the experiments, BERT base is used as the pre-trained embedding model. For single-label tasks, the label with the highest similarity score is chosen. For multi-label tasks, any label with a similarity score greater than $0.5$ is chosen.
    \item \textbf{Entail-single} and \textbf{Entail-ensemble} correspond to the best \textit{label-fully-unseen} ZSL single and ensemble results in ~\cite{yin2019benchmarking}. As \textbf{HumAID} and \textbf{Situation} were not used in that paper, their methodology is followed to create a similar setup by fine-tuning three variants of BERT on three inference datasets (RTE/MNLI/FEVER) and then being tested on the datasets. The best of these in each category is reported as ``entail-single'' and the ensemble of the three variants is reported as ``entail-ensemble''.
    \item \textbf{Entail-Distil} attempts to overcome the inference inefficiency issue in the entailment matching-based methods that require all-versus-all pairwise matching. The training data is pseudo-labelled first by the matching model (\texttt{bert-base-uncased} fine-tuned on MNLI) and then the pseudo-labelled data is used for downstream model fine-tuning (\texttt{bert-base-uncased}).

    \item \textbf{ConWea} is a contextualised weakly-supervised approach for text classification, which uses a small quantity of human-provided seed words for label expansion. In the experiments, their released code\footnote{\url{https://github.com/dheeraj7596/ConWea}} is re-run on the selected datasets using at least 3 seed words. As a comparison, weak human supervision (few seed words) entails this approach unlike P-ZSC using label names only.

    \item \textbf{X-Class} uses label names only by building class-oriented document representations first and then using a Gaussian Mixture Model (GMM) to obtain the pseudo-labelled data. In the experiments, their released code\footnote{\url{https://github.com/ZihanWangKi/XClass}} is run on the selected datasets to obtain their corresponding pseudo-labelled for model fine-tuning (\texttt{bert-base-uncased}) and the performance is reported on the fine-tuned model.

    \item \textbf{WeSTClass} is configurable to accept up to three sources of supervision. In the experiments, their released code~\footnote{\url{https://github.com/yumeng5/WeSTClass}} is re-run on the selected datasets using label names as the only supervision resource so as to be consistent with P-ZSC.

    \item \textbf{LOTClass} is another weakly-supervised approach using only label names followed by a self-training component for text classification. In the experiments, their officially-released code~\footnote{\url{https://github.com/yumeng5/LOTClass}} is re-run for the selected datasets.

    \item \textbf{Supervised BERT (Sup. BERT)} is included so that a comparison with a fully-supervised approach can be done. This uses \texttt{bert-base-uncased}, fine-tuned on the training sets with labels, validated by the developments sets and tested on test sets. While a ZSL approach will be unlikely to match the performance of a fully-supervised approach, it is important to illustrate how large that performance gap actually is and to illustrate the degree to which P-ZSC contributes towards closing that gap.
\end{itemize}

\subsection{Results and discussion}

In this section, P-ZSC is compared with the baselines in a crisis zero-shot classification setting. To dissect each component of P-ZSC, an ablation study is conducted. Following this, an additional experiment is conducted to investigate the quality of the pseudo-labelled data that is generated by the system. Finally, its generalisation to other domains such as emotion classification is tested and verified.

\subsubsection{Comparison with baselines}

\begin{table*}[]
\centering
\renewcommand{\arraystretch}{1.2}
\begin{tabular}{lll}
\toprule
                      & HumAID     & Situation     \\
                      \midrule
\multicolumn{3}{l}{Indirectly supervised runs} \\
Label similarity~\cite{veeranna2016using}      & 57.89      & 40.75         \\
Entail-single~\cite{yin2019benchmarking}         & 40.96      & 37.20         \\
Entail-ensemble~\cite{yin2019benchmarking}       & 41.34      & 38.00         \\
Entail-Distil~\cite{joe2021}         & 45.81      & 38.85         \\
                      \midrule

\multicolumn{3}{l}{Weakly-supervised runs}        \\
ConWea~\cite{mekala2020contextualized}                & 46.35      & 25.91         \\
X-Class~\cite{wang2021x}               & 45.70      & 39.27         \\
WeSTClasss~\cite{meng2018weakly}            & 35.39      & 28.40         \\
LOTClass~\cite{meng-etal-2020-text}              & 15.63      & 5.85          \\

P-ZSC                 & \textbf{67.09}      & \textbf{55.02}         \\
\midrule
Sup. BERT             & 88.89      & 85.27        \\
\bottomrule
\end{tabular}
\caption{Performance of P-ZSC and baselines on test sets of crisis datasets}
\label{tab:pzsc-crisis-results}
\end{table*}

In testing the effectiveness of P-ZSC on the two selected crisis categorisation tasks, the baselines are divided into two categories: indirectly-supervised runs and weakly-supervised runs, as presented in Table~\ref{tab:pzsc-crisis-results}. P-ZSC falls into the category of weakly-supervised runs. The results show that P-ZSC achieves state-of-the-art $67.09$ accuracy and $55.02$ F1 score for \textbf{HumAID} and \textbf{Situation} respectively.

P-ZSC substantially outperforms the weakly-supervised methods including Label similarity, Entail-single, Entail-ensemble and Entail-Distil. Label similarity performs well among the baselines of this kind. Although it is a simple matching between the sentence embeddings of label names and document phrases, interestingly it outperforms the entailment methods and Entail-Distil for both datasets. This indicates that semantically matching a document's phrases with the label names using pre-trained word embeddings can help determine the document's class. It should also be noted that label similarity and the entailment runs are around $n$ (number of labels) times slower than Entail-Distil and the weakly-supervised runs. This indicates that P-ZSC as a weakly-supervised approach has advantages in terms of inference efficiency also. Compared to the best results of weakly-supervised runs ($46.35$ for \textbf{HumAID} by ConWea and $39.27$ for \textbf{Situation} by X-Class), P-ZSC shows performance improvements of $+20.74$ and $+15.75$ for \textbf{HumAID} and \textbf{Situation} respectively. However, as can be seen from the table, there is still a gap between P-ZSC (unsupervised) and Sup. BERT (the supervised BERT run). This suggests that although the results of P-ZSC constitute substantial progress in crisis zero-shot (label-fully-unseen) text classification, crisis zero-shot learning remains challenging and cannot yet be considered to be a solved problem. On the other hand, the necessity for significant quantities of expensive annotations (both in terms of time and effort) means that a fully-supervised approach is impractical for real-world crisis events.


\begin{table}[]
\centering
\begin{tabular}{lll}
\toprule
              & HumAID & Situation \\
              \midrule
P-ZSC         & 67.09  & 55.02     \\
\midrule
- self-training & 60.73  & 50.51     \\
- PLA           & 56.43  & 46.20      \\
- PLA+LE        & 55.03  & 43.26    \\
\bottomrule
\end{tabular}
\caption{Ablation study of P-ZSC on crisis datasets}
\label{tab:pzsc-crisis-ablation-study}
\end{table}

\subsubsection{Ablation study}

To examine the contribution of each component of P-ZSC, an ablation study is conducted, with the results reported in Table~\ref{tab:pzsc-crisis-ablation-study}. This shows the performance of the entire P-ZSC system, and also separate results with the self-training step omitted, with the pseudo-label assignment (PLA) omitted and with both PLA and label expansion (LE) omitted. In each case, the removal of any phase results in a decline in performance for all datasets. This decrease becomes slight when only the self-training phase is removed and becomes much greater when both the PLA and LE phases are removed. This indicates that all phases are important to maintain peak effectiveness.

\subsubsection{Pseudo-labelled data quality}

\begin{figure*}[!h]
     \centering
      \begin{subfigure}[b]{0.45\textwidth}
         \centering
         \includegraphics[width=\textwidth]{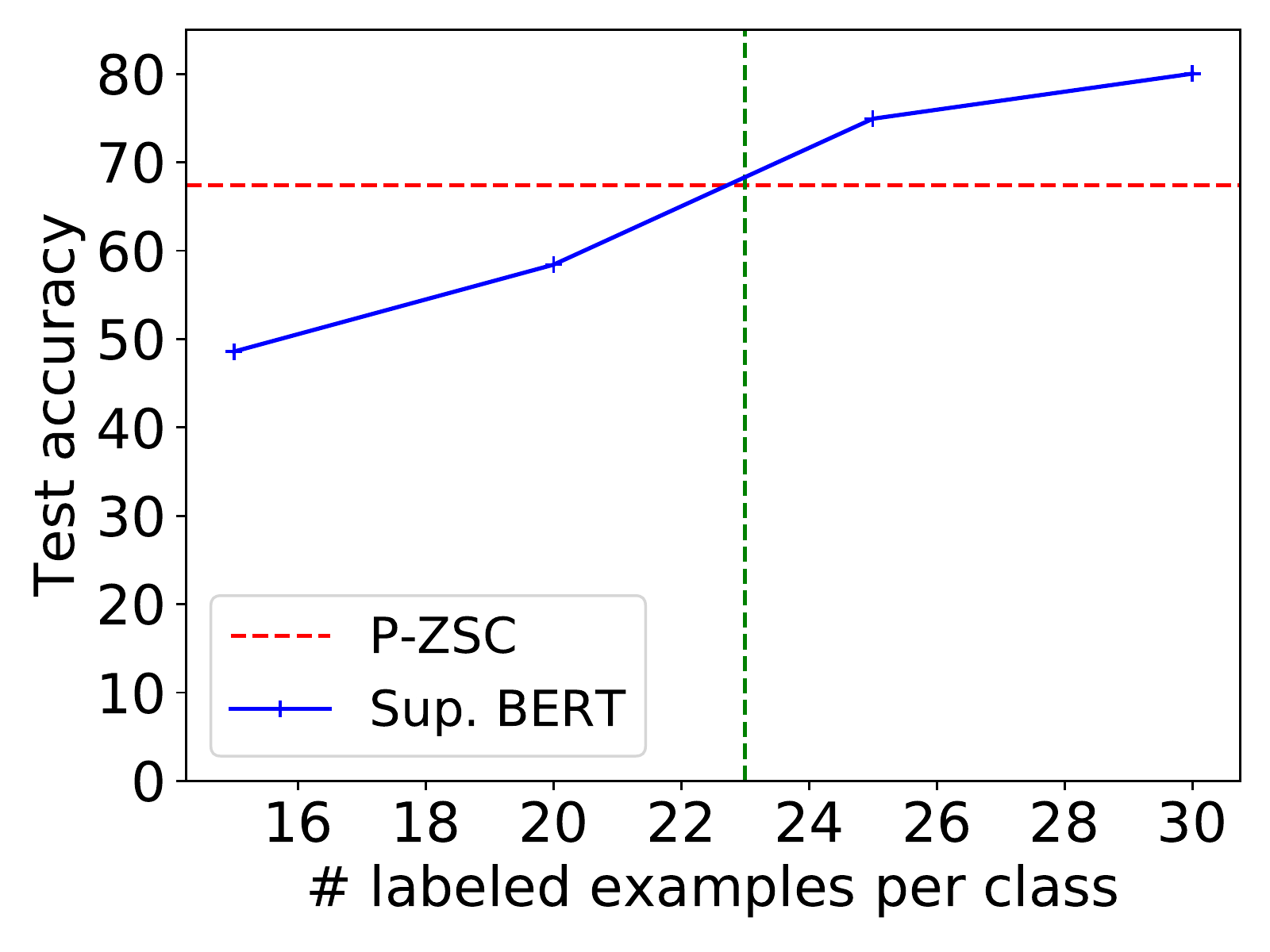}
         \caption{\textbf{HumAID}}
         \label{fig:humaid_cpc}
     \end{subfigure}
     \begin{subfigure}[b]{0.45\textwidth}
         \centering
         \includegraphics[width=\textwidth]{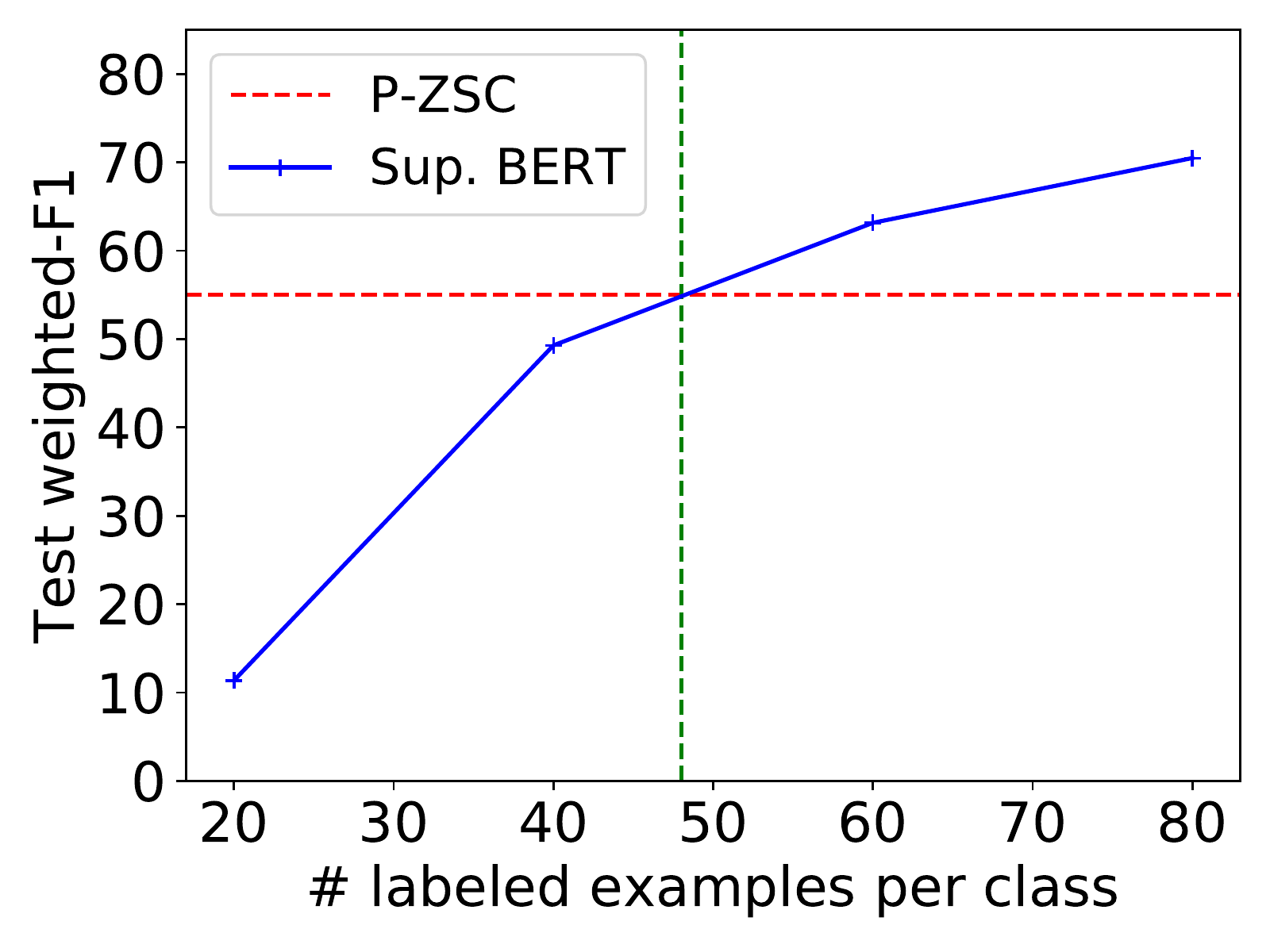}
         \caption{\textbf{Situation}}
         \label{fig:situation_cpc}
     \end{subfigure}
        \caption{The investigation of pseudo-labelled data by P-ZSC on crisis datasets}
        \label{fig:crisis-cpc}
\end{figure*}

In the system, the only ``supervision'' resources for the downstream model training is from the pseudo-labelled data that is obtained via LE and PLA. In the pipeline of P-ZSC, pseudo-labelled data is obtained without any human supervision, using only the label names and an unlabelled corpus relating to the target task. Given the importance of the pseudo-labelled data in the final performance, the pseudo-labelled data is constructed to be a subset of the unlabelled corpus (i.e., only high-confidence label allocations are included in the pseudo-labelled set). To quantify the quality of pseudo-labelled data, an extra experiment is conducted for this purpose using similar methodology to~\cite{meng-etal-2020-text}. In this methodology, P-ZSC is compared with the Sup. BERT run fine-tuned on different numbers of actually-labelled examples per class. From Figure~\ref{fig:crisis-cpc}, it is noticed that the pseudo-labelling of P-ZSC can equal the performance of $23$ and $48$ actually-labelled examples per class on \textbf{HumAID} and \textbf{Situation} respectively. Considering that \textbf{HumAID} has 8 classes and \textbf{Situation} has 11 classes, it therefore accounts for a total of 184 annotated examples for \textbf{HumAID} and 528 for \textbf{Situation}. This indicates how much human annotation effort can be saved when using P-ZSC for the two crisis categorisation tasks as compared to supervised methods.


\subsubsection{Generalisation to domains beyond crisis}

Having verified P-ZSC's effectiveness on two crisis categorisation tasks, it is interesting to also examine the extent to which it can be generalised to domains beyond crises. Three additional datasets are selected for this purpose: \textbf{Topic}~\cite{zhang2015character}, \textbf{UnifyEmotion}~\cite{klinger2018analysis} and \textbf{Emotion}~\cite{saravia-etal-2018-carer}. More details on the selected datasets can be found in Section\ref{sec:rds-beyond-crisis}. The former two are the benchmarking datasets used in~\cite{yin2019benchmarking} and \textbf{Emotion} is another emotion dataset that does not overlap with \textbf{UnifyEmotion}. It is noted that \textbf{Topic} is class-balanced while the other two are not, and thus the evaluation uses accuracy for measuring P-ZSC's performance on \textbf{Topic} and weighted F1 for the other two. In the experiments, P-ZSC and the selected benchmark approaches are re-run on the three datasets following the same experimental configuration as for the crisis datasets. 

\begin{table*}[!h]
\centering
\renewcommand{\arraystretch}{1.2}
\begin{tabular}{lrrr}
\toprule
                            & Topic &  UnifyEmotion & Emotion \\
                            \midrule
\textit{Indirectly-supervised runs}\\
Label similarity~\cite{veeranna2016using}              & 34.62           & 26.21            & 56.03          \\
Entail-single~\cite{yin2019benchmarking}               & 43.80        & 24.70         & 49.60      \\
Entail-ensemble~\cite{yin2019benchmarking}             & 45.70          & 25.20         & 50.16          \\
Entail-Distil~\cite{joe2021} &	44.47 &		29.43&	48.87\\

\midrule
\textit{Weakly-supervised runs} \\
ConWea~\cite{mekala2020contextualized}&	49.81	&	21.39&	47.34\\
X-Class~\cite{wang2021x}&	48.12	&	15.19&	42.21\\
WeSTClass~\cite{meng2018weakly}&	34.96	&	15.45&	22.54\\
LOTClass~\cite{meng-etal-2020-text} &	\textbf{52.07}&	7.19&	16.82\\

P-ZSC  & 50.68         & \textbf{30.22}           & \textbf{64.47}     \\
\midrule
Sup. BERT~\cite{BERT2018}                & 74.86             & 40.10            & 92.02       \\
\bottomrule
\end{tabular}
\caption{Performance of P-ZSC and baselines on test sets of datasets beyond crisis. Due to differing levels of label imbalance, accuracy is used to evaluate the Topic dataset, and weighted F1 is used for the others.}
\label{tab:pzsc-other-results}
\end{table*}

Regarding the comparison to baselines, Table~\ref{tab:pzsc-other-results} shows the performance of P-ZSC and baselines on the test sets of the three selected datasets. Overall, P-ZSC outperforms both the indirectly-supervised runs and the weakly-supervised runs. Although it can be seen  that the accuracy of P-ZSC on \textbf{Topic} is slightly below the best-performing LOTClass ($50.48$ vs $52.07$), it achieves the highest performance on the other two datasets. On \textbf{UnifyEmotion} its performance is similar to the best-performing Entail-Distil, and its \textbf{Emotion} performance is substantially better than all of the other approaches. P-ZSC consistently performs well across all datasets and achieves the best overall performance. In contrast, other baselines may perform well on some datasets but not as well on others.

For example, LOTCLass performs well in \textbf{Topic} but exhibits poorer performance in the other datasets, suggesting that LOTClass does not generalise particularly well. By analysis, to expand labels, LOTClass identifies unlabelled samples with exact-word matches to label names. These are then expanded using BERT masked language modelling (MLM). Masked Category Prediction (MCP) is then used to pseudo-label the unlabelled samples at the token level. For some tasks, this works well since the label names (e.g. ``education'', ``sports'') are straightforward and usually have enough exact-word matches within unlabelled samples. Thus LOTClass performs well in the \textbf{Topic} dataset. However, for datasets like Emotion detection, classes such as ``sadness'' are more abstract and have a more unbalanced distribution, and are not contained directly in unlabelled samples. This leads to few samples in the label name data subsequently used by MLM and MCP. Thus LOTClass obtains relatively poor results for \textbf{UnifyEmotion} and \textbf{Emotion}. 

Concerning the ablation study on the datasets, Table~\ref{tab:pzsc-general-ablation-study} presents the results. The results look similar to those of P-ZSC on the crisis datasets. This further demonstrates that each component of P-ZSC is essential to the final performance. As the results show, the label expansion (LE) component stands out as being particularly important, indicated by the fact that the performance substantially decreases without it.

\begin{figure*}[!h]
     \centering
      \begin{subfigure}[b]{0.45\textwidth}
         \centering
         \includegraphics[width=\textwidth]{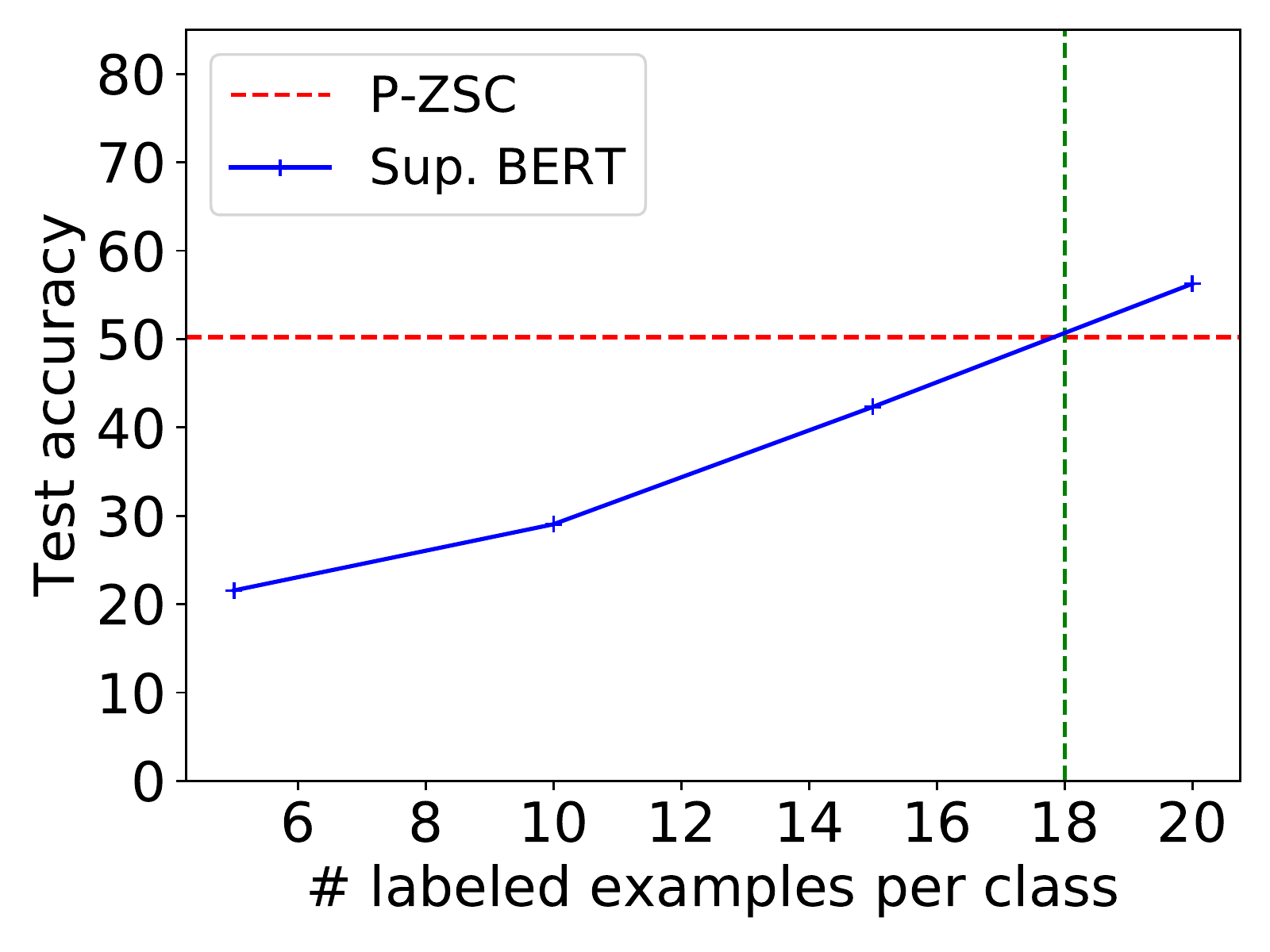}
         \caption{\textbf{Topic}}
         \label{fig:topic_cpc}
     \end{subfigure}
     \begin{subfigure}[b]{0.45\textwidth}
         \centering
         \includegraphics[width=\textwidth]{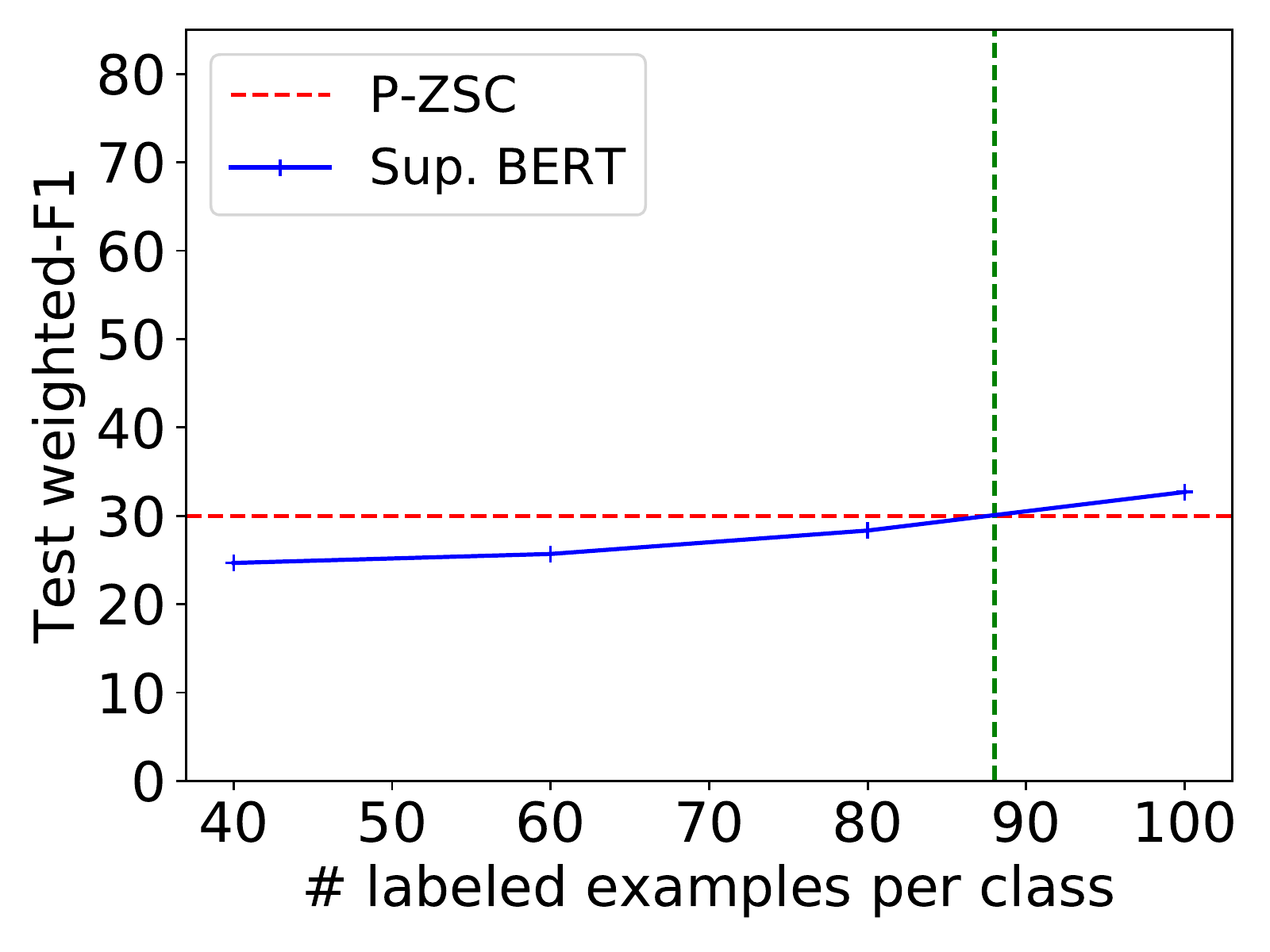}
         \caption{\textbf{UnifyEmotion}}
         \label{fig:unifyemo_cpc}
     \end{subfigure}
          \begin{subfigure}[b]{0.45\textwidth}
         \centering
         \includegraphics[width=\textwidth]{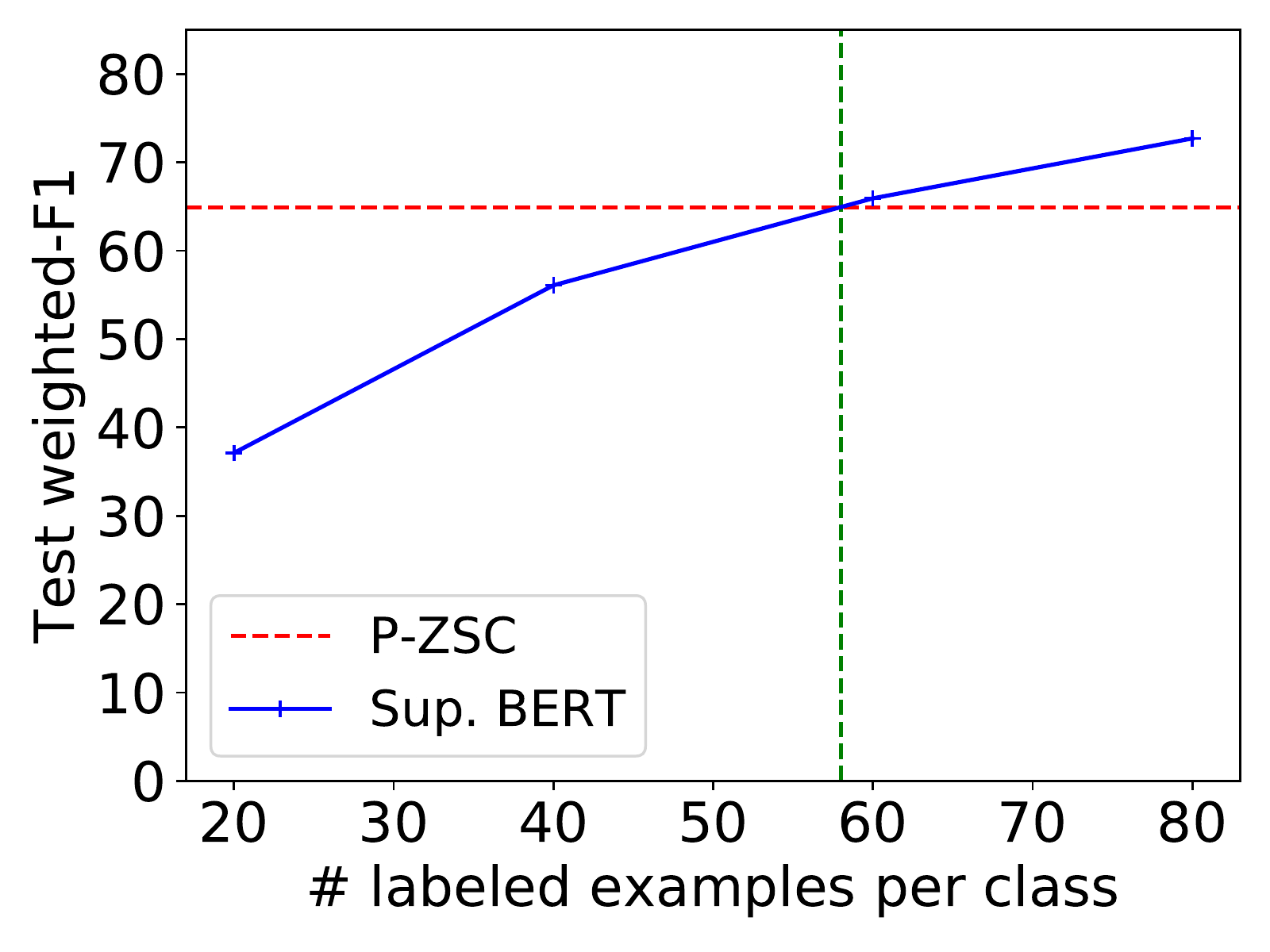}
         \caption{\textbf{Emotion}}
         \label{fig:twiemo_cpc}
     \end{subfigure}
        \caption{The investigation of pseudo-labelled data by P-ZSC on datasets beyond crisis}
        \label{fig:general-cpc}
\end{figure*}

Finally, the investigation of pseudo-labelled data by P-ZSC on the datasets is carried out and the results are plotted in Figure~\ref{fig:general-cpc}. To summarise the results, the pseudo-labelling of P-ZSC can equal the performance of a fully-supervised BERT classifier with $18$, $88$ and $58$ actually-labelled samples per class on \textbf{Topic}, \textbf{UnifyEmotion} and \textbf{Emotion} respectively. In practical terms, this represents a total of 40, 880 and 348 annotated samples as the datasets have 4, 10, and 6 classes respectively. This shows that there is some room to improve the pseudo-labelled data on datasets like \textbf{Topic}. However, there is nonetheless a clear indication that expensive annotations can be dramatically reduced by the use of P-ZSC.



\begin{table*}[!h]
\normalsize
\centering
\renewcommand{\arraystretch}{1.2}
\begin{tabular}{llll}
\hline 
                & Topic                     & UnifyEmotion & Emotion \\
                \hline
P-ZSC         & 50.68                          & 30.22        & 64.47      \\
\hline
- Self-training & 44.18                     &  25.35        & 59.83      \\
- PLA           & 49.83      & 29.86        & 56.35      \\
- PLA + LE      & 46.33                         & 20.97        & 48.59     \\
\hline
\end{tabular}
\caption{Ablation study of P-ZSC on datasets beyond crisis}
\label{tab:pzsc-general-ablation-study}
\end{table*}

\section{Conclusion}

Motivated by the fact that the class space is usually not known until a crisis occurs and that annotation is a burden (particularly when the class space is large), this chapter presented a novel approach called P-ZSC that used pseudo-labelled data for zero-shot crisis message categorisation. To classify a crisis message into new classes that are not represented in any previously-available training data, P-ZSC obtained a pseudo-labelled data set through a matching mechanism between an unlabelled corpus of the target task and the label vocabularies of new classes. In contrast with crisis few-short learning, P-ZSC can be considered to be a form of crisis zero-shot learning because it does not rely on any annotated data for model training but only the pseudo-labelled data, which helps eliminate the annotation cost in real-world crisis message categorisation. By comparing P-ZSC with baselines, it achieved state-of-the-art performance not only in the crisis categorisation task but also was shown to generalise better to tasks of other domains. The investigation of pseudo-labelled data showed that the pseudo-labelled data produced by P-ZSC can perform as well as using a supervised approach with hundreds of manually-annotated samples, indicating the potential for significant time and effort savings in training a model for the task of crisis message categorisation.

\chapter{Conclusion}
\label{ch:conclusion}
\section{Summary of contributions}

This thesis has focused on using messages circulated on social media during crisis situations for the purpose of supporting emergency response. The research has specifically developed computational techniques for categorising the messages in text forms (i.e., crisis tweets) based on the types of aid being requested, in order to provide useful information for emergency responders. In existing curated datasets, human annotators have associated large quantities of crisis-related messages with appropriate labels, and so it is common for researchers to experiment with supervised approaches to this classification problem. However, the nature of emerging crises means that this scenario is unrealistic. Primarily, since crises are time-critical, it is not feasible to spend the time required for the levels of annotation that would be required to train a supervised model. Additionally, there is a significant cost (both in terms of money and expertise) associated with organising sufficient levels of annotation. Therefore, in real-world crisis message categorisation, there is usually low availability of labelled data relating to emerging crisis events, leading to the difficulty of model training in a fully supervised way. Hence, this research was carried out in addressing the challenge of low data availability for practical crisis message categorisation through three scenarios. 

The first, crisis domain adaptation, involves adapting a categorisation model trained on one domain (past crisis events) to perform well on a different but related domain (emerging events). This can be particularly useful in situations where for the same categorisation task, labelled data from past events are available even though no labelled data relating to emerging events are available.

The second, crisis few-shot learning, involves training a categorisation model to learn from a small number of labelled examples, often just a few labelled examples per class. This can be useful in situations where there is a limited amount of labelled data from emerging events available and a new categorisation task is required.

The third, crisis zero-shot learning, involves training a categorisation model to recognise and classify new classes (aid types) without any labelled examples of those classes. This can be useful in situations when there is no labelled data from emerging events available and a new categorisation task is required for emerging events. The following summarises major contributions with proposed approaches for the three research scenarios.

In Chapter~\ref{ch:adaptation} and~\ref{ch:adaptation2}, the focus was on crisis domain adaptation and the findings of this research have been published in~\cite{congcong2020cls,Wang2021b,Wang2021c,Wang2021a}. Specifically, this research first put effort into many-to-many adaptation, which helped to address the need for categorisation models trained on past events being adapted to emerging events without requiring detailed knowledge of those events. To this end, the research presented a multi-task learning (MTL) approach that fine-tuned pre-trained language models (e.g. BERT) for crisis message categorisation, achieving superior performance compared to baselines. An ensemble version of this approach, combining multiple MTL models, was also introduced and demonstrated state-of-the-art performance in experiments. The research also addressed many/one-to-one crisis domain adaptations: developing adaptation models that can learn from a previous crisis or multiple crises in order to categorise messages related to an emerging crisis. For this purpose, the research proposed the use of sequence-to-sequence (seq2seq) pre-trained language models with the embedding of event information (CAST), without the need for any target data. Experimental results showed that CAST outperformed existing state-of-the-art crisis domain adaptation approaches, particularly when it had the least dependence on target data. In addition, the research investigated the selection of source events for a target event adaptation based on the characteristics of the events, and found that combining similar events tended to yield the best performance, as compared to adding dissimilar events to the training set. Overall, the contributions include the development of state-of-the-art approaches for crisis domain adaptation that address user needs for both many-to-many and many/one-to-one adaptations, as well as insights into how to select appropriate source events for a target event adaptation.

In Chapter~\ref{ch:sta-isa}, the research examined crisis few-shot learning, which addresses the need for emergency response systems to handle crisis message categorisation in real-world scenarios where there may be new predefined classes and limited annotated data for emerging events. To address this challenge, the research introduced two novel augmentation approaches: STA and ISA. STA is a self-controlled method that uses seq2seq pre-trained language models to generate new crisis messages based on a small number of labelled messages per class. To improve the quality of generated messages in terms of both lexical diversity and semantic fidelity (label-alignment), the messages were first generated by the seq2seq model to be conditional on the class names and then these messages were fidelity-checked by the model itself. This approach was found to generate messages with superior lexical diversity and semantic fidelity compared to existing baselines. ISA was then proposed as an optimised version of STA that includes an iterative mechanism and duplication mechanism to further improve the quality of generated messages. The results demonstrated that ISA outperformed STA in both lexical diversity and semantic fidelity and achieved better performance in few-shot crisis message categorisation. In addition, the experiments revealed that both STA and ISA exhibited strong generalisation capabilities across other domains such as emotion and topic classification. Overall, the major contributions of this chapter include the development of two state-of-the-art augmentation approaches for crisis few-shot learning and insights into how to improve the quality of augmented crisis messages in terms of lexical diversity and semantic fidelity to enhance performance in few-shot crisis message categorisation.

While crisis few-shot learning addresses the need for emergency response systems to handle crisis message categorisation when there are new predefined classes and limited annotated data for emerging events, it still requires costly annotation when there is a large number of predefined classes and does not account for unequal class distributions, which are common in real-world scenarios. In Chapter~\ref{ch:pzsc}, the research explored crisis zero-shot learning, which addresses the challenge of categorising crisis messages without depending on any annotated data and the findings of this research have been published in~\cite{wang2022using}. To address this challenge, the research proposed a novel approach called P-ZSC that uses pseudo-labelled data for zero-shot crisis message categorisation. P-ZSC obtained the pseudo-labelled data set by matching an unlabelled corpus of the target event with the label vocabularies of new classes, allowing for the categorisation of crisis messages into new classes without the need for labelled data. The results of comparing P-ZSC to baselines showed that it achieved state-of-the-art performance not only in the crisis categorisation domain but also demonstrated strong generalisation capabilities across other domains such as emotion detection and topic classification. It was also found that the pseudo-labelled data produced by P-ZSC was as effective as hundreds of manually annotated samples in both crisis categorisation and non-crisis classification tasks, indicating the possibility of saving a significant amount of time and effort when training a model with P-ZSC for these tasks. When comparing P-ZSC to a fully-supervised BERT model, the results show that a gap still exists in terms of the performance of P-ZSC. However, the fully-supervised scenario is artificial because it relies on pre-annotated data (i.e. whole training dataset) in a crisis situation, which is not practical as a crisis emerges. Due to time constraints and the workload involved, making a dataset like this is not trivial. Hence, a zero-shot approach like P-ZSC is a crucial area of research because it can be realistically deployed in a real crisis situation where users can set up labels to reflect what they are looking for and then start matching messages almost immediately. While P-ZSC does not fully solve this problem, it makes a significant improvement to the state of the art.

In summary, this thesis makes good contributions to the field of crisis informatics by addressing the challenge of limited annotated data availability for categorising crisis messages on social media during emerging events. The approaches presented in this thesis have been developed in close collaboration with real-world scenarios and have demonstrated state-of-the-art performance in experiments. These approaches have the potential to be applied in practice to provide timely and effective humanitarian aid responses.

\section{Future directions}

This research has made progress in improving the categorisation of social media crisis messages in low data settings. However, there is still much work to be done to fully solve this research problem. There are several key areas of future work that should be considered in order to continue advancing in this area. The following identifies limitations of this research and presents a list of future work to consider. 


\begin{itemize}
    \item In the many-to-many adaptation work presented in this thesis, the MTL-based ensemble approach achieved the best performance to date. However, this approach is not very efficient because it involves combining multiple MTL models for crisis message categorisation. Inference efficiency is a critical concern during crises, when messages may come in rapidly, and hence it is an area of future work to optimise the efficiency of the ensemble approach. Techniques such as distillation, pruning, or compression of categorisation models could potentially improve the speed at which messages can be processed, leading to more efficient and timely emergency responses.
    \item In crisis few-shot learning, STA and ISA use human-designed templates to ensure lexical diversity and semantic fidelity for the generated messages. This makes it challenging to determine which templates are the most effective for optimal performance. As a result, future work could focus on automating the process of finding the optimal set of templates for STA and ISA.
    \item In P-ZSC, although the pseudo-labelled data represents high quality in some aspects, there remains some room to improve it. For example, the existing matching mechanism simply assigns a label to a message when they are locally matched with each other with high confidence. Hence, the future work can seek to explore better solutions such as global semantic meaning matching for obtaining better quality of pseudo-labelled data.
    \item It is necessary to develop a dynamic system that can adapt to different levels of data availability. While this research has presented state-of-the-art approaches for crisis domain adaptation, few-shot, and zero-shot learning, it is important to combine these approaches into a system that can handle both high and low levels of data availability, or that can adapt to changing data availability over time. This includes integrating approaches for crisis message categorisation depending on the amount of available data. For instance, in situations where few labeled data and larger quantities of unlabeled data for emerging events are available, it would be valuable to explore how the zero-shot P-ZSC approach could be used in a semi-supervised manner in conjunction with the augmentation-based few-shot approaches (STA and ISA) to further improve crisis message categorisation.
    \item The proposed approaches used in this research focus on the raw text content of crisis messages. It is important to incorporate additional sources of information beyond the text of the messages themselves. This could include incorporating images, locations, or knowledge graphs to provide additional context and help improve the accuracy of categorisation.
    \item This research so far has just handled English crisis messages. It is important to extend this work to other languages, as crisis messages are often written in a variety of languages. Cross-lingual crisis message categorisation is therefore a critical area of future work.
    \item The proposed approaches operate on the assumption that the crisis messages being learned from, and being classified, are credible and reliable sources of information. Verifying the credibility of crisis messages is also critical, as false or misleading information can cause confusion and hinder effective response. Hence, it would be helpful to add a credibility-checking component before feeding the messages to the categorisation system. 
    \item This research has made strides in decreasing the need for labelled data in the categorisation of real-world social media crisis messages. However, it still requires some labelled and/or unlabelled data for emerging events. One potential area of future work is to optimise large-scale language models like GPT-3 and ChatGPT, which are trained on vast amounts of text data. If these models can be made more efficient, they have the potential to be used for crisis message categorisation of any event without the need for additional training on domain-specific data.
\end{itemize}

Overall, these are key areas of future work that are important not only for this specific research, but also for the broader community working on social media crisis message categorisation in the face of low data availability.

\bibliographystyle{plain}
\begin{flushleft}
\bibliography{bibtex}
\end{flushleft}
\appendix
\chapter{Appendix}

\section{Information types of TREC-IS}
\label{appendix:trecis-its}

\begin{table*}[!h]
\footnotesize
\def\arraystretch{1.0}
\centering
\begin{tabular}{|c|l|p{8cm}|}
\hline
\multicolumn{1}{|l|}{}           & Category                     & Description                                                                                                      \\ \hline
\multirow{6}{*}{Actionable}      & Request-GoodsServices        & The user is asking for a particular service or physical good                                                   \\ \cline{2-3} 
                                 & Request-SearchAndRescue      & The user is requesting a rescue (for themselves or others)                                                      \\ \cline{2-3} 
                                 & Report-NewSubEvent           & The user is reporting a new occurence that public safety   officers need to respond to                          \\ \cline{2-3} 
                                 & Report-ServiceAvailable      & The user is reporting that they or someone else is providing a  service                                         \\ \cline{2-3} 
                                 & CallToAction-MovePeople      & The user is asking people to leave an area or go to another   area                                               \\ \cline{2-3} 
                                 & Report-EmergingThreats       & The user is reporting a potential problem that may cause   future loss of life or damage                         \\ \hline
\multirow{19}{*}{Non-actionable} & CallToAction-Volunteer       & The user is asking people to volunteer to help the response   effort                                             \\ \cline{2-3} 
                                 & CallToAction-Donations       & The user is asking people to donate goods/money                                                                  \\ \cline{2-3} 
                                 & Report-Weather               & The user is providing a weather report (current or forcast)                                                      \\ \cline{2-3} 
                                 & Report-Location              & The post contains information about the user or observation   location                                           \\ \cline{2-3} 
                                 & Request-InformationWanted    & The user is requesting information                                                                               \\ \cline{2-3} 
                                 & Report-FirstPartyObservation & The user is giving an eye-witness account                                                                        \\ \cline{2-3} 
                                 & Report-ThirdPartyObservation & The user is reporting a information that they recieved from   someone else                                       \\ \cline{2-3} 
                                 & Report-MultimediaShare       & The user is sharing images or video                                                                              \\ \cline{2-3} 
                                 & Report-Factoid               & The user is relating some facts, typically  numerical                                                             \\ \cline{2-3} 
                                 & Report-Official              & An official report by a government or public safety   representative                                             \\ \cline{2-3} 
                                 & Report-News                  & The post is a news report providing/linking to   current/continious coverage of the event                        \\ \cline{2-3} 
                                 & Report-CleanUp               & A report of the clean up after the event                                                                         \\ \cline{2-3} 
                                 & Report-Hashtags              & Reporting which hashtags correspond to each event                                                                \\ \cline{2-3} 
                                 & Report-OriginalEvent         & A report of the original event occuring.      \\ \cline{2-3} 
                                 & Other-ContextualInformation  & The post contains contextual information that can help understand the event, but is not about the event itself\\ \cline{2-3} 
                                 & Other-Advice                 & The user is providing some advice to the public                                                                \\ \cline{2-3} 
                                 & Other-Sentiment              & The post is expressing some sentiment about the event                                                            \\ \cline{2-3} 
                                 & Other-Discussion             & Users are discussing the event                                                                                   \\ \cline{2-3} 
                                 & Other-Irrelevant             & The post is unrelated to the event or contains no information                                                   \\ \hline
\end{tabular}

\caption{Actionable and non-actionable information types in the TREC-IS dataset~\cite{mccreadie2019trec}}
\label{tab:actionable-versus-non}
\end{table*}


\end{document}